# A General Theory of Hypothesis Tests and Confidence Regions for Sparse High Dimensional Models

Yang Ning[*]   Han Liu[†]

**Abstract**

We consider the problem of uncertainty assessment for low dimensional components in high dimensional models. Specifically, we propose a decorrelated score function to handle the impact of high dimensional nuisance parameters. We consider both hypothesis tests and confidence regions for generic penalized M-estimators. Unlike most existing inferential methods which are tailored for individual models, our approach provides a general framework for high dimensional inference and is applicable to a wide range of applications. From the testing perspective, we develop general theorems to characterize the limiting distributions of the decorrelated score test statistic under both null hypothesis and local alternatives. These results provide asymptotic guarantees on the type I errors and local powers of the proposed test. Furthermore, we show that the decorrelated score function can be used to construct point estimators that are semi-parametrically efficient and the optimal confidence regions. We also generalize this framework to broaden its applications. First, we extend it to handle high dimensional null hypothesis, where the number of parameters of interest can increase exponentially fast with the sample size. Second, we establish the theory for model misspecification. Third, we go beyond the likelihood framework, by introducing the generalized score test based on general loss functions. Thorough numerical studies are conducted to back up the developed theoretical results.

## 1 Introduction

Given $n$ independent and identically distributed multivariate random variables $\boldsymbol{U}_1,...,\boldsymbol{U}_n$, assume that they are generated from a statistical model $\mathcal{P} = \{\mathbb{P}_{\boldsymbol{\beta}} : \boldsymbol{\beta} \in \Omega\}$, where $\boldsymbol{\beta}$ is a vector of unknown parameter with dimension $d \gg n$ and $\Omega$ is the parameter space. In this high dimensional problem, one general approach to estimate $\boldsymbol{\beta}$ is given by the following penalized M-estimator,

$$\widehat{\boldsymbol{\beta}} = \operatorname*{argmin}_{\boldsymbol{\beta}} \ell(\boldsymbol{\beta}) + P_\lambda(\boldsymbol{\beta}), \qquad (1.1)$$

[*]Department of Operations Research and Financial Engineering, Princeton University, Princeton, NJ 08544, USA; e-mail: yning@princeton.edu.
[†]Department of Operations Research and Financial Engineering, Princeton University, Princeton, NJ 08544, USA; e-mail: hanliu@princeton.edu.

where $\ell(\boldsymbol{\beta})$ in many cases corresponds to the negative log-likelihood and $P_\lambda(\boldsymbol{\beta})$ is a penalty function with a tuning parameter $\lambda$. There has been much literature on studying different penalty functions, which can be classified into convex penalties and nonconvex penalties. The most popular convex penalty is the $L_1$ penalty, also known as the Lasso penalty (Tibshirani, 1996). The theoretical properties of the estimator with the Lasso penalty have been extensively studied in the literature. For instance, the sparsity oracle inequalities and statistical rate of convergence of the Lasso estimator are established by Meinshausen and Yu (2009); Bickel et al. (2009); Bunea et al. (2007); van de Geer (2008); Zhang (2009); Negahban et al. (2012), and the variable selection consistency is studied by Meinshausen and Bühlmann (2006); Zhao and Yu (2006); Wainwright (2009). The class of nonconvex penalties includes MCP (Zhang, 2010a), SCAD (Fan and Li, 2001), and capped-$L_1$ penalty (Zhang, 2010b), among others. The estimation consistency and oracle properties of the nonconvex estimators are investigated by Fan et al. (2012); Fan and Lv (2011); Loh and Wainwright (2013); Wang et al. (2013a); Zhang et al. (2013); Wang et al. (2013b).

Though significant progress has been made towards understanding the estimation theory for high dimensional models, how to quantify the uncertainty of the obtained penalized estimators remains largely unexplored. To bridge this gap, this paper proposes a novel device, named as decorrelated score function, to test statistical hypotheses and construct confidence regions for low dimensional components in high dimensional models. In particular, we partition the parameter $\boldsymbol{\beta}$ as $\boldsymbol{\beta} = (\theta, \boldsymbol{\gamma})$, where $\theta$ is a univariate parameter of interest and $\boldsymbol{\gamma}$ is a $(d-1)$ dimensional nuisance parameter. We aim to test the null hypothesis $H_0 : \theta^* = 0$, where $\theta^*$ is the true value of $\theta$. Compared to its well-studied low dimensional counterpart, the main challenge of this problem is the presence of high dimensional nuisance parameters. To handle this challenge, we apply a decorrelation operation on the high dimensional score function, so that the obtained decorrelated score function for $\theta$ becomes uncorrelated with the nuisance score functions. Unlike the classical score function, the decorrelated score function can handle the impact of high dimensional nuisance parameters. With the decorrelated score function as a key ingredient, our framework is quite general. It allows us to use a wide range of estimation procedures for the high dimensional nuisance parameters. Therefore, it provides valid inference for penalized M-estimators with both convex and the nonconvex penalties.

Theoretically, from the hypothesis testing perspective, we prove the limiting distributions of the decorrelated score test statistic under both the null hypothesis and local alternatives. These results characterize the asymptotic type I errors and local powers of the test. We further establish the uniform weak convergence of the test statistic, which implies honesty of the score test in terms of the type I errors and powers. From the confidence interval perspective, we show that the decorrelated score function can be used to construct an estimator that achieves the information lower bound and therefore it leads to the optimal confidence region. These theorems are established under full generality with mild conditions imposed on the likelihood and the penalized M-estimators. Thus, this paper provides a general theory for hypothesis tests and confidence regions in high dimensional models. To illustrate the applicability of the theory, we verify the validity of these conditions under high dimensional linear models and generalized linear models.

To further broaden our framework, we consider several extensions of the proposed score test.

First, we propose a score test for high dimensional null hypothesis $H_0^{d_0} : \boldsymbol{\theta}^* = \mathbf{0}$, where $\boldsymbol{\theta}^*$ has dimension $d_0$. Under some general conditions, the limiting distribution of the score test statistic is established and the validity of a practical multiplier bootstrap procedure is justified, even if $d_0$ has the same scaling as $d$. These conditions are then verified under high dimensional linear models. Second, we investigate the properties of the score test under misspecified models. This provides a formal justification for the inference on oracle parameters (i.e., least false parameters) in the high dimensional setting. Finally, we propose a generalized score test based on the loss function other than the negative log-likelihood. This result can be useful for inference in high dimensional semiparametric models, whose log-likelihoods can be complicated or infeasible to maximize.

We point out that the decorrelated score test can be viewed as a high dimensional extension of the Rao's score test in statistics (Cox and Hinkley, 1979) and the Lagrange multiplier test in econometrics (Godfrey, 1991). In particular, the decorrelated score test is asymptotically equivalent to these two tests in low dimensional models. However, in high dimensions, the type I error can be controlled by the decorrelated score test rather than the two classical tests.

## 1.1 Related Works

In the literature, there are very limited discussions about the uncertainty assessment for the regularized estimators in high dimensional models. In an early work, Knight and Fu (2000) showed that the limiting distribution of the Lasso estimator is not normal even in the low dimensional setting. Recently, in the high dimensional setting, several authors including Meinshausen et al. (2009); Wasserman and Roeder (2009); Meinshausen and Bühlmann (2010); Shah and Samworth (2013) considered p-values based on the sample splitting technique or subsampling. Unlike these approaches, our approach avoids sample splitting and is potentially more powerful. For the Lasso estimator in the linear regression, Lockhart et al. (2012); Taylor et al. (2014); Lee et al. (2013) considered the conditional inference given the event that some covariates have been selected, which is philosophically different from our unconditional inference. An instrumental variable approach together with a double selection procedure is proposed by Belloni et al. (2012, 2013). From a different perspective, Zhang and Zhang (2011); van de Geer et al. (2014); Javanmard and Montanari (2013) proposed a bias correction method also named as the LDPE or de-sparsifying method for constructing confidence intervals for linear models or generalized linear models with the Lasso penalty. Compared with these works, our method can incorporate the general family of penalized M-estimators (including nonconvex penalties or estimators obtained by greedy algorithms). More importantly, our score test is applicable to general models and can be used to infer the oracle parameter under model misspecification. With a class of nonconvex penalty functions, Fan and Lv (2011); Bradic et al. (2011) established the asymptotic normality results for low dimensional nonzero parameters based on the oracle properties. However, such oracle results require strong minimal signal conditions which may not hold in many applications and the uncertainty of the estimation for small signals cannot be evaluated. In contrast, our method does not require variable selection consistency and can provide valid inference for small parameters. From the hypothesis testing perspective, Voorman et al. (2014) proposed a penalized score test. However, they focused on a null hypothesis depending on the tuning parameter and their test is biased for $H_0 : \theta^* = 0$. In addition, the validity of their test hinges on an irrepresentable type condition, which is not needed here.

## 1.2 Organization of the Paper

In Section 2, we propose the decorrelated score function. Section 3 is devoted to establishing general results for the score tests and confidence regions. In Sections 4 and 5, we provide implications of these general results for high dimensional linear and generalized linear models. In Section 6, we consider several extensions of the score test. Section 7 provides numerical results and Section 8 contains more discussions. We defer most of the proofs to the Appendices.

## 1.3 Notations

The following notation is adopted throughout this paper. For $\mathbf{v} = (v_1, ..., v_d)^T \in \mathbb{R}^d$, and $1 \leq q \leq \infty$, we define $||\mathbf{v}||_q = (\sum_{i=1}^{d} |v_i|^q)^{1/q}$, $||\mathbf{v}||_0 = |\text{supp}(\mathbf{v})|$, where $\text{supp}(\mathbf{v}) = \{j : v_j \neq 0\}$ and $|A|$ is the cardinality of a set $A$. Denote $||\mathbf{v}||_\infty = \max_{1 \leq i \leq d} |v_i|$ and $\mathbf{v}^{\otimes 2} = \mathbf{v}\mathbf{v}^T$. For a matrix $\mathbf{M} = [M_{jk}]$, let $||\mathbf{M}||_{\max} = \max_{jk} |M_{jk}|$, $||\mathbf{M}||_1 = \sum_{jk} |M_{jk}|$, $||\mathbf{M}||_{\ell_\infty} = \max_j \sum_k |M_{jk}|$. If the matrix $\mathbf{M}$ is symmetric, then $\lambda_{\min}(\mathbf{M})$ and $\lambda_{\max}(\mathbf{M})$ are the minimal and maximal eigenvalues of $\mathbf{M}$. We denote $\mathbf{I}_d$ as the $d \times d$ identity matrix. For $S \subseteq \{1, ..., d\}$, let $\mathbf{v}_S = \{v_j : j \in S\}$ and $\bar{S}$ be the complement of $S$. The gradient and subgradient of a function $f(\mathbf{x})$ are denoted by $\nabla f(\mathbf{x})$ and $\partial f(\mathbf{x})$ respectively. Let $\nabla_S f(\mathbf{x})$ denote the gradient of $f(\mathbf{x})$ with respect to $\mathbf{x}_S$. For two positive sequences $a_n$ and $b_n$, we write $a_n \asymp b_n$ if $C \leq a_n/b_n \leq C'$ for some $C, C' > 0$. Similarly, we use $a \lesssim b$ to denote $a \leq Cb$ for some constant $C > 0$. For a sequence of random variables $X_n$, we write $X_n \rightsquigarrow X$, for some random variable $X$, if $X_n$ converges weakly to $X$, and write $X_n \rightarrow_p a$, for some constant $a$, if $X_n$ converges in probability to $a$. Given $a, b \in \mathbb{R}$, let $a \vee b$ and $a \wedge b$ denote the maximum and minimum of $a$ and $b$. For notational simplicity, we use $C, C', C''$ to denote generic constants, whose values can change from line to line. To characterize the tail behavior of random variables, we introduce the following definitions.

**Definition 1.1** (Sub-exponential variable and sub-exponential norm). A random variable $X$ is called sub-exponential if there exists some positive constant $K_1$ such that $\mathbb{P}(|X| > t) \leq \exp(1 - t/K_1)$ for all $t \geq 0$. The sub-exponential norm of $X$ is defined as $||X||_{\psi_1} = \sup_{p \geq 1} p^{-1} (\mathbb{E}|X|^p)^{1/p}$.

**Definition 1.2** (Sub-Gaussian variable and sub-Gaussian norm). A random variable $X$ is called sub-Gaussian if there exists some positive constant $K_2$ such that $\mathbb{P}(|X| > t) \leq \exp(1 - t^2/K_2^2)$ for all $t \geq 0$. The sub-Gaussian norm of $X$ is defined as $||X||_{\psi_2} = \sup_{p \geq 1} p^{-1/2} (\mathbb{E}|X|^p)^{1/p}$.

## 2 Score Function for High Dimensional Models

In this section, we first introduce a general modeling framework. Then, we review the classical Rao's score test for low dimensional models, and highlight the difficulty for directly applying it to models with high dimensional nuisance parameters. Finally, we propose a new device, named as decorrelated score function, to construct valid tests and confidence regions in high dimensions.

## 2.1 Modeling Framework

Let $U$ denote a multivariate random variable and assume that the probability distribution of $U$ follows from a high dimensional statistical model $\mathcal{P} = \{\mathbb{P}_{\boldsymbol{\beta}} : \boldsymbol{\beta} \in \Omega\}$, where $\boldsymbol{\beta}$ is a $d$ dimensional unknown parameter and $\Omega$ is the parameter space. To infer the true value of $\boldsymbol{\beta}$, denoted by $\boldsymbol{\beta}^*$, we assume that there exist $n$ independently identically distributed copies of $U$, that is $U_1, ..., U_n$. In many statistical problems, the unknown parameter $\boldsymbol{\beta}$ can be partitioned as $\boldsymbol{\beta} = (\theta, \boldsymbol{\gamma})$, where $\theta$ is a univariate parameter of interest and $\boldsymbol{\gamma}$ is a $(d-1)$ dimensional nuisance parameter. The statistical inference problem can be formulated as testing the validity of the null hypothesis $H_0 : \theta^* = 0$ or constructing confidence intervals for $\theta^*$. The likelihood plays a pivotal role for inference in statistics. For simplicity of presentation, we introduce the notation of the negative log-likelihood,

$$\ell(\theta, \boldsymbol{\gamma}) = \frac{1}{n} \sum_{i=1}^{n} \ell_i(\theta, \boldsymbol{\gamma}), \quad \text{where} \quad \ell_i(\theta, \boldsymbol{\gamma}) = -\log f(\boldsymbol{U}_i; \theta, \boldsymbol{\gamma}), \tag{2.1}$$

denotes the negative log-likelihood for the $i$th observation and $f$ is the density function corresponding to the model $\mathbb{P}_{\boldsymbol{\beta}}$. Given the negative log-likelihood $\ell(\boldsymbol{\beta})$, the information matrix for $\boldsymbol{\beta}$ is defined as $\mathbf{I} = \mathbb{E}_{\boldsymbol{\beta}}(\nabla^2 \ell(\boldsymbol{\beta}))$, and the partial information matrix is defined as $I_{\theta|\boldsymbol{\gamma}} = I_{\theta\theta} - \mathbf{I}_{\theta\boldsymbol{\gamma}} \mathbf{I}_{\boldsymbol{\gamma}\boldsymbol{\gamma}}^{-1} \mathbf{I}_{\boldsymbol{\gamma}\theta}$, where $I_{\theta\theta}, \mathbf{I}_{\theta\boldsymbol{\gamma}}, \mathbf{I}_{\boldsymbol{\gamma}\boldsymbol{\gamma}}$ and $\mathbf{I}_{\boldsymbol{\gamma}\theta}$ are the corresponding partitions of $\mathbf{I}$. Similarly, denote $\mathbf{I}^* = \mathbb{E}_{\boldsymbol{\beta}^*}(\nabla^2 \ell(\boldsymbol{\beta}^*))$. Hereafter, we use $\mathbb{P}_{\boldsymbol{\beta}^*}(\cdot)$ and $\mathbb{E}_{\boldsymbol{\beta}^*}(\cdot)$ to denote the probability and expectation evaluated under the joint probability density of $(U_1, ..., U_n)$ indexed by the parameter $\boldsymbol{\beta}^*$.

## 2.2 Challenges of Score Test in High Dimensional Models

When the dimension of the parameters is fixed and much smaller than the sample size, the classical Rao's score test for $H_0 : \theta^* = 0$ versus $H_1 : \theta^* \neq 0$ is based on the profile score function $\nabla_\theta \ell(0, \widehat{\boldsymbol{\gamma}}(0))$, where $\widehat{\boldsymbol{\gamma}}(\theta) = \mathrm{argmin}_{\boldsymbol{\gamma}} \ell(\theta, \boldsymbol{\gamma})$ is the constrained maximum likelihood estimator (MLE) of $\boldsymbol{\gamma}$ for fixed $\theta$. Under the null hypothesis, it is well known that (Cox and Hinkley, 1979),

$$n^{1/2} \nabla_\theta \ell(0, \widehat{\boldsymbol{\gamma}}(0)) \rightsquigarrow N(0, I^*_{\theta|\boldsymbol{\gamma}}). \tag{2.2}$$

The Rao's score test statistic is given by $S_c = n\{\nabla_\theta \ell(0, \widehat{\boldsymbol{\gamma}}(0))\}^2 \widehat{I}_{\theta|\boldsymbol{\gamma}}^{-1}$, where $\widehat{I}_{\theta|\boldsymbol{\gamma}}$ is some empirical estimate of $I^*_{\theta|\boldsymbol{\gamma}}$. The score test is obtained by rejecting $H_0$, if and only if the value of $S_c$ is large. In low dimensions, the score test is known to be asymptotically optimal against local alternative hypothesis.

In this paper, we are mainly interested in testing parameters for high dimensional models, in which $d$ can be much larger than $n$. When the nuisance parameter $\boldsymbol{\gamma}$ is of high dimension, the constrained maximum likelihood estimator (MLE) $\widehat{\boldsymbol{\gamma}}(\theta)$ is no longer consistent (Portnoy, 1984, 1985). Even though the corresponding maximum penalized likelihood estimator (MPLE) such as the Lasso estimator is consistent under certain conditions, it does not have a tractable limiting distribution even in the fixed dimensional case (Knight and Fu, 2000). To illustrate the infeasibility of the Rao's score test for high dimensional models, for any estimator $\widetilde{\boldsymbol{\gamma}}$, consider the following

Taylor expansion,

$$n^{1/2}\nabla_\theta\ell(0,\widetilde{\gamma}) = \underbrace{n^{1/2}\nabla_\theta\ell(0,\gamma^*)}_{I_1} + \underbrace{n^{1/2}\nabla^2_{\theta\gamma}\ell(0,\gamma^*)(\widetilde{\gamma} - \gamma^*)}_{I_2} + \text{Rem}, \qquad (2.3)$$

where Rem represents the remainder terms. Although $I_1$ usually converges weakly to a normal distribution due to the Central Limit Theorem, the asymptotic normality of $n^{1/2}\nabla_\theta\ell(0,\widetilde{\gamma})$ fails due to the incompatible effect of sparsity on $I_2$ and Rem. First, to ensure Rem is asymptotically ignorable, $\widetilde{\gamma}$ must have a fast convergence rate, which rules out the nonsparse MLE estimators. Second, for those sparse estimators such as MPLE, similar to the argument in Knight and Fu (2000), $I_2$ may converge to some intractable limiting distribution, if it exists. Hence, the score function with $\gamma$ estimated by either MLE or MPLE does not have a simple limiting distribution in the high dimensional setting.

## 2.3 A Decorrelated Score Method for High Dimensional Models

As seen in the previous section, the score function with estimated nuisance parameters cannot be used for inference in high dimensional models. This motivates us to develop a new type of score function applicable in this more challenging regime. Assume that $\beta$ can be estimated by the penalized M-estimator (1.1). In many applications, the penalty function $P_\lambda(\beta)$ in (1.1) is decomposable in the sense that $P_\lambda(\beta) = \sum_{j=1}^{d} p_\lambda(\beta_j)$. In our framework, we allow both convex and nonconvex penalties. For instance, $p_\lambda(\beta_j)$ can be taken as the $L_1$ penalty (Tibshirani, 1996) $p_\lambda(\beta_j) = \lambda|\beta_j|$, the SCAD penalty (Fan and Li, 2001),

$$p_\lambda(\beta_j) = \int_0^{|\beta_j|} \left\{\lambda I(z \leq \lambda) + \frac{(a\lambda - z)_+}{a-1} I(z > \lambda)\right\} dz,$$

for some $a > 2$ and the MCP penalty (Zhang, 2010a),

$$p_\lambda(\beta_j) = \lambda \int_0^{|\beta_j|} \left(1 - \frac{z}{\lambda b}\right)_+ dz,$$

for some $b > 0$. To infer the value of $\theta$, we propose a new type of score function given by

$$S(\theta,\gamma) = \nabla_\theta\ell(\theta,\gamma) - \mathbf{w}^T\nabla_\gamma\ell(\theta,\gamma), \quad \text{with} \quad \mathbf{w}^T = \mathbf{I}_{\theta\gamma}\mathbf{I}_{\gamma\gamma}^{-1}. \qquad (2.4)$$

We call $S(\theta,\gamma)$ as the decorrelated score function for $\theta$. This name comes from the fact that $S(\theta,\gamma)$ is uncorrelated with the nuisance score functions $\nabla_\gamma\ell(\beta)$, i.e., $\mathbb{E}_\beta(S(\beta)\nabla_\gamma\ell(\beta)) = 0$. We can show that the decorrelation operation is crucial to control the variability of high order terms in the Taylor expansions, similar to $I_2$ in (2.3). Further geometric insight of $S(\theta,\gamma)$ and its connection with the profile score will be discussed in Section 2.5.

To construct a score test for $\theta^* = 0$ based on $S(\theta,\gamma)$, one needs to estimate the nuisance parameter $\gamma$ and unknown quantity $\mathbf{w}$. The whole estimation procedure is described in Algorithm 1. The output of this algorithm is the estimated decorrelated score function $\widehat{S}(\theta,\widehat{\gamma})$, where $\gamma$ is estimated by $\widehat{\gamma}$ in (1.1) and $\mathbf{w}$ is estimated by $\widehat{\mathbf{w}}$ in (2.5). Hence, we can calculate the value of the estimated decorrelated score function $\widehat{S}(\theta, \widehat{\gamma})$ at $\theta = 0$ to evaluate the validity of the null hypothesis. Note that the key step in the algorithm is to estimate $\mathbf{w}$, which essentially searches for the best sparse linear combination of the nuisance score functions $\nabla_{\gamma}\ell(\theta, \gamma)$ to approximate the score function $\nabla_{\theta}\ell(\theta, \gamma)$, in a computationally efficient way. This can be also seen from the alternative formulation in (2.8). Since the nuisance score functions $\nabla_{\gamma}\ell(\boldsymbol{\beta})$ and $\mathbf{w}$ have dimension $d - 1$, we may need to impose the sparsity assumption on $\mathbf{w}$ to control the estimation error. The implication of this assumption and the comparison with existing methods are discussed in Section 4.

---

**Algorithm 1** Calculate the estimated decorrelated score function $\widehat{S}(\theta, \widehat{\gamma})$

**Require:** : Negative log-likelihood $\ell(\theta, \gamma)$, penalty function $P(\cdot)$ and tuning parameters $\lambda$ and $\lambda'$.

(i): Calculate $\widehat{\boldsymbol{\beta}}$ in (1.1) and partition $\widehat{\boldsymbol{\beta}}$ as $\widehat{\boldsymbol{\beta}} = (\widehat{\theta}, \widehat{\gamma})$;

(ii): Estimate $\mathbf{w}$ by the Dantzig type estimator $\widehat{\mathbf{w}}$,

$$\widehat{\mathbf{w}} = \operatorname{argmin} ||\mathbf{w}||_1, \quad \text{s.t.} \quad ||\nabla^2_{\theta\gamma}\ell(\widehat{\boldsymbol{\beta}}) - \mathbf{w}^T\nabla^2_{\gamma\gamma}\ell(\widehat{\boldsymbol{\beta}})||_\infty \leq \lambda', \quad (2.5)$$

(iii): Calculate the estimated decorrelated score function

$$\widehat{S}(\theta, \widehat{\gamma}) = \nabla_\theta \ell(\theta, \widehat{\gamma}) - \widehat{\mathbf{w}}^T \nabla_\gamma \ell(\theta, \widehat{\gamma}). \quad (2.6)$$

**return** $\widehat{S}(\theta, \widehat{\gamma})$.

---

**Remark 2.1.** Our method allows a wide range of penalty functions $P(\cdot)$ including nonconvex penalties, whereas most of the existing works mainly consider the Lasso penalty. For instance, the method in van de Geer et al. (2014) is based on inverting Karush-Kuhn-Tucker (KKT) conditions of Lasso, which is not directly applicable for the nonconvex penalties obtained by a statistical optimization algorithm.

**Remark 2.2.** There exist alternative procedures for estimating $\mathbf{w}$ in step (ii). Some examples are given by the following penalized M-estimators $\widetilde{\mathbf{w}}$ and $\bar{\mathbf{w}}$,

$$\widetilde{\mathbf{w}} = \operatorname*{argmin}_{\mathbf{w}} \frac{1}{2n} \sum_{i=1}^{n} \left\{ \mathbf{w}^T \nabla_{\gamma\gamma}\ell_i(\widehat{\boldsymbol{\beta}})\mathbf{w} - 2\mathbf{w}^T \nabla_{\gamma\theta}\ell_i(\widehat{\boldsymbol{\beta}}) \right\} + Q_{\lambda'}(\mathbf{w}), \quad (2.7)$$

$$\bar{\mathbf{w}} = \operatorname*{argmin}_{\mathbf{w}} \frac{1}{2n} \sum_{i=1}^{n} \left\{ \nabla_\theta \ell_i(\widehat{\boldsymbol{\beta}}) - \mathbf{w}^T \nabla_\gamma \ell_i(\widehat{\boldsymbol{\beta}}) \right\}^2 + Q_{\lambda'}(\mathbf{w}), \quad (2.8)$$

where $\ell_i(\cdot)$ is defined in (2.1) and $Q(\cdot)$ is a general penalty function. The theoretical properties of these estimators will be established in Section 4 for linear models and Section 5 for generalized linear models.

**Remark 2.3.** For notational simplicity, we assume that $\boldsymbol{\beta} = (\theta, \gamma)$ is estimated under the full model and $\widehat{\gamma}$ is the corresponding component of $\widehat{\boldsymbol{\beta}}$. It is well known that the classical Rao's score test only requires the estimation under the null hypothesis. Indeed, we can use the similar strategy to estimate the nuisance parameters. Specifically, instead of plugging $\widehat{\boldsymbol{\gamma}}$ into the decorrelated score function, we can estimate $\boldsymbol{\gamma}$ by $\widehat{\boldsymbol{\gamma}}_0 = \operatorname{argmin}_{\boldsymbol{\gamma}} \{\ell(0, \boldsymbol{\gamma}) + P_\lambda(\boldsymbol{\gamma})\}$, and replace $\widehat{\boldsymbol{\beta}}$ in (2.7) and (2.5) with $(0, \widehat{\boldsymbol{\gamma}}_0)$. As shown in Theorem 3.5, under mild conditions, no matter which estimator ($\widehat{\boldsymbol{\gamma}}$ or $\widehat{\boldsymbol{\gamma}}_0$) is plugged in, the estimated decorrelated score function is asymptotically equivalent to $S(\theta^*, \boldsymbol{\gamma}^*)$.

**Remark 2.4.** In this section, we confine our attention to the univariate parameter of interest. Similar to (2.4), we can define the decorrelated score function for multi-dimensional parameter of interest, even if the dimensionality of the parameter of interest is comparable to $d$. This extension will be investigated in Section 6.

### 2.4 One-Step Estimator and Confidence Regions

Though the decorrelated score device is motivated from the hypothesis testing perspective, in this section, we consider how to use the decorrelated score function to construct a valid confidence interval for the parameter of interest $\theta$. This is based on the key observation that the estimated decorrelated score function $\widehat{S}(\theta, \widehat{\boldsymbol{\gamma}})$ in (2.6) can be regarded as an approximately unbiased estimating function for $\theta$ (Godambe and Kale, 1991). Thus, one general approach to define an estimator through the estimating function $\widehat{S}(\theta, \widehat{\boldsymbol{\gamma}})$ is to solve this equation, i.e., $\widehat{S}(\theta, \widehat{\boldsymbol{\gamma}}) = 0$. As commented in Chapter 5 of Van der Vaart (2000), this Z-estimation approach may have several drawbacks. For instance, the estimating equation $\widehat{S}(\theta, \widehat{\boldsymbol{\gamma}})$ may have multiple roots, such that the estimator becomes ill-posed. To overcome these issues, similar to Bickel (1975), we consider the one-step method, which solves a linear approximation to $\widehat{S}(\theta, \widehat{\boldsymbol{\gamma}}) = 0$. Given the penalized M-estimator $\widehat{\theta}$, a one-step estimator $\widetilde{\theta}$ is the solution to $\widehat{S}(\widehat{\boldsymbol{\beta}}) + \widehat{I}_{\theta|\boldsymbol{\gamma}}(\theta - \widehat{\theta}) = 0$, defined as

$$\widetilde{\theta} = \widehat{\theta} - \widehat{S}(\widehat{\boldsymbol{\beta}})/\widehat{I}_{\theta|\boldsymbol{\gamma}}, \quad \text{where} \quad \widehat{I}_{\theta|\boldsymbol{\gamma}} = \nabla^2_{\theta\theta}\ell(\widehat{\boldsymbol{\beta}}) - \widehat{\mathbf{w}}^T \nabla^2_{\boldsymbol{\gamma}\theta}\ell(\widehat{\boldsymbol{\beta}}). \tag{2.9}$$

In Section 3, we show that, under mild conditions on the likelihood and the penalized M-estimator, the one-step estimator $\widetilde{\theta}$ is asymptotically normal with mean $\theta^*$ and is semiparametrically efficient. Based on the asymptotic normality of $\widetilde{\theta}$, we can easily construct the optimal confidence interval for $\theta^*$. Note that the similar method can be straightforwardly used to construct the optimal confidence region, if $\theta^*$ is a multivariate vector with fixed dimension.

### 2.5 Geometric Interpretation and Further Discussion

Given the random variable $\boldsymbol{U}$, consider $S_U(\theta, \boldsymbol{\gamma}) = \nabla_\theta \log f(\boldsymbol{U}; \boldsymbol{\beta}) - \mathbf{w}^T \nabla_{\boldsymbol{\gamma}} \log f(\boldsymbol{U}; \boldsymbol{\beta})$, where $f(\boldsymbol{U}, \boldsymbol{\beta})$ is the probability density of $\boldsymbol{U}$ under the model $\mathbb{P}_{\boldsymbol{\beta}}$, and the decorrelated score function is $S(\theta, \boldsymbol{\gamma}) = n^{-1} \sum_{i=1}^n S_{U_i}(\theta, \boldsymbol{\gamma})$. For simplicity, we focus on the geometric interpretation for $S_U(\theta, \boldsymbol{\gamma})$. For any $\boldsymbol{\beta} = (\theta, \boldsymbol{\gamma})$, the linear space spanned by the score functions can be expressed by $T = \{a_{\boldsymbol{\beta}} \nabla_\theta \log f(\boldsymbol{U}; \boldsymbol{\beta}) + \mathbf{b}_{\boldsymbol{\beta}}^T \nabla_{\boldsymbol{\gamma}} \log f(\boldsymbol{U}; \boldsymbol{\beta})\}$, where $a_{\boldsymbol{\beta}}$ is a nonrandom scalar and $\mathbf{b}_{\boldsymbol{\beta}}$ is a nonrandom $(d-1)$ dimensional vector. As suggested by the notation, $a_{\boldsymbol{\beta}}$ and $\mathbf{b}_{\boldsymbol{\beta}}$ can depend on $\boldsymbol{\beta}$. It is shown by Small and McLeish (2011) that, the space $T$ is a Hilbert space with inner product given by $\langle g_1(\boldsymbol{U}; \boldsymbol{\beta}), g_2(\boldsymbol{U}; \boldsymbol{\beta}) \rangle = \mathbb{E}_{\boldsymbol{\beta}}(g_1(\boldsymbol{U}; \boldsymbol{\beta}) g_2(\boldsymbol{U}; \boldsymbol{\beta}))$, for any $g_1(\boldsymbol{U}; \boldsymbol{\beta}), g_2(\boldsymbol{U}; \boldsymbol{\beta}) \in T$. Similarly, consider the linear space spanned by the nuisance score functions $T_N = \{\mathbf{b}_{\boldsymbol{\beta}}^T \nabla_{\boldsymbol{\gamma}} \log f(\boldsymbol{U}; \boldsymbol{\beta})\}$, where $\mathbf{b}_{\boldsymbol{\beta}}$ is a nonrandom $(d-1)$ dimensional vector, and its orthogonal complement $T_N^\perp = \{g(\boldsymbol{U};\boldsymbol{\beta}) \in T, \langle g(\boldsymbol{U};\boldsymbol{\beta}), s(\boldsymbol{U};\boldsymbol{\beta})\rangle = 0, \forall s(\boldsymbol{U};\boldsymbol{\beta}) \in T_N\}$. Since $\nabla_\theta \log f(\boldsymbol{U};\boldsymbol{\beta}) \in T$, the projection of $\nabla_\theta \log f(\boldsymbol{U};\boldsymbol{\beta})$ to the closed space $T_N^\perp$ is well defined and identical to the decorrelated score function $S_U(\theta,\boldsymbol{\gamma})$. Note that, for the estimation purpose, we assume that the projection of $\nabla_\theta \log f(\boldsymbol{U};\boldsymbol{\beta})$ to $T_N$ is identical to the projection of $\nabla_\theta \log f(\boldsymbol{U};\boldsymbol{\beta})$ to a low dimensional subspace $T_N(S)$ of $T_N$. Here, $T_N(S)$ is defined as $T_N(S) = \{\mathbf{c}_{\boldsymbol{\beta}}^T \nabla_{\boldsymbol{\gamma}_S} \log f(\boldsymbol{U};\boldsymbol{\beta})\}$, where $S$ is a subset of $\{1,...,d-1\}$ with $|S| \ll n$, and $\mathbf{c}_{\boldsymbol{\beta}}$ is a nonrandom $|S|$ dimensional vector. In other words, $T_N(S)$ is a linear space spanned by the components of nuisance score functions corresponding to $\boldsymbol{\gamma}_S$. This low dimensional structure $T_N(S)$ within the high dimensional Hilbert space $T_N$ makes our decorrelated score function different from the efficient score in Van der Vaart (2000).

Finally, we comment on the connection between the decorrelated score device and profile score function. Recall that the profile score function is given by $\nabla_\theta \ell(\theta, \widehat{\boldsymbol{\gamma}}_\theta)$, where $\ell(\theta, \widehat{\boldsymbol{\gamma}}_\theta)$ is the negative profile log-likelihood, and $\widehat{\boldsymbol{\gamma}}_\theta$ is the constrained maximum likelihood estimator, i.e., $\widehat{\boldsymbol{\gamma}}_\theta = \operatorname{argmin}_{\boldsymbol{\gamma}} \ell(\theta, \boldsymbol{\gamma})$. It is easily seen that, the decorrelated score function $S(\theta, \widehat{\boldsymbol{\gamma}}_\theta)$ with $\boldsymbol{\gamma}$ estimated by $\widehat{\boldsymbol{\gamma}}_\theta$ is identical to the profile score function, due to $\nabla_{\boldsymbol{\gamma}} \ell(\theta, \widehat{\boldsymbol{\gamma}}_\theta) = 0$. Hence, in low dimensional problems, the profile score function implicitly performs decorrelation. From this perspective, we view our decorrelated score function as a natural high dimensional extension of the profile score function.

## 3 A General Theory for Tests and Confidence Regions

In this section, under full generality, we establish the theoretical guarantees for the score test and the confidence regions. Here, we assume that the statistical model is correctly specified.

### 3.1 Weak Convergence of Score Test under Null Hypothesis

To study the properties of the decorrelated score test under the null hypothesis, we impose the following conditions.

**Assumption 3.1** (Estimation Error Bound). Assume that

$$\lim_{n\to\infty} \mathbb{P}_{\boldsymbol{\beta}^*}\big(\|\widehat{\boldsymbol{\gamma}} - \boldsymbol{\gamma}^*\|_1 \lesssim \eta_1(n)\big) = 1 \quad \text{and} \quad \lim_{n\to\infty} \mathbb{P}_{\boldsymbol{\beta}^*}\big(\|\widehat{\mathbf{w}} - \mathbf{w}^*\|_1 \lesssim \eta_2(n)\big) = 1,$$

where $\mathbf{w}^* = \mathbf{I}_{\gamma\gamma}^{*-1} \mathbf{I}_{\gamma\theta}^*$, and $\eta_1(n)$ and $\eta_2(n)$ converge to 0, as $n \to \infty$.

**Assumption 3.2** (Noise Condition). Assume that $\lim_{n\to\infty} \mathbb{P}_{\boldsymbol{\beta}^*}\big(\|\nabla_{\boldsymbol{\gamma}} \ell(0, \boldsymbol{\gamma}^*)\|_\infty \lesssim \eta_3(n)\big) = 1$, for some $\eta_3(n) \to 0$, as $n \to \infty$.

**Assumption 3.3** (Stability Condition). For $\boldsymbol{\gamma}_v = v\boldsymbol{\gamma}^* + (1-v)\widehat{\boldsymbol{\gamma}}$ with $v \in [0,1]$,

$$\lim_{n\to\infty} \mathbb{P}_{\boldsymbol{\beta}^*}\big(\sup_{v\in[0,1]} \|\nabla_{\theta\boldsymbol{\gamma}}^2 \ell(0,\boldsymbol{\gamma}_v) - \widehat{\mathbf{w}}^T \nabla_{\boldsymbol{\gamma}\boldsymbol{\gamma}}^2 \ell(0,\boldsymbol{\gamma}_v)\|_\infty \lesssim \eta_4(n)\big) = 1,$$

for some $\eta_4(n) \to 0$, as $n \to \infty$.

**Assumption 3.4** (CLT)**.** Recall that $\mathbf{I}^* = \mathbb{E}_{\boldsymbol{\beta}^*}(\nabla^2 \ell(0, \boldsymbol{\gamma}^*))$. For $\mathbf{v}^* = (1, -\mathbf{w}^{*T})^T$, it holds that

$$\frac{\sqrt{n}\mathbf{v}^{*T}\nabla\ell(0, \boldsymbol{\gamma}^*)}{\sqrt{\mathbf{v}^T\mathbf{I}^*\mathbf{v}}} \rightsquigarrow N(0, 1).$$

Assume that $C' \leq I^*_{\theta|\gamma} < \infty$, where $I^*_{\theta|\gamma} = I^*_{\theta\theta} - \mathbf{w}^{*T}\mathbf{I}^*_{\gamma\theta}$, and $C' > 0$ is a constant.

In Assumption 3.1, the estimation error bounds for $\boldsymbol{\beta}^*$ in terms of the $L_1$ norm have been thoroughly studied for a variety of estimators, including the Lasso estimator, nonconvex estimator and Dantzig selector; see Candes and Tao (2007); Bickel et al. (2009); Zhang (2009); Negahban et al. (2012); Loh and Wainwright (2013); Wang et al. (2013a); Zhang et al. (2013); Wang et al. (2013b), among many others. The estimation error bound for $\mathbf{w}^*$ can be derived by the similar method. But the key difference is that $\widehat{\mathbf{w}}$ depends on the estimator $\widehat{\boldsymbol{\beta}}$, which may cause extra technical difficulty in bounding $\|\widehat{\mathbf{w}} - \mathbf{w}^*\|_1$. In Section 4, we show that, for the linear model, we have $\eta_1(n) \asymp s^*\sqrt{\log d/n}$ and $\eta_2(n) \asymp s'\sqrt{\log d/n}$, where $s^* = \|\boldsymbol{\gamma}^*\|_0$ and $s' = \|\mathbf{w}^*\|_0$. Assumption 3.2 quantifies the concentration of the $(d-1)$ dimensional score function, which reduces to the average of the residual in the least square loss. Hence, we name it as the noise condition. In Section 4, we show that, for the linear model, $\eta_3(n) \asymp \sqrt{\log d/n}$. Assumption 3.3 is related to the stability of the Hessian matrix. To see the connection, if the Hessian matrices $\nabla^2_{\theta\gamma}\ell(\boldsymbol{\beta})$ and $\nabla^2_{\gamma\gamma}\ell(\boldsymbol{\beta})$ are sufficiently smooth, compared with the constraint in (2.5), we may expect that $\eta_4(n)$ can be controlled by the tuning parameter $\lambda'$ together with the estimation error bound of $\widehat{\boldsymbol{\beta}}$. In Section 4, we show that, for the linear model, $\eta_4(n) \asymp \sqrt{\log d/n}$ with $\lambda' \asymp \sqrt{\log d/n}$. Assumption 3.4 is the central limit theorem for a linear combination of the score functions, which can be obtained by verifying the Lindeberg's condition. This assumption also implicitly assumes the information equality or the second Bartlett identity, which says that the information matrix is identical to the covariance of the score function (Lindsay, 1982; Bartlett, 1953a,b). This property holds for the log-likelihood under the regular model, and a variety of alternative likelihood type functions (Mykland, 1999). The extension of the score test to the situation in which the information equality fails will be studied in Section 6.

The following theorem shows that the estimated decorrelated score function is asymptotically normal, with $\sqrt{n}$ convergence rate.

**Theorem 3.5.** Under the Assumptions 3.1–3.4, with probability tending to one

$$n^{1/2}\big|\widehat{S}(0, \widehat{\boldsymbol{\gamma}}) - S(0, \boldsymbol{\gamma}^*)\big| \lesssim n^{1/2}\big(\eta_2(n)\eta_3(n) + \eta_1(n)\eta_4(n)\big). \tag{3.1}$$

If $n^{1/2}(\eta_2(n)\eta_3(n) + \eta_1(n)\eta_4(n)) = o(1)$, we have

$$n^{1/2}\widehat{S}(0, \widehat{\boldsymbol{\gamma}})I^{*-1/2}_{\theta|\gamma} \rightsquigarrow N(0, 1). \tag{3.2}$$

*Proof of Theorem 3.5.* By the definition of $\widehat{S}(0, \widehat{\boldsymbol{\gamma}})$ and the mean value theorem, we obtain

$$\begin{aligned}\widehat{S}(0, \widehat{\boldsymbol{\gamma}}) &= \nabla_\theta\ell(0, \widehat{\boldsymbol{\gamma}}) - \widehat{\mathbf{w}}^T\nabla_\gamma\ell(0, \widehat{\boldsymbol{\gamma}}) \\ &= \nabla_\theta\ell(0, \boldsymbol{\gamma}^*) - \widehat{\mathbf{w}}^T\nabla_\gamma\ell(0, \boldsymbol{\gamma}^*) + \big\{\nabla^2_{\theta\gamma}\ell(0, \widetilde{\boldsymbol{\gamma}}) - \widehat{\mathbf{w}}^T\nabla^2_{\gamma\gamma}\ell(0, \widetilde{\boldsymbol{\gamma}})\big\}(\widehat{\boldsymbol{\gamma}} - \boldsymbol{\gamma}^*),\end{aligned}$$

where $\widetilde{\boldsymbol{\gamma}} = v\boldsymbol{\gamma}^* + (1-v)\widehat{\boldsymbol{\gamma}}$ for some $v \in [0,1]$. Thus, by rearranging terms, we can show that

$$\widehat{S}(0,\widehat{\boldsymbol{\gamma}}) = S(0,\boldsymbol{\gamma}^*) + \underbrace{(\mathbf{w}^* - \widehat{\mathbf{w}})^T \nabla_{\boldsymbol{\gamma}}\ell(0,\boldsymbol{\gamma}^*)}_{I_1} + \underbrace{\{\nabla^2_{\theta\boldsymbol{\gamma}}\ell(0,\widetilde{\boldsymbol{\gamma}}) - \widehat{\mathbf{w}}^T \nabla^2_{\boldsymbol{\gamma}\boldsymbol{\gamma}}\ell(0,\widetilde{\boldsymbol{\gamma}})\}(\widehat{\boldsymbol{\gamma}} - \boldsymbol{\gamma}^*)}_{I_2}. \quad (3.3)$$

By Assumptions 3.1 and 3.2, we have with probability tending to one,

$$|I_1| \leq ||\mathbf{w}^* - \widehat{\mathbf{w}}||_1 ||\nabla_{\boldsymbol{\gamma}}\ell(0,\boldsymbol{\gamma}^*)||_\infty \lesssim \eta_2(n)\eta_3(n). \quad (3.4)$$

For $I_2$, by Assumptions 3.1 and 3.3, we have with probability tending to one,

$$|I_2| \leq ||\nabla^2_{\theta\boldsymbol{\gamma}}\ell(0,\widetilde{\boldsymbol{\gamma}}) - \widehat{\mathbf{w}}^T \nabla^2_{\boldsymbol{\gamma}\boldsymbol{\gamma}}\ell(0,\widetilde{\boldsymbol{\gamma}})||_\infty ||\widehat{\boldsymbol{\gamma}} - \boldsymbol{\gamma}^*||_1 \lesssim \eta_1(n)\eta_4(n).$$

Together with (3.3) and (3.4), we have shown that inequality (3.1) holds in probability. Note that $S(0,\boldsymbol{\gamma}^*) = (1,-\mathbf{w}^{*T})\nabla\ell(0,\boldsymbol{\gamma}^*)$ and $I^*_{\theta|\boldsymbol{\gamma}} = (1,-\mathbf{w}^{*T})\mathbf{I}^*(1,-\mathbf{w}^{*T})^T$. Combining with Assumption 3.4, we have $n^{1/2}S(0,\boldsymbol{\gamma}^*)I^{*-1/2}_{\theta|\boldsymbol{\gamma}} \rightsquigarrow N(0,1)$. By $I^*_{\theta|\boldsymbol{\gamma}} \geq C'$ in Assumption 3.4 and inequality (3.1), we obtain that $n^{1/2}|\widehat{S}(0,\widehat{\boldsymbol{\gamma}})I^{*-1/2}_{\theta|\boldsymbol{\gamma}} - S(0,\boldsymbol{\gamma}^*)I^{*-1/2}_{\theta|\boldsymbol{\gamma}}| = o_\mathbb{P}(1)$. We complete the proof of (3.2) by applying the Slutsky's theorem. $\square$

Since $n^{1/2}\widehat{S}(0,\widehat{\boldsymbol{\gamma}})/I^{*1/2}_{\theta|\boldsymbol{\gamma}}$ depends on the partial information matrix $I^*_{\theta|\boldsymbol{\gamma}}$, to define the test statistic, one needs to estimate $I^*_{\theta|\boldsymbol{\gamma}}$. Such an estimator $\widehat{I}_{\theta|\boldsymbol{\gamma}}$ is given by (2.9). The following assumption ensures that the element of the Fisher information matrix can be consistently estimated, which can be used to show the consistency of $\widehat{I}_{\theta|\boldsymbol{\gamma}}$.

**Assumption 3.6** (Convergence of Hessian Matrix). It holds that $\lim_{n\to\infty} \mathbb{P}_{\boldsymbol{\beta}^*}(||\nabla^2\ell(\widehat{\boldsymbol{\beta}}) - \mathbf{I}^*||_{\max} \lesssim \eta_5(n)) = 1$, for some $\eta_5(n) \to 0$, as $n \to \infty$.

In Section 4, we show that, for the linear model, $\eta_5(n) \asymp \sqrt{\log d/n}$. Now, we define the decorrelated score test statistic as

$$\widehat{U}_n = n^{1/2}\widehat{S}(0,\widehat{\boldsymbol{\gamma}})\widehat{I}^{-1/2}_{\theta|\boldsymbol{\gamma}}. \quad (3.5)$$

The following corollary establishes the asymptotic distribution of $\widehat{U}_n$ under $H_0 : \theta^* = 0$.

**Corollary 3.7.** Assume that the Assumptions 3.1–3.4, and 3.6 hold. It also holds that $||\mathbf{w}^*||_1\eta_5(n) = o(1)$, $\eta_2(n)||\mathbf{I}^*_{\theta\boldsymbol{\gamma}}||_\infty = o(1)$, and $n^{1/2}(\eta_2(n)\eta_3(n) + \eta_1(n)\eta_4(n)) = o(1)$. Under $H_0 : \theta^* = 0$, we have for any $t \in \mathbb{R}$,

$$\lim_{n\to\infty} |\mathbb{P}_{\boldsymbol{\beta}^*}(\widehat{U}_n \leq t) - \Phi(t)| = 0. \quad (3.6)$$

*Proof.* A detailed proof is shown in Appendix A. $\square$

**Remark 3.8.** Based on the test statistic $\widehat{U}_n$, the score test with significance level $\alpha$, for the null hypothesis $H_0 : \theta^* = 0$ versus the two-sided alternative $H_1 : \theta \neq 0$ is given by

$$T_n = \begin{cases} 0 & \text{if } |\widehat{U}_n| \leq \Phi^{-1}(1-\alpha/2), \\ 1, & \text{if } |\widehat{U}_n| > \Phi^{-1}(1-\alpha/2), \end{cases} \quad (3.7)$$

where $\Phi(\cdot)$ is the cdf of a standard normal distribution. Given the value of $T_n$, we reject the null hypothesis if and only if $T_n = 1$. To demonstrate the validity of the proposed score test, we need to show that the type I error of $T_n$, that is the probability of rejecting $H_0$ (i.e., $T_n = 1$) when $H_0$ is true, can be controlled by $\alpha$ asymptotically. This is true by Corollary 3.7, i.e., $\lim_{n\to\infty} |\mathbb{P}_{\boldsymbol{\beta}^*}(T_n = 1) - \alpha| = 0$.

## 3.2 Uniform Weak Convergence of Score Test under Null Hypothesis

Although in the previous section the limiting distribution of the score test statistic $\widehat{U}_n$ is established, the convergence is shown under the fixed probability distribution $\mathbb{P}_{\boldsymbol{\beta}^*} = \mathbb{P}_{(0,\boldsymbol{\gamma}^*)}$. However, in practice, $\boldsymbol{\gamma}^*$ is unknown. To guarantee that the convergence properties are not affected by the values of $\boldsymbol{\gamma}^*$, it is of interest to strengthen the weak convergence results in Theorem 3.5 and Corollary 3.7 to the weak convergence uniformly over the values of $\boldsymbol{\gamma}^*$. In particular, consider the following parameter space

$$\Omega_0 = \{(0, \boldsymbol{\gamma}) : \|\boldsymbol{\gamma}\|_0 \leq s^*, \text{ for some } s^* \ll n\}.$$

To ensure that the parameter $\boldsymbol{\beta}$ can be still consistently estimated, we assume $\Omega_0$ only contains sparse parameters. Similarly, to study the weak convergence uniformly over $\boldsymbol{\beta}^* \in \Omega_0$, we impose the following conditions.

**Assumption 3.9** (Uniform Estimation Error Bound). It holds that $\lim_{n\to\infty} \inf_{\boldsymbol{\beta}^* \in \Omega_0} \mathbb{P}_{\boldsymbol{\beta}^*}(\mathcal{F}_1^{\boldsymbol{\beta}^*}) = 1$ and $\lim_{n\to\infty} \inf_{\boldsymbol{\beta}^* \in \Omega_0} \mathbb{P}_{\boldsymbol{\beta}^*}(\mathcal{F}_2^{\boldsymbol{\beta}^*}) = 1$, where

$$\mathcal{F}_1^{\boldsymbol{\beta}^*} = \{\|\widehat{\boldsymbol{\gamma}} - \boldsymbol{\gamma}^*\|_1 \lesssim \eta_1(n)\} \quad \text{and} \quad \mathcal{F}_2^{\boldsymbol{\beta}^*} = \{\|\widehat{\mathbf{w}} - \mathbf{w}^*\|_1 \lesssim \eta_2(n)\},$$

for some $\eta_1(n)$ and $\eta_2(n)$ converging to 0, as $n \to \infty$.

**Assumption 3.10** (Uniform Noise Condition). Assume that $\lim_{n\to\infty} \inf_{\boldsymbol{\beta}^* \in \Omega_0} \mathbb{P}_{\boldsymbol{\beta}^*}(\mathcal{F}_3^{\boldsymbol{\beta}^*}) = 1$, where $\mathcal{F}_3^{\boldsymbol{\beta}^*} = \{\|\nabla_{\boldsymbol{\gamma}} \ell(0, \boldsymbol{\gamma}^*)\|_\infty \lesssim \eta_3(n)\}$, for some $\eta_3(n) \to 0$, as $n \to \infty$.

**Assumption 3.11** (Uniform Stability Condition). Assume that $\lim_{n\to\infty} \inf_{\boldsymbol{\beta}^* \in \Omega_0} \mathbb{P}_{\boldsymbol{\beta}^*}(\mathcal{F}_4^{\boldsymbol{\beta}^*}) = 1$, where

$$\mathcal{F}_4^{\boldsymbol{\beta}^*} = \left\{ \sup_{v \in [0,1]} \|\nabla^2_{\theta\boldsymbol{\gamma}} \ell(0, \boldsymbol{\gamma}_v) - \widehat{\mathbf{w}}^T \nabla^2_{\boldsymbol{\gamma}\boldsymbol{\gamma}} \ell(0, \boldsymbol{\gamma}_v)\|_\infty \lesssim \eta_4(n) \right\}.$$

Here $\boldsymbol{\gamma}_v = v\boldsymbol{\gamma}^* + (1-v)\widehat{\boldsymbol{\gamma}}$ with $v \in [0, 1]$, and $\eta_4(n) \to 0$, as $n \to \infty$.

**Assumption 3.12** (Uniform CLT). For $\mathbf{v}^* = (1, -\mathbf{w}^{*T})^T$, it holds that,

$$\lim_{n\to\infty} \sup_{\boldsymbol{\beta}^* \in \Omega_0} \sup_{t \in \mathbb{R}} \left| \mathbb{P}_{\boldsymbol{\beta}^*} \left( \frac{\sqrt{n} \mathbf{v}^{*T} \nabla \ell(\boldsymbol{\beta}^*)}{\sqrt{\mathbf{v}^{*T} \mathbf{I}^* \mathbf{v}^*}} \leq t \right) - \Phi(t) \right| = 0.$$

Assume that $C' \leq I^*_{\theta|\gamma} < \infty$, for some $C' > 0$.

**Assumption 3.13** (Convergence of Hessian Matrix). Assume that $\lim_{n\to\infty} \inf_{\boldsymbol{\beta}^* \in \Omega_0} \mathbb{P}_{\boldsymbol{\beta}^*}(\mathcal{F}_5^{\boldsymbol{\beta}^*}) = 1$, where $\mathcal{F}_5^{\boldsymbol{\beta}^*} = \{\|\nabla^2 \ell(\widehat{\boldsymbol{\beta}}) - \mathbf{I}^*\|_{\max} \lesssim \eta_5(n)\}$, for some $\eta_5(n) \to 0$, as $n \to \infty$.

Note that $\mathbf{w}^*$ in Assumption 3.9 also implicitly depends on $\boldsymbol{\beta}^*$. For notational simplicity, we suppress this dependence. Here, $\eta_1(n), ..., \eta_5(n)$ are deterministic and do not depend on $\boldsymbol{\beta}^*$. Assumptions 3.9, 3.10, 3.11, 3.12 and 3.13 are similar but stronger than Assumptions 3.1, 3.2, 3.3, 3.4 and 3.6, respectively. Since in the previous section the theoretical guarantees are provided under fixed $\mathbb{P}_{\boldsymbol{\beta}^*} = \mathbb{P}_{(0,\boldsymbol{\gamma}^*)}$, it is sufficient to assume 3.1, 3.2, 3.3, 3.12 and 3.6 hold under the probability distribution $\mathbb{P}_{\boldsymbol{\beta}^*}$. However, in this section, to study the uniform convergence, we assume that the events $\mathcal{F}_1^{\boldsymbol{\beta}^*}, ..., \mathcal{F}_5^{\boldsymbol{\beta}^*}$ hold under the distribution $\mathbb{P}_{\boldsymbol{\beta}^*}$ uniformly over $\boldsymbol{\beta}^* \in \Omega_0$. The following theorem establishes the uniform convergence of the score test statistic $\widehat{U}_n$ in (3.5).

**Theorem 3.14.** Assume that the Assumptions 3.9–3.13 hold. It holds that $\sup_{\boldsymbol{\beta}^* \in \Omega_0} ||\mathbf{w}^*||_1 \eta_5(n) = o(1)$, $\sup_{\boldsymbol{\beta}^* \in \Omega_0} \eta_2(n) ||\mathbf{I}_{\theta\gamma}^*||_\infty = o(1)$, and $n^{1/2}(\eta_2(n)\eta_3(n) + \eta_1(n)\eta_4(n)) = o(1)$. Then, we have

$$\lim_{n \to \infty} \sup_{\boldsymbol{\beta}^* \in \Omega_0} \sup_{t \in \mathbb{R}} |\mathbb{P}_{\boldsymbol{\beta}^*}(\widehat{U}_n \leq t) - \Phi(t)| = 0. \tag{3.8}$$

*Proof.* A detailed proof is shown in Appendix A. □

**Remark 3.15.** Theorem 3.14 implies that the type I error of the score test $T_n$ in (3.7) converges to its significance level $\alpha$ uniformly over $\boldsymbol{\beta}^* \in \Omega_0$, i.e.,

$$\lim_{n \to \infty} \sup_{\boldsymbol{\beta}^* \in \Omega_0} \sup_{\alpha \in (0,1)} \left|\mathbb{P}_{\boldsymbol{\beta}^*}(T_n = 1) - \alpha\right| = 0.$$

The hypothesis test with such uniform convergence property is called honest test. Some examples of honest tests and confidence intervals are considered by Belloni et al. (2013); van de Geer et al. (2014); Javanmard and Montanari (2013).

### 3.3 Uniform Weak Convergence of Score Test under Alternative Hypothesis

In this section, we consider the power of the score test for detecting the alternative hypothesis. In particular, we are interested in the limiting behavior of $T_n$ under the sequence of alternative hypothesis $H_{1n} : \theta^* = \widetilde{C} n^{-\phi}$, where $\widetilde{C}$ is a constant, and $\phi$ is a positive constant. Consider the following parameter space

$$\Omega_1(\widetilde{C}, \phi) = \{(\theta, \boldsymbol{\gamma}) : \theta = \widetilde{C} n^{-\phi}, \|\boldsymbol{\gamma}\|_0 \leq s^*, \text{ for some } s^* \ll n\}.$$

The parameter space $\Omega_1(\widetilde{C}, \phi)$ describes the local alternative around the null hypothesis $\theta^* = 0$, in the sense that $\theta^* = \widetilde{C} n^{-\phi}$ gradually shrinks to 0 as $n \to \infty$. Similar to $\Omega_0$, we only consider sparse local alternatives.

**Assumption 3.16** (Uniform Local Approximation). It holds that $\lim_{n\to\infty} \inf_{\boldsymbol{\beta}^* \in \Omega_1(\widetilde{C},\phi)} \mathbb{P}_{\boldsymbol{\beta}^*}(\mathcal{F}_6^{\boldsymbol{\beta}^*}) = 1$, where

$$\mathcal{F}_6^{\boldsymbol{\beta}^*} = \{\sqrt{n}|S(\theta^*, \boldsymbol{\gamma}^*) - S(0, \boldsymbol{\gamma}^*) - \theta^* \mathbf{I}_{\theta|\gamma}^*| \lesssim \eta_6(n)\}, \tag{3.9}$$

for some $\eta_6(n) \to 0$, as $n \to \infty$.

**Assumption 3.17** (Uniform Estimation Error Bound). It holds that $\lim_{n\to\infty} \inf_{\boldsymbol{\beta}^* \in \Omega_1(\widetilde{C},\phi)} \mathbb{P}_{\boldsymbol{\beta}^*}(\mathcal{F}_1^{\boldsymbol{\beta}^*}) = 1$ and $\lim_{n\to\infty} \inf_{\boldsymbol{\beta}^* \in \Omega_1(\widetilde{C},\phi)} \mathbb{P}_{\boldsymbol{\beta}^*}(\mathcal{F}_2^{\boldsymbol{\beta}^*}) = 1$, where $\mathcal{F}_1^{\boldsymbol{\beta}^*}$ and $\mathcal{F}_2^{\boldsymbol{\beta}^*}$ are given in Assumption 3.9.

**Assumption 3.18** (Uniform Noise Condition). Assume that $\lim_{n\to\infty} \inf_{\boldsymbol{\beta}^* \in \Omega_1(\widetilde{C},\phi)} \mathbb{P}_{\boldsymbol{\beta}^*}(\mathcal{F}_3^{\boldsymbol{\beta}^*}) = 1$, where $\mathcal{F}_3^{\boldsymbol{\beta}^*}$ is given in Assumption 3.10.

**Assumption 3.19** (Uniform Stability Condition). Assume that $\lim_{n\to\infty} \inf_{\boldsymbol{\beta}^* \in \Omega_1(\widetilde{C},\phi)} \mathbb{P}_{\boldsymbol{\beta}^*}(\mathcal{F}_4^{\boldsymbol{\beta}^*}) = 1$, where $\mathcal{F}_4^{\boldsymbol{\beta}^*}$ is given in Assumption 3.11.

**Assumption 3.20** (Uniform CLT). For $\mathbf{v}^* = (1, -\mathbf{w}^{*T})^T$, it holds that,

$$\lim_{n\to\infty} \sup_{\boldsymbol{\beta}^* \in \Omega_1(\widetilde{C},\phi)} \sup_{t\in\mathbb{R}} \left| \mathbb{P}_{\boldsymbol{\beta}^*}\left( \frac{\sqrt{n}\mathbf{v}^{*T}\nabla\ell(\boldsymbol{\beta}^*)}{\sqrt{\mathbf{v}^{*T}\mathbf{I}^*\mathbf{v}^*}} \leq t \right) - \Phi(t) \right| = 0.$$

Assume that $C' \leq I^*_{\theta|\boldsymbol{\gamma}} \leq C'' < \infty$, where $C', C''$ are positive constants.

**Assumption 3.21** (Convergence of Hessian Matrix). Assume that $\lim_{n\to\infty} \inf_{\boldsymbol{\beta}^* \in \Omega_1(\widetilde{C},\phi)} \mathbb{P}_{\boldsymbol{\beta}^*}(\mathcal{F}_5^{\boldsymbol{\beta}^*}) = 1$, where $\mathcal{F}_5^{\boldsymbol{\beta}^*}$ is given in Assumption 3.13.

Assumption 3.16 is part of the uniformly locally asymptotic normality (ULAN) condition for $\theta$, commonly used to study the asymptotic properties of hypothesis tests in low dimensional models (Van der Vaart, 2000). Since we focus on the score test, we rewrite the ULAN condition in terms of the score functions instead of the likelihood ratio. Essentially, Assumption 3.16 specifies the linear approximation of $\nabla\ell(\widetilde{C}n^{-\phi}, \boldsymbol{\gamma}^*)$ locally around $\theta^* = 0$. Moreover, due to the presence of the nuisance parameter $\boldsymbol{\gamma}$, it is necessary to assume that the local linear approximation holds uniformly over $\boldsymbol{\beta}^* \in \Omega_1(\widetilde{C}, \phi)$. Therefore we call this condition as the uniform local approximation condition. Assumptions 3.17, 3.18, 3.19, 3.20 and 3.21 are parallel to Assumptions 3.9, 3.10, 3.11, 3.12 and 3.13, respectively. The key difference is that we now evaluate the events $\mathcal{F}_1^{\boldsymbol{\beta}^*}, ..., \mathcal{F}_5^{\boldsymbol{\beta}^*}$ at $\mathbb{P}_{\boldsymbol{\beta}^*}$, where $\boldsymbol{\beta}^*$ belongs to $\Omega_1(\widetilde{C}, \phi)$ instead of $\Omega_0$. The following theorem characterizes the limiting distributions of the score test statistic $\widehat{U}_n = n^{1/2}\widehat{S}(0, \widehat{\boldsymbol{\gamma}})\widehat{I}_{\theta|\boldsymbol{\gamma}}^{-1/2}$, with respect to different values of $\phi$.

**Theorem 3.22.** Under the Assumptions 3.16–3.21, we also assume, $\sup_{\boldsymbol{\beta}^* \in \Omega_1(\widetilde{C},\phi)} \|\mathbf{w}^*\|_1 \eta_5(n) = o(1)$, $\eta_2(n) \sup_{\boldsymbol{\beta}^* \in \Omega_1(\widetilde{C},\phi)} \|\mathbf{I}^*_{\theta\boldsymbol{\gamma}}\|_\infty = o(1)$, and $n^{1/2}(\eta_2(n)\eta_3(n) + \eta_1(n)\eta_4(n)) = o(1)$. Then

$$\lim_{n\to\infty} \sup_{\boldsymbol{\beta}^* \in \Omega_1(\widetilde{C},\phi)} \sup_{t\in\mathbb{R}} \left| \mathbb{P}_{\boldsymbol{\beta}^*}(\widehat{U}_n \leq t) - \Phi(t) \right| = 0, \quad \text{if} \quad \phi > 1/2, \quad (3.10)$$

$$\lim_{n\to\infty} \sup_{\boldsymbol{\beta}^* \in \Omega_1(\widetilde{C},\phi)} \sup_{t\in\mathbb{R}} \left| \mathbb{P}_{\boldsymbol{\beta}^*}(\widehat{U}_n \leq t) - \Phi(t + \widetilde{C}I_{\theta|\boldsymbol{\gamma}}^{*1/2}) \right| = 0, \quad \text{if} \quad \phi = 1/2, \quad (3.11)$$

$$\lim_{n\to\infty} \sup_{\boldsymbol{\beta}^* \in \Omega_1(\widetilde{C},\phi)} \mathbb{P}_{\boldsymbol{\beta}^*}(|\widehat{U}_n| \leq t) = 0, \quad \text{if} \quad \phi < 1/2. \quad (3.12)$$

Here, (3.12) holds for any fixed $t \in \mathbb{R}$ and $\widetilde{C} \neq 0$.

*Proof.* A detailed proof is shown in Appendix A. □

**Remark 3.23.** This theorem implies that the score test statistic $\widehat{U}_n$ has distinct limiting behaviors in terms of the magnitude of $\phi$. In particular, (3.10) implies that $\widehat{U}_n \rightsquigarrow N(0,1)$ if $\phi > 1/2$ and (3.11) implies that $\widehat{U}_n \rightsquigarrow N(-\widetilde{C}I_{\theta|\boldsymbol{\gamma}}^{*1/2}, 1)$ if $\phi = 1/2$. These results agree with the classical Rao's score test for low dimensional $\boldsymbol{\gamma}$.

**Remark 3.24.** Note that the power of the two-sided test $T_n$ in (3.7) is given by the probability of $T_n = 1$ when $\boldsymbol{\beta}^* \in \Omega_1(\widetilde{C}, \phi)$. Given the fact that the type I error of $T_n$ can be controlled at level $\alpha$ asymptotically, Theorem 3.22 characterizes the uniform asymptotic power of $T_n$ under the alternative hypothesis $H_{1n} : \theta^* = \widetilde{C} n^{-\phi}$. In particular, Theorem 3.22 implies

$$\lim_{n \to \infty} \sup_{\boldsymbol{\beta}^* \in \Omega_1(\widetilde{C}, \phi)} \sup_{\alpha \in (0,1)} \left| \mathbb{P}_{\boldsymbol{\beta}^*}(T_n = 1) - \alpha \right| = 0, \quad \text{if} \quad \phi > 1/2, \tag{3.13}$$

$$\lim_{n \to \infty} \sup_{\boldsymbol{\beta}^* \in \Omega_1(\widetilde{C}, \phi)} \sup_{\alpha \in (0,1)} \left| \mathbb{P}_{\boldsymbol{\beta}^*}(T_n = 1) - \psi_\alpha \right| = 0, \quad \text{if} \quad \phi = 1/2, \tag{3.14}$$

$$\lim_{n \to \infty} \inf_{\boldsymbol{\beta}^* \in \Omega_1(\widetilde{C}, \phi)} \mathbb{P}_{\boldsymbol{\beta}^*}(T_n = 1) = 1, \quad \text{if} \quad \phi < 1/2, \tag{3.15}$$

where $\psi_\alpha = 1 - \Phi(\Phi^{-1}(1-\alpha/2) + \widetilde{C} I_{\theta|\boldsymbol{\gamma}}^{*1/2}) + \Phi(-\Phi^{-1}(1-\alpha/2) + \widetilde{C} I_{\theta|\boldsymbol{\gamma}}^{*1/2})$. Here, (3.15) holds for any $\alpha \in [\delta, 1)$ with some constant $\delta > 0$ and $\widetilde{C} \neq 0$. In particular, (3.13) implies that the test $T_n$ has no power beyond the type I error to distinguish $H_0$ from $H_{1n}$ if $\phi > 1/2$. Moreover, it is seen that $\psi_\alpha > \alpha$ for any $\widetilde{C} \neq 0$. Hence, (3.14) shows that the test $T_n$ is asymptotically unbiased. That means the proposed score test $T_n$ has asymptotic power larger than the type I error for detecting $H_{1n} : \theta^* = \widetilde{C} n^{-1/2}$. Finally, (3.15) implies that the minimal power of $T_n$ increases to 1 as $n \to \infty$, if $\phi < 1/2$.

**Remark 3.25.** Recall that the hypothesis test $T_n$ is for $H_0 : \theta^* = 0$ versus $H_1 : \theta^* \neq 0$, which is two-sided. To test the one-sided alternative hypothesis, say $H_1' : \theta^* > 0$, with the significance level $\alpha$, we can define the score test $T_n'$, such that $T_n' = 1$ if and only if $\widehat{U}_n < -\Phi^{-1}(1-\alpha)$. Theorem 3.14 shows that the type I error of $T_n'$ converges to its significance level $\alpha$ uniformly. In addition, by Theorem 3.22, the uniform asymptotic power of $T_n'$ under the alternative hypothesis $H_{1n} : \theta^* = \widetilde{C} n^{-1/2}$ for some $\widetilde{C} > 0$ is given by (3.14) with $T_n$ replaced by $T_n'$ and $\psi_\alpha$ replaced by $\psi_\alpha' = 1 - \Phi(\Phi^{-1}(1-\alpha) - \widetilde{C} I_{\theta|\boldsymbol{\gamma}}^{*1/2})$.

**Remark 3.26.** Note that the asymptotic power studied in this section is uniform over $\boldsymbol{\beta}^* \in \Omega_1(\widetilde{C}, \phi)$, and the required conditions are also uniform over the same parameter space. Similar to the theory in Section 3.1, one can derive the non-uniform (pointwise) asymptotic power of the score test under $\mathbb{P}_{\boldsymbol{\beta}^*}$, where $\boldsymbol{\beta}^* = (\widetilde{C} n^{-\phi}, \boldsymbol{\gamma}^*)$, based on the non-uniform types of conditions.

### 3.4 Theoretical Results for Optimal Confidence Regions

To study the properties of the one-step estimator $\widetilde{\theta}$ in (2.9) and the related confidence region, we need similar assumptions to those in Section 3.1.

**Assumption 3.27** (Noise Condition). Assume that $\lim_{n \to \infty} \mathbb{P}_{\boldsymbol{\beta}^*}(\|\nabla_{\boldsymbol{\gamma}} \ell(\boldsymbol{\beta}^*)\|_\infty \lesssim \eta_3(n)) = 1$, for some $\eta_3(n) \to 0$, as $n \to \infty$.

**Assumption 3.28** (Stability Condition). For $\boldsymbol{\gamma}_v = v\boldsymbol{\gamma}^* + (1-v)\widehat{\boldsymbol{\gamma}}$ with $v \in [0,1]$,

$$\lim_{n \to \infty} \mathbb{P}_{\boldsymbol{\beta}^*} \Big( \sup_{v \in [0,1]} \|\nabla_{\theta \boldsymbol{\gamma}}^2 \ell(\theta^*, \boldsymbol{\gamma}_v) - \widehat{\mathbf{w}}^T \nabla_{\boldsymbol{\gamma}\boldsymbol{\gamma}}^2 \ell(\theta^*, \boldsymbol{\gamma}_v)\|_\infty \lesssim \eta_4(n) \Big) = 1,$$

for some $\eta_4(n) \to 0$, as $n \to \infty$.

**Assumption 3.29** (CLT). Recall that $\mathbf{I}^* = \mathbb{E}_{\boldsymbol{\beta}^*}(\nabla^2 \ell(\boldsymbol{\beta}^*))$. For $\mathbf{v}^* = (1, -\mathbf{w}^{*T})^T$, it holds that

$$\frac{\sqrt{n}\mathbf{v}^{*T}\nabla\ell(\boldsymbol{\beta}^*)}{\sqrt{\mathbf{v}^{*T}\mathbf{I}^*\mathbf{v}^*}} \rightsquigarrow N(0,1).$$

Assume that $C' \leq I^*_{\theta|\boldsymbol{\gamma}} \leq C < \infty$, where $C > C' > 0$ are some positive constants.

**Assumption 3.30** (Convergence of Hessian Matrix). For $\theta_v = v\theta^* + (1-v)\widehat{\theta}$ with $v \in [0,1]$,

$$\lim_{n \to \infty} \mathbb{P}_{\boldsymbol{\beta}^*}\big(\sup_{v \in [0,1]} ||\nabla^2 \ell(\theta_v, \widehat{\boldsymbol{\gamma}}) - \mathbf{I}^*||_{\max} \lesssim \eta_5(n)\big) = 1,$$

for some $\eta_5(n) \to 0$, as $n \to \infty$.

Since we use $\theta^*$ to denote the truth instead of 0 as stated in the null hypothesis, the Assumptions 3.27, 3.28, and 3.29 are essentially the same as Assumptions 3.2, 3.3, and 3.4, respectively. Compared to the Assumption 3.6, we need slightly stronger form in Assumption 3.30, which is necessary for applying the mean value theorem in the proof. The following main theorem of this section shows that the one-step estimator $\widetilde{\theta}$ is asymptotically normal.

**Theorem 3.31.** Assume that the Assumptions 3.1, 3.27–3.30 hold. It also holds that $||\mathbf{w}^*||_1 \eta_5(n) = o(1)$, $\eta_2(n)||\mathbf{I}^*_{\theta\boldsymbol{\gamma}}||_\infty = o(1)$, and $n^{1/2}(\eta_2(n)\eta_3(n) + \eta_1(n)\eta_4(n)) = o(1)$. In addition, assume that $n^{1/2}(\widehat{\theta} - \theta^*)||\mathbf{w}^*||_1 \eta_5(n) = o_\mathbb{P}(1)$. Then, we have

$$n^{1/2}(\widetilde{\theta} - \theta^*)\widehat{I}_{\theta|\boldsymbol{\gamma}}^{1/2} = n^{1/2}(\widetilde{\theta} - \theta^*)I_{\theta|\boldsymbol{\gamma}}^{*1/2} + o_\mathbb{P}(1) = N + o_\mathbb{P}(1), \quad (3.16)$$

where $N \sim N(0,1)$ and $\widehat{I}_{\theta|\boldsymbol{\gamma}} = \nabla^2_{\theta\theta}\ell(\widehat{\boldsymbol{\beta}}) - \widehat{\mathbf{w}}^T \nabla^2_{\boldsymbol{\gamma}\theta}\ell(\widehat{\boldsymbol{\beta}})$.

*Proof.* A detailed proof is shown in Appendix A. □

**Remark 3.32.** Note that the conditions in Theorem 3.31 are slightly stronger than those in Corollary 3.7. In particular, we need to assume $n^{1/2}(\widehat{\theta} - \theta^*)||\mathbf{w}^*||_1 \eta_5(n) = o_\mathbb{P}(1)$ holds in Theorem 3.31 instead of Corollary 3.7. Since the following inequalities hold $|\widehat{\theta} - \theta^*| \leq ||\widehat{\boldsymbol{\beta}} - \boldsymbol{\beta}^*||_\infty \leq ||\widehat{\boldsymbol{\beta}} - \boldsymbol{\beta}^*||_2$, we can replace $(\widehat{\theta} - \theta^*)$ by $||\widehat{\boldsymbol{\beta}} - \boldsymbol{\beta}^*||_\infty$ or $||\widehat{\boldsymbol{\beta}} - \boldsymbol{\beta}^*||_2$. For instance, in the linear model, Wang et al. (2013b) showed that $||\widehat{\boldsymbol{\beta}} - \boldsymbol{\beta}^*||_\infty = \mathcal{O}_\mathbb{P}(\sqrt{\frac{\log s^*}{n}})$, with the nonconvex penalty. Similarly, with the Lasso penalty and the irrepresentable condition, Wainwright (2009) showed that $||\widehat{\boldsymbol{\beta}} - \boldsymbol{\beta}^*||_\infty = \mathcal{O}_\mathbb{P}(\sqrt{\frac{\log d}{n}})$. If such strong model assumptions are not desired, we can further relax it to $||\widehat{\boldsymbol{\beta}} - \boldsymbol{\beta}^*||_2$, whose magnitude is well understood; see Bickel et al. (2009); Negahban et al. (2012); Wang et al. (2013b).

**Remark 3.33.** One implication of (3.16) is that the asymptotic variance of $\widetilde{\theta}$ is identical to the inverse of the partial information matrix $I^*_{\theta|\boldsymbol{\gamma}}$, which is also known as the efficient information in the presence of nuisance parameters; see Chapter 25 of Van der Vaart (2000). Hence, we conclude that the one-step estimator $\widetilde{\theta}$ is semiparametrically efficient. Another implication of (3.16) is that a $(1-\alpha) \times 100\%$ confidence interval for $\theta^*$ is given by $[\widetilde{\theta} - n^{-1/2}\Phi^{-1}(1-\alpha/2)\widehat{I}_{\theta|\boldsymbol{\gamma}}^{-1/2}, \widetilde{\theta} + n^{-1/2}\Phi^{-1}(1-\alpha/2)\widehat{I}_{\theta|\boldsymbol{\gamma}}^{-1/2}]$. Inherited from the semiparametric efficiency of $\widetilde{\theta}$, this confidence interval is also optimal. A Similar criterion on optimality for confidence intervals under the linear model is considered by van de Geer et al. (2014).

# 4 High Dimensional Linear Models

In this section, we study the consequences of the general results in Section 3 for the linear model. We first consider the case where the variance of the noise is known and illustrate the optimality property of the score test. Then, we show that the score test with unknown variance of the noise has the same asymptotic properties. Unlike van de Geer et al. (2014); Javanmard and Montanari (2013), throughout this section, we do not need the Gaussian noise assumption.

## 4.1 Sub-Gaussian Noise with Known Variance

Consider the linear regression, $Y_i = \theta^* Z_i + \boldsymbol{\gamma}^{*T} \boldsymbol{X}_i + \epsilon_i$, where $Z_i \in \mathbb{R}$, $\boldsymbol{X}_i \in \mathbb{R}^{d-1}$, and the error $\epsilon_i$ satisfies $\mathbb{E}(\epsilon_i) = 0$, $\mathbb{E}(\epsilon_i^2) = \sigma^2$ for $i = 1, ..., n$. Let $\boldsymbol{Q}_i = (Z_i, \boldsymbol{X}_i^T)^T$ denote the collection of all covariates for subject $i$. In this section, we assume that the variance $\sigma^2$ is known. To be concrete, we consider the following Lasso estimator,

$$\widehat{\boldsymbol{\beta}} = \operatorname*{argmin}_{\boldsymbol{\beta}} \left\{ \frac{1}{2n} \sum_{i=1}^n (Y_i - \boldsymbol{\beta}^T \boldsymbol{Q}_i)^2 + \lambda \|\boldsymbol{\beta}\|_1 \right\}. \tag{4.1}$$

Based on the Gaussian quasi-likelihood, the decorrelated score function is

$$S(\theta, \boldsymbol{\gamma}) = -\frac{1}{\sigma^2 n} \sum_{i=1}^n (Y_i - \theta Z_i - \boldsymbol{\gamma}^T \boldsymbol{X}_i)(Z_i - \mathbf{w}^T \boldsymbol{X}_i),$$

where $\mathbf{w} = \mathbb{E}_{\boldsymbol{\beta}}(\boldsymbol{X}_i^{\otimes 2})^{-1} \mathbb{E}_{\boldsymbol{\beta}}(Z_i \boldsymbol{X}_i)$. Since the distribution of the design matrix does not depend on $\boldsymbol{\beta}$, we can replace $\mathbb{E}_{\boldsymbol{\beta}}(\cdot)$ by $\mathbb{E}(\cdot)$ for notational simplicity. In practice, under the null hypothesis $H_0 : \theta^* = 0$, the decorrelated score function can be estimated by

$$\widehat{S}(0, \widehat{\boldsymbol{\gamma}}) = -\frac{1}{\sigma^2 n} \sum_{i=1}^n (Y_i - \widehat{\boldsymbol{\gamma}}^T \boldsymbol{X}_i)(Z_i - \widehat{\mathbf{w}}^T \boldsymbol{X}_i), \tag{4.2}$$

where

$$\widehat{\mathbf{w}} = \operatorname{argmin} \|\mathbf{w}\|_1, \quad \text{s.t.} \quad \left\| \frac{1}{n} \sum_{i=1}^n \boldsymbol{X}_i (Z_i - \mathbf{w}^T \boldsymbol{X}_i) \right\|_\infty \leq \lambda'. \tag{4.3}$$

The (partial) information matrices are

$$\mathbf{I}^* = \sigma^{-2} \mathbb{E}(\boldsymbol{Q}_i^{\otimes 2}), \quad \text{and} \quad I^*_{\theta|\boldsymbol{\gamma}} = \sigma^{-2}(\mathbb{E}(Z_i^2) - \mathbb{E}(Z_i \boldsymbol{X}_i^T) \mathbb{E}(\boldsymbol{X}_i^{\otimes 2})^{-1} \mathbb{E}(\boldsymbol{X}_i Z_i)),$$

which can be estimated by

$$\widehat{\mathbf{I}} = \frac{1}{\sigma^2 n} \sum_{i=1}^n \boldsymbol{Q}_i^{\otimes 2}, \quad \text{and} \quad \widehat{I}_{\theta|\boldsymbol{\gamma}} = \frac{1}{\sigma^2} \left\{ \frac{1}{n} \sum_{i=1}^n Z_i^2 - \widehat{\mathbf{w}}^T \left( \frac{1}{n} \sum_{i=1}^n \boldsymbol{X}_i Z_i \right) \right\},$$

respectively. This leads to the score test statistic $\widehat{U}_n = n^{1/2} \widehat{S}(0, \widehat{\boldsymbol{\gamma}}) \widehat{I}_{\theta|\boldsymbol{\gamma}}^{-1/2}$. The following theorem establishes the asymptotic null distribution of the score test statistic $\widehat{U}_n$.

**Theorem 4.1.** Assume that (1) $\lambda_{\min}(\mathbb{E}(\boldsymbol{Q}_i^{\otimes 2})) \geq 2\kappa$ for some constant $\kappa > 0$, (2) $S = \text{supp}(\boldsymbol{\beta}^*)$ and $S' = \text{supp}(\mathbf{w}^*)$ satisfy $|S| = s^*$ and $|S'| = s'$, (3) $\epsilon_i$, $\mathbf{w}^{*T}\boldsymbol{X}_i$, $Q_{ij}$ are all sub-Gaussian with $\|\epsilon_i\|_{\psi_2} \leq C$, $\|\mathbf{w}^{*T}\boldsymbol{X}_i\|_{\psi_2} \leq C$ and $\|Q_{ij}\|_{\psi_2} \leq C$, where $C$ is a positive constant. If $n^{-1/2}(s' \vee s^*) \log d = o(1)$ and $\lambda \asymp \lambda' \asymp \sqrt{\frac{\log d}{n}}$, then under $H_0 : \theta^* = 0$ for each $t \in \mathbb{R}$,

$$\lim_{n \to \infty} |\mathbb{P}_{\boldsymbol{\beta}^*}(\widehat{U}_n \leq t) - \Phi(t)| = 0. \tag{4.4}$$

*Proof.* This theorem follows from the general results in Theorem 3.5 and Corollary 3.7. A detailed proof is shown in Appendix B. □

Note that the estimator $\widehat{\mathbf{w}}$ in (4.3) has the same form as the Dantzig selector (Candes and Tao, 2007). With the $L_1$ penalty, the alternative estimator $\widetilde{\mathbf{w}}$ defined in (2.7) is equivalent to the following Lasso estimator,

$$\widetilde{\mathbf{w}} = \operatorname*{argmin}_{\mathbf{w}} \left\{ \frac{1}{2n} \sum_{i=1}^{n} \left( Z_i - \mathbf{w}^T \boldsymbol{X}_i \right)^2 + \lambda' \|\mathbf{w}\|_1 \right\}. \tag{4.5}$$

**Corollary 4.2.** Under the same conditions in Theorem 4.1, the score test statistic $\widehat{U}_n$ with $\widehat{\mathbf{w}}$ replaced by $\widetilde{\mathbf{w}}$ in (4.5) satisfies $\lim_{n \to \infty} |\mathbb{P}_{\boldsymbol{\beta}^*}(\widehat{U}_n \leq t) - \Phi(t)| = 0$, for any $t \in \mathbb{R}$.

*Proof.* A detailed proof is shown in Appendix B. □

Since the estimated decorrelated score function $\widehat{S}(\theta, \widehat{\gamma})$ is linear in $\theta$, the solution to $\widehat{S}(\theta, \widehat{\gamma}) = 0$ is identical to the one-step estimator, which is given by

$$\widetilde{\theta} = \frac{\sum_{i=1}^{n}(Y_i - \widehat{\gamma}^T \boldsymbol{X}_i)(Z_i - \widehat{\mathbf{w}}^T \boldsymbol{X}_i)}{\sum_{i=1}^{n} Z_i(Z_i - \widehat{\mathbf{w}}^T \boldsymbol{X}_i)}.$$

Similarly, we can replace $\widehat{\mathbf{w}}$ in $\widetilde{\theta}$ by the Lasso estimator $\widetilde{\mathbf{w}}$ in (4.5). The following corollary illustrates the consequence of the general results in Theorem 3.31 for linear models.

**Corollary 4.3.** Assume that conditions (1), (2) and (3) in Theorem 4.1 hold. If $n^{-1/2}(s' \vee s^*) \log d = o(1)$ and $\lambda \asymp \lambda' \asymp \sqrt{\frac{\log d}{n}}$, then $n^{1/2}(\widetilde{\theta} - \theta^*)\widehat{I}_{\theta|\gamma}^{1/2} \rightsquigarrow N(0,1)$, where $\widetilde{\theta}$ and $\widehat{I}_{\theta|\gamma}$ are constructed based on either $\widehat{\mathbf{w}}$ or $\widetilde{\mathbf{w}}$.

*Proof.* This result follows from the general results in Theorem 3.31 and the proof of Theorem 4.1. See Appendix B for a detailed proof. □

**Remark 4.4.** In Theorem 4.1, the condition (1) ensures that the covariance of the design matrix has a bounded minimal eigenvalue, which is a typical condition for the high dimensional design matrix. The condition (2) assumes the sparsity of $\boldsymbol{\beta}^*$ and $\mathbf{w}^*$. Note that, by the block matrix inversion formula, the sparsity of $\mathbf{w}^*$ is implied by the sparsity of the precision matrix $(\mathbb{E}(\boldsymbol{Q}_i^{\otimes 2}))^{-1}$, assumed in van de Geer et al. (2014). This suggests that the sparsity of $\mathbf{w}^*$ is weaker than the assumptions in van de Geer et al. (2014). Finally, the condition (3) is a technical condition for applying concentration inequalities. Similar assumptions are also made by Bickel et al. (2009); van de Geer et al. (2014).

**Remark 4.5.** By Theorem 2.4 of van de Geer et al. (2014) and Theorem 4.1 of Javanmard and Montanari (2013), it can be seen that their estimators are asymptotically equivalent to the one-step estimator in Corollary 4.3 and their Wald tests are also asymptotically equivalent to the decorrelated score test in Theorem 4.1, if the linear model is correctly specified. On the other hand, when the linear model is misspecified, the results in van de Geer et al. (2014) and Javanmard and Montanari (2013) are invalid, and the score test for the oracle parameter can be still applied; see Section 6.

**Remark 4.6.** By inspecting the proof of Theorem 4.1, it can be shown that Assumptions 3.9–3.13 hold. Hence, by the general results in Theorem 3.14, under the same conditions as those in Theorem 4.1, we can obtain the following uniform convergence result,

$$\lim_{n\to\infty} \sup_{\boldsymbol{\beta}^*\in\Omega_0} \sup_{t\in\mathbb{R}} |\mathbb{P}_{\boldsymbol{\beta}^*}(\widehat{U}_n \leq t) - \Phi(t)| = 0, \tag{4.6}$$

where $\Omega_0 = \{(0, \boldsymbol{\gamma}) : \|\boldsymbol{\gamma}\|_0 \leq s^*\}$. Note that, van de Geer et al. (2014) and Javanmard and Montanari (2013) derived the similar results to (4.6) for their Wald tests, under the Gaussian noise assumption. Here, we follow the general results in Theorem 3.14 and do not need this assumption.

Given the fact that the score test can control the type I error asymptotically uniformly over $\Omega_0$, we now derive the asymptotic power of the score test.

**Theorem 4.7.** Assume that (1) $\lambda_{\min}(\mathbb{E}(\boldsymbol{Q}_i^{\otimes 2})) \geq 2\kappa$ for some constant $\kappa > 0$, (2) $\|\mathbf{w}^*\|_0 = s'$, (3) $\epsilon_i$, $\mathbf{w}^{*T}\boldsymbol{X}_i$, $Q_{ij}$ are all sub-Gaussian with $\|\epsilon_i\|_{\psi_2} \leq C$, $\|\mathbf{w}^{*T}\boldsymbol{X}_i\|_{\psi_2} \leq C$ and $\|Q_{ij}\|_{\psi_2} \leq C$, where $C$ is a positive constant. Let $\Omega_1(\widetilde{C}, \phi) = \{(\theta, \boldsymbol{\gamma}) : \theta = \widetilde{C}n^{-\phi}, \|\boldsymbol{\gamma}\|_0 \leq s^*\}$. If $n^{-1/2}(s' \vee s^*) \log d = o(1)$, $n^{-\phi}\sqrt{\log n} = o(1)$ and $n^{-\phi}s'\sqrt{\log d} = o(1)$, then

$$\lim_{n\to\infty} \sup_{\boldsymbol{\beta}^*\in\Omega_1(\widetilde{C},\phi)} \sup_{t\in\mathbb{R}} |\mathbb{P}_{\boldsymbol{\beta}^*}(\widehat{U}_n \leq t) - \Phi(t)| = 0, \quad \text{if} \quad \phi > 1/2,$$

$$\lim_{n\to\infty} \sup_{\boldsymbol{\beta}^*\in\Omega_1(\widetilde{C},\phi)} \sup_{t\in\mathbb{R}} |\mathbb{P}_{\boldsymbol{\beta}^*}(\widehat{U}_n \leq t) - \Phi(t + \widetilde{C}I_{\theta|\gamma}^{*1/2})| = 0, \quad \text{if} \quad \phi = 1/2,$$

$$\lim_{n\to\infty} \inf_{\boldsymbol{\beta}^*\in\Omega_1(\widetilde{C},\phi)} \mathbb{P}_{\boldsymbol{\beta}^*}(|\widehat{U}_n| > t) = 1, \quad \text{if} \quad \phi < 1/2,$$

where the last equation holds for any fixed $t \in \mathbb{R}$ and $\widetilde{C} \neq 0$.

*Proof.* This theorem follows from the general results in Theorem 3.22. See Appendix B for a detailed proof. □

**Remark 4.8.** We can see the asymptotic power of the score test can be established under the nearly identical conditions to those in Theorem 4.7. The extra condition $n^{-\phi}\sqrt{\log n} = o(1)$ ensures that the uniform local approximation in Assumption 3.16 holds. In addition, $n^{-\phi}s'\sqrt{\log d} = o(1)$ is needed to control the noise level $\|\nabla_{\boldsymbol{\gamma}}\ell(0,\boldsymbol{\gamma})\|_\infty$ in Assumption 3.18.

## 4.2 Minimax Optimality of Score Test

In this section, we study the optimality of the score test for the linear models. In particular, we compare the asymptotic powers of the score test with the optimal test. One such criterion for the optimality of tests is in terms of the minimax power for testing $H_0 : \theta^* = 0$ versus the local alternative $H_{1n} : \theta^* = n^{-1/2}\widetilde{C}$, for some constant $\widetilde{C}$.

Recall that $\Omega_0 = \{(0, \boldsymbol{\gamma}) : \|\boldsymbol{\gamma}\|_0 \leq s^*\}$ and $\Omega_1(\widetilde{C}, 1/2) = \{(\widetilde{C}n^{-1/2}, \boldsymbol{\gamma}) : \|\boldsymbol{\gamma}\|_0 \leq s^*\}$. A univariate statistic $T$ is called hypothesis test, if it is a function of random variables independent of parameters and only takes values in $\{0, 1\}$. Without loss of generality, we assume that $H_0$ is rejected if and only if $T = 1$. Hence, the type I error of $T$ is given by $\mathbb{P}_{\boldsymbol{\beta}^*}(T = 1)$ if $\boldsymbol{\beta}^* \in \Omega_0$ and the power of $T$ is given by $\mathbb{P}_{\boldsymbol{\beta}^*}(T = 1)$ if $\boldsymbol{\beta}^* \in \Omega_1(\widetilde{C}, 1/2)$. To rule out the trivial test, such as $T = 1$, a.s, we only consider the test with type I error (asymptotically) controlled at level $\alpha$, for some $0 < \alpha < 1$. The following theorem characterizes the minimax power for testing $H_0 : \theta^* = 0$ versus $H_{1n} : \theta^* = n^{-1/2}\widetilde{C}$.

**Theorem 4.9.** Assume that $\epsilon_i \sim N(0, \sigma^2)$, and $\boldsymbol{Q}_i \sim N(0, \boldsymbol{\Sigma})$. For a test $T$, denote $\alpha(T) = \lim_{n \to \infty} \sup_{\boldsymbol{\beta}^* \in \Omega_0} \mathbb{P}_{\boldsymbol{\beta}^*}(T = 1)$, and $T_\alpha = \{T : \alpha(T) \leq \alpha\}$. The minimax power for testing $\Omega_0 = \{(0, \boldsymbol{\gamma}) : \|\boldsymbol{\gamma}\|_0 \leq s^*\}$ versus $\Omega_1(\widetilde{C}, 1/2) = \{(\widetilde{C}n^{-1/2}, \boldsymbol{\gamma}) : \|\boldsymbol{\gamma}\|_0 \leq s^*\}$ is

$$\lim_{n \to \infty} \sup_{T \in T_\alpha} \inf_{\boldsymbol{\beta}^* \in \Omega_1(\widetilde{C}, 1/2)} \mathbb{P}_{\boldsymbol{\beta}^*}(T = 1) \leq 1 - \Phi\big(\Phi^{-1}(1 - \alpha) - \widetilde{C} I_{\theta|S}^{*1/2}\big),$$

where $I_{\theta|S}^* = \sigma^{-2}(\mathbb{E}(Z_i^2) - \mathbb{E}(Z_i \boldsymbol{X}_{i,S}^T)\mathbb{E}(\boldsymbol{X}_{i,S}^{\otimes 2})^{-1}\mathbb{E}(\boldsymbol{X}_{i,S} Z_i))$, and $S$ represents the support set of $\boldsymbol{\gamma}^*$.

*Proof.* A detailed proof is shown in Appendix B. □

**Remark 4.10.** This theorem shows that the power corresponding to the optimal test uniformly over the one-sided alternative $H_{1n}$ cannot exceed $1 - \Phi\big(\Phi^{-1}(1 - \alpha) - \widetilde{C} I_{\theta|S}^{*1/2}\big)$. A similar minimax upper bound for two-sided alternatives is shown in Theorem 3.6 of Javanmard and Montanari (2013).

**Remark 4.11.** By Remark 3.25, the limiting power of the one-sided score test $T'_n$ for testing $H_0$ versus $H_{1n}$ is $1 - \Phi\big(\Phi^{-1}(1 - \alpha) - \widetilde{C} I_{\theta|\gamma}^{*1/2}\big)$, which is typically smaller than the minimax power in Theorem 4.9, due to $I_{\theta|\gamma}^* \leq I_{\theta|S}^*$. If $(Z_i, \boldsymbol{X}_{i,S})$ is uncorrelated with $\boldsymbol{X}_{i,\bar{S}}$, then the score test $T'_n$ achieves the minimax bound. In general, the score test is not minimax optimal. This is mainly due to the unknown support of $\boldsymbol{\beta}^*$.

In the following, we will show that the upper bound in Theorem 4.9 can be achieved in the high dimensional setting with some additional assumptions. Let $\widehat{S} = \text{supp}(\widehat{\boldsymbol{\beta}})$ denote the support set of the estimator $\widehat{\boldsymbol{\beta}}$. Consider the following truncated decorrelated score function,

$$\widehat{S}_{TC}(0, \widehat{\boldsymbol{\gamma}}) = -\frac{1}{\sigma^2 n} \sum_{i=1}^n (Y_i - \widehat{\boldsymbol{\gamma}}^T \boldsymbol{X}_i)(Z_i - \widehat{\boldsymbol{v}}^T \boldsymbol{X}_{i,\widehat{S}}),$$

where

$$\widehat{\boldsymbol{v}} = \arg\min \|\boldsymbol{v}\|_1, \quad \text{s.t.} \quad \left\| \frac{1}{n} \sum_{i=1}^n \boldsymbol{X}_{i,\widehat{S}} (Z_i - \boldsymbol{v}^T \boldsymbol{X}_{i,\widehat{S}}) \right\|_\infty \leq \lambda'.$$

The key difference between $\widehat{S}_{TC}(0, \widehat{\boldsymbol{\gamma}})$ and $\widehat{S}(0, \widehat{\boldsymbol{\gamma}})$ is that $\widehat{\boldsymbol{v}}$ in $\widehat{S}_{TC}(0, \boldsymbol{\gamma})$ belongs to $\mathbb{R}^{|\widehat{S}|}$, whose dimension can be much smaller than the dimension of $\widehat{\boldsymbol{w}}$. Recall that the estimand of $\widehat{\boldsymbol{w}}$ is $\boldsymbol{w}^* =$ $\mathbb{E}(\boldsymbol{X}_i^{\otimes 2})^{-1}\mathbb{E}(Z_i\boldsymbol{X}_i) \in \mathbb{R}^d$. In contrast, the estimand of $\widehat{\mathbf{v}}$ is $\mathbf{v}^* = \mathbb{E}(\boldsymbol{X}_{i,S}^{\otimes 2})^{-1}\mathbb{E}(Z_i\boldsymbol{X}_{i,S}) \in \mathbb{R}^{|S|}$. The Dantzig type estimator $\widehat{\mathbf{v}}$ essentially solves a low dimensional problem by truncating the covariate $\boldsymbol{X}_i$, and therefore we call the resulting score function as the truncated decorrelated score function. Denote $\widehat{U}_{TC} = n^{1/2}\widehat{S}_{TC}(0,\widehat{\boldsymbol{\gamma}})\widehat{I}_{\theta|\widehat{S}}^{-1/2}$ to be the truncated decorrelated score statistic, where

$$\widehat{I}_{\theta|\widehat{S}} = \frac{1}{\sigma^2}\left\{\frac{1}{n}\sum_{i=1}^n Z_i^2 - \widehat{\mathbf{v}}^T\left(\frac{1}{n}\sum_{i=1}^n \boldsymbol{X}_{i,\widehat{S}}Z_i\right)\right\}.$$

Recall that the irrepresentable condition introduced by Zhao and Yu (2006) is the following, $\|(\sum_{i=1}^n \boldsymbol{Q}_{i,\bar{S}}\boldsymbol{Q}_{i,S}^T)(\sum_{i=1}^n \boldsymbol{Q}_{i,S}^{\otimes 2})^{-1}\|_{\ell_\infty} \leq 1 - \eta$, for some $\eta > 0$. The next proposition shows that the minimax power can be achieved by $\widehat{U}_{TC}$.

**Proposition 4.12.** Assume that the irrepresentable condition holds. Under the same conditions as those in Theorem 4.1 with $\mathbf{w}^*$ replaced by $\mathbf{v}^*$ and $\min_{j\in S}|\boldsymbol{\gamma}_j^*| \geq C\sqrt{\log d/n}$ for some constant $C > 0$, for any $t \in \mathbb{R}$, we have

$$\lim_{n\to\infty}|\mathbb{P}_{\boldsymbol{\beta}^*}(\widehat{U}_{TC} \leq t) - \Phi(t)| = 0.$$

Furthermore, if the same conditions as those in Theorem 4.7 with $\phi = 1/2$ and $\mathbf{w}^*$ replaced by $\mathbf{v}^*$ also hold, we have

$$\lim_{n\to\infty}\sup_{t\in\mathbb{R}}\left|\mathbb{P}_{\boldsymbol{\beta}^*}(\widehat{U}_{TC} \leq t) - \Phi(t + \widetilde{C}I_{\theta|S}^{*1/2})\right| = 0,$$

where $\boldsymbol{\beta}^* = (\widetilde{C}n^{-1/2},\boldsymbol{\gamma}^*)$ with $\min_{j\in S}|\boldsymbol{\gamma}_j^*| \geq C\sqrt{\log d/n}$ for some constant $C > 0$.

*Proof.* A detailed proof is shown in Appendix B. □

**Remark 4.13.** By this proposition, we find that the hypothesis test based on $\widehat{U}_{TC}$ for testing $\theta^* = 0$ versus $\theta^* = \widetilde{C}n^{-1/2}$ achieves the minimax power described in Theorem 4.9. This proposition depends on the key assumptions: (1) the minimal nonzero signal $\min_{j\in S}|\boldsymbol{\gamma}_j^*|$ must be large enough and (2) the irrepresentable condition holds. These two conditions together imply that the estimated support set of $\boldsymbol{\gamma}^*$ agrees with the truth (i.e., $\operatorname{supp}(\widehat{\boldsymbol{\gamma}}) = \operatorname{supp}(\boldsymbol{\gamma}^*)$) with high probability. On this event, Proposition 4.12 follows directly from the Theorems 4.1 and 4.7.

**Remark 4.14.** There exist alternative tests that can also achieve the optimal power. For instance, in $\widehat{S}_{TC}(0,\widehat{\boldsymbol{\gamma}})$, to estimate $\mathbf{v}^*$, instead of the Dantzig type estimator $\widehat{\mathbf{v}}$, we can use the empirical estimator $\widetilde{\mathbf{v}} = (\sum_{i=1}^n \boldsymbol{X}_{i,\widehat{S}}^{\otimes 2})^{-1}(\sum_{i=1}^n Z_i\boldsymbol{X}_{i,\widehat{S}})$. This provides an alternative procedure to construct the truncated decorrelated score test. Indeed, the decorrelation operation is asymptotically equivalent to the classical profile based method for low dimensional models. Hence, the score test based on the profile score function under $H_0$, i.e., $\frac{1}{\sigma^2 n}\sum_{i=1}^n Z_i(Y_i - \widetilde{\boldsymbol{\gamma}}_{0,\widehat{S}}^T\boldsymbol{X}_{i,\widehat{S}})$, is also optimal, where $\widetilde{\boldsymbol{\gamma}}_{0,\widehat{S}} = (\sum_{i=1}^n \boldsymbol{X}_{i,\widehat{S}}^{\otimes 2})^{-1}(\sum_{i=1}^n Y_i\boldsymbol{X}_{i,\widehat{S}})$ is the least square estimator under the selected model. Although these tests can be locally most powerful, they all hinge on the model selection consistency, which limits their applications in practice.

## 4.3 Estimation of Unknown Variance $\sigma^2$

In the previous section, we assume that the variance $\sigma^2$ is known. In this section, we consider the estimation of $\sigma^2$ and the asymptotic properties of the score test with estimated $\sigma^2$. With the Lasso estimator $\widehat{\boldsymbol{\beta}}$, one can estimate $\sigma^2$ by $\widehat{\sigma}^2 = \frac{1}{n}\sum_{i=1}^n (Y_i - \widehat{\boldsymbol{\beta}}^T \boldsymbol{Q}_i)^2$. Consider the following score statistic with $\widehat{\sigma}^2$,

$$\widetilde{U}_n = -\frac{1}{\widehat{\sigma} n^{1/2}} \sum_{i=1}^n (Y_i - \widehat{\boldsymbol{\gamma}}^T \boldsymbol{X}_i)(Z_i - \widehat{\mathbf{w}}^T \boldsymbol{X}_i)(H_Z - \widehat{\mathbf{w}}^T \mathbf{H}_{XZ})^{-1/2},$$

where $H_Z = n^{-1} \sum_{i=1}^n Z_i^2$ and $\mathbf{H}_{XZ} = n^{-1} \sum_{i=1}^n Z_i \boldsymbol{X}_i$. The following theorem characterizes the asymptotic null distribution of $\widetilde{U}_n$. In particular, we show that $\widetilde{U}_n$ and $\widehat{U}_n$ are uniformly asymptotically equivalent, where $\widehat{U}_n$ is the score test statistic with known $\sigma^2$.

**Theorem 4.15.** Assume that the conditions in Theorem 4.1 hold and the true value $\sigma^{*2} \geq C$ for some constant $C > 0$. Then $\widetilde{U}_n$ and $\widehat{U}_n$ are uniformly asymptotically equivalent, i.e., for any $\epsilon > 0$,

$$\lim_{n \to \infty} \sup_{\boldsymbol{\beta}^* \in \Omega_0} \mathbb{P}_{\boldsymbol{\beta}^*}\left(|\widetilde{U}_n - \widehat{U}_n| \geq \epsilon\right) = 0. \tag{4.7}$$

Moreover, for $\Omega_0 = \{(0, \boldsymbol{\gamma}) : \|\boldsymbol{\gamma}\|_0 \leq s^*\}$, we have,

$$\lim_{n \to \infty} \sup_{\boldsymbol{\beta}^* \in \Omega_0} \sup_{t \in \mathbb{R}} |\mathbb{P}_{\boldsymbol{\beta}^*}(\widetilde{U}_n \leq t) - \Phi(t)| = 0.$$

*Proof.* A detailed proof is shown in Appendix B. □

**Corollary 4.16.** Under the same conditions as in Corollary 4.3, if $\sigma^{*2} \geq C$ for some constant $C > 0$, then we have $n^{1/2}(\widetilde{\theta} - \theta^*)V \rightsquigarrow N(0,1)$, where

$$V = \frac{1}{\widehat{\sigma}} \left\{ \frac{1}{n} \sum_{i=1}^n Z_i^2 - \widehat{\mathbf{w}}^T \left(\frac{1}{n} \sum_{i=1}^n \boldsymbol{X}_i Z_i\right) \right\}^{1/2}.$$

*Proof.* A detailed proof is shown in Appendix B. □

**Remark 4.17.** Note that the asymptotic equivalence (4.7) also holds under the local alternative. Together with Theorem 4.7, we can obtain that $\widetilde{U}_n$ has the same asymptotic power as $\widehat{U}_n$, i.e.,

$$\lim_{n \to \infty} \sup_{\boldsymbol{\beta}^* \in \Omega_1(\widetilde{C}, 1/2)} \sup_{t \in \mathbb{R}} \left|\mathbb{P}_{\boldsymbol{\beta}^*}(\widetilde{U}_n \leq t) - \Phi(t + \widetilde{C} I_{\theta|\boldsymbol{\gamma}}^{*1/2})\right| = 0,$$

where $\Omega_1(\widetilde{C}, 1/2) = \{(\widetilde{C} n^{-1/2}, \boldsymbol{\gamma}) : \|\boldsymbol{\gamma}\|_0 \leq s^*\}$.

**Remark 4.18.** In general, the asymptotic equivalence (4.7) follows from the orthogonality of parameters $\boldsymbol{\beta}$ and $\sigma^2$ in the log-likelihood function, rather than the use of the Lasso estimators $\widehat{\boldsymbol{\beta}}$ and $\widehat{\sigma}^2$. This implies that, for the estimation of $(\boldsymbol{\beta}, \sigma)$, we can use alternative estimators such as the scaled Lasso (Sun and Zhang, 2012),

$$(\widetilde{\boldsymbol{\beta}}_s, \widetilde{\sigma}_s) = \underset{\boldsymbol{\beta} \in \mathbb{R}^d, \sigma > 0}{\operatorname{argmin}} \left\{ \frac{1}{2\sigma n} \sum_{i=1}^n (Y_i - \boldsymbol{\beta}^T \boldsymbol{Q}_i)^2 + \frac{\sigma}{2} + \lambda \|\boldsymbol{\beta}\|_1 \right\}.$$

By Theorem 1 and Corollary 1 of Sun and Zhang (2012), we can show that the corresponding score test with $(\widetilde{\boldsymbol{\beta}}_s, \widetilde{\sigma}_s)$ is asymptotically equivalent to $\widetilde{U}_n$ and $\widehat{U}_n$, and therefore has the same type I error and local asymptotic power. Moreover, the estimator $(\widetilde{\boldsymbol{\beta}}_s, \widetilde{\sigma}_s)$ has the additional advantage of being tuning insensitive. We refer to Sun and Zhang (2012); Belloni et al. (2011) for more detailed discussions.

## 5 High Dimensional Generalized Linear Models

In this section, we illustrate the consequences of the general results in Section 3 for the generalized linear model. Assume that $Y_i$ given $\boldsymbol{Q}_i = (Z_i, \boldsymbol{X}_i)$ follows from the generalized linear model, whose negative log-likelihood is

$$\ell(\theta, \boldsymbol{\gamma}) = -\frac{1}{n}\sum_{i=1}^{n} \frac{1}{a(\phi)} \left\{ Y_i(\theta Z_i + \boldsymbol{\gamma}^T \boldsymbol{X}_i) - b(\theta Z_i + \boldsymbol{\gamma}^T \boldsymbol{X}_i) \right\},$$

where $a(\cdot)$ and $b(\cdot)$ are known functions. For simplicity, we set $a(\phi) = 1$. Assume that $\widehat{\boldsymbol{\beta}}$ is given by (1.1) with some generic penalty function. After some simple algebra, the estimated decorrelated score function under the null hypothesis is

$$\widehat{S}(0, \widehat{\boldsymbol{\gamma}}) = -\frac{1}{n}\sum_{i=1}^{n}(Y_i - b'(\widehat{\boldsymbol{\gamma}}^T \boldsymbol{X}_i))(Z_i - \widehat{\mathbf{w}}^T \boldsymbol{X}_i),$$

where $\widehat{\mathbf{w}}$ in the context of generalized linear models is given by

$$\widehat{\mathbf{w}} = \operatorname{argmin} ||\mathbf{w}||_1, \quad \text{s.t.} \quad \left\| \frac{1}{n}\sum_{i=1}^{n} b''(\widehat{\boldsymbol{\beta}}^T \boldsymbol{Q}_i)(Z_i - \mathbf{w}^T \boldsymbol{X}_i)\boldsymbol{X}_i \right\|_\infty \leq \lambda'. \qquad (5.1)$$

In this example, the Fisher information matrix is $\mathbf{I}^* = \mathbb{E}_{\boldsymbol{\beta}^*}(b''(\boldsymbol{\beta}^{*T}\boldsymbol{Q}_i)\boldsymbol{Q}_i^{\otimes 2})$, and the partial Fisher information matrix is $I^*_{\theta|\boldsymbol{\gamma}} = \mathbb{E}_{\boldsymbol{\beta}^*}(b''(\boldsymbol{\beta}^{*T}\boldsymbol{Q}_i)Z_i(Z_i - \mathbf{w}^{*T}\boldsymbol{X}_i))$. The corresponding estimators are

$$\widehat{\mathbf{I}} = \frac{1}{n}\sum_{i=1}^{n} b''(\widehat{\boldsymbol{\beta}}^T\boldsymbol{Q}_i)\boldsymbol{Q}_i^{\otimes 2}, \quad \text{and} \quad \widehat{I}_{\theta|\boldsymbol{\gamma}} = \frac{1}{n}\sum_{i=1}^{n} b''(\widehat{\boldsymbol{\beta}}^T\boldsymbol{Q}_i)Z_i^2 - \widehat{\mathbf{w}}^T\left(\frac{1}{n}\sum_{i=1}^{n} b''(\widehat{\boldsymbol{\beta}}^T\boldsymbol{Q}_i)\boldsymbol{X}_iZ_i\right).$$

Thus, the score test statistic is given by $\widehat{U}_n = n^{1/2}\widehat{S}(0, \widehat{\boldsymbol{\gamma}})\widehat{I}_{\theta|\boldsymbol{\gamma}}^{-1/2}$. To study the limiting property of $\widehat{U}_n$, we consider the following regularity conditions for the generalized linear model.

**Assumption 5.1.** Assume that (1) $\lambda_{\min}(\mathbf{I}^*) \geq \kappa^2$ for some constant $\kappa > 0$, (2) $S = \operatorname{supp}(\boldsymbol{\beta}^*)$ and $S' = \operatorname{supp}(\mathbf{w}^*)$ satisfy $|S| = s^*$ and $|S'| = s'$, (3) $\max_{1 \leq i \leq n} \|\boldsymbol{Q}_i\|_\infty \leq K$, $\max_{1 \leq i \leq n} |\mathbf{w}^{*T}\boldsymbol{X}_i| \leq K$, $\max_{1 \leq i \leq n} |\boldsymbol{\beta}^{*T}\boldsymbol{Q}_i| \leq K$ and $\max_{1 \leq i \leq n} |Y_i - b'(\boldsymbol{Q}_i^T\boldsymbol{\beta}^*)| \leq K'$, for some $K$ and $K'$, (4) for any $t, t_1, t_2 \in [-2K, 2K]$, $0 < C' \leq |b''(t)| \leq C$ and $|b''(t_1) - b''(t_2)| \leq C|t_1 - t_2|$, where $C$ and $C'$ are some constants.

Note that conditions (1) and (2) in Assumption 5.1 are similar to conditions (1) and (2) in Theorem 4.1. For technical simplicity, in the condition (3), we directly assume the bounds for the covariates, the projection of the covariates, the covariate effect and the noise. This condition can be further relaxed to the sub-Gaussian type conditions considered in Theorem 4.1. Finally, the condition (4) characterizes the nonlinearity of the generalized linear model. In particular, we assume that the function $b''(\cdot)$ is bounded away from 0 and infinity and also Lipschitz in a closed interval. This condition holds for a large class of generalized linear models including the logistic regression and Poisson regression.

**Assumption 5.2.** Assume that

$$||\widehat{\boldsymbol{\beta}} - \boldsymbol{\beta}^*||_1 \leq C' s^* \lambda, \quad ||\widehat{\boldsymbol{\beta}} - \boldsymbol{\beta}^*||_2 \leq C' \sqrt{s^*} \lambda, \quad \text{and} \quad (\widehat{\boldsymbol{\beta}} - \boldsymbol{\beta}^*)^T \mathbf{H}_Q (\widehat{\boldsymbol{\beta}} - \boldsymbol{\beta}^*) \leq C' s^* \lambda^2,$$

where $\mathbf{H}_Q = \frac{1}{n} \sum_{i=1}^n \boldsymbol{Q}_i^{\otimes 2}$, and $C'$ is a positive constant.

With the Lasso penalty, this assumption can be verified by modifying the argument in Bickel et al. (2009). With the nonconvex penalty, we refer the details to Wang et al. (2013b). Bases on these assumptions, we establish the asymptotic null distribution of the score test statistic $\widehat{U}_n$, in the following theorem.

**Theorem 5.3.** Assume that the Assumptions 5.1 and 5.2 hold. With $\lambda \asymp \sqrt{\frac{\log d}{n}}$ and $\lambda' \asymp \sqrt{\frac{\log d}{n}}$, if $K = \mathcal{O}(1)$, $K' = \mathcal{O}(1)$, and $n^{-1/2}(s' \vee s^*) \log d = o(1)$, then under the null hypothesis, for each $t \in \mathbb{R}$, $\lim_{n \to \infty} |\mathbb{P}_{\boldsymbol{\beta}^*}(\widehat{U}_n \leq t) - \Phi(t)| = 0$.

*Proof.* A detailed proof is shown in Appendix C. □

Similarly, we can use alternative estimators of $\mathbf{w}^*$ to construct the decorrelated score function. For instance, the estimators $\widetilde{\mathbf{w}}$ in (2.7) and $\bar{\mathbf{w}}$ in (2.8) with the $L_1$ penalty, are equivalent to the following weighted Lasso estimators,

$$\widetilde{\mathbf{w}} = \underset{\mathbf{w}}{\operatorname{argmin}} \left\{ \frac{1}{n} \sum_{i=1}^n b''(\widehat{\boldsymbol{\beta}}^T \boldsymbol{Q}_i)(Z_i - \mathbf{w}^T \boldsymbol{X}_i)^2 + \lambda' ||\mathbf{w}||_1 \right\}. \tag{5.2}$$

$$\bar{\mathbf{w}} = \underset{\mathbf{w}}{\operatorname{argmin}} \left\{ \frac{1}{n} \sum_{i=1}^n (Y_i - b'(\widehat{\boldsymbol{\beta}}^T \boldsymbol{Q}_i))^2 (Z_i - \mathbf{w}^T \boldsymbol{X}_i)^2 + \lambda' ||\mathbf{w}||_1 \right\}. \tag{5.3}$$

The following two theorems establish the theoretical results for the decorrelated score test statistic with $\widetilde{\mathbf{w}}$ and $\bar{\mathbf{w}}$.

**Theorem 5.4.** Assume that the Assumptions 5.1 and 5.2 hold. With $\lambda \asymp \sqrt{\frac{\log d}{n}}$ and $\lambda' \asymp \sqrt{\frac{\log d}{n}}$, it also holds that $K = \mathcal{O}(1)$, $K' = \mathcal{O}(1)$, and $n^{-1/2}(s' \vee s^*) \log d = o(1)$. Then under the null hypothesis, the decorrelated score test statistic $\widehat{U}_n$ with $\widehat{\mathbf{w}}$ replaced by $\widetilde{\mathbf{w}}$ in (5.2) satisfies, $\lim_{n \to \infty} |\mathbb{P}_{\boldsymbol{\beta}^*}(\widehat{U}_n \leq t) - \Phi(t)| = 0$, for each $t \in \mathbb{R}$.

*Proof.* A detailed proof is shown in Appendix C. □

**Theorem 5.5.** Assume that the Assumptions 5.1 and 5.2 hold. In addition, $\max_{1\leq i\leq n} |Y_i - b'(\boldsymbol{\beta}^{*T}\boldsymbol{Q}_i)| \geq C > 0$ for some constant $C$. With $\lambda \asymp \sqrt{\frac{\log d}{n}}$ and $\lambda' \asymp \sqrt{\frac{\log d}{n}}$, it also holds that $K = \mathcal{O}(1)$, $K' = \mathcal{O}(1)$, and $n^{-1/2}(s' \vee s^*)\log d = o(1)$. Then under the null hypothesis, the decorrelated score test statistic $\widehat{U}_n$ with $\widehat{\mathbf{w}}$ replaced by $\bar{\mathbf{w}}$ in (5.3) satisfies, $\lim_{n\to\infty} |\mathbb{P}_{\boldsymbol{\beta}^*}(\widehat{U}_n \leq t) - \Phi(t)| = 0$, for each $t \in \mathbb{R}$.

*Proof.* A detailed proof is shown in Appendix C. □

To compare the conditions and results for the decorrelated score test based on different estimators of $\mathbf{w}^*$, we find that the assumption on the growth of $s^*, s', d$ and $n$ in Theorems 5.3, 5.4, and 5.5 are identical, which is also same as that for the linear regression in Theorem 4.1. Moreover, the scaling we derived agrees with the best existing results for the generalized linear model obtained by van de Geer et al. (2014). Note that, since the score function is nonlinear for parameters, to get the desired scaling in Theorems 5.4 and 5.5, we conduct a refined analysis, by modifying the proof of our master Theorem 3.5. The details can be found in the proofs of Theorems 5.4 and 5.5 in Appendix C. In terms of the conditions, we need to assume $|Y_i - b'(\boldsymbol{\beta}^{*T}\boldsymbol{Q}_i)|^2$ is away from 0 in Theorem 5.5. This condition is used to lower bound the weights in the Lasso type estimator (5.3). As a corollary of Theorems 5.3, 5.4 and 5.5, we can show that, under the same conditions, the one-step estimator is also asymptotically normal.

**Corollary 5.6.** Assume that the Assumptions 5.1 and 5.2 hold. In addition, for any $i = 1, ..., n$, $|Y_i - b'(\boldsymbol{\beta}^{*T}\boldsymbol{Q}_i)| \geq C > 0$ for some constant $C$. With $\lambda \asymp \sqrt{\frac{\log d}{n}}$ and $\lambda' \asymp \sqrt{\frac{\log d}{n}}$, if $K = \mathcal{O}(1)$, $K' = \mathcal{O}(1)$, and $n^{-1/2}(s' \vee s^*)\log d = o(1)$, then we have $n^{1/2}(\widetilde{\theta} - \theta^*)\widehat{I}_{\theta|\gamma}^{1/2} \rightsquigarrow N(0,1)$, where $\widetilde{\theta}$ is given by equation (2.9) with either $\widehat{\mathbf{w}}$ in (5.1), $\widetilde{\mathbf{w}}$ in (5.2) or $\bar{\mathbf{w}}$ in (5.3).

*Proof.* A detailed proof is shown in Appendix C. □

## 6 Extensions of High Dimensional Score Test

In this section, we consider several important extensions of the score test. First, we extend the score test to handle high dimensional parameters of interest. Second, we study the properties of the score test under misspecified model. Finally, we propose the generalized score test based on the loss function other than the negative log-likelihood.

### 6.1 Score Test for High Dimensional Null Hypothesis

In this section, we assume that the parameter of interest $\boldsymbol{\theta}$ is of dimension $d_0$, say $\boldsymbol{\theta} = (\theta_1, ..., \theta_{d_0})$ and the nuisance parameter $\boldsymbol{\gamma}$ is of dimension $d_1 = d - d_0$. We are interested in testing the $d_0$ dimensional null hypothesis $H_0^{d_0} : \boldsymbol{\theta}^* = (0, ..., 0)$, where $d_0$ can increase with $n$. This result can be useful for testing the regression coefficients, which have the group structure.

Similar to the test for one dimensional hypothesis, the building block for the high dimensional test remains the decorrelated score function. Recall that the $d_0$ dimensional decorrelated score function for $\boldsymbol{\theta}$ is

$$\mathbf{S}(\boldsymbol{\theta}, \boldsymbol{\gamma}) = \nabla_\theta \ell(\boldsymbol{\theta}, \boldsymbol{\gamma}) - \mathbf{W}^T \nabla_\gamma \ell(\boldsymbol{\theta}, \boldsymbol{\gamma}),$$

where $\mathbf{W}^T = \mathbf{I}_{\theta\gamma} \mathbf{I}_{\gamma\gamma}^{-1} \in \mathbb{R}^{d_0 \times d_1}$. Similar to Section 2, we can estimate $\mathbf{W}$ by the Dantzig type estimator $\widehat{\mathbf{W}}$. In particular, $\widehat{\mathbf{W}}$ can be solved column-wisely, i.e., $\widehat{\mathbf{W}} = (\widehat{\mathbf{W}}_{*1}, ..., \widehat{\mathbf{W}}_{*d_0})$, where

$$\widehat{\mathbf{W}}_{*j} = \operatorname{argmin} ||\mathbf{w}||_1, \quad \text{s.t.} \quad ||\nabla^2_{\theta_j \gamma} \ell(\widehat{\boldsymbol{\beta}}) - \mathbf{w}^T \nabla^2_{\gamma\gamma} \ell(\widehat{\boldsymbol{\beta}})||_\infty \leq \lambda', \tag{6.1}$$

and $\lambda'$ is a tuning parameter, for $j = 1, ..., d_0$. Given the penalized M-estimator $\widehat{\boldsymbol{\beta}}$ in (1.1) and $\widehat{\mathbf{W}}$, the estimated decorrelated score function under the null hypothesis $H_0^{d_0} : \boldsymbol{\theta}^* = (0, ..., 0)$ is,

$$\widehat{\mathbf{S}}(\mathbf{0}, \widehat{\boldsymbol{\gamma}}) = \nabla_\theta \ell(\mathbf{0}, \widehat{\boldsymbol{\gamma}}) - \widehat{\mathbf{W}}^T \nabla_\gamma \ell(\mathbf{0}, \widehat{\boldsymbol{\gamma}}).$$

As shown in Section 3, under the null hypothesis, we expect that each component of the score function $\widehat{\mathbf{S}}(\mathbf{0}, \widehat{\boldsymbol{\gamma}})$ converges weakly to a mean 0 Gaussian random variable. To combine these score functions for testing the null hypothesis, we consider their extreme values, i.e., $\|\widehat{\mathbf{S}}(\mathbf{0}, \widehat{\boldsymbol{\gamma}})\|_\infty$. In particular, define the test statistic as $\|\widehat{\mathbf{T}}\|_\infty$, where $\widehat{\mathbf{T}} = \sqrt{n} \widehat{\mathbf{S}}(\mathbf{0}, \widehat{\boldsymbol{\gamma}})$. In Lemma D.1 of Appendix D, we show that the distribution of $\|\widehat{\mathbf{T}}\|_\infty$ can be approximated by that of $\|\mathbf{N}\|_\infty$, where $\mathbf{N}$ is a $d_0$ dimensional multivariate normal random variable. However, the distribution of $\mathbf{N}$ still depends on the unknown parameters. Following the pioneering work by Chernozhukov et al. (2013), we consider a multiplier bootstrap approach to directly approximate the distribution of $\|\widehat{\mathbf{T}}\|_\infty$. Specifically, consider the following multiplier bootstrapped statistic

$$\widehat{\mathbf{N}}_e = \frac{1}{\sqrt{n}} \sum_{i=1}^n e_i (\nabla_\theta \ell_i(\mathbf{0}, \widehat{\boldsymbol{\gamma}}) - \widehat{\mathbf{W}}^T \nabla_\gamma \ell_i(\mathbf{0}, \widehat{\boldsymbol{\gamma}})),$$

where $e_1, ..., e_n$ are i.i.d samples from $N(0, 1)$ and independent of the data. Denote the $\alpha$-quantile of $\|\widehat{\mathbf{N}}_e\|_\infty$ by

$$c'_N(\alpha) = \inf\{t \in \mathbb{R} : \mathbb{P}_e(\|\widehat{\mathbf{N}}_e\|_\infty \leq t) \geq \alpha\}, \tag{6.2}$$

where $\mathbb{P}_e(\mathcal{A})$ denotes the probability of the event $\mathcal{A}$ with respect to $e_1, ..., e_n$. The null hypothesis is rejected if and only if $\|\widehat{\mathbf{T}}\|_\infty \geq c'_N(\alpha)$, where $\widehat{\mathbf{T}} = \sqrt{n} \widehat{\mathbf{S}}(\mathbf{0}, \widehat{\boldsymbol{\gamma}})$. Compared with the standard resampling based methods, the multiplier bootstrap avoids the estimation of $\boldsymbol{\beta}$ and $\mathbf{W}$ based on resampled data and is computationally much more efficient. To justify the validity of the multiplier bootstrap, we consider the following assumptions.

**Assumption 6.1** (Estimation Error Bound). Assume that

$$\lim_{n \to \infty} \mathbb{P}_{\boldsymbol{\beta}^*}\big(\|\widehat{\boldsymbol{\gamma}} - \boldsymbol{\gamma}^*\|_1 \lesssim \eta_1(n)\big) = 1 \quad \text{and} \quad \lim_{n \to \infty} \mathbb{P}_{\boldsymbol{\beta}^*}\big(\max_{1 \leq j \leq d_0} \|\widehat{\mathbf{W}}_{*j} - \mathbf{W}^*_{*j}\|_1 \lesssim \eta_2(n)\big) = 1,$$

where $\eta_1(n)$ and $\eta_2(n)$ converge to 0, as $n \to \infty$.

**Assumption 6.2** (Noise Condition). Assume that $\lim_{n \to \infty} \mathbb{P}_{\boldsymbol{\beta}^*}\big(\|\nabla_\gamma \ell(\mathbf{0}, \boldsymbol{\gamma}^*)\|_\infty \lesssim \eta_3(n)\big) = 1$, for some $\eta_3(n) \to 0$, as $n \to \infty$.

**Assumption 6.3** (Stability Condition). For $\gamma_v = v\gamma^* + (1-v)\widehat{\gamma}$ with $v \in [0,1]$,

$$\lim_{n\to\infty} \mathbb{P}_{\boldsymbol{\beta}^*}\Big(\max_{1\leq j\leq d_0} \sup_{v\in[0,1]} \|\nabla^2_{\theta_j\gamma}\ell(\mathbf{0},\gamma_v) - \widehat{\mathbf{W}}^T_{*j}\nabla^2_{\gamma\gamma}\ell(\mathbf{0},\gamma_v)\|_\infty \lesssim \eta_4(n)\Big) = 1,$$

for some $\eta_4(n) \to 0$, as $n \to \infty$.

**Assumption 6.4** (Tail Condition). Denote $S_{ij} = \nabla_{\theta_j}\ell_i(\mathbf{0},\gamma^*) - \mathbf{W}^{*T}_{*j}\nabla_\gamma\ell_i(\mathbf{0},\gamma^*)$, where $i = 1,...,n$ and $j = 1,...,d_0$. Assume that for any $j = 1,...,d_0$, $S_{ij}$ is sub-exponential with $\|S_{ij}\|_{\psi_1} \leq C$, and $\mathbb{E}_{\boldsymbol{\beta}^*}(S^2_{ij}) \geq C_{\min}$, where $C_{\min}$ is a positive constant.

**Assumption 6.5.** Denote $\widehat{S}_{ij} = \nabla_{\theta_j}\ell_i(\mathbf{0},\widehat{\gamma}) - \widehat{\mathbf{W}}^T_{*j}\nabla_\gamma\ell_i(\mathbf{0},\widehat{\gamma})$, where $i = 1,...,n$ and $j = 1,...,d_0$. Assume that

$$\lim_{n\to\infty} \mathbb{P}_{\boldsymbol{\beta}^*}\left(\max_{1\leq j\leq d_0} \sqrt{\frac{1}{n}\sum_{i=1}^n (\widehat{S}_{ij} - S_{ij})^2} \lesssim \eta_7(n)\right) = 1,$$

where $\eta_7(n) \to 0$, as $n \to \infty$.

Assumptions 6.1-6.3 are the natural extensions of Assumptions 3.1-3.3 for testing one dimensional null hypothesis. For instance, Assumption 3.1 specifies the $L_1$-convergence rate of the vector $\widehat{\mathbf{w}}$, whereas Assumption 6.1 specifies the convergence rate of $\widehat{\mathbf{W}}$ in terms of the matrix $L_1$ norm. Assumptions 6.4 and 6.5 are the technical assumptions needed for studying the multiplier bootstrap (Chernozhukov et al., 2013). For simplicity, we assume that $S_{ij}$ in Assumption 6.4 is sub-exponential with finite $\psi_1$ norm. This assumption can be relaxed to the moment conditions; see Chernozhukov et al. (2013). The following main theorem of this section provides theoretical guarantees for the multiplier bootstrap, even if $d_0$ has the same order as $d$.

**Theorem 6.6.** Assume that Assumptions 6.1–6.5, and

$$(q(n) \vee q'(n))\left(1 \vee \log\frac{d_0}{q(n) \vee q'(n)}\right)^{1/2} = o(1), \tag{6.3}$$

hold, where $q(n) = n^{1/2}(\eta_2(n)\eta_3(n) \vee \eta_1(n)\eta_4(n))$ and $q'(n) = \eta_7(n)\sqrt{\log d_0}$. If $(\log(d_0 n))^9/n = o(1)$, we obtain

$$\lim_{n\to\infty} \sup_{\alpha\in(0,1)} \left|\mathbb{P}_{\boldsymbol{\beta}^*}\big(\|\widehat{\mathbf{T}}\|_\infty \leq c'_N(\alpha)\big) - \alpha\right| = 0,$$

where $c'_N(\alpha)$ is defined in (6.2).

*Proof.* A detailed proof is shown in Appendix D. □

**Remark 6.7.** This theorem establishes the validity of multiplier bootstrap for the test statistic $\|\sqrt{n}\widehat{\mathbf{S}}(\mathbf{0},\widehat{\gamma})\|_\infty$. Compared to the score test statistic, we do not standardize the score function by the Fisher information. Hence, the assumption similar to 3.6 is not necessary. In practice, the standardization leads to a rescaled statistic and can improve the performance of the test. In particular, denote $\mathbf{I}^*_{\theta|\gamma} = \mathbf{I}^*_{\theta\theta} - \mathbf{I}^*_{\theta\gamma}\mathbf{I}^{*-1}_{\gamma\gamma}\mathbf{I}^*_{\gamma\theta}$, and $\mathbf{D}^* = \text{diag}((\mathbf{I}^*_{\theta|\gamma})_{11},...,(\mathbf{I}^*_{\theta|\gamma})_{d_0 d_0})$, which is a $d_0 \times d_0$ diagonal matrix. An estimator of $\mathbf{D}^*$ is given by $\widehat{\mathbf{D}} = \text{diag}((\widehat{\mathbf{I}}_{\theta|\gamma})_{11},...,(\widehat{\mathbf{I}}_{\theta|\gamma})_{d_0 d_0})$, where

$\widehat{\mathbf{I}}_{\boldsymbol{\theta}|\boldsymbol{\gamma}} = \nabla^2_{\boldsymbol{\theta}\boldsymbol{\theta}}\ell(\widehat{\boldsymbol{\beta}}) - \widehat{\mathbf{W}}^T \nabla^2_{\boldsymbol{\gamma}\boldsymbol{\theta}}\ell(\widehat{\boldsymbol{\beta}})$. Denote $\widehat{\mathbf{T}}_R = \sqrt{n}\widehat{\mathbf{D}}^{-1/2}\widehat{\mathbf{S}}(\mathbf{0}, \widehat{\boldsymbol{\gamma}})$. Similar to Theorem 6.6, the theoretical properties of the rescaled test statistic $\|\widehat{\mathbf{T}}_R\|_\infty$ are established. The detailed results are presented in Appendix D.

As an illustration of Theorem 6.6, we now consider the linear regression as a concrete example. Assume that $Y_i = \boldsymbol{\theta}^{*T}\mathbf{Z}_i + \boldsymbol{\gamma}^{*T}\mathbf{X}_i + \epsilon_i$, where $\mathbb{E}(\epsilon_i) = 0$, $\mathbb{E}(\epsilon_i^2) = \sigma^2$ and the covariate $\mathbf{Z}_i$ is $d_0$ dimensional, that is $\mathbf{Z}_i = (Z_{i1}, ..., Z_{id_0})$. Here, $\sigma^2$ is also unknown. Consider the null hypothesis $\boldsymbol{\theta}^* = \mathbf{0}$. To avoid the estimation uncertainty of $\sigma^2$, we eliminate $\sigma^2$ in the estimated decorrelated score function (4.2), and redefine it as,

$$\widehat{\mathbf{S}}(\mathbf{0}, \widehat{\boldsymbol{\gamma}}) = -\frac{1}{n}\sum_{i=1}^n (Y_i - \widehat{\boldsymbol{\gamma}}^T\mathbf{X}_i)(\mathbf{Z}_i - \widehat{\mathbf{W}}^T\mathbf{X}_i),$$

where

$$\widehat{\mathbf{W}}_{*j} = \operatorname{argmin} \|\mathbf{w}\|_1, \quad \text{s.t.} \quad \left\|\frac{1}{n}\sum_{i=1}^n \mathbf{X}_i\left(Z_{ij} - \mathbf{w}^T\mathbf{X}_i\right)\right\|_\infty \leq \lambda'.$$

In the linear model example, the multiplier bootstrapped statistic is

$$\widehat{\mathbf{N}}_e = \frac{1}{\sqrt{n}}\sum_{i=1}^n e_i(Y_i - \widehat{\boldsymbol{\gamma}}^T\mathbf{X}_i)(\mathbf{Z}_i - \widehat{\mathbf{W}}^T\mathbf{X}_i),$$

where $e_i \sim N(0,1)$. The following corollary characterizes the consequence of Theorem 6.6 for the linear regression.

**Corollary 6.8.** Assume that (1) $\lambda_{\min}(\mathbb{E}(\mathbf{Q}_i^{\otimes 2})) \geq 2\kappa$ for some constant $\kappa > 0$, (2) $\|\mathbf{w}^*\|_0 = s'$ and $\|\boldsymbol{\beta}^*\|_0 = s^*$, (3) $\epsilon_i$, $\mathbf{W}_{*j}^{*T}\mathbf{X}_i$, $Q_{ij}$ are all sub-Gaussian with $\|\epsilon_i\|_{\psi_2} \leq C$, $\|\mathbf{W}_{*j}^{*T}\mathbf{X}_i\|_{\psi_2} \leq C$ and $\|Q_{ij}\|_{\psi_2} \leq C$, where $C$ is a positive constant. In addition, assume that $\sigma^{*2} \geq C$. If $n^{-1/2}(s' \vee s^*)(\log(nd))^{3/2}\sqrt{\log d_0} = o(1)$, $(\log(d_0 n))^9/n = o(1)$ and $\lambda \asymp \lambda' \asymp \sqrt{\frac{\log d}{n}}$, then

$$\lim_{n\to\infty} \sup_{\alpha \in (0,1)} \left|\mathbb{P}_{\boldsymbol{\beta}^*}\left(\|\widehat{\mathbf{T}}\|_\infty \leq c'_N(\alpha)\right) - \alpha\right| = 0,$$

where $c'_N(\alpha)$ is defined in (6.2) and $\widehat{\mathbf{T}} = \sqrt{n}\widehat{\mathbf{S}}(\mathbf{0}, \widehat{\boldsymbol{\gamma}})$.

*Proof.* A detailed proof is shown in Appendix D. □

**Remark 6.9.** When $d_0$ is fixed, Corollary 6.8 holds if $n^{-1/2}(s' \vee s^*)(\log(nd))^{3/2} = o(1)$. Compared with $n^{-1/2}(s' \vee s^*)\log d = o(1)$ in Theorem 4.1, we need an extra $\log d$ factor to ensure the validity of the multiplier bootstrap. When $d_0$ has the same order as $d$, Corollary 6.8 holds if $\log(nd) = o(n^{1/9} \vee \frac{n^{1/4}}{(s'\vee s^*)^{1/2}})$. Moreover, in Corollary D.9 of Appendix D, we show that under the same conditions, the multiplier bootstrap works for the rescaled test statistic $\|\widehat{\mathbf{T}}_R\|_\infty$. Hence, the proposed method can be used for testing high dimensional null hypothesis.

## 6.2 Score Test with Model Misspecification

In the previous sections, an implicit assumption is that the probability model for $Y_i$ given the covariate $\boldsymbol{Q}_i$ is correctly specified. In this section, we establish the theoretical properties of the score test, if the true probability distribution denoted by $\mathbb{P}^*$ does not belong to the assumed statistical model $\mathcal{P} = \{\mathbb{P}_{\boldsymbol{\beta}}, \boldsymbol{\beta} \in \Omega\}$. To this end, define the Kullback-Leibler divergence as

$$\mathrm{KL}(\boldsymbol{\beta}) = \mathbb{E}^* \left\{ \log \frac{f^*(Y_i, \boldsymbol{Q}_i)}{f(Y_i, \boldsymbol{Q}_i; \boldsymbol{\beta})} \right\},$$

where $f^*(Y_i, \boldsymbol{Q}_i)$ is the true density function of $(Y_i, \boldsymbol{Q}_i)$, and $f(Y_i, \boldsymbol{Q}_i; \boldsymbol{\beta})$ is the density corresponding to the model $\mathbb{P}_{\boldsymbol{\beta}}$. Here, we use $\mathbb{P}^*(\cdot)$ and $\mathbb{E}^*(\cdot)$ to denote the probability and the expectation with respect to the true density function $f^*(Y_i, \boldsymbol{Q}_i)$. Let $\boldsymbol{\beta}^o$ denote the oracle parameter (or least false parameter) that minimizes the Kullback-Leibler divergence, i.e., $\boldsymbol{\beta}^o = \mathrm{argmin}_{\boldsymbol{\beta}} \mathrm{KL}(\boldsymbol{\beta})$, where $\boldsymbol{\beta}^o = (\theta^o, \boldsymbol{\gamma}^o)$. Note that, if the model is correctly specified, we have $f^*(Y_i, \boldsymbol{Q}_i) = f(Y_i, \boldsymbol{Q}_i; \boldsymbol{\beta}^*)$ and the oracle parameter reduces to $\boldsymbol{\beta}^*$. Although, under the misspecified model, the true distribution is not estimable, it is often of interest to understand the behavior of the oracle parameter. In particular, assume that the inferential problem can be formulated as testing $H_0^o : \theta^o = 0$ versus $H_1^o : \theta^o \neq 0$. Similar to the previous sections, we define $\mathbf{I}^o = \mathbb{E}^*(\nabla^2 \ell(\boldsymbol{\beta}^o))$, and $\mathbf{w}^{oT} = \mathbf{I}^o_{\theta\gamma} \mathbf{I}^{o-1}_{\gamma\gamma}$. To study the properties of the estimated decorrelated score function $\widehat{S}(0, \widehat{\boldsymbol{\gamma}})$ in (2.6) under model misspecification, the following assumptions are imposed.

**Assumption 6.10** (Estimation Error Bound). Assume that

$$\lim_{n \to \infty} \mathbb{P}^* \left( \|\widehat{\boldsymbol{\gamma}} - \boldsymbol{\gamma}^o\|_1 \lesssim \eta_1(n) \right) = 1 \quad \text{and} \quad \lim_{n \to \infty} \mathbb{P}^* \left( \|\widehat{\mathbf{w}} - \mathbf{w}^o\|_1 \lesssim \eta_2(n) \right) = 1,$$

where $\eta_1(n)$ and $\eta_2(n)$ converge to 0, as $n \to \infty$.

**Assumption 6.11** (Noise Condition). Assume that $\lim_{n \to \infty} \mathbb{P}^* \left( \|\nabla_{\boldsymbol{\gamma}} \ell(0, \boldsymbol{\gamma}^o)\|_\infty \lesssim \eta_3(n) \right) = 1$, for some $\eta_3(n) \to 0$, as $n \to \infty$.

**Assumption 6.12** (Stability Condition). For $\boldsymbol{\gamma}_v = v \boldsymbol{\gamma}^o + (1-v) \widehat{\boldsymbol{\gamma}}$ with $v \in [0,1]$,

$$\lim_{n \to \infty} \mathbb{P}^* \left( \sup_{v \in [0,1]} \|\nabla^2_{\theta\gamma} \ell(0, \boldsymbol{\gamma}_v) - \widehat{\mathbf{w}}^T \nabla^2_{\gamma\gamma} \ell(0, \boldsymbol{\gamma}_v)\|_\infty \lesssim \eta_4(n) \right) = 1,$$

for some $\eta_4(n) \to 0$, as $n \to \infty$.

**Assumption 6.13** (CLT). Let $\boldsymbol{\Sigma}^o = \mathbb{E}^*(\nabla \ell_i^{\otimes 2}(0, \boldsymbol{\gamma}^o))$. For $\mathbf{v}^o = (1, -\mathbf{w}^{oT})^T$, it holds that

$$\frac{\sqrt{n} \mathbf{v}^{oT} \nabla \ell(0, \boldsymbol{\gamma}^o)}{\sqrt{\mathbf{v}^{oT} \boldsymbol{\Sigma}^o \mathbf{v}^o}} \rightsquigarrow N(0, 1).$$

Assume that $C' \leq \mathbf{v}^{oT} \boldsymbol{\Sigma}^o \mathbf{v}^o < \infty$, where $C' > 0$ is a constant.

**Assumption 6.14** (Estimator of $\boldsymbol{\Sigma}^o$). There exists an estimator $\widehat{\boldsymbol{\Sigma}}$ such that $\lim_{n \to \infty} \mathbb{P}^* \left( \|\widehat{\boldsymbol{\Sigma}} - \boldsymbol{\Sigma}^o\|_{\max} \lesssim \eta_7(n) \right) = 1$, for some $\eta_7(n) \to 0$, as $n \to \infty$.

Note that the Assumptions 6.10–6.13 are essentially identical to Assumptions 3.1–3.4 with $\gamma^*$ replaced by the oracle parameter $\gamma^o$. Since the variance of score function is given by $\Sigma^o$ in Assumption 6.13, we assume that there exists an estimator with theoretical guarantees described in Assumption 6.14. The theoretical properties of the decorrelated score function under misspecified models is shown in the following theorem.

**Theorem 6.15.** Under the Assumptions 6.10–6.13, we have that with probability tending to one

$$n^{1/2}|\widehat{S}(0,\widehat{\gamma}) - S(0,\gamma^o)| \lesssim n^{1/2}\big(\eta_2(n)\eta_3(n) + \eta_1(n)\eta_4(n)\big). \tag{6.4}$$

If $n^{1/2}(\eta_2(n)\eta_3(n) + \eta_1(n)\eta_4(n)) = o(1)$, we have

$$n^{1/2}\widehat{S}(0,\widehat{\gamma})/\sqrt{\mathbf{v}^{oT}\Sigma^o\mathbf{v}^o} \rightsquigarrow N(0,1), \tag{6.5}$$

where $\mathbf{v}^o = (1, -\mathbf{w}^{oT})^T$.

*Proof.* A detailed proof is shown in Appendix E. □

As seen in Theorem 6.15, in the misspecified model, we need to standardize the score function $\widehat{S}(0,\widehat{\gamma})$ by $\sqrt{\mathbf{v}^{oT}\Sigma^o\mathbf{v}^o}$, which is different from the factor $I^{*1/2}_{\theta|\gamma}$ shown in Theorem 3.5 for the correctly specified model. This is one of the main consequences of the model misspecification. Thus, we need to redefine the decorrelated score test statistic as $\widehat{U}_n^o = n^{1/2}\widehat{S}(0,\widehat{\gamma})/\sqrt{\widehat{\mathbf{v}}^T\widehat{\Sigma}\widehat{\mathbf{v}}}$, where $\widehat{\mathbf{v}} = (1, -\widehat{\mathbf{w}}^T)^T$. The following corollary establishes the asymptotic distribution of $\widehat{U}_n^o$ under the null hypothesis $H_0 : \theta^* = 0$.

**Corollary 6.16.** Assume that the Assumptions 6.10–6.14 hold. It also holds that $\|\Sigma^o\|_{\max} = \mathcal{O}(1)$, $\|\mathbf{w}^o\|_1^2 \eta_7(n) = o(1)$, $\|\Sigma^o\mathbf{v}^o\|_\infty \eta_2(n) = \mathcal{O}_\mathbb{P}(1)$, and $n^{1/2}(\eta_2(n)\eta_3(n) + \eta_1(n)\eta_4(n)) = o(1)$. Under $H_0 : \theta^o = 0$, we have for any $t \in \mathbb{R}$

$$\lim_{n\to\infty}|\mathbb{P}^*(\widehat{U}_n^o \le t) - \Phi(t)| = 0.$$

*Proof.* A detailed proof is shown in Appendix E. □

As an illustration of the general results in Theorem 6.15 and Corollary 6.16, we now consider the linear regression under model misspecification. Since the linear model assumption is no longer true, we cannot use the simple identity $\epsilon_i = Y_i - \boldsymbol{\beta}^{*T}\mathbf{Q}_i$. Hence, some of the technical details will be different from the correctly specified model considered in Section 4.

Assume that $\widehat{\boldsymbol{\beta}}$ is the Lasso estimator in (4.1). By definition, the oracle parameter $\boldsymbol{\beta}^o$ is defined as $\boldsymbol{\beta}^o = \operatorname{argmin}_{\boldsymbol{\beta}} \mathbb{E}^*(Y_i - \boldsymbol{\beta}^T\mathbf{Q}_i)^2$, and the decorrelated score function for testing $\theta^o = 0$ is

$$\widehat{S}(0,\widehat{\gamma}) = -\frac{1}{n}\sum_{i=1}^{n}(Y_i - \widehat{\gamma}^T\mathbf{X}_i)(Z_i - \widehat{\mathbf{w}}^T\mathbf{X}_i).$$

Without loss of generality, we ignore $\sigma$ in the log-likelihood function, because $\sigma$ serves as a scaling factor and can be canceled in the score test statistic. By definition, $\Sigma^o = \mathbb{E}^*(\mathbf{Q}_i^{\otimes 2}(Y_i - \boldsymbol{\beta}^{oT}\mathbf{Q}_i)^2)$, which can be estimated by

$$\widehat{\Sigma} = \frac{1}{n}\sum_{i=1}^{n}\mathbf{Q}_i^{\otimes 2}(Y_i - \widehat{\boldsymbol{\beta}}^T\mathbf{Q}_i)^2.$$

Thus, the score test statistic is given by $\widehat{U}_n^o = n^{1/2}\widehat{S}(0,\widehat{\gamma})/\sqrt{\widehat{\mathbf{v}}^T\widehat{\boldsymbol{\Sigma}}\widehat{\mathbf{v}}}$, where $\widehat{\mathbf{v}} = (1, -\widehat{\mathbf{w}}^T)^T$ and $\widehat{\mathbf{w}}$ is defined in (4.3). We can obtain the following corollary.

**Corollary 6.17.** Assume that (1) $\lambda_{\min}(\mathbb{E}^*(\mathbf{Q}_i^{\otimes 2})) \geq 2\kappa$ and $\lambda_{\min}(\boldsymbol{\Sigma}^o) \geq 2\kappa$ for some constant $\kappa > 0$, (2) $\|\mathbf{w}^o\|_0 = s'$ and $\|\boldsymbol{\beta}^o\|_0 = s^*$, (3) $Y_i - \boldsymbol{\gamma}^{oT}\mathbf{X}_i$, $\mathbf{w}^{oT}\mathbf{X}_i$, $Q_{ij}$ are all sub-Gaussian with $\|Y_i - \boldsymbol{\gamma}^{oT}\mathbf{X}_i\|_{\psi_2} \leq C$, $\|\mathbf{w}^{oT}\mathbf{X}_i\|_{\psi_2} \leq C$ and $\|Q_{ij}\|_{\psi_2} \leq C$, where $C$ is a positive constant. If $n^{-1}s^*(\log(nd))^5 = o(1)$, $n^{-1/2}(s' \vee s^*)\log d = o(1)$ and $\lambda \asymp \lambda' \asymp \sqrt{\frac{\log d}{n}}$, then under $H_0^o: \theta^o = 0$, for each $t \in \mathbb{R}$,
$$\lim_{n\to\infty} |\mathbb{P}^*(\widehat{U}_n^o \leq t) - \Phi(t)| = 0.$$

*Proof.* A detailed proof is shown in Appendix E. □

**Remark 6.18.** To further understand the consequences of model misspecification under high dimensional setting, it is of interest to compare Corollary 6.17 with Theorem 4.1. As expected, the conditions in Corollary 6.17 for the misspecified model is stronger. Specifically, in condition (1) of Corollary 6.17, we also need the bound for the minimal eigenvalue of $\boldsymbol{\Sigma}^o$, since it corresponds to the variance of the score function. However, we find that the scaling assumptions in Corollary 6.17 for $s^*, s', d$ and $n$ are identical to those in Theorem 4.1 for the correctly specified linear model. To the best of our knowledge, this paper for the first time derived the sharp theoretical properties for the misspecified linear models.

### 6.3 Generalized Score Test

In the previous sections, we assume that $\ell(\boldsymbol{\beta})$ corresponds to the negative log-likelihood. However, in many cases, the inference based on the likelihood may be infeasible or undesirable. Instead, there may exist some convenient loss function $\ell(\boldsymbol{\beta})$ satisfying $\boldsymbol{\beta}^* = \arg\min_{\boldsymbol{\beta}} \mathbb{E}_{\boldsymbol{\beta}^*}(\ell(\boldsymbol{\beta}))$; see van de Geer et al. (2012). In this section, we consider the score test based on the general loss function. For notational simplicity, here, we use $\ell(\theta)$ to denote a general loss function.

Similar to that in Algorithm 1, the generalized decorrelated score function is defined as
$$\widehat{S}(\theta, \widehat{\boldsymbol{\gamma}}) = \nabla_\theta \ell(\theta, \widehat{\boldsymbol{\gamma}}) - \widehat{\mathbf{w}}^T \nabla_{\boldsymbol{\gamma}} \ell(\theta, \widehat{\boldsymbol{\gamma}}),$$

where $\widehat{\mathbf{w}}$ is either obtained by (2.5) or (2.7). Define $\mathbf{w}^* = \mathbf{I}_{\gamma\gamma}^{*-1}\mathbf{I}_{\gamma\theta}^*$, where $\mathbf{I}^* = \mathbb{E}_{\boldsymbol{\beta}^*}(\nabla^2\ell(0, \boldsymbol{\gamma}^*))$. Assume the following assumptions hold.

**Assumption 6.19** (CLT). Let $\boldsymbol{\Sigma}^* = \mathbb{E}_{\boldsymbol{\beta}^*}(\nabla\ell_i^{\otimes 2}(0, \boldsymbol{\gamma}^*))$. For $\mathbf{v}^* = (1, -\mathbf{w}^{*T})^T$, it holds that
$$\frac{\sqrt{n}\mathbf{v}^{*T}\nabla\ell(0, \boldsymbol{\gamma}^*)}{\sqrt{\mathbf{v}^{*T}\boldsymbol{\Sigma}^*\mathbf{v}^*}} \rightsquigarrow N(0, 1).$$

Assume that $C' \leq \mathbf{v}^{*T}\boldsymbol{\Sigma}^*\mathbf{v}^* < \infty$, where $C' > 0$ is a constant.

**Assumption 6.20.** Assume that there exists an estimator $\widehat{\boldsymbol{\Sigma}}$ such that $\lim_{n\to\infty} \mathbb{P}_{\boldsymbol{\beta}^*}(\|\widehat{\boldsymbol{\Sigma}} - \boldsymbol{\Sigma}^*\|_{\max} \lesssim \eta_7(n)) = 1$, for some $\eta_7(n) \to 0$, as $n \to \infty$.

Define the decorrelated score test statistic as $\widehat{U}_n = n^{1/2}\widehat{S}(0,\widehat{\boldsymbol{\gamma}})/\sqrt{\widehat{\mathbf{v}}^T\widehat{\boldsymbol{\Sigma}}\widehat{\mathbf{v}}}$. The main result of this section establishes the asymptotic distribution of $\widehat{U}_n$ under the null hypothesis $H_0 : \theta^* = 0$.

**Theorem 6.21.** Assume that the Assumptions 3.1–3.3, 6.19, and 6.20 hold. It also holds that $\|\boldsymbol{\Sigma}^*\|_{\max} = \mathcal{O}(1)$, $\|\mathbf{w}^*\|_1^2 \eta_7(n) = o(1)$, $\|\boldsymbol{\Sigma}^*\mathbf{v}^*\|_\infty \eta_2(n) = o_\mathbb{P}(1)$, and $n^{1/2}(\eta_2(n)\eta_3(n) + \eta_1(n)\eta_4(n)) = o(1)$, where $\mathbf{v}^* = (1, -\mathbf{w}^{*T})^T$. Under $H_0 : \theta^* = 0$, we have for any $t \in \mathbb{R}$

$$\lim_{n\to\infty} |\mathbb{P}_{\boldsymbol{\beta}^*}(\widehat{U}_n \leq t) - \Phi(t)| = 0.$$

*Proof.* A detailed proof is shown in Appendix F. □

The conditions in Theorem 6.21 are similar to those in Corollary 3.7 for the likelihood based score test. However, Theorem 6.21 requires $\|\mathbf{w}^*\|_1^2 \eta_7(n) = o(1)$. This assumption is needed to ensure the consistent estimation of the sandwich type variance $\mathbf{v}^{*T}\boldsymbol{\Sigma}^*\mathbf{v}^*$. In specific examples, this assumption can be relaxed. Due to the space constraint, we defer the analysis of one example based on the general loss to Appendix F.

# 7 Numerical Results

In this section, we conduct simulation studies to investigate the finite sample performance of the proposed score test. In particular, we simulate the response from the following two models: the linear regression with the standard Gaussian noise and the logistic regression. To generate the covariates, we simulate $n = 200$ independent samples from a multivariate Gaussian distribution $N_d(\mathbf{0}, \boldsymbol{\Sigma})$, where $d = 100, 200, 500$ and $\boldsymbol{\Sigma}$ is a Toeplitz matrix with $\Sigma_{jk} = \rho^{|j-k|}$. Here, we consider four possible values for $\rho$, i.e., $0.25, 0.4, 0.6, 0.75$. The true value $\boldsymbol{\beta}^*$ satisfies $\|\boldsymbol{\beta}^*\|_0 = s$, with $s = 2, 3$. We consider two scenarios for generating $\boldsymbol{\beta}^*$ on its support set $S$. In the first setting, we set $\boldsymbol{\beta}^*_S = (1, ..., 1)$, which is a Dirac measure. In the second setting, we generate each component of $\boldsymbol{\beta}^*_S$ from a uniform distribution on $[0, 2]$. Our goal is to test $H_0 : \beta_1 = 0$ versus $H_1 : \beta_1 \neq 0$. To check the validity of the type I error of the score test, we keep $\beta_1^*$ as 0. The number of simulations is 500.

The explicit forms of the score test for the linear regression and the generalized linear model are described in Sections 4 and 5. The tuning parameters $\lambda$ and $\lambda'$ are chosen by cross-validations. In the linear regression, we compare the performance of the score test with the de-sparsifying method (Lasso-Pro) in van de Geer et al. (2014) and the de-bias method (SSLasso) in Javanmard and Montanari (2013). Both of their methods are equivalent to certain types of Wald tests. The type I errors of the three tests are reported in Tables 1 and 2 for $s = 2$ and $s = 3$, respectively. We find that all three tests have similar performance and their type I errors are close to its significance level, which is consistent with the asymptotic equivalence among these three tests. To evaluate the power of the tests, we regenerate the data with the values of $\boldsymbol{\beta}_1^*$ ranging from 0 to 0.55. The power of the three tests is shown in Figure 1. We find that the score test is at least as powerful as the existing methods. Moreover, in many scenarios, the score test can be much more powerful; see the top panels of Figure 1. This agrees with the intuition, in the statistical literature, that the score test can be more powerful than the Wald test. To conclude this section, we note that the score test also performs well in the logistic regression; see Table 3.

Table 1: Average Type I error of the decorrelated score test, Lasso-Pro and SSLasso for the linear regression at 5% significance level, with $s = 2$.

|         |     | $\rho = 0.25$ |           | $\rho = 0.4$ |           | $\rho = 0.6$ |           | $\rho = 0.75$ |           |
|---------|-----|---------|-----------|-------|-----------|-------|-----------|-------|-----------|
| Method  | $d$ | Dirac   | Unif[0,2] | Dirac | Unif[0,2] | Dirac | Unif[0,2] | Dirac | Unif[0,2] |
| Score   | 100 | 5.3%    | 4.9%      | 5.1%  | 4.8%      | 5.2%  | 5.5%      | 5.3%  | 5.0%      |
|         | 200 | 5.1%    | 4.8%      | 5.3%  | 4.8%      | 5.9%  | 5.6%      | 4.7%  | 5.2%      |
|         | 500 | 5.7%    | 5.7%      | 5.8%  | 5.8%      | 5.4%  | 5.7%      | 4.2%  | 4.3%      |
| Lasso-Pro | 100 | 5.1%  | 5.2%      | 5.0%  | 4.7%      | 5.4%  | 5.1%      | 4.9%  | 5.1%      |
|         | 200 | 5.3%    | 4.9%      | 4.8%  | 5.1%      | 5.4%  | 5.1%      | 4.9%  | 5.3%      |
|         | 500 | 5.6%    | 5.7%      | 5.3%  | 4.7%      | 5.1%  | 4.6%      | 3.9%  | 4.1%      |
| SSLasso | 100 | 5.0%    | 5.1%      | 5.2%  | 4.8%      | 4.8%  | 4.7%      | 5.2%  | 5.4%      |
|         | 200 | 5.2%    | 4.7%      | 4.6%  | 5.4%      | 4.7%  | 5.1%      | 5.2%  | 4.8%      |
|         | 500 | 5.4%    | 5.5%      | 4.5%  | 4.4%      | 4.5%  | 4.8%      | 6.2%  | 5.9%      |

## 8 Discussion

In this paper, we propose a general framework for high dimensional inference based on the decorrelated score function. It can be used to test statistical hypothesis and construct confidence intervals. To broaden the applicability of the method, the theory is presented under full generality. In principle, the inference problem for many high dimensional models can be analyzed by using the current framework, although the technical details can be different case by case; see Ning and Liu (2014); Fang et al. (2014).

Similar to many existing procedures (Belloni et al., 2013; Zhang and Zhang, 2011; van de Geer et al., 2014; Javanmard and Montanari, 2013), the construction of the score function depends on the tuning parameters $\lambda$ and $\lambda'$. To reduce the sensitivity of tuning parameters, we can construct the tuning free score function by replacing the standard Lasso with the scaled Lasso (Sun and Zhang, 2012; Belloni et al., 2011) and the standard Dantzig selector with the calibrated Dantzig selector (Gautier and Tsybakov, 2011). It is of interest to develop the properties of this tuning free score function.

## Acknowledgments

We thank Ethan X. Fang for helping with the numerical studies, and Heather Battey for useful comments. This research is partially supported by NSF CAREER Award DMS1454377, NSF

Table 2: Average Type I error of the decorrelated score test, Lasso-Pro and SSLasso for the linear regression at 5% significance level, with $s = 3$.

| Method | $d$ | $\rho = 0.25$ | | $\rho = 0.4$ | | $\rho = 0.6$ | | $\rho = 0.75$ | |
| --- | --- | --- | --- | --- | --- | --- | --- | --- | --- |
| | | Dirac | Unif[0,2] | Dirac | Unif[0,2] | Dirac | Unif[0,2] | Dirac | Unif[0,2] |
| Score | 100 | 5.1% | 5.3% | 5.2% | 4.9% | 5.0% | 4.8% | 4.4% | 4.6% |
| | 200 | 4.7% | 4.8% | 5.1% | 5.4% | 5.3% | 4.9% | 5.1% | 4.8% |
| | 500 | 4.4% | 4.1% | 4.3% | 4.0% | 4.1% | 4.3% | 4.0% | 4.1% |
| Lasso-Pro | 100 | 5.0% | 4.9% | 5.2% | 4.8% | 5.3% | 5.2% | 4.7% | 4.6% |
| | 200 | 5.4% | 5.3% | 5.3% | 5.2% | 4.7% | 5.6% | 5.4% | 5.5% |
| | 500 | 5.5% | 5.9% | 5.1% | 4.6% | 4.7% | 5.3% | 6.2% | 6.3% |
| SSLasso | 100 | 5.4% | 5.3% | 4.9% | 4.7% | 5.1% | 5.0% | 5.1% | 4.9% |
| | 200 | 5.3% | 5.2% | 4.9% | 4.8% | 5.3% | 4.8% | 4.5% | 4.7% |
| | 500 | 5.8% | 5.6% | 5.5% | 5.7% | 5.3% | 5.6% | 6.5% | 6.1% |

# A  Proofs of Main Theorems

In this appendix, we present the proofs of the main results in Section 3, namely, Corollary 3.7, Theorems 3.14, 3.22 and 3.31.

## A.1  Proof of Corollary 3.7

*Proof of Corollary 3.7.* Since (3.6) is equivalent to $\widehat{U}_n \rightsquigarrow N(0,1)$, we only need to show that $\widehat{I}_{\theta|\gamma} = I^*_{\theta|\gamma} + o_\mathbb{P}(1)$ holds. Then, the weak convergence follows by Theorem 3.5 and the Slutsky's Theorem. It is easily seen that

$$
\begin{aligned}
|\widehat{I}_{\theta|\gamma} - I^*_{\theta|\gamma}| &\leq |\nabla^2_{\theta\theta}\ell(\widehat{\boldsymbol{\beta}}) - I^*_{\theta\theta}| + |\mathbf{w}^{*T}(\mathbf{I}^*_{\gamma\theta} - \nabla^2_{\gamma\theta}\ell(\widehat{\boldsymbol{\beta}}))| + |(\widehat{\mathbf{w}} - \mathbf{w}^*)^T \nabla^2_{\gamma\theta}\ell(\widehat{\boldsymbol{\beta}})| \\
&\leq |\nabla^2_{\theta\theta}\ell(\widehat{\boldsymbol{\beta}}) - I^*_{\theta\theta}| + ||\mathbf{w}^*||_1 ||\mathbf{I}^*_{\theta\gamma} - \nabla^2_{\theta\gamma}\ell(\widehat{\boldsymbol{\beta}})||_\infty + ||\widehat{\mathbf{w}} - \mathbf{w}^*||_1 ||\nabla^2_{\theta\gamma}\ell(\widehat{\boldsymbol{\beta}})||_\infty \\
&= \mathcal{O}_\mathbb{P}(\eta_5(n)) + \mathcal{O}_\mathbb{P}(||\mathbf{w}^*||_1 \eta_5(n)) + \mathcal{O}_\mathbb{P}(\eta_2(n)||\mathbf{I}^*_{\theta\gamma}||_\infty) + \mathcal{O}_\mathbb{P}(\eta_2(n)\eta_5(n)).
\end{aligned}
$$

Hence, $|\widehat{I}_{\theta|\gamma} - I^*_{\theta|\gamma}| = o_\mathbb{P}(1)$. This completes the proof. □

## A.2  Proof of Theorem 3.14

To prove Theorem 3.14, we first present the following Lemma.

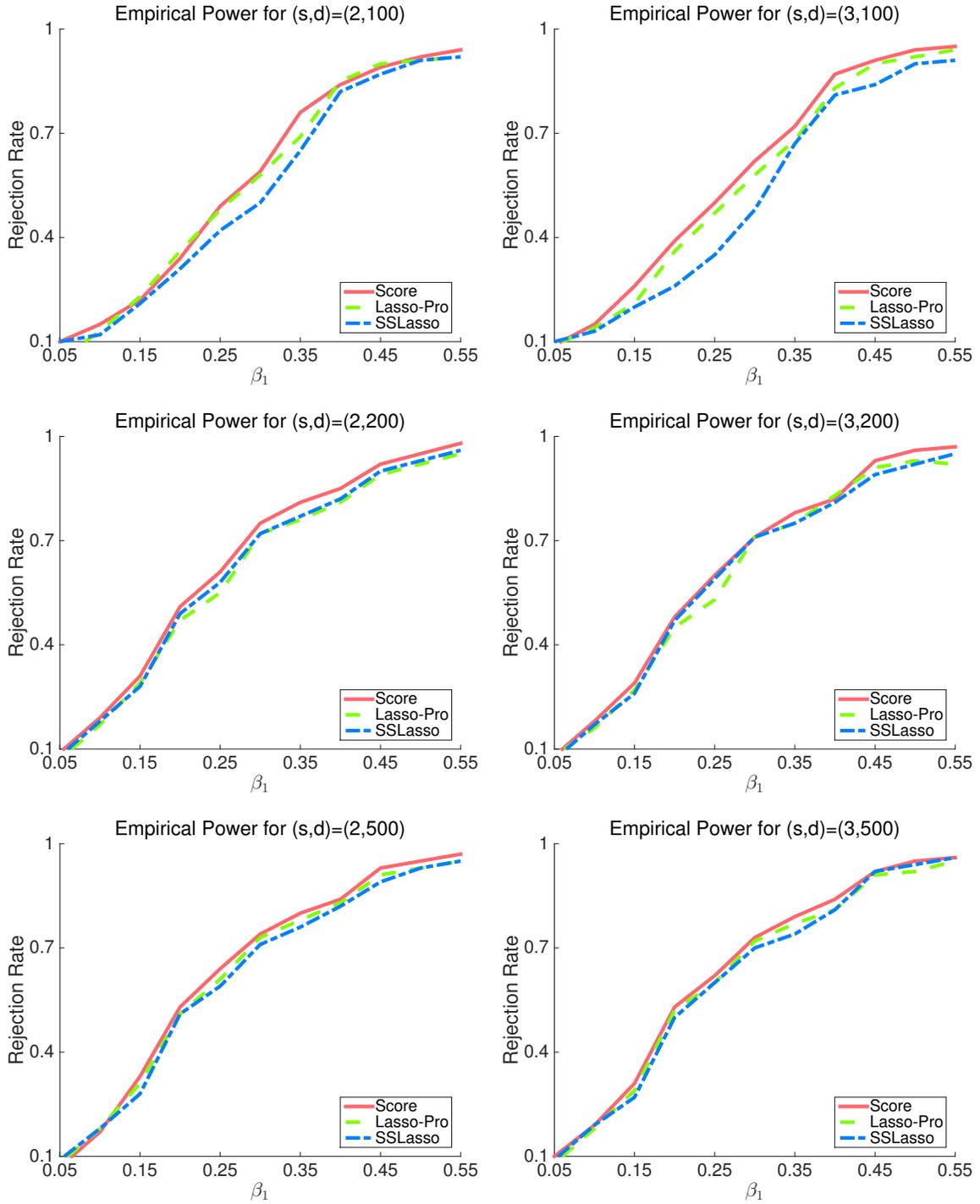

Figure 1: Power of the decorrelated score test, Lasso-Pro and SSLasso for the linear regression at 5% significance level.

Table 3: Average Type I error of the decorrelated score test for the logistic regression at 5% significance level.

|  |  | $\rho = 0.25$ | | $\rho = 0.4$ | | $\rho = 0.6$ | | $\rho = 0.75$ | |
| --- | --- | --- | --- | --- | --- | --- | --- | --- | --- |
| Method | $d$ | Dirac | Unif[0,2] | Dirac | Unif[0,2] | Dirac | Unif[0,2] | Dirac | Unif[0,2] |
| $s=2$ | 100 | 5.4% | 5.3% | 4.8% | 4.9% | 5.0% | 4.9% | 4.8% | 5.1% |
|  | 200 | 5.1% | 4.5% | 4.9% | 5.4% | 4.6% | 4.7% | 4.4% | 4.2% |
|  | 500 | 3.7% | 4.2% | 4.7% | 4.4% | 6.7% | 5.9% | 6.9% | 6.1% |
| $s=3$ | 100 | 5.6% | 5.2% | 5.4% | 5.5% | 4.8% | 4.5% | 4.7% | 5.0% |
|  | 200 | 4.3% | 4.5% | 4.7% | 4.9% | 5.3% | 5.5% | 4.8% | 4.6% |
|  | 500 | 3.6% | 3.4% | 3.6% | 4.1% | 3.7% | 3.2% | 5.5% | 5.2% |

**Lemma A.1.** Under the Assumption 3.9–3.12, we have that with probability tending to one

$$\lim_{n\to\infty} \inf_{\boldsymbol{\beta}^* \in \Omega_0} \mathbb{P}_{\boldsymbol{\beta}^*}\left(\left|\widehat{S}(0,\widehat{\boldsymbol{\gamma}}) - S(\boldsymbol{\beta}^*)\right| \lesssim \left(\eta_2(n)\eta_3(n) + \eta_1(n)\eta_4(n)\right)\right) = 1. \quad \text{(A.1)}$$

If $n^{1/2}(\eta_2(n)\eta_3(n) + \eta_1(n)\eta_4(n)) = o(1)$, we have

$$\lim_{n\to\infty} \sup_{\boldsymbol{\beta}^* \in \Omega_0} \sup_{t \in \mathbb{R}} \left|\mathbb{P}_{\boldsymbol{\beta}^*}\left(n^{1/2}\widehat{S}(0,\widehat{\boldsymbol{\gamma}})I_{\theta|\gamma}^{*-1/2} \leq t\right) - \Phi(t)\right| = 0. \quad \text{(A.2)}$$

*Proof.* See Appendix G.1 for a detailed proof. □

We now present the proof of Theorem 3.14.

*Proof of Theorem 3.14.* Similar to the proof of Corollary 3.7, by the triangle inequality, we can show that on the event $\mathcal{F}^{\boldsymbol{\beta}^*} = \mathcal{F}_1^{\boldsymbol{\beta}^*} \cap ... \cap \mathcal{F}_5^{\boldsymbol{\beta}^*}$,

$$\begin{aligned}
|\widehat{I}_{\theta|\gamma} - I_{\theta|\gamma}^*| &\leq |\nabla_{\theta\theta}^2 \ell(\widehat{\boldsymbol{\beta}}) - I_{\theta\theta}^*| + ||\mathbf{w}^*||_1 ||\mathbf{I}_{\theta\gamma}^* - \nabla_{\theta\gamma}^2 \ell(\widehat{\boldsymbol{\beta}})||_\infty + ||\widehat{\mathbf{w}} - \mathbf{w}^*||_1 ||\nabla_{\theta\gamma}^2 \ell(\widehat{\boldsymbol{\beta}})||_\infty \\
&\lesssim \eta_5(n)(1 + ||\mathbf{w}^*||_1) + \eta_2(n)(||\mathbf{I}_{\theta\gamma}^*||_\infty + \eta_5(n)). \quad \text{(A.3)}
\end{aligned}$$

Denote $\psi_n = \eta_5(n) \sup_{\boldsymbol{\beta}^* \in \Omega_0} ||\mathbf{w}^*||_1 + \eta_2(n) \sup_{\boldsymbol{\beta}^* \in \Omega_0} ||\mathbf{I}_{\theta\gamma}^*||_\infty$. By assumptions in the corollary, $\psi_n = o(1)$. Denote $U_n = n^{1/2}\widehat{S}(0,\widehat{\boldsymbol{\gamma}})I_{\theta|\gamma}^{*-1/2}$. To show (3.8), for any $t$ and a sequence of positive $\delta_n \to 0$ to be determined later, we have

$$\mathbb{P}_{\boldsymbol{\beta}^*}(\widehat{U}_n \leq t) - \Phi(t) = \underbrace{\left\{\mathbb{P}_{\boldsymbol{\beta}^*}(\widehat{U}_n \leq t) - \mathbb{P}_{\boldsymbol{\beta}^*}(U_n \leq t + \delta_n)\right\}}_{I_1} + \underbrace{\left\{\mathbb{P}_{\boldsymbol{\beta}^*}(U_n \leq t + \delta_n) - \Phi(t + \delta_n)\right\}}_{I_2}$$

$$+ \underbrace{\left\{\Phi(t + \delta_n) - \Phi(t)\right\}}_{I_3}.$$

In the following, we first show that $\limsup_{n\to\infty} \sup_{\boldsymbol{\beta}^*\in\Omega_0} \sup_{t\in\mathbb{R}} I_1 \leq 0$. By the triangle inequality, it is easily seen that

$$
\begin{aligned}
\sup_{t\in\mathbb{R}} I_1 &\leq \mathbb{P}_{\boldsymbol{\beta}^*}(|\widehat{U}_n - U_n| \geq \delta_n) = \mathbb{P}_{\boldsymbol{\beta}^*}\big(|U_n||1 - \widehat{I}_{\theta|\gamma}^{-1/2} I_{\theta|\gamma}^{*1/2}| \geq \delta_n\big) \\
&\leq \mathbb{P}_{\boldsymbol{\beta}^*}(|U_n| \geq \delta_n^{-1}) + \mathbb{P}_{\boldsymbol{\beta}^*}\big(|1 - \widehat{I}_{\theta|\gamma}^{-1/2} I_{\theta|\gamma}^{*1/2}| \geq \delta_n^2\big).
\end{aligned}
\tag{A.4}
$$

The first term of (A.4) can be bounded by

$$
\mathbb{P}_{\boldsymbol{\beta}^*}(|U_n| \geq \delta_n^{-1}) \leq \big|\mathbb{P}_{\boldsymbol{\beta}^*}(|U_n| \geq \delta_n^{-1}) - \mathbb{P}(|N| \geq \delta_n^{-1})\big| + \mathbb{P}(|N| \geq \delta_n^{-1}),
$$

where $N \sim N(0,1)$. By Lemma A.1,

$$
\limsup_{n\to\infty} \sup_{\boldsymbol{\beta}^*\in\Omega_0} \sup_{\delta_n\in\mathbb{R}} \big|\mathbb{P}_{\boldsymbol{\beta}^*}(|U_n| \geq \delta_n^{-1}) - \mathbb{P}(|N| \geq \delta_n^{-1})\big| = 0.
$$

The tail bound for the standard normal distribution yields $\mathbb{P}(|N| \geq \delta_n^{-1}) \leq 2\frac{\delta_n}{\sqrt{2\pi}} \exp(-\frac{1}{2\delta_n^2}) \to 0$, as $\delta_n \to 0$. Thus, $\limsup_{n\to\infty} \sup_{\boldsymbol{\beta}^*\in\Omega_0} \sup_{\delta_n\in\mathbb{R}} \mathbb{P}_{\boldsymbol{\beta}^*}(|U_n| \geq \delta_n^{-1}) \leq 0$. That means, the first term of (A.4) is bounded above by 0. For the second term of (A.4), we have

$$
\mathbb{P}_{\boldsymbol{\beta}^*}\big(|1 - \widehat{I}_{\theta|\gamma}^{-1/2} I_{\theta|\gamma}^{*1/2}| \geq \delta_n^2\big) = \mathbb{P}_{\boldsymbol{\beta}^*}\bigg(\frac{|\widehat{I}_{\theta|\gamma} - I_{\theta|\gamma}^*|}{(\widehat{I}_{\theta|\gamma}^{1/2} + I_{\theta|\gamma}^{*1/2})\widehat{I}_{\theta|\gamma}^{1/2}} \geq \delta_n^2\bigg).
$$

On the event $\mathcal{F}^{\boldsymbol{\beta}^*}$, by (A.3) and the assumption $C' \leq I_{\theta|\gamma}^*$, we can show that

$$
|\widehat{I}_{\theta|\gamma} - I_{\theta|\gamma}^*|\big\{(\widehat{I}_{\theta|\gamma}^{1/2} + I_{\theta|\gamma}^{*1/2})\widehat{I}_{\theta|\gamma}^{1/2}\big\}^{-1} \lesssim |\widehat{I}_{\theta|\gamma} - I_{\theta|\gamma}^*| \lesssim \psi_n,
$$

since $\widehat{I}_{\theta|\gamma} \geq C' - \psi_n \geq C'/2$ for $n$ large enough. Hence, with $\delta_n = C\psi_n^{1/2}$, for some sufficiently large constant $C$, we obtain that the second term of (A.4), $\mathbb{P}_{\boldsymbol{\beta}^*}(|1 - \widehat{I}_{\theta|\gamma}^{-1/2} I_{\theta|\gamma}^{*1/2}| \geq \delta_n^2) = 0$, for $n$ large enough. As a result, $\limsup_{n\to\infty} \sup_{\boldsymbol{\beta}^*\in\Omega_0} \sup_{t\in\mathbb{R}} I_1 \leq 0$. By Lemma A.1, we can show that

$$
\limsup_{n\to\infty} \sup_{\boldsymbol{\beta}^*\in\Omega_0} \sup_{t\in\mathbb{R}} I_2 \leq \limsup_{n\to\infty} \sup_{\boldsymbol{\beta}^*\in\Omega_0} \sup_{t'\in\mathbb{R}} \big|\mathbb{P}_{\boldsymbol{\beta}^*}(U_n \leq t') - \Phi(t')\big| = 0.
$$

Finally, $I_3 \leq (2\pi)^{-1/2}\delta_n$, which implies that $\limsup_{n\to\infty} \sup_{\boldsymbol{\beta}^*\in\Omega_0} \sup_{t\in\mathbb{R}} I_3 \leq 0$. Combining these results, we obtain

$$
\limsup_{n\to\infty} \sup_{\boldsymbol{\beta}^*\in\Omega_0} \sup_{t\in\mathbb{R}} \{\mathbb{P}_{\boldsymbol{\beta}^*}(\widehat{U}_n \leq t) - \Phi(t)\} \leq 0,
$$

Similar arguments yields the bound for the minimum,

$$
\liminf_{n\to\infty} \inf_{\boldsymbol{\beta}^*\in\Omega_0} \inf_{t\in\mathbb{R}} \{\mathbb{P}_{\boldsymbol{\beta}^*}(\widehat{U}_n \leq t) - \Phi(t)\} \geq 0.
$$

This completes the proof of (3.8).

$\square$

## A.3 Proof of Theorem 3.22

To prove Theorem 3.22, we start from the Lemma A.2.

**Lemma A.2.** *Under the Assumption 3.17–3.20, we have*

$$\lim_{n\to\infty} \inf_{\boldsymbol{\beta}^*\in\Omega_1(\widetilde{C},\phi)} \mathbb{P}_{\boldsymbol{\beta}^*}\left(n^{1/2}|\widehat{S}(0,\widehat{\boldsymbol{\gamma}}) - S(\boldsymbol{\beta}^*) + \widetilde{C}n^{-\phi}I^*_{\theta|\boldsymbol{\gamma}}| \lesssim \eta_6(n) + n^{1/2}\psi_n\right) = 1, \tag{A.5}$$

*where $\psi_n = \eta_2(n)\eta_3(n) + \eta_1(n)\eta_4(n)$. If $n^{1/2}\psi_n = o(1)$ and $\eta_6(n) = o(1)$, we have*

$$\lim_{n\to\infty} \sup_{\boldsymbol{\beta}^*\in\Omega_1(\widetilde{C},\phi)} \sup_{t\in\mathbb{R}} \left|\mathbb{P}_{\boldsymbol{\beta}^*}\left(n^{1/2}\widehat{S}(0,\widehat{\boldsymbol{\gamma}})I^{*-1/2}_{\theta|\boldsymbol{\gamma}} \le t\right) - \Phi(t)\right| = 0, \quad \text{if} \quad \phi > 1/2, \tag{A.6}$$

$$\lim_{n\to\infty} \sup_{\boldsymbol{\beta}^*\in\Omega_1(\widetilde{C},\phi)} \sup_{t\in\mathbb{R}} \left|\mathbb{P}_{\boldsymbol{\beta}^*}\left(n^{1/2}\widehat{S}(0,\widehat{\boldsymbol{\gamma}})I^{*-1/2}_{\theta|\boldsymbol{\gamma}} \le t\right) - \Phi(t + \widetilde{C}I^{*1/2}_{\theta|\boldsymbol{\gamma}})\right| = 0, \quad \text{if} \quad \phi = 1/2, \tag{A.7}$$

*and for any fixed $t \in \mathbb{R}$ and $\widetilde{C} \ne 0$*

$$\lim_{n\to\infty} \sup_{\boldsymbol{\beta}^*\in\Omega_1(\widetilde{C},\phi)} \mathbb{P}_{\boldsymbol{\beta}^*}\left(|n^{1/2}\widehat{S}(0,\widehat{\boldsymbol{\gamma}})I^{*-1/2}_{\theta|\boldsymbol{\gamma}}| \le t\right) = 0, \quad \text{if} \quad \phi < 1/2. \tag{A.8}$$

*Proof.* See Appendix G.1 for a detailed proof. □

Given Lemma A.2, we now prove Theorem 3.22.

*Proof of Theorem 3.22.* The proof is similar to that of Theorem 3.14. To highlight the difference, we only present the proofs of (3.11) and (3.12). By the triangle inequality, we can show that on the event $\mathcal{F}_1^{\boldsymbol{\beta}^*} \cap \dots \cap \mathcal{F}_6^{\boldsymbol{\beta}^*}$,

$$\begin{aligned}
|\widehat{I}_{\theta|\boldsymbol{\gamma}} - I^*_{\theta|\boldsymbol{\gamma}}| &\le |\nabla^2_{\theta\theta}\ell(\widehat{\boldsymbol{\beta}}) - I^*_{\theta\theta}| + ||\mathbf{w}^*||_1||\mathbf{I}^*_{\theta\boldsymbol{\gamma}} - \nabla^2_{\theta\boldsymbol{\gamma}}\ell(\widehat{\boldsymbol{\beta}})||_\infty + ||\widehat{\mathbf{w}} - \mathbf{w}^*||_1||\nabla^2_{\theta\boldsymbol{\gamma}}\ell(\widehat{\boldsymbol{\beta}})||_\infty \\
&\lesssim \eta_5(n)||\mathbf{w}^*||_1 + \eta_2(n)||\mathbf{I}^*_{\theta\boldsymbol{\gamma}}||_\infty.
\end{aligned}$$

Denote $\psi_n = \eta_5(n)\sup_{\boldsymbol{\beta}^*\in\Omega_1(\widetilde{C},\phi)}||\mathbf{w}^*||_1 + \eta_2(n)\sup_{\boldsymbol{\beta}^*\in\Omega_1(\widetilde{C},\phi)}||\mathbf{I}^*_{\theta\boldsymbol{\gamma}}||_\infty$. Note that $\psi_n = o(1)$. Denote $U_n = n^{1/2}\widehat{S}(0,\widehat{\boldsymbol{\gamma}})I^{*-1/2}_{\theta|\boldsymbol{\gamma}}$. For any $t$ and a sequence of positive $\delta_n \to 0$ to be determined later,

$$\mathbb{P}_{\boldsymbol{\beta}^*}(\widehat{U}_n \le t) - \Phi(t + \widetilde{C}I^{*1/2}_{\theta|\boldsymbol{\gamma}}) = \underbrace{\left\{\mathbb{P}_{\boldsymbol{\beta}^*}(\widehat{U}_n \le t) - \mathbb{P}_{\boldsymbol{\beta}^*}(U_n \le t + \delta_n)\right\}}_{I_1}$$
$$+ \underbrace{\left\{\mathbb{P}_{\boldsymbol{\beta}^*}(U_n \le t + \delta_n) - \Phi(t + \widetilde{C}I^{*1/2}_{\theta|\boldsymbol{\gamma}} + \delta_n)\right\}}_{I_2} + \underbrace{\left\{\Phi(t + \widetilde{C}I^{*1/2}_{\theta|\boldsymbol{\gamma}} + \delta_n) - \Phi(t + \widetilde{C}I^{*1/2}_{\theta|\boldsymbol{\gamma}})\right\}}_{I_3}.$$

By the triangle inequality, it is easily seen that

$$\begin{aligned}
\sup_{t\in\mathbb{R}} I_1 &\le \mathbb{P}_{\boldsymbol{\beta}^*}(|\widehat{U}_n - U_n| \ge \delta_n) = \mathbb{P}_{\boldsymbol{\beta}^*}\left(|U_n||1 - \widehat{I}^{-1/2}_{\theta|\boldsymbol{\gamma}}I^{*1/2}_{\theta|\boldsymbol{\gamma}}| \ge \delta_n\right) \\
&\le \mathbb{P}_{\boldsymbol{\beta}^*}\left(|U_n| \ge \delta_n^{-1}\right) + \mathbb{P}_{\boldsymbol{\beta}^*}\left(|1 - \widehat{I}^{-1/2}_{\theta|\boldsymbol{\gamma}}I^{*1/2}_{\theta|\boldsymbol{\gamma}}| \ge \delta_n^2\right). \tag{A.9}
\end{aligned}$$

The first term of (A.9) can be further bounded by

$$\mathbb{P}_{\boldsymbol{\beta}^*}(|U_n| \geq \delta_n^{-1}) \leq \left|\mathbb{P}_{\boldsymbol{\beta}^*}(|U_n| \geq \delta_n^{-1}) - \mathbb{P}(|N - \widetilde{C}I_{\theta|\gamma}^{*1/2}| \geq \delta_n^{-1})\right| + \mathbb{P}(|N - \widetilde{C}I_{\theta|\gamma}^{*1/2}| \geq \delta_n^{-1}),$$

where $N \sim N(0,1)$. By Lemma A.2,

$$\limsup_{n \to \infty} \sup_{\boldsymbol{\beta}^* \in \Omega_1(\widetilde{C}, \phi)} \sup_{\delta_n \in \mathbb{R}} \left|\mathbb{P}_{\boldsymbol{\beta}^*}(|U_n| \geq \delta_n^{-1}) - \mathbb{P}(|N - \widetilde{C}I_{\theta|\gamma}^{*1/2}| \geq \delta_n^{-1})\right| = 0,$$

and the tail bound for the standard normal distribution yields

$$\mathbb{P}(|N - \widetilde{C}I_{\theta|\gamma}^{*1/2}| \geq \delta_n^{-1}) \leq \mathbb{P}(|N| \geq \delta_n^{-1} - |\widetilde{C}|I_{\theta|\gamma}^{*1/2})$$
$$\leq \frac{2}{\sqrt{2\pi}(\delta_n^{-1} - |\widetilde{C}|I_{\theta|\gamma}^{*1/2})} \exp\left(-\frac{(\delta_n^{-1} - |\widetilde{C}|I_{\theta|\gamma}^{*1/2})^2}{2}\right) \to 0,$$

as $\delta \to 0$, uniformly over $\boldsymbol{\beta}^*$, due to $I_{\theta|\gamma}^* \leq C''$. Thus, the first term of (A.9) is bounded above by 0. For the second term of (A.9), we have

$$\mathbb{P}_{\boldsymbol{\beta}^*}\left(|1 - \widehat{I}_{\theta|\gamma}^{-1/2}I_{\theta|\gamma}^{*1/2}| \geq \delta_n^2\right) = \mathbb{P}_{\boldsymbol{\beta}^*}\left(\frac{|\widehat{I}_{\theta|\gamma} - I_{\theta|\gamma}^*|}{(\widehat{I}_{\theta|\gamma}^{1/2} + I_{\theta|\gamma}^{*1/2})\widehat{I}_{\theta|\gamma}^{1/2}} \geq \delta_n^2\right).$$

On the event $\mathcal{F}^{\boldsymbol{\beta}^*}$, by $C' \leq I_{\theta|\gamma}^*$, there exists a constant $C'''$ such that $|\widehat{I}_{\theta|\gamma} - I_{\theta|\gamma}^*|\{(\widehat{I}_{\theta|\gamma}^{1/2} + I_{\theta|\gamma}^{*1/2})\widehat{I}_{\theta|\gamma}^{1/2}\}^{-1} \leq C'''\psi_n$, where $\widehat{I}_{\theta|\gamma} \geq C' - \psi_n \geq C'/2$ for $n$ large enough. Hence, with $\delta_n = C\psi_n^{1/2}$, for some sufficiently large constant $C$, we obtain that the second term of (A.9) goes to 0. As a result, $\limsup_{n\to\infty} \sup_{\boldsymbol{\beta}^* \in \Omega_0} \sup_{t \in \mathbb{R}} I_1 \leq 0$. By Lemma A.2, we can show that

$$\limsup_{n\to\infty} \sup_{\boldsymbol{\beta}^* \in \Omega_1(\widetilde{C},\phi)} \sup_{t \in \mathbb{R}} I_2 \leq \limsup_{n\to\infty} \sup_{\boldsymbol{\beta}^* \in \Omega_1(\widetilde{C},\phi)} \sup_{t' \in \mathbb{R}} \left|\mathbb{P}_{\boldsymbol{\beta}^*}(U_n \leq t') - \Phi(t' + \widetilde{C}I_{\theta|\gamma}^{*1/2})\right| = 0.$$

Finally, $I_3 \leq (2\pi)^{-1/2}\delta_n$, which implies that $\limsup_{n\to\infty} \sup_{\boldsymbol{\beta}^* \in \Omega_1(\widetilde{C},\phi)} \sup_{t \in \mathbb{R}} I_3 \leq 0$. Combining these results, we obtain

$$\limsup_{n\to\infty} \sup_{\boldsymbol{\beta}^* \in \Omega_1(\widetilde{C},\phi)} \sup_{t \in \mathbb{R}} \{\mathbb{P}_{\boldsymbol{\beta}^*}(\widehat{U}_n \leq t) - \Phi(t + \widetilde{C}I_{\theta|\gamma}^{*1/2})\} \leq 0,$$

Similar arguments yield the following lower bound

$$\liminf_{n\to\infty} \inf_{\boldsymbol{\beta}^* \in \Omega_1(\widetilde{C},\phi)} \inf_{t \in \mathbb{R}} \{\mathbb{P}_{\boldsymbol{\beta}^*}(\widehat{U}_n \leq t) - \Phi(t + \widetilde{C}I_{\theta|\gamma}^{*1/2})\} \geq 0.$$

This completes the proof of (3.11). For (3.12), since $\sup_{\boldsymbol{\beta}^* \in \Omega_1(\widetilde{C},\phi)} |\widehat{I}_{\theta|\gamma} - I_{\theta|\gamma}^*| = o_{\mathbb{P}}(1)$ and $C' \leq I_{\theta|\gamma}^*$, we have $|\widehat{I}_{\theta|\gamma}/I_{\theta|\gamma}^* - 1| \leq 3$, for $n$ large enough (not depending on $\boldsymbol{\beta}^*$). Given any $t \in \mathbb{R}$, for $n$ sufficiently large,

$$\mathbb{P}_{\boldsymbol{\beta}^*}(|\widehat{U}_n| \leq t) = \mathbb{P}_{\boldsymbol{\beta}^*}\left(|U_n| \leq t(\widehat{I}_{\theta|\gamma}/I_{\theta|\gamma}^*)^{1/2}\right) \leq \mathbb{P}_{\boldsymbol{\beta}^*}(|U_n| \leq 2t).$$

Hence, by (A.8) in Lemma A.2, we finally obtain

$$\lim_{n\to\infty} \sup_{\boldsymbol{\beta}^* \in \Omega_1(\widetilde{C},\phi)} \mathbb{P}_{\boldsymbol{\beta}^*}(|\widehat{U}_n| \leq t) \leq \lim_{n\to\infty} \sup_{\boldsymbol{\beta}^* \in \Omega_1(\widetilde{C},\phi)} \mathbb{P}_{\boldsymbol{\beta}^*}(|U_n| \leq 2t) = 0.$$

□

## A.4 Proofs of Theorems 3.31

*Proof of Theorem 3.31.* By the definition of $\widetilde{\theta}$ and the mean value theorem, we obtain

$$\begin{aligned}
n^{1/2}(\widetilde{\theta} - \theta^*) &= n^{1/2}(\widehat{\theta} - \theta^* - \widehat{I}_{\theta|\gamma}^{-1}\widehat{S}(\widehat{\boldsymbol{\beta}})) \\
&= n^{1/2}(\widehat{\theta} - \theta^* - \widehat{I}_{\theta|\gamma}^{-1}\widehat{S}(\theta^*, \widehat{\gamma}) - \widehat{I}_{\theta|\gamma}^{-1}\bar{I}_{\theta|\gamma}(\widehat{\theta} - \theta^*)).
\end{aligned}$$

Rearranging terms, we can get

$$n^{1/2}(\widetilde{\theta} - \theta^*) = -n^{1/2}\widehat{I}_{\theta|\gamma}^{-1}\widehat{S}(\theta^*, \widehat{\gamma}) - n^{1/2}(\widehat{I}_{\theta|\gamma}^{-1}\bar{I}_{\theta|\gamma} - 1)(\widehat{\theta} - \theta^*), \tag{A.10}$$

where $\bar{I}_{\theta|\gamma} = \nabla^2_{\theta\theta}\ell(\bar{\theta}, \widehat{\gamma}) - \widehat{\mathbf{w}}^T \nabla^2_{\gamma\theta}\ell(\bar{\theta}, \widehat{\gamma})$ with $\bar{\theta} = v\theta^* + (1-v)\widehat{\theta}$ for some $v \in [0, 1]$. The proof of Corollary 3.7 implies $n^{1/2}\widehat{I}_{\theta|\gamma}^{-1}\widehat{S}(\theta^*, \widehat{\gamma}) = n^{1/2}I_{\theta|\gamma}^{*-1}\widehat{S}(\theta^*, \widehat{\gamma}) + o_{\mathbb{P}}(1)$. So, it remains to show the last term in (A.10) is of order $o_{\mathbb{P}}(1)$. Note that $\widehat{I}_{\theta|\gamma}^{-1}\bar{I}_{\theta|\gamma} - 1 = \widehat{I}_{\theta|\gamma}^{-1}(\bar{I}_{\theta|\gamma} - \widehat{I}_{\theta|\gamma})$. By the proof of Corollary 3.7 and Assumption 3.29, $\widehat{I}_{\theta|\gamma}^{-1} = I_{\theta|\gamma}^{*-1} + o_{\mathbb{P}}(1) = \mathcal{O}_{\mathbb{P}}(1)$. Moreover, by Assumption 3.30, following the similar arguments to those in the proof of Corollary 3.7, we can obtain

$$|\bar{I}_{\theta|\gamma} - \widehat{I}_{\theta|\gamma}| \leq |\nabla^2_{\theta\theta}\ell(\bar{\theta}, \widehat{\gamma}) - \nabla^2_{\theta\theta}\ell(\widehat{\boldsymbol{\beta}})| + |\widehat{\mathbf{w}}^T(\nabla^2_{\gamma\theta}\ell(\bar{\theta}, \widehat{\gamma}) - \nabla^2_{\gamma\theta}\ell(\widehat{\boldsymbol{\beta}}))| = \mathcal{O}_{\mathbb{P}}(||\mathbf{w}^*||_1 \eta_5(n)).$$

Together with the assumptions in Theorem 3.31 and equation (A.10), we obtain that

$$n^{1/2}(\widetilde{\theta} - \theta^*) = -n^{1/2}I_{\theta|\gamma}^{*-1}\widehat{S}(\theta^*, \widehat{\gamma}) + n^{1/2}(\widehat{\theta} - \theta^*)\mathcal{O}_{\mathbb{P}}(||\mathbf{w}^*||_1 \eta_5(n)) = -n^{1/2}I_{\theta|\gamma}^{*-1}\widehat{S}(\theta^*, \widehat{\gamma}) + o_{\mathbb{P}}(1).$$

Hence, $n^{1/2}(\widetilde{\theta} - \theta^*)I_{\theta|\gamma}^{*1/2} \rightsquigarrow N(0, 1)$, by Theorem 3.5. Since $n^{1/2}(\widetilde{\theta} - \theta^*) = \mathcal{O}_{\mathbb{P}}(1)$ and $\widehat{I}_{\theta|\gamma} = I_{\theta|\gamma}^* + o_{\mathbb{P}}(1)$, we complete the proof, by invoking the Slutsky's Theorem. $\square$

## B Proofs for the Linear Model

In this appendix, we first prove Theorem 4.1, Corollary 4.2, 4.3 and Theorem 4.7. Then, we prove the optimality results, and finally prove Theorem 4.15 and Corollary 4.16.

### B.1 Proofs of Theorem 4.1, Corollary 4.2, 4.3 and Theorem 4.7

The proofs of Theorems 4.1 and 4.7 are the direct consequences of the general results based on the following auxiliary lemmas. For the reader's convenience, we first present the conclusions from the lemmas and prove Theorems 4.1 and 4.7. The detailed proofs of Lemmas are presented afterward.

**Lemma B.1.** Under the conditions of Theorem 4.1, with probability at least $1 - d^{-1}$,

$$\left\|\frac{1}{n}\sum_{i=1}^{n} \boldsymbol{X}_i \epsilon_i \right\|_{\infty} \leq C\sqrt{\frac{\log d}{n}},$$

for some constant $C > 0$.

*Proof.* See Appendix G.2 for a detailed proof. $\square$

**Lemma B.2.** Under the conditions of Theorem 4.1, with probability at least $1 - d^{-1}$,

$$\left\| \frac{1}{n} \sum_{i=1}^{n} (Z_i \boldsymbol{X}_i - \widehat{\mathbf{w}}^T \boldsymbol{X}_i^{\otimes 2}) \right\|_\infty \leq C\sqrt{\frac{\log d}{n}},$$

for some constant $C > 0$.

*Proof.* See Appendix G.2 for a detailed proof. □

**Lemma B.3.** Under the conditions of Theorem 4.1, with probability at least $1 - (1 + C')d^{-1}$,

$$\|\widehat{\boldsymbol{\beta}} - \boldsymbol{\beta}^*\|_1 \leq \frac{12Cs^*}{\kappa}\sqrt{\frac{\log d}{n}}, \quad \text{and} \quad (\widehat{\boldsymbol{\beta}} - \boldsymbol{\beta}^*)^T \mathbf{H}_Q (\widehat{\boldsymbol{\beta}} - \boldsymbol{\beta}^*) \leq \frac{36C^2}{\kappa} \frac{s^* \log d}{n}$$

where $\mathbf{H}_Q = n^{-1} \sum_{i=1}^{n} \boldsymbol{Q}_i^{\otimes 2}$ and the constants $C$ and $C'$ are given by Lemmas B.1 and G.1.

*Proof.* See Appendix G.2 for a detailed proof. □

**Lemma B.4.** Under the conditions of Theorem 4.1, with probability at least $1 - (C' + 1)d^{-1}$,

$$\|\widehat{\mathbf{w}} - \mathbf{w}^*\|_1 \leq 8C\kappa^{-1} s' \sqrt{\frac{\log d}{n}},$$

where the constants $C$ and $C'$ are given by Lemmas B.2 and G.1.

*Proof.* See Appendix G.2 for a detailed proof. □

**Lemma B.5.** Under the conditions of Theorem 4.1, it holds that $T^* \rightsquigarrow N(0, 1)$, and

$$\sup_{x \in \mathbb{R}} \left| \mathbb{P}_{\boldsymbol{\beta}^*}(T^* \leq x) - \Phi(x) \right| \leq Cn^{-1/2}, \tag{B.1}$$

where $T^* = n^{1/2} S(0, \boldsymbol{\gamma}^*) / I_{\theta|\gamma}^{*1/2}$ and $C$ is a positive constant not depending on $\boldsymbol{\beta}^*$.

*Proof.* See Appendix G.2 for a detailed proof. □

**Lemma B.6.** Under the conditions of Theorem 4.1, with probability at least $1 - d^{-1}$, we have $\|\mathbf{H}_Q - \mathbb{E}(\boldsymbol{Q}_i^{\otimes 2})\|_{\max} \leq C\sqrt{\frac{\log d}{n}}$, for some constant $C > 0$, where $\mathbf{H}_Q = n^{-1} \sum_{i=1}^{n} \boldsymbol{Q}_i^{\otimes 2}$.

*Proof.* See Appendix G.2 for a detailed proof. □

Given the Lemmas B.1-B.6, we now prove Theorem 4.1.

*Proof of Theorem 4.1.* Note that Lemma B.3 implies $\eta_1(n) \asymp s^* \sqrt{\frac{\log d}{n}}$, Lemma B.4 implies $\eta_2(n) \asymp s' \sqrt{\frac{\log d}{n}}$, Lemma B.1 implies $\eta_3(n) \asymp \sqrt{\frac{\log d}{n}}$, Lemma B.2 implies $\eta_4(n) \asymp \sqrt{\frac{\log d}{n}}$, the central limit theorem holds by Lemma B.5 and Lemma B.6 implies $\eta_5(n) \asymp \sqrt{\frac{\log d}{n}}$. In addition, note that $\|\mathbf{w}^*\|_1 = \mathcal{O}(s')$, $|\mathbf{I}_{\theta\gamma_j}^*| = |\mathbb{E}(X_{ij} Z_i)| \leq \{\mathbb{E}(X_{ij}^2)\mathbb{E}(Z_i^2)\}^{1/2} \leq 2C^2$, where $C$ is the constant in Theorem 4.1. Then, after some algebra, the conditions in Theorem 3.5 and Corollary 3.7 hold. This completes the proof. □

To prove Theorem 4.7, we need the following additional lemmas.

**Lemma B.7.** *Under the conditions of Theorem 4.7, it holds that $\lim_{n\to\infty} \inf_{\boldsymbol{\beta}^* \in \Omega_1(\widetilde{C},\phi)} \mathbb{P}_{\boldsymbol{\beta}^*}(\mathcal{F}_6^{\boldsymbol{\beta}^*}) = 1$, where*

$$\mathcal{F}_6^{\boldsymbol{\beta}^*} = \{\sqrt{n}|S(\theta^*,\boldsymbol{\gamma}^*) - S(0,\boldsymbol{\gamma}^*) - \theta^*\mathbf{I}^*_{\theta|\boldsymbol{\gamma}}| \leq Cn^{-\phi}\sqrt{\log n}\},$$

*for some positive constant $C$.*

*Proof.* See Appendix G.2 for a detailed proof. □

**Lemma B.8.** *Under the conditions of Theorem 4.7, it holds that $\lim_{n\to\infty} \inf_{\boldsymbol{\beta}^* \in \Omega_1(\widetilde{C},\phi)} \mathbb{P}_{\boldsymbol{\beta}^*}(\mathcal{F}_3^{\boldsymbol{\beta}^*}) = 1$, where*

$$\mathcal{F}_3^{\boldsymbol{\beta}^*} = \{\|\nabla_{\boldsymbol{\gamma}}\ell(0,\boldsymbol{\gamma}^*)\|_\infty \leq C(\sqrt{n^{-1}\log d} \vee n^{-\phi})\},$$

*for some positive constant $C$.*

*Proof.* See Appendix G.2 for a detailed proof. □

*Proof of Theorem 4.7.* In addition to the results in Theorem 4.1, we also need Lemmas B.7 and B.8. Specifically, Lemma B.7 implies $\eta_6(n) \asymp n^{-\phi}\sqrt{\log n}$ and the uniform noise condition is implied by Lemma B.8 with $\eta_3(n) \asymp \sqrt{n^{-1}\log d} \vee n^{-\phi}$. After some algebra, the conditions in Theorem 3.22 hold, and we finish the proof. □

Finally, we present the proofs of Corollaries 4.2 and 4.3.

*Proof of Corollary 4.2.* Since $\widetilde{\mathbf{w}}$ is essentially a Lasso estimator, similar to the proof of Lemma B.3, we can show that $\eta_2(n) \asymp s'\sqrt{\frac{\log d}{n}}$. In addition, Lemma B.1 implies $\eta_3(n) \asymp \sqrt{\frac{\log d}{n}}$. Thus, $n^{-1/2}\eta_2(n)\eta_3(n) = o_{\mathbb{P}}(1)$. To get the best scaling, one cannot directly control $\eta_1(n)$ and $\eta_4(n)$ in Assumptions 3.1 and 3.3. Instead, we should combine them and directly bound $I_2$ in the proof of Theorem 3.5. In the context of linear models, $I_2$ in the proof of Theorem 3.5 is

$$\frac{1}{n}\sum_{i=1}^n (Z_i - \widehat{\mathbf{w}}^T\boldsymbol{X}_i)\boldsymbol{X}_i^T(\widehat{\boldsymbol{\gamma}} - \boldsymbol{\gamma}^*)$$

$$= \underbrace{\frac{1}{n}\sum_{i=1}^n (Z_i - \mathbf{w}^{*T}\boldsymbol{X}_i)\boldsymbol{X}_i^T(\widehat{\boldsymbol{\gamma}} - \boldsymbol{\gamma}^*)}_{J_1} - \underbrace{\frac{1}{n}\sum_{i=1}^n (\widehat{\mathbf{w}} - \mathbf{w}^*)^T\boldsymbol{X}_i\boldsymbol{X}_i^T(\widehat{\boldsymbol{\gamma}} - \boldsymbol{\gamma}^*)}_{J_2}.$$

The Hoeffding inequality, and Lemma B.3 imply,

$$|J_1| \leq \left\|\frac{1}{n}\sum_{i=1}^n (Z_i - \mathbf{w}^{*T}\boldsymbol{X}_i)\boldsymbol{X}_i\right\|_\infty \cdot \|(\widehat{\boldsymbol{\gamma}} - \boldsymbol{\gamma}^*)\|_1 = \mathcal{O}_{\mathbb{P}}\left(\frac{s^*\log d}{n}\right).$$

The Cauchy-Schwartz inequality and Lemma B.3 imply

$$|J_2| \leq \left|\frac{1}{n}\sum_{i=1}^n ((\widehat{\mathbf{w}} - \mathbf{w}^*)^T\boldsymbol{X}_i)^2\right|^{1/2} \cdot \left|\frac{1}{n}\sum_{i=1}^n ((\widehat{\boldsymbol{\gamma}} - \boldsymbol{\gamma}^*)^T\boldsymbol{X}_i)^2\right|^{1/2} = \mathcal{O}_{\mathbb{P}}\left(\frac{\sqrt{s^*s'}\log d}{n}\right).$$

Hence, $n^{1/2}\cdot I_2$ in the proof of Theorem 3.5 is of order $\mathcal{O}_{\mathbb{P}}(n^{-1/2}(s^*\vee s')\log d) = o_{\mathbb{P}}(1)$. The remaining proof is identical to that of Theorem 4.1. For simplicity, we do not repeat the details. □

*Proof of Corollary 4.3.* Compared to Theorem 4.1, we only need to verify the extra condition $n^{1/2}(\widehat{\theta} - \theta^*)\|\mathbf{w}^*\|_1 \eta_5(n) = o_{\mathbb{P}}(1)$ in Theorem 3.31. Indeed, this is trivial, because this condition comes from the upper bound for $|\bar{I}_{\theta|\gamma} - \widehat{I}_{\theta|\gamma}|$ in the proof of Theorem 3.31. Note that in the linear model, the information matrix does not depend on the parameter. Thus, we have $|\bar{I}_{\theta|\gamma} - \widehat{I}_{\theta|\gamma}| = 0$. This finishes the proof of Corollary 4.3. $\square$

## B.2 Proofs for Optimality

The proof of Theorem 4.9 requires the following auxiliary lemma.

**Lemma B.9.** *Under the same conditions of Theorem 4.9, as $k, n \to \infty$*

$$\frac{1}{n}\mathbf{Z}^T(k\mathbf{X}_S\mathbf{X}_S^T + \sigma^2 \mathbf{I}_n)^{-1}\mathbf{Z} = I^*_{\theta|S} + o_{\mathbb{P}}(1).$$

*Proof.* See Appendix G.2 for a detailed proof. $\square$

*Proof of Theorem 4.9.* Consider the prior distribution for $\boldsymbol{\gamma}_S$, $\pi(\boldsymbol{\gamma}_S) \sim N(0, k\mathbf{I}_{s^*})$, where $\mathbf{I}_s$ is a $s \times s$ identity matrix, $k$ is a sequence of positive constants to be chosen later. Denote $\mathbf{Y} = (Y_1, ..., Y_n)^T$ and $\mathbf{Q} = (\boldsymbol{Q}_1, ..., \boldsymbol{Q}_n)^T$ to be the response and the design matrix, where $\mathbf{Q} \in \mathbb{R}^{n \times d}$. Let $\mathbf{X}_S$ be the $n \times s^*$ submatrix of $\mathbf{Q}$ corresponding to the support set $S$. With the prior $\pi(\cdot)$, we have $\mathbf{Y} \mid \mathbf{Q} \sim N(0, \boldsymbol{\Sigma}_n)$ under $H_0$ and $\mathbf{Y} \mid \mathbf{Q} \sim N(n^{-1/2}\widetilde{C}\mathbf{Z}, \boldsymbol{\Sigma}_n)$ under $H_1$, where $\mathbf{Z} = (Z_1, ..., Z_n)^T$ and $\boldsymbol{\Sigma}_n = k\mathbf{X}_S\mathbf{X}_S^T + \sigma^2 \mathbf{I}_n$. Without loss of generality, we denote the parameters under $H_0$ and $H_1$ by $(0, \boldsymbol{\gamma}_S, 0)$ and $(n^{-1/2}\widetilde{C}, \boldsymbol{\gamma}_S, 0)$. For any test $T$, we have

$$\inf_{\boldsymbol{\beta}^* \in \Omega_1(\widetilde{C}, 1/2)} \mathbb{P}_{\boldsymbol{\beta}^*}(T = 1) \leq \int \mathbb{P}_{(n^{-1/2}\widetilde{C}, \boldsymbol{\gamma}_S, 0)}(T = 1) d\pi(\boldsymbol{\gamma}_S) = \mathbb{E}^1 I(T = 1), \tag{B.2}$$

where $\mathbb{E}^1 A = \int \mathbb{E}_{(n^{-1/2}\widetilde{C}, \boldsymbol{\gamma}_S, 0)} A d\pi(\boldsymbol{\gamma}_S)$ for some random variable $A$. Similarly, denote $\mathbb{E}^0 A = \int \mathbb{E}_{(0, \boldsymbol{\gamma}_S, 0)} A d\pi(\boldsymbol{\gamma}_S)$ to be the expectation of $(\mathbf{Y}, \mathbf{Q})$ under $H_0$. Then

$$\mathbb{E}^1 I(T = 1) = \mathbb{E}^0 \left\{ I(T = 1) \frac{d\mathbb{P}_{(n^{-1/2}\widetilde{C}, \boldsymbol{\gamma}_S, 0)}}{d\mathbb{P}_{(0, \boldsymbol{\gamma}_S, 0)}} \right\},$$

which is maximized when $T = T^{\text{opt}}$, where $T^{\text{opt}} = I\{\frac{d\mathbb{P}_{(n^{-1/2}\widetilde{C}, \boldsymbol{\gamma}_S, 0)}}{d\mathbb{P}_{(0, \boldsymbol{\gamma}_S, 0)}} \geq t'\}$, for some $t'$. Since $\mathbf{Q}$ follows from the same distribution under the null and alternative hypothesis, an equivalent form of $T^{\text{opt}}$ is given by $T^{\text{opt}} = I\{\mathbf{a}^T \mathbf{Y} \geq t\}$, where $\mathbf{a}^T = n^{-1/2}\widetilde{C}\mathbf{Z}^T(k\mathbf{X}_S\mathbf{X}_S^T + \sigma^2 \mathbf{I}_n)^{-1}$, and $t$ is chosen such that the type I error is controlled at level $\alpha$. Specifically, since $\mathbf{a}^T \mathbf{Y} \mid \mathbf{Q} \sim N(0, \mathbf{a}^T \boldsymbol{\Sigma}_n \mathbf{a})$ under $H_0$, by the dominated convergence theorem, we have

$$\lim_{n \to \infty} \mathbb{P}_{(0, \boldsymbol{\gamma}_S, 0)}(\mathbf{a}^T \mathbf{Y} \geq t) = 1 - \lim_{n \to \infty} \mathbb{E}\Phi\left(\frac{t}{\sqrt{\mathbf{a}^T \boldsymbol{\Sigma}_n \mathbf{a}}}\right) = 1 - \mathbb{E}\Phi\left(\lim_{n \to \infty} \frac{t}{\sqrt{\mathbf{a}^T \boldsymbol{\Sigma}_n \mathbf{a}}}\right) \leq \alpha.$$

Together with Lemma B.9, we obtain $t \geq \Phi^{-1}(1 - \alpha)\widetilde{C}I^{*1/2}_{\theta|S}$, asymptotically. Since $\mathbf{a}^T \mathbf{Y} \mid \mathbf{Q} \sim$ $N(n^{-1/2}\widetilde{C}\mathbf{a}^T\mathbf{Z}, \mathbf{a}^T\boldsymbol{\Sigma}_n\mathbf{a})$ under $H_1$, we have,

$$\begin{aligned}
\mathbb{E}^1 I(T=1) &\leq \mathbb{P}_{(n^{-1/2}\widetilde{C},\boldsymbol{\gamma}_S,0)}(\mathbf{a}^T\mathbf{Y} \geq t) = 1 - \mathbb{E}\Phi\left(\frac{t - n^{-1/2}\widetilde{C}\mathbf{a}^T\mathbf{Z}}{\sqrt{\mathbf{a}^T\boldsymbol{\Sigma}_n\mathbf{a}}}\right) \\
&\leq 1 - \mathbb{E}\Phi\left(\Phi^{-1}(1-\alpha)\frac{\widetilde{C}I_{\theta|S}^{*1/2}}{\sqrt{\mathbf{a}^T\boldsymbol{\Sigma}_n\mathbf{a}}} - \frac{n^{-1/2}\widetilde{C}\mathbf{a}^T\mathbf{Z}}{\sqrt{\mathbf{a}^T\boldsymbol{\Sigma}_n\mathbf{a}}}\right) \\
&= 1 - \mathbb{E}\Phi\left(\Phi^{-1}(1-\alpha)\frac{\widetilde{C}I_{\theta|S}^{*1/2}}{\sqrt{\mathbf{a}^T\boldsymbol{\Sigma}_n\mathbf{a}}} - n^{-1/2}\widetilde{C}\sqrt{\mathbf{Z}^T(k\mathbf{X}_S\mathbf{X}_S^T + \sigma^2\mathbf{I}_n)^{-1}\mathbf{Z}}\right).
\end{aligned}$$

As $k, n \to \infty$, by the dominated convergence theorem, and Lemma B.9, we obtain

$$\mathbb{E}^1 I(T=1) \leq 1 - \Phi\left(\Phi^{-1}(1-\alpha) - \widetilde{C}I_{\theta|S}^{*1/2}\right).$$

Together with (B.2), we finish the proof.

$\square$

*Proof of Proposition 4.12.* By Theorem 1 of Wainwright (2009), the support of $\boldsymbol{\gamma}^*$ can be recovered with high probability, i.e., $\text{supp}(\widehat{\boldsymbol{\gamma}}) = \text{supp}(\boldsymbol{\gamma}^*)$. Denote the event $\mathcal{A}$ to be $\mathcal{A} = \{\text{supp}(\widehat{\boldsymbol{\gamma}}) = \text{supp}(\boldsymbol{\gamma}^*)\}$. Given this event, Proposition 4.12 follows directly from the Theorems 4.1 and 4.7. As an illustration, we prove Lemma G.2 in Appendix G.2. The proofs of the remaining Lemmas are essentially the same. To save space, we do not replicate the details. $\square$

## B.3 Proofs of Theorem 4.15 and Corollary 4.16

*Proof of Theorem 4.15.* It is shown that $|\widetilde{U}_n - \widehat{U}_n| = |\widehat{U}_n| \cdot \left|1 - \frac{\sigma^*}{\widehat{\sigma}}\right|$. By the remark following Theorem 4.1, for a sequence of positive constants $t_n \to 0$ to be chosen later, we can show that $\lim_{n\to\infty} \sup_{\boldsymbol{\beta}^* \in \Omega_0} \mathbb{P}_{\boldsymbol{\beta}^*}\left(|\widehat{U}_n| \geq t_n^{-1}\right) = 0$. It remains to show that

$$\lim_{n\to\infty} \sup_{\boldsymbol{\beta}^* \in \Omega_0} \mathbb{P}_{\boldsymbol{\beta}^*}\left(\left|1 - \frac{\sigma^*}{\widehat{\sigma}}\right| \geq t_n\right) = 0. \tag{B.3}$$

Note that

$$\widehat{\sigma}^2 - \sigma^{*2} = \left(\frac{1}{n}\sum_{i=1}^n \epsilon_i^2 - \sigma^{*2}\right) + \widehat{\boldsymbol{\Delta}}^T \mathbf{H}_Q \widehat{\boldsymbol{\Delta}} - 2\widehat{\boldsymbol{\Delta}}^T \frac{1}{n}\sum_{i=1}^n \epsilon_i \boldsymbol{Q}_i, \tag{B.4}$$

where $\widehat{\boldsymbol{\Delta}} = \widehat{\boldsymbol{\beta}} - \boldsymbol{\beta}^*$. Since $\|\epsilon_i^2\|_{\psi_1} \leq 2C^2$, by Lemma H.2, $|n^{-1}\sum_{i=1}^n \epsilon_i^2 - \sigma^{*2}| \leq C\sqrt{\frac{\log n}{n}}$, for some constant $C$, with probability tending to one. For the second term of (B.4), Lemma B.3 yields $\widehat{\boldsymbol{\Delta}}^T \mathbf{H}_Q \widehat{\boldsymbol{\Delta}} \leq C\frac{s^* \log d}{n}$, for some constant $C$, with probability tending to one. Finally, following the similar proof of Lemma B.1, the last term of (B.4) satisfies

$$\left|\widehat{\boldsymbol{\Delta}}^T \frac{1}{n}\sum_{i=1}^n \epsilon_i \boldsymbol{Q}_i\right| \leq \|\widehat{\boldsymbol{\Delta}}\|_1 \cdot \left\|\frac{1}{n}\sum_{i=1}^n \epsilon_i \boldsymbol{Q}_i\right\|_\infty \leq C\frac{s^* \log d}{n},$$

for some constant $C$. Combining these results with equation (B.4), we have $|\widehat{\sigma}^2 - \sigma^{*2}| \leq C\sqrt{\frac{\log n}{n}} \vee \frac{s^* \log d}{n}$, for some constant $C$, with probability tending to one. Note that

$$\left|1 - \frac{\sigma^*}{\widehat{\sigma}}\right| = \widehat{\sigma}^{-2}\left|1 + \frac{\sigma^*}{\widehat{\sigma}}\right|^{-1} \cdot |\widehat{\sigma}^2 - \sigma^{*2}| \lesssim |\widehat{\sigma}^2 - \sigma^{*2}| \lesssim \sqrt{\frac{\log n}{n}} \vee \frac{s^* \log d}{n},$$

in probability, because $\sigma^{*2} \geq C$ for some constant $C > 0$ and $\widehat{\sigma}^2 = \sigma^{*2} + o_{\mathbb{P}}(1)$. Hence, if we take $t_n$ such that $\sqrt{\frac{\log n}{n}} \vee \frac{s^* \log d}{n} \lesssim t_n$, then (B.3) is true, and thus (4.7) holds. By Theorem 4.1, we finish the proof. □

*Proof of Corollary 4.16.* We only need to show that the asymptotic variance of $\widetilde{\theta}$ can be consistently estimated. This is true, i.e., $\widehat{\sigma}^2 = \sigma^{*2} + o_{\mathbb{P}}(1)$, by the proof of Theorem 4.15. Invoking Corollary 4.3 and Slutsky's Theorem, we complete the proof. □

## C  Proofs for the Generalized Linear Model

In this appendix, we present the proofs of Theorems 5.3, 5.4, and 5.5 and Corollary 5.6.

### C.1  Proof of Theorem 5.3

For the reader's convenience, we first present some auxiliary lemmas used to prove Theorem 5.3.

**Lemma C.1.** *Assume that Assumptions 5.1 and 5.2 hold. Then,*

$$\left\|\frac{1}{n}\sum_{i=1}^{n}(-Y_i + b'(\boldsymbol{Q}_i^T\boldsymbol{\beta}^*))\boldsymbol{X}_i\right\|_{\infty} = \mathcal{O}_{\mathbb{P}}\left(\sqrt{\frac{\log d}{n}}\right).$$

*Proof.* See Appendix G.3 for a detailed proof. □

**Lemma C.2.** *Under the Assumption 5.1, it holds that,*

$$\sup_{x \in \mathbb{R}}\left|\mathbb{P}_{\boldsymbol{\beta}^*}(T^* \leq x) - \Phi(x)\right| \leq Cn^{-1/2},$$

*where $T^* = n^{1/2}S(0, \boldsymbol{\gamma}^*)/I_{\theta|\gamma}^{*1/2}$ and $C$ is a positive constant.*

*Proof.* See Appendix G.3 for a detailed proof. □

**Lemma C.3.** *Assume that Assumptions 5.1 and 5.2 hold. Then*

$$\|\widehat{\mathbf{w}} - \mathbf{w}^*\|_1 = \mathcal{O}_{\mathbb{P}}\left(s'\sqrt{\frac{\log d}{n}}\right), \quad \text{and} \quad \frac{1}{n}\sum_{i=1}^{n}b''(\boldsymbol{Q}_i^T\widehat{\boldsymbol{\beta}})\left((\widehat{\mathbf{w}} - \mathbf{w}^*)^T\boldsymbol{X}_i\right)^2 = \mathcal{O}_{\mathbb{P}}\left(\frac{s^* \log d}{n}\right).$$

*Proof.* See Appendix G.3 for a detailed proof. □

*Proof of Theorem 5.3.* Note that, Assumption 5.2 implies $\eta_1(n) \asymp s^*\lambda$, Lemma C.1 implies $\eta_3(n) \asymp \sqrt{\frac{\log d}{n}}$, Lemma C.2 implies the central limit theorem in Assumption 3.4 holds, and Lemma C.3 implies $\eta_2(n) \asymp s'\sqrt{\frac{\log d}{n}}$. In addition, the proof of Lemma C.3 implies $\eta_4(n) \asymp \sqrt{\frac{\log d}{n}}$. In addition, $\|\mathbf{I}^*_{\theta\gamma}\|_\infty \leq K^2$ and $\|\mathbf{I}^*_{\theta\gamma}\|_\infty \eta_2(n) = o(1)$. Finally, by the proof of Corollary 3.7, we only need to show $\mathbf{w}^{*T}(\nabla^2_{\gamma\theta}\ell(\widehat{\boldsymbol{\beta}}) - \mathbf{I}^*_{\gamma\theta}) = o_{\mathbb{P}}(1)$. In the context of generalized linear model, this term equals to

$$\left|\frac{1}{n}\sum_{i=1}^n b''(\boldsymbol{Q}_i^T\widehat{\boldsymbol{\beta}})\mathbf{w}^{*T}\boldsymbol{X}_i Z_i - \mathbf{w}^{*T}\mathbf{I}^*_{\gamma\theta}\right| \leq \underbrace{\left|\frac{1}{n}\sum_{i=1}^n (b''(\boldsymbol{Q}_i^T\widehat{\boldsymbol{\beta}}) - b''(\boldsymbol{Q}_i^T\boldsymbol{\beta}^*))\mathbf{w}^{*T}\boldsymbol{X}_i Z_i\right|}_{I_1}$$
$$+ \underbrace{\left|\frac{1}{n}\sum_{i=1}^n b''(\boldsymbol{Q}_i^T\boldsymbol{\beta}^*)\mathbf{w}^{*T}\boldsymbol{X}_i Z_i - \mathbf{w}^{*T}\mathbf{I}^*_{\gamma\theta}\right|}_{I_2}.$$

To bound the first term $I_1$, we can show that $I_1$ is equal to

$$\frac{1}{n}\sum_{i=1}^n (b''(\boldsymbol{Q}_i^T\widehat{\boldsymbol{\beta}}) - b''(\boldsymbol{Q}_i^T\boldsymbol{\beta}^*))\mathbf{w}^{*T}\boldsymbol{X}_i Z_i = \frac{1}{n}\sum_{i=1}^n \frac{b''(\boldsymbol{Q}_i^T\widehat{\boldsymbol{\beta}}) - b''(\boldsymbol{Q}_i^T\boldsymbol{\beta}^*)}{\boldsymbol{Q}_i^T(\widehat{\boldsymbol{\beta}} - \boldsymbol{\beta}^*)}\boldsymbol{Q}_i^T(\widehat{\boldsymbol{\beta}} - \boldsymbol{\beta}^*)\mathbf{w}^{*T}\boldsymbol{X}_i Z_i,$$

Hence, by the Hölder inequality,

$$|I_1| \leq \left(\frac{1}{n}\sum_{i=1}^n \left(\frac{b''(\boldsymbol{Q}_i^T\widehat{\boldsymbol{\beta}}) - b''(\boldsymbol{Q}_i^T\boldsymbol{\beta}^*)}{\boldsymbol{Q}_i^T(\widehat{\boldsymbol{\beta}} - \boldsymbol{\beta}^*)}\right)^2 (\widehat{\boldsymbol{\beta}} - \boldsymbol{\beta}^*)\boldsymbol{Q}_i\boldsymbol{Q}_i^T(\widehat{\boldsymbol{\beta}} - \boldsymbol{\beta}^*)(\mathbf{w}^{*T}\boldsymbol{X}_i Z_i)^2\right)^{1/2}$$
$$\leq \left(\frac{1}{n}\sum_{i=1}^n C^2 K^4 (\boldsymbol{\beta}^* - \widehat{\boldsymbol{\beta}})\boldsymbol{Q}_i\boldsymbol{Q}_i^T(\boldsymbol{\beta}^* - \widehat{\boldsymbol{\beta}})\right)^{1/2} \leq CC'^{1/2}K^2 s^{*1/2}\lambda.$$

This implies that $|I_1| = \mathcal{O}_{\mathbb{P}}(s^{*1/2}\lambda) = o_{\mathbb{P}}(1)$. For the second term $I_2$, by the law of large numbers, we have $|I_2| = o_{\mathbb{P}}(1)$. These together imply $\mathbf{w}^{*T}(\nabla^2_{\gamma\theta}\ell(\widehat{\boldsymbol{\beta}}) - \mathbf{I}^*_{\gamma\theta}) = o_{\mathbb{P}}(1)$. This completes the proof. □

## C.2 Proof of Theorem 5.4

We introduce the following Lemma to prove Theorem 5.4.

**Lemma C.4.** Under the same conditions as in Theorem 5.4, we obtain

$$\|\widetilde{\mathbf{w}} - \mathbf{w}^*\|_1 = \mathcal{O}_{\mathbb{P}}\left((s' \vee s^*)\sqrt{\frac{\log d}{n}}\right), \quad \text{and} \tag{C.1}$$

$$\frac{1}{n}\sum_{i=1}^n b''(\boldsymbol{Q}_i^T\widehat{\boldsymbol{\beta}})((\widetilde{\mathbf{w}} - \mathbf{w}^*)^T \boldsymbol{X}_i)^2 = \mathcal{O}_{\mathbb{P}}\left(\frac{(s^* \vee s')\log d}{n}\right). \tag{C.2}$$

*Proof.* See Appendix G.3 for a detailed proof. □

*Proof of Theorem 5.4.* By Assumption 5.2, we have $\eta_1(n) \asymp s^*\lambda$. In addition, Lemma C.1 implies $\eta_3(n) \asymp \sqrt{\frac{\log d}{n}}$, and Lemma C.2 implies that the central limit theorem in Assumption 3.4 holds. Moreover, Lemma C.4 implies $\eta_2(n) \asymp (s^* \vee s')\sqrt{\frac{\log d}{n}}$. We comment that if you directly plug these rates into the conditions in Theorem 3.5, we can only obtain a suboptimal scaling on $s^*, s', d$ and $n$. To get a better scaling, we conduct a refined analysis. In particular, we do not control $\eta_1(n)$ and $\eta_4(n)$ in Assumptions 3.1 and 3.3, separately. Instead, we combine them and directly bound $I_2$ in the proof of Theorem 3.5. In the context of generalized linear models, $I_2$ in the proof of Theorem 3.5 is

$$
\begin{aligned}
I_2 &= \frac{1}{n}\sum_{i=1}^{n} b''(\boldsymbol{Q}_i^T\widehat{\boldsymbol{\beta}})(Z_i - \widetilde{\mathbf{w}}^T\boldsymbol{X}_i)\boldsymbol{X}_i^T(\widehat{\boldsymbol{\gamma}} - \boldsymbol{\gamma}^*) \\
&= \underbrace{\frac{1}{n}\sum_{i=1}^{n} b''(\boldsymbol{Q}_i^T\widehat{\boldsymbol{\beta}})(Z_i - \mathbf{w}^{*T}\boldsymbol{X}_i)\boldsymbol{X}_i^T(\widehat{\boldsymbol{\gamma}} - \boldsymbol{\gamma}^*)}_{J_1} - \underbrace{\frac{1}{n}\sum_{i=1}^{n} b''(\boldsymbol{Q}_i^T\widehat{\boldsymbol{\beta}})(\widetilde{\mathbf{w}} - \mathbf{w}^*)^T\boldsymbol{X}_i\boldsymbol{X}_i^T(\widehat{\boldsymbol{\gamma}} - \boldsymbol{\gamma}^*)}_{J_2}.
\end{aligned}
$$

The triangle inequality implies,

$$
|J_1| \leq \underbrace{\left|\frac{1}{n}\sum_{i=1}^{n} b''(\boldsymbol{Q}_i^T\boldsymbol{\beta}^*)(Z_i - \mathbf{w}^{*T}\boldsymbol{X}_i)\boldsymbol{X}_i^T(\widehat{\boldsymbol{\gamma}} - \boldsymbol{\gamma}^*)\right|}_{J_{11}} + \underbrace{\left|\frac{1}{n}\sum_{i=1}^{n}(b''(\boldsymbol{Q}_i^T\widehat{\boldsymbol{\beta}}) - b''(\boldsymbol{Q}_i^T\boldsymbol{\beta}^*))(Z_i - \mathbf{w}^{*T}\boldsymbol{X}_i)\boldsymbol{X}_i^T(\widehat{\boldsymbol{\gamma}} - \boldsymbol{\gamma}^*)\right|}_{J_{12}}.
$$

The Hoeffding inequality, and Assumption 5.2, yield that the first term of $|J_1|$ is less than

$$
|J_{11}| \leq \left\|\frac{1}{n}\sum_{i=1}^{n} b''(\boldsymbol{Q}_i^T\boldsymbol{\beta}^*)(Z_i - \mathbf{w}^{*T}\boldsymbol{X}_i)\boldsymbol{X}_i^T\right\|_{\infty} \cdot \|(\widehat{\boldsymbol{\gamma}} - \boldsymbol{\gamma}^*)\|_1 = \mathcal{O}_{\mathbb{P}}\left(s^*\frac{\log d}{n}\right).
$$

Following the similar arguments to the proof of Theorem 5.3, by the Cauchy-Schwartz inequality,

$$
|J_{12}| \lesssim \left|\frac{1}{n}\sum_{i=1}^{n}((\widehat{\boldsymbol{\gamma}} - \boldsymbol{\gamma}^*)^T\boldsymbol{X}_i)^2\right|^{1/2} \cdot \left|\frac{1}{n}\sum_{i=1}^{n}((\widehat{\boldsymbol{\beta}} - \boldsymbol{\beta}^*)^T\boldsymbol{Q}_i)^2\right|^{1/2} = \mathcal{O}_{\mathbb{P}}\left(s^*\frac{\log d}{n}\right).
$$

Moreover, by the Cauchy-Schwartz inequality, Assumption 5.2, and Lemma C.4,

$$
|J_2| \leq \left|\frac{1}{n}\sum_{i=1}^{n} b''(\boldsymbol{Q}_i^T\widehat{\boldsymbol{\beta}})((\widetilde{\mathbf{w}}-\mathbf{w}^*)^T\boldsymbol{X}_i)^2\right|^{1/2} \cdot \left|\frac{1}{n}\sum_{i=1}^{n} b''(\boldsymbol{Q}_i^T\widehat{\boldsymbol{\beta}})((\widehat{\boldsymbol{\gamma}}-\boldsymbol{\gamma}^*)^T\boldsymbol{X}_i)^2\right|^{1/2} = \mathcal{O}_{\mathbb{P}}\left(\frac{(s^* \vee s)\log d}{n}\right).
$$

Combining the bounds for $J_1$ and $J_2$, we can show that $n^{1/2} \cdot I_2$ in the proof of Theorem 3.5 is of order $\mathcal{O}_{\mathbb{P}}(n^{-1/2}(s^* \vee s')\log d) = o_{\mathbb{P}}(1)$. Similar to the proof of Theorem 5.3, we can show that $\mathbf{w}^{*T}(\nabla^2_{\gamma\theta}\ell(\widehat{\boldsymbol{\beta}}) - \mathbf{I}^*_{\gamma\theta}) = o_{\mathbb{P}}(1)$. This completes the proof. $\square$

## C.3 Proof of Theorem 5.5

**Lemma C.5.** Under the same conditions as in Theorem 5.5, we obtain

$$||\bar{\mathbf{w}} - \mathbf{w}^*||_1 = \mathcal{O}_\mathbb{P}\left((s' \vee s^*)\sqrt{\frac{\log d}{n}}\right), \quad \text{and} \quad \frac{1}{n}\sum_{i=1}^n ((\widetilde{\mathbf{w}} - \mathbf{w}^*)^T \mathbf{X}_i)^2 = \mathcal{O}_\mathbb{P}\left(\frac{(s^* \vee s')\log d}{n}\right). \quad (C.3)$$

*Proof.* See Appendix G.3 for a detailed proof. □

*Proof of Theorem 5.5.* By Assumption 5.2, we have $\eta_1(n) \asymp s^*\sqrt{\frac{\log d}{n}}$. In addition, Lemma C.1 implies $\eta_3(n) \asymp \sqrt{\frac{\log d}{n}}$, and Lemma C.2 implies that the central limit theorem in Assumption 3.4 holds. Moreover, Lemma C.5 implies $\eta_2(n) \asymp (s^* \vee s')\sqrt{\frac{\log d}{n}}$. Similar to the proof of Theorem 5.4, we conduct a refined analysis to get the best scaling. Here, we directly bound $I_2$ in the proof of Theorem 3.5. In the context of generalized linear models, $I_2$ in the proof of Theorem 3.5 is

$$I_2 = \underbrace{\frac{1}{n}\sum_{i=1}^n b''(\mathbf{Q}_i^T\widehat{\boldsymbol{\beta}})(Z_i - \mathbf{w}^{*T}\mathbf{X}_i)\mathbf{X}_i^T(\widehat{\boldsymbol{\gamma}} - \boldsymbol{\gamma}^*)}_{J_1} - \underbrace{\frac{1}{n}\sum_{i=1}^n b''(\mathbf{Q}_i^T\widehat{\boldsymbol{\beta}})(\bar{\mathbf{w}} - \mathbf{w}^*)^T\mathbf{X}_i\mathbf{X}_i^T(\widehat{\boldsymbol{\gamma}} - \boldsymbol{\gamma}^*)}_{J_2}.$$

Denote $b''_i = b''(\mathbf{Q}_i^T\boldsymbol{\beta}^*)$ and $\widehat{b}''_i = b''(\mathbf{Q}_i^T\widehat{\boldsymbol{\beta}})$. Furthermore, let

$$J_{11} = \frac{1}{n}\sum_{i=1}^n b''_i(Z_i - \mathbf{w}^{*T}\mathbf{X}_i)\mathbf{X}_i^T(\widehat{\boldsymbol{\gamma}} - \boldsymbol{\gamma}^*), \quad \text{and} \quad J_{12} = \frac{1}{n}\sum_{i=1}^n (\widehat{b}''_i - b''_i)(Z_i - \mathbf{w}^{*T}\mathbf{X}_i)\mathbf{X}_i^T(\widehat{\boldsymbol{\gamma}} - \boldsymbol{\gamma}^*).$$

Similar to the proof of Theorem 5.4, we get $|J_{11}| = \mathcal{O}_\mathbb{P}(s^*\frac{\log d}{n})$ and $|J_{12}| = \mathcal{O}_\mathbb{P}(s^*\frac{\log d}{n})$. The triangle inequality implies $|J_1| \leq |J_{11}| + |J_{12}| = \mathcal{O}_\mathbb{P}(s^*\frac{\log d}{n})$. Moreover, by the Cauchy-Schwartz inequality, Assumption 5.2, and Lemma C.5,

$$|J_2| \leq \left|\frac{1}{n}\sum_{i=1}^n b''(\mathbf{Q}_i^T\widehat{\boldsymbol{\beta}})((\bar{\mathbf{w}} - \mathbf{w}^*)^T\mathbf{X}_i)^2\right|^{1/2} \cdot \left|\frac{1}{n}\sum_{i=1}^n b''(\mathbf{Q}_i^T\widehat{\boldsymbol{\beta}})((\widehat{\boldsymbol{\gamma}} - \boldsymbol{\gamma}^*)^T\mathbf{X}_i)^2\right|^{1/2} = \mathcal{O}_\mathbb{P}\left(\frac{(s^* \vee s)\log d}{n}\right).$$

Here, in the last step, we use the fact that $b''(\mathbf{Q}_i^T\widehat{\boldsymbol{\beta}}) \leq C$ for some constant $C$ asymptotically. Combining the bounds for $J_1$ and $J_2$, we can show that $n^{1/2} \cdot I_2$ in the proof of Theorem 3.5 is of order $\mathcal{O}_\mathbb{P}(n^{-1/2}(s^* \vee s')\log d) = o_\mathbb{P}(1)$. Similar to the proof of Theorem 5.3, we can show that $\mathbf{w}^{*T}(\nabla^2_{\gamma\theta}\ell(\widehat{\boldsymbol{\beta}}) - \mathbf{I}^*_{\gamma\theta}) = o_\mathbb{P}(1)$. This completes the proof. □

## C.4 Proof of Corollary 5.6

*Proof of Corollary 5.6.* To get a better scaling, instead of directly verifying the extra condition $n^{1/2}(\widehat{\theta} - \theta^*)||\mathbf{w}^*||_1\eta_5(n) = o_\mathbb{P}(1)$ in Theorem 3.31, we consider the upper bound for $|\bar{I}_{\theta|\gamma} - \widehat{I}_{\theta|\gamma}|$ needed in the proof of Theorem 3.31. It is seen that

$$|\bar{I}_{\theta|\gamma} - \widehat{I}_{\theta|\gamma}| \leq \left|\frac{1}{n}\sum_{i=1}^n (b''(\mathbf{Q}_i^T\widehat{\boldsymbol{\beta}}) - b''(\mathbf{Q}_i^T\bar{\boldsymbol{\beta}}))Z_i^2\right| + \left|\frac{1}{n}\sum_{i=1}^n (b''(\mathbf{Q}_i^T\widehat{\boldsymbol{\beta}}) - b''(\mathbf{Q}_i^T\bar{\boldsymbol{\beta}}))Z_i\widehat{\mathbf{w}}^T\mathbf{X}_i\right|, \quad (C.4)$$

where $\bar{\boldsymbol{\beta}}$ lies between $\boldsymbol{\beta}^*$ and $\widehat{\boldsymbol{\beta}}$. Using the similar arguments for bounding $J_{12}$ in the proof of Theorem 5.4, we can get the right hand side of the inequality (C.4) is of order $\mathcal{O}_{\mathbb{P}}\big(\sqrt{\frac{(s^*\vee s)\log d}{n}}\big)$. Since $n^{1/2}(\widehat{\theta} - \theta^*) \leq n^{1/2}\|(\widehat{\boldsymbol{\beta}} - \boldsymbol{\beta}^*)\|_2 = \mathcal{O}_{\mathbb{P}}(\sqrt{s^*\log d})$ by Assumption 5.2, therefore, $n^{1/2}(\widehat{\theta} - \theta^*)|\bar{I}_{\theta|\gamma} - \widehat{I}_{\theta|\gamma}| = o_{\mathbb{P}}(1)$, which completes the proof. $\square$

## D Proofs for High Dimensional Null Hypothesis

In this appendix, we first present the general theorem and then illustrate the consequences of the theorem for the linear model. Finally, we present the results for the rescaled test statistic.

### D.1 Proof of Theorem 6.6

Recall that $\widehat{\mathbf{T}} = \sqrt{n}\widehat{\mathbf{S}}(\mathbf{0}, \widehat{\boldsymbol{\gamma}})$, and $\mathbf{T}^* = \sqrt{n}\mathbf{S}(\mathbf{0}, \boldsymbol{\gamma}^*)$. The following lemma characterizes the distribution of the test statistic $\|\widehat{\mathbf{T}}\|_\infty$.

**Lemma D.1.** Assume that Assumptions 6.1–6.4 hold. Assume that $q(n)\sqrt{1 \vee \log(d_0/q(n))} = o(1)$, where $q(n) = n^{1/2}(\eta_2(n)\eta_3(n) \vee \eta_1(n)\eta_4(n))$. If $(\log(d_0 n))^7/n = o(1)$, then,

$$\lim_{n \to \infty} \sup_{t \in \mathbb{R}} \left| \mathbb{P}_{\boldsymbol{\beta}^*}(\|\widehat{\mathbf{T}}\|_\infty \leq t) - \mathbb{P}(\|\mathbf{N}\|_\infty \leq t) \right| = 0,$$

where $\mathbf{N} \sim N_{d_0}(\mathbf{0}, \mathbb{E}_{\boldsymbol{\beta}^*}(\mathbf{T}^{*\otimes 2}))$.

*Proof.* See Appendix G.4 for a detailed proof. $\square$

Recall that

$$\widehat{\mathbf{N}}_e = \frac{1}{\sqrt{n}} \sum_{i=1}^n e_i(\nabla_{\boldsymbol{\theta}}\ell_i(\mathbf{0}, \widehat{\boldsymbol{\gamma}}) - \widehat{\mathbf{W}}^T\nabla_{\boldsymbol{\gamma}}\ell(\mathbf{0}, \widehat{\boldsymbol{\gamma}})),$$

$$\mathbf{N}_e = \frac{1}{\sqrt{n}} \sum_{i=1}^n e_i(\nabla_{\boldsymbol{\theta}}\ell_i(\mathbf{0}, \boldsymbol{\gamma}^*) - \mathbf{W}^{*T}\nabla_{\boldsymbol{\gamma}}\ell(\mathbf{0}, \boldsymbol{\gamma}^*)),$$

where $e_i \sim N(0,1)$. Let $\mathbb{P}_e(\mathcal{A})$ denote the probability of the event $\mathcal{A}$ with respect to $e_1,...,e_n$

**Lemma D.2.** Under the conditions of Lemma D.1 and Assumption 6.5, as $n \to \infty$,

$$\mathbb{P}_{\boldsymbol{\beta}^*}\bigg(\big|\|\widehat{\mathbf{T}}\|_\infty - \|\mathbf{T}^*\|_\infty\big| \geq Cq(n)\bigg) \to 0, \text{ and } \mathbb{P}_e\bigg(\big|\|\widehat{\mathbf{N}}_e\|_\infty - \|\mathbf{N}_e\|_\infty\big| \geq Cq'(n)\bigg) \to_p 0,$$

where $q(n) = n^{1/2}(\eta_2(n)\eta_3(n) \vee \eta_1(n)\eta_4(n))$ and $q'(n) = \eta_7(n)\sqrt{\log d_0}$, for some constant $C$ sufficiently large.

*Proof.* See Appendix G.4 for a detailed proof. $\square$

*Proof of Theorem 6.6.* The proof follows from the general results for the multiplier bootstrap in Chernozhukov et al. (2013). Here, we adapt their result and rewrite it in a suitable form for our analysis. It is now presented in Lemma H.7 of Appendix H. With this lemma and Lemmas D.1, and D.2, we complete the proof. $\square$

## D.2 Proofs for the Linear Model

To apply Theorem 6.6 to the linear model, we need the following auxiliary lemmas, whose proofs are deferred to Appendix G.4.

**Lemma D.3.** *Under the conditions of Corollary 6.8, with probability at least $1 - d^{-1}$,*

$$\max_{1 \leq j \leq d_0} \|\widehat{\mathbf{W}}_{*j} - \mathbf{W}^*_{*j}\|_1 \leq Cs'\sqrt{\frac{\log d}{n}},$$

*for some constant $C > 0$.*

*Proof.* See Appendix G.4 for a detailed proof. □

**Lemma D.4.** *Under the conditions of Corollary 6.8, with probability at least $1 - d^{-1}$,*

$$\max_{1 \leq j \leq d_0} \left\| \frac{1}{n} \sum_{i=1}^n (Z_{ij}\mathbf{X}_i - \widehat{\mathbf{W}}_{*j}^T \mathbf{X}_i^{\otimes 2}) \right\|_\infty \leq C\sqrt{\frac{\log d}{n}},$$

*for some constant $C > 0$.*

*Proof.* See Appendix G.4 for a detailed proof. □

**Lemma D.5.** *Recall that $\widehat{S}_{ij} = (Y_i - \widehat{\boldsymbol{\gamma}}^T \mathbf{X}_i)(Z_{ij} - \widehat{\mathbf{W}}_{*j}^T \mathbf{X}_i)$, and $S_{ij} = \epsilon_i(Z_{ij} - \mathbf{W}_{*j}^{*T}\mathbf{X}_i)$. If $(s^* \vee s')\sqrt{\frac{(\log(nd))^3}{n}} = o(1)$, then*

$$\max_{1 \leq j \leq d_0} \sqrt{\frac{1}{n} \sum_{i=1}^n (\widehat{S}_{ij} - S_{ij})^2} = \mathcal{O}_{\mathbb{P}}\left((s^* \vee s')\sqrt{\frac{(\log(nd))^3}{n}}\right),$$

*Proof.* See Appendix G.4 for a detailed proof. □

Combining Lemmas D.3, D.4 and D.5, we can prove Corollary 6.8.

*Proof of Corollary 6.8.* It remains to check the validity of the conditions in Theorem 6.6. Note that Lemma B.3 implies $\eta_1(n) \asymp s^*\sqrt{\frac{\log d}{n}}$, Lemma D.3 implies $\eta_2(n) \asymp s'\sqrt{\frac{\log d}{n}}$, Lemma B.1 implies $\eta_3(n) \asymp \sqrt{\frac{\log d}{n}}$, Lemma D.4 implies $\eta_4(n) \asymp \sqrt{\frac{\log d}{n}}$. Recall that $S_{ij} = \epsilon_i(Z_{ij} - \mathbf{W}_{*j}^{*T}\mathbf{X}_i)$. Therefore, $\|S_{ij}\|_{\psi_1} \leq 2C^2$. In addition, the $d_0 \times d_0$ matrix $\mathbb{E}_{\boldsymbol{\beta}^*}(\mathbf{S}_i^{\otimes 2})$ satisfies $\mathbb{E}_{\boldsymbol{\beta}^*}(\mathbf{S}_i^{\otimes 2}) = \sigma^2[\mathbb{E}(\mathbf{Z}_i^{\otimes 2}) - \mathbb{E}(\mathbf{Z}_i\mathbf{X}_i)\{\mathbb{E}(\mathbf{X}_i^{\otimes 2})\}^{-1}\mathbb{E}(\mathbf{X}_i\mathbf{Z}_i)]$. Since $\{\mathbb{E}_{\boldsymbol{\beta}^*}(\mathbf{S}_i^{\otimes 2})\}^{-1} = \sigma^{-2}\{\mathbb{E}(\mathbf{Q}_i^{\otimes 2})^{-1}\}_{\theta\theta}$, we get $\mathbb{E}_{\boldsymbol{\beta}^*}(S_{ij}^2) \geq \sigma^2\lambda_{\min}(\mathbb{E}(\mathbf{Q}_i^{\otimes 2})) \geq C$, for some constant $C > 0$. Finally, Lemma D.5 implies that $\eta_7(n) \asymp (s^* \vee s')\sqrt{\frac{(\log(nd))^3}{n}}$. After some algebra, we obtain the results. □

### D.3 Theoretical Properties for Rescaled Test Statistic $\|\widehat{\mathbf{T}}_R\|_\infty$

Recall that $\widehat{\mathbf{D}} = \text{diag}((\widehat{\mathbf{I}}_{\boldsymbol{\theta}|\boldsymbol{\gamma}})_{11}, ..., (\widehat{\mathbf{I}}_{\boldsymbol{\theta}|\boldsymbol{\gamma}})_{d_0 d_0})$, where $\widehat{\mathbf{I}}_{\boldsymbol{\theta}|\boldsymbol{\gamma}} = \nabla^2_{\boldsymbol{\theta}\boldsymbol{\theta}}\ell(\widehat{\boldsymbol{\beta}}) - \widehat{\mathbf{W}}^T \nabla^2_{\boldsymbol{\gamma}\boldsymbol{\theta}}\ell(\widehat{\boldsymbol{\beta}})$. The rescaled test statistic is given by $\|\widehat{\mathbf{T}}_R\|_\infty$, where $\widehat{\mathbf{T}}_R = \sqrt{n}\widehat{\mathbf{D}}^{-1/2}\widehat{\mathbf{S}}(\mathbf{0}, \widehat{\boldsymbol{\gamma}})$. Recall that $\mathbf{I}^*_{\boldsymbol{\theta}|\boldsymbol{\gamma}} = \mathbf{I}^*_{\boldsymbol{\theta}\boldsymbol{\theta}} - \mathbf{I}^*_{\boldsymbol{\theta}\boldsymbol{\gamma}}\mathbf{I}^{*-1}_{\boldsymbol{\gamma}\boldsymbol{\gamma}}\mathbf{I}^*_{\boldsymbol{\gamma}\boldsymbol{\theta}}$, and $\mathbf{D}^* = \text{diag}((\mathbf{I}^*_{\boldsymbol{\theta}|\boldsymbol{\gamma}})_{11}, ..., (\mathbf{I}^*_{\boldsymbol{\theta}|\boldsymbol{\gamma}})_{d_0 d_0})$. Denote $\mathbf{T}^*_R = \sqrt{n}\mathbf{D}^{*-1/2}\mathbf{S}(\mathbf{0}, \boldsymbol{\gamma}^*)$. The following lemma characterizes the distribution of the test statistic $\|\widehat{\mathbf{T}}_R\|_\infty$.

**Lemma D.6.** Assume that Assumptions 6.1–6.4, and 3.6 hold. Assume that

$$\left(q(n) + r(n)\right)\sqrt{1 \vee \log\left(\frac{d_0}{q(n) + r(n)}\right)} = o(1),$$

where $q(n) = n^{1/2}(\eta_2(n)\eta_3(n) \vee \eta_1(n)\eta_4(n))$ and $r(n) = (C_W \eta_5(n) + C_I \eta_2(n))\log d_0$. Here, $C_W = \max_{1\leq j\leq d_0} \|\mathbf{W}^*_{*j}\|_1$ and $C_I = \max_{1\leq j\leq d_0} \|\mathbf{I}^*_{\boldsymbol{\theta}_j \boldsymbol{\gamma}}\|_\infty$. If $(\log(d_0 n))^7 / n = o(1)$, then,

$$\lim_{n\to\infty} \sup_{t\in\mathbb{R}} \left|\mathbb{P}_{\boldsymbol{\beta}^*}(\|\widehat{\mathbf{T}}_R\|_\infty \leq t) - \mathbb{P}(\|\mathbf{N}_R\|_\infty \leq t)\right| = 0,$$

where $\mathbf{N}_R \sim N_{d_0}(\mathbf{0}, \mathbb{E}_{\boldsymbol{\beta}^*}(\mathbf{T}^{*\otimes 2}_R))$.

*Proof.* See Appendix G.4 for a detailed proof. □

Denote the rescaled bootstrap statistics as

$$\widehat{\mathbf{N}}^R_e = \frac{1}{\sqrt{n}} \sum_{i=1}^n e_i \widehat{\mathbf{D}}^{-1/2}(\nabla_{\boldsymbol{\theta}}\ell_i(\mathbf{0}, \widehat{\boldsymbol{\gamma}}) - \widehat{\mathbf{W}}^T \nabla_{\boldsymbol{\gamma}}\ell(\mathbf{0}, \widehat{\boldsymbol{\gamma}})),$$

$$\mathbf{N}^R_e = \frac{1}{\sqrt{n}} \sum_{i=1}^n e_i \mathbf{D}^{*-1/2}(\nabla_{\boldsymbol{\theta}}\ell_i(\mathbf{0}, \boldsymbol{\gamma}^*) - \mathbf{W}^{*T}\nabla_{\boldsymbol{\gamma}}\ell(\mathbf{0}, \boldsymbol{\gamma}^*)),$$

where $e_i \sim N(0,1)$. Let $\mathbb{P}_e(\mathcal{A})$ denote the probability of the event $\mathcal{A}$ with respect to $e_1, ..., e_n$

**Lemma D.7.** Under the conditions of Lemma D.6 and Assumption 6.5, as $n \to \infty$,

$$\mathbb{P}_{\boldsymbol{\beta}^*}\left(\left|\|\widehat{\mathbf{T}}_R\|_\infty - \|\mathbf{T}^*_R\|_\infty\right| \geq C(q(n)+r(n))\right) \to 0, \quad \mathbb{P}_e\left(\left|\|\widehat{\mathbf{N}}^R_e\|_\infty - \|\mathbf{N}^R_e\|_\infty\right| \geq C(q'(n)+r'(n))\right) \to_p 0,$$

where $q(n), r(n)$ are given in Lemma D.6, $q'(n) = \eta_7(n)\sqrt{\log d_0}$ and $r'(n) = r(n)\sqrt{\log d_0}$, for some constant $C$ sufficiently large.

*Proof.* See Appendix G.4 for a detailed proof. □

We now state the main results for the rescaled test statistic $\|\widehat{\mathbf{T}}_R\|_\infty$.

**Theorem D.8.** Assume that Assumptions 6.1–6.5, 3.6, and

$$(q(n) \vee q'(n) \vee r'(n))\left(1 \vee \log \frac{d_0}{q(n) \vee q'(n) \vee r'(n)}\right)^{1/2} = o(1), \tag{D.1}$$

hold, where $q(n) = n^{1/2}(\eta_2(n)\eta_3(n) \vee \eta_1(n)\eta_4(n))$, $q'(n) = \eta_7(n)\sqrt{\log d_0}$ and $r'(n) = (C_W\eta_5(n) + C_I\eta_2(n))(\log d_0)^{3/2}$. Here, $C_W = \max_{1\leq j\leq d_0}||\mathbf{W}^*_{*j}||_1$ and $C_I = \max_{1\leq j\leq d_0}||\mathbf{I}^*_{\theta_j\gamma}||_\infty$. Moreover, under $(\log(d_0 n))^9/n = o(1)$, we obtain

$$\lim_{n\to\infty} \sup_{\alpha\in(0,1)} \left|\mathbb{P}_{\boldsymbol{\beta}^*}\left(||\widehat{\mathbf{T}}_R||_\infty \leq c'^R_N(\alpha)\right) - \alpha\right| = 0,$$

where

$$c'^R_N(\alpha) = \inf\{t \in \mathbb{R} : \mathbb{P}_e(||\widehat{\mathbf{N}}^R_e||_\infty \leq t) \geq \alpha\}. \tag{D.2}$$

*Proof of Theorem D.8.* Similar to the proof of Theorem 6.6, the proof follows from Lemmas H.7, D.6, and D.7. To save space, we do not replicate the details. □

Let us consider the linear regression example. Recall that

$$\widehat{\mathbf{S}}(\mathbf{0},\widehat{\boldsymbol{\gamma}}) = -\frac{1}{n}\sum_{i=1}^n (Y_i - \widehat{\boldsymbol{\gamma}}^T \mathbf{X}_i)(\mathbf{Z}_i - \widehat{\mathbf{W}}^T\mathbf{X}_i),$$

where

$$\widehat{\mathbf{W}}_{*j} = \operatorname{argmin} ||\mathbf{w}||_1, \quad \text{s.t.} \quad \left\|\frac{1}{n}\sum_{i=1}^n \mathbf{X}_i\left(Z_{ij} - \mathbf{w}^T\mathbf{X}_i\right)\right\|_\infty \leq \lambda'.$$

We have $\widehat{\mathbf{D}} = \operatorname{diag}((\widehat{\mathbf{I}}_{\boldsymbol{\theta}|\boldsymbol{\gamma}})_{11}, ..., (\widehat{\mathbf{I}}_{\boldsymbol{\theta}|\boldsymbol{\gamma}})_{d_0 d_0})$, where $\widehat{\mathbf{I}}_{\boldsymbol{\theta}|\boldsymbol{\gamma}} = n^{-1}\sum_{i=1}^n \mathbf{Z}_i^{\otimes 2} - \widehat{\mathbf{W}}^T(n^{-1}\sum_{i=1}^n \mathbf{Z}_i\mathbf{X}_i)$. The rescaled test statistic is $||\widehat{\mathbf{T}}_R||_\infty$, where $\widehat{\mathbf{T}}_R = \sqrt{n}\widehat{\mathbf{D}}^{-1/2}\widehat{\mathbf{S}}(\mathbf{0},\widehat{\boldsymbol{\gamma}})$. In this case, the multiplier bootstrapped statistic is

$$\widehat{\mathbf{N}}^R_e = \frac{1}{\sqrt{n}}\sum_{i=1}^n e_i\widehat{\mathbf{D}}^{-1/2}(Y_i - \widehat{\boldsymbol{\gamma}}^T \mathbf{X}_i)(\mathbf{Z}_i - \widehat{\mathbf{W}}^T\mathbf{X}_i),$$

where $e_i \sim N(0,1)$. The implication of Theorem D.8 to the high dimensional linear regression is given in the following corollary.

**Corollary D.9.** Assume that (1) $\lambda_{\min}(\mathbb{E}(\boldsymbol{Q}_i^{\otimes 2})) \geq 2\kappa$ for some constant $\kappa > 0$, (2) $\|\mathbf{w}^*\|_0 = s'$ and $\|\boldsymbol{\beta}^*\|_0 = s^*$, (3) $\epsilon_i$, $\mathbf{W}^{*T}_{*j}\mathbf{X}_i$, $Q_{ij}$ are all sub-Gaussian with $\|\epsilon_i\|_{\psi_2} \leq C$, $\|\mathbf{W}^{*T}_{*j}\mathbf{X}_i\|_{\psi_2} \leq C$ and $\|Q_{ij}\|_{\psi_2} \leq C$, where $C$ is a positive constant. In addition, assume that $\sigma^{*2} \geq C$. If $n^{-1/2}(s' \vee s^*)(\log(nd))^{3/2}\sqrt{\log d_0} = o(1)$, $(\log(d_0 n))^9/n = o(1)$ and $\lambda \asymp \lambda' \asymp \sqrt{\frac{\log d}{n}}$, then

$$\lim_{n\to\infty} \sup_{\alpha\in(0,1)} \left|\mathbb{P}_{\boldsymbol{\beta}^*}\left(||\widehat{\mathbf{T}}_R||_\infty \leq c'^R_N(\alpha)\right) - \alpha\right| = 0,$$

where $c'^R_N(\alpha)$ is defined in (D.2).

*Proof of Corollary D.9.* Note that Lemma B.3 implies $\eta_1(n) \asymp s^*\sqrt{\frac{\log d}{n}}$, Lemma D.3 implies $\eta_2(n) \asymp s'\sqrt{\frac{\log d}{n}}$, Lemma B.1 implies $\eta_3(n) \asymp \sqrt{\frac{\log d}{n}}$, and Lemma D.4 implies $\eta_4(n) \asymp \sqrt{\frac{\log d}{n}}$. Recall that $S_{ij} = \epsilon_i(Z_{ij} - \mathbf{W}^{*T}_{*j}\mathbf{X}_i)$. Therefore, $\|S_{ij}\|_{\psi_1} \leq 2C^2$. In addition, we can show that $\mathbb{E}_{\boldsymbol{\beta}^*}(S_{ij}^2) \geq \sigma^2\lambda_{\min}(\mathbb{E}(\boldsymbol{Q}_i^{\otimes 2})) \geq C$, for some constant $C > 0$. Lemma D.5 implies that $\eta_7(n) \asymp (s^* \vee s')\sqrt{\frac{(\log(nd))^3}{n}}$. Note that $r'(n) \asymp n^{-1/2}(s^* \vee s')\sqrt{\log d}(\log d_0)^{3/2} = o_{\mathbb{P}}(n^{-1/2}(s' \vee s^*)(\log(nd))^{3/2}\sqrt{\log d_0})$. Thus, under the same conditions as those in Corollary 6.8, the multiplier bootstrap is valid for rescaled statistic $||\widehat{\mathbf{T}}_R||_\infty$. □

# E  Proofs for the Misspecified Model

In this appendix, we first present the proofs for the general theorem, namely, Theorem 6.15 and Corollary 6.16. Then, we present the technical details for applying these results to misspecified linear models.

## E.1  Proofs for General Results

*Proof of Theorem 6.15.* Following the proof of Theorem 3.5, we can obtain (6.4). Note that $S(0, \gamma^o) = (1, -\mathbf{w}^{oT})\nabla \ell(0, \gamma^o)$. Combining with Assumption 6.13, we have $n^{1/2} S(0, \gamma^o)/\sqrt{\mathbf{v}^{oT} \mathbf{\Sigma}^o \mathbf{v}^o}$ converges weakly to $N(0, 1)$. By $\mathbf{v}^{oT}\mathbf{\Sigma}^o\mathbf{v}^o \geq C'$ in Assumption 6.13 and inequality (6.4), we obtain that

$$n^{1/2}\big|\widehat{S}(0,\widehat{\gamma})/\sqrt{\mathbf{v}^{oT}\mathbf{\Sigma}^o\mathbf{v}^o} - S(0,\gamma^o)/\sqrt{\mathbf{v}^{oT}\mathbf{\Sigma}^o\mathbf{v}^o}\big| = o_\mathbb{P}(1).$$

We complete the proof of (6.5) by applying the Slutsky's Theorem.  □

*Proof of Corollary 6.16.* Similar to Corollary 3.7, it suffices to show that $\widehat{\mathbf{v}}^T\widehat{\mathbf{\Sigma}}\widehat{\mathbf{v}} = \mathbf{v}^{oT}\mathbf{\Sigma}^o\mathbf{v}^o + o_\mathbb{P}(1)$ holds. Then, the weak convergence follows by Theorem F.1 and the Slutsky's Theorem. By the following Lemma E.1, it is easily seen that

$$\begin{aligned}|\widehat{\mathbf{v}}^T\widehat{\mathbf{\Sigma}}\widehat{\mathbf{v}} - \mathbf{v}^{oT}\mathbf{\Sigma}^o\mathbf{v}^o| &\leq |\widehat{\mathbf{v}}^T\widehat{\mathbf{\Sigma}}\widehat{\mathbf{v}} - \mathbf{v}^{oT}\widehat{\mathbf{\Sigma}}\mathbf{v}^o| + |\mathbf{v}^{oT}\mathbf{\Sigma}^o\mathbf{v}^o - \mathbf{v}^{oT}\widehat{\mathbf{\Sigma}}\mathbf{v}^o| \\ &\leq ||\widehat{\mathbf{\Sigma}}||_{\max}||\widehat{\mathbf{v}} - \mathbf{v}^o||_1^2 + 2||\widehat{\mathbf{\Sigma}}\mathbf{v}^o||_\infty ||\widehat{\mathbf{v}} - \mathbf{v}^o||_1 + ||\mathbf{v}^o||_1^2 ||\widehat{\mathbf{\Sigma}} - \mathbf{\Sigma}^o||_{\max}. \quad \text{(E.1)}\end{aligned}$$

Note that $||\widehat{\mathbf{\Sigma}}||_{\max}||\widehat{\mathbf{v}} - \mathbf{v}^o||_1^2 = \mathcal{O}_\mathbb{P}(\eta_2^2(n))$, and $||\widehat{\mathbf{\Sigma}}\mathbf{v}^o||_\infty ||\widehat{\mathbf{v}} - \mathbf{v}^o||_1 = o_\mathbb{P}(1)$. In addition,

$$\|\mathbf{v}^o\|_1^2 \|\widehat{\mathbf{\Sigma}} - \mathbf{\Sigma}^o\|_{\max} \leq \|\mathbf{v}^o\|_1^2 C_7 \eta_7(n) = o_\mathbb{P}(1).$$

This complete the proof.  □

**Lemma E.1.** Let $\mathbf{W}$ be a symmetric $(d \times d)$-matrix and $\widehat{\mathbf{v}}$ and $\mathbf{v} \in \mathbb{R}^d$. Then

$$|\widehat{\mathbf{v}}^T\mathbf{W}\widehat{\mathbf{v}} - \mathbf{v}^T\mathbf{W}\mathbf{v}| \leq ||\mathbf{W}||_{\max}||\widehat{\mathbf{v}} - \mathbf{v}||_1^2 + 2||\mathbf{W}\mathbf{v}||_\infty ||\widehat{\mathbf{v}} - \mathbf{v}||_1.$$

*Proof of Lemma E.1.* Note that

$$\begin{aligned}|\widehat{\mathbf{v}}^T\mathbf{W}\widehat{\mathbf{v}} - \mathbf{v}^T\mathbf{W}\mathbf{v}| &\leq |(\widehat{\mathbf{v}} - \mathbf{v})^T\mathbf{W}(\widehat{\mathbf{v}} - \mathbf{v})| + 2|\mathbf{v}^T\mathbf{W}(\widehat{\mathbf{v}} - \mathbf{v})| \\ &\leq ||\mathbf{W}||_{\max}||\widehat{\mathbf{v}} - \mathbf{v}||_1^2 + 2||\mathbf{W}\mathbf{v}||_\infty ||\widehat{\mathbf{v}} - \mathbf{v}||_1.\end{aligned}$$

The proof is complete.  □

## E.2  Proofs for the Misspecified Linear Model

In the following, we first present auxiliary Lemmas E.2, E.3, E.4, E.5, E.6, E.7, and then we prove Corollary 6.17 based on these lemmas. The proofs of these lemmas are deferred to Appendix G.5.

**Lemma E.2.** Under the conditions of Corollary 6.17, with probability at least $1 - d^{-1}$,

$$\left\|\frac{1}{n}\sum_{i=1}^n \boldsymbol{X}_i(Y_i - \boldsymbol{\gamma}^{oT}\boldsymbol{X}_i)\right\|_\infty \leq C\sqrt{\frac{\log d}{n}}.$$

*Proof.* See Appendix G.5 for a detailed proof. □

**Lemma E.3.** Under the conditions of Corollary 6.17, with probability at least $1 - (1 + C')d^{-1}$,

$$\|\widehat{\boldsymbol{\beta}} - \boldsymbol{\beta}^o\|_1 \leq \frac{12Cs^*}{\kappa}\sqrt{\frac{\log d}{n}}, \quad \text{and} \quad (\widehat{\boldsymbol{\beta}} - \boldsymbol{\beta}^o)^T \mathbf{H}_Q (\widehat{\boldsymbol{\beta}} - \boldsymbol{\beta}^o) \leq \frac{36C^2}{\kappa}\frac{s^* \log d}{n},$$

where $\mathbf{H}_Q = n^{-1}\sum_{i=1}^n \boldsymbol{Q}_i^{\otimes 2}$ and the constants $C$ and $C'$ are given by Lemmas E.2 and G.1.

*Proof.* See Appendix G.5 for a detailed proof. □

**Lemma E.4.** Under the conditions of Corollary 6.17, with probability at least $1 - (C' + 1)d^{-1}$,

$$\|\widehat{\mathbf{w}} - \mathbf{w}^o\|_1 \leq 8C\kappa^{-1}s'\sqrt{\frac{\log d}{n}},$$

where the constants $C$ and $C'$ are given by Lemmas E.2 and G.1.

*Proof.* See Appendix G.5 for a detailed proof. □

**Lemma E.5.** Under the conditions of Corollary 6.17, with probability at least $1 - d^{-1}$,

$$\left\|\frac{1}{n}\sum_{i=1}^n (Z_i \boldsymbol{X}_i - \widehat{\mathbf{w}}^T \boldsymbol{X}_i^{\otimes 2})\right\|_\infty \leq C\sqrt{\frac{\log d}{n}},$$

for some constant $C > 0$.

*Proof.* See Appendix G.5 for a detailed proof. □

Recall that $S(\theta, \boldsymbol{\gamma}) = -\frac{1}{n}\sum_{i=1}^n (Y_i - \theta Z_i - \boldsymbol{\gamma}^T \boldsymbol{X}_i)(Z_i - \mathbf{w}^{oT}\boldsymbol{X}_i)$.

**Lemma E.6.** Under the conditions of Corollary 6.17, $\sqrt{n}S(0, \boldsymbol{\gamma}^o)/\sqrt{\mathbf{v}^{oT}\boldsymbol{\Sigma}^o \mathbf{v}^o} \rightsquigarrow N(0,1)$.

*Proof.* See Appendix G.5 for a detailed proof. □

**Lemma E.7.** Under the conditions of Corollary 6.17,

$$\|\widehat{\boldsymbol{\Sigma}} - \boldsymbol{\Sigma}^o\|_{\max} = \mathcal{O}_\mathbb{P}\left(\sqrt{\frac{(\log d)^5}{n}} \vee \sqrt{\frac{s^* \log d}{n}}\right), \quad \|(\widehat{\boldsymbol{\Sigma}} - \boldsymbol{\Sigma}^o)\mathbf{v}^o\|_\infty = \mathcal{O}_\mathbb{P}\left(\sqrt{\frac{(\log d)^5}{n}} \vee \sqrt{\frac{s^* \log d}{n}}\right).$$

*Proof.* See Appendix G.5 for a detailed proof. □

*Proof of Corollary 6.17.* This Corollary follows from the general results in Theorem 6.15 and Corollary 6.16. We now verify the conditions in Corollary 6.16 hold. By Lemma E.4, we have $\eta_2(n) \asymp s'\sqrt{\frac{\log d}{n}}$. In addition, Lemma E.2 implies $\eta_3(n) \asymp \sqrt{\frac{\log d}{n}}$, Lemma E.3 implies $\eta_1(n) \asymp s^*\sqrt{\frac{\log d}{n}}$, Lemma E.5 implies $\eta_4(n) \asymp \sqrt{\frac{\log d}{n}}$, and Lemma E.6 implies the central limit theorem in Assumption 6.13 holds. Finally, we need to show that the asymptotic variance can be consistently estimated, i.e., $|\widehat{\mathbf{v}}^T\widehat{\mathbf{\Sigma}}\widehat{\mathbf{v}} - \mathbf{v}^{oT}\mathbf{\Sigma}^o\mathbf{v}^o| = o_{\mathbb{P}}(1)$. To get a better scaling, we will verify that the last three terms in (E.1) in the proof of Corollary 6.16 are all asymptotically ignorable. First, for any $j, k$, $|\mathbf{\Sigma}^o_{ij}|^2 = |\mathbb{E}^*(Q_{ij}Q_{ik}(Y_i - \boldsymbol{\beta}^{oT}\mathbf{Q}_i)^2)|^2 \leq \mathbb{E}^*(Q_{ij}^2 Q_{ik}^2)\mathbb{E}^*(Y_i - \boldsymbol{\beta}^{oT}\mathbf{Q}_i)^2$, which is bounded by a constant due to condition (3). Hence, $\|\mathbf{\Sigma}^o\|_{\max} = \mathcal{O}(1)$, and $\|\mathbf{\Sigma}^o\|_{\max}\eta_2^2(n) = o_{\mathbb{P}}(1)$. Similar arguments yield that $\|\mathbf{\Sigma}^o\mathbf{v}^o\|_\infty = \mathcal{O}(1)$. Together with Lemma E.7, we have $\|\widehat{\mathbf{\Sigma}}\mathbf{v}^o\|_\infty \leq \|\mathbf{\Sigma}^o\mathbf{v}^o\|_\infty + \|(\widehat{\mathbf{\Sigma}} - \mathbf{\Sigma}^o)\mathbf{v}^o\|_\infty = \mathcal{O}_{\mathbb{P}}(1)$, and therefore $\|\widehat{\mathbf{\Sigma}}\mathbf{v}^o\|_\infty \eta_2(n) = o_{\mathbb{P}}(1)$. Finally, similar to the proof of Lemma E.7, we can show that $|\mathbf{v}^{oT}(\mathbf{\Sigma}^o - \widehat{\mathbf{\Sigma}})\mathbf{v}^o| = o_{\mathbb{P}}(1)$. Combining these results, we get $|\widehat{\mathbf{v}}^T\widehat{\mathbf{\Sigma}}\widehat{\mathbf{v}} - \mathbf{v}^{oT}\mathbf{\Sigma}^o\mathbf{v}^o| = o_{\mathbb{P}}(1)$, and this completes the proof. □

# F  Proofs for Generalized Score Test

In this appendix, we first prove Theorem 6.21 and then present one example for the generalized score test.

## F.1  Proof of Theorem 6.21

To prove Theorem 6.21, we first present the following Lemma, which is parallel to Theorem 3.5.

**Lemma F.1.** Under the Assumptions 3.1–3.3 and 6.19, with probability tending to one

$$n^{1/2}|\widehat{S}(0,\widehat{\boldsymbol{\gamma}}) - S(0,\boldsymbol{\gamma}^*)| \lesssim n^{1/2}(\eta_2(n)\eta_3(n) + \eta_1(n)\eta_4(n)). \tag{F.1}$$

If $n^{1/2}(\eta_2(n)\eta_3(n) + \eta_1(n)\eta_4(n)) = o(1)$, we have $n^{1/2}\widehat{S}(0,\widehat{\boldsymbol{\gamma}})/\sqrt{\mathbf{v}^{*T}\mathbf{\Sigma}^*\mathbf{v}^*} \rightsquigarrow N(0,1)$, where $\mathbf{v}^* = (1, -\mathbf{w}^{*T})^T$.

*Proof of Lemma F.1.* Following the proof of Theorem 3.5, we can obtain (F.1). Note that $S(0,\boldsymbol{\gamma}^*) = (1, -\mathbf{w}^{*T})\nabla\ell(0,\boldsymbol{\gamma}^*)$. Combining with Assumption 6.19, we have $n^{1/2}S(0,\boldsymbol{\gamma}^*)/\sqrt{\mathbf{v}^{*T}\mathbf{\Sigma}^*\mathbf{v}^*}$ converges weakly to $N(0,1)$. By $\mathbf{v}^{*T}\mathbf{\Sigma}^*\mathbf{v}^* \geq C'$ in Assumption 6.19 and inequality (F.1), we obtain that

$$n^{1/2}|\widehat{S}(0,\widehat{\boldsymbol{\gamma}})/\sqrt{\mathbf{v}^{*T}\mathbf{\Sigma}^*\mathbf{v}^*} - S(0,\boldsymbol{\gamma}^*)/\sqrt{\mathbf{v}^{*T}\mathbf{\Sigma}^*\mathbf{v}^*}| = o_{\mathbb{P}}(1).$$

We complete the proof, by applying the Slutsky's Theorem. □

*Proof of Theorem 6.21.* Similar to Corollary 3.7, it suffices to show that $\widehat{\mathbf{v}}^T\widehat{\mathbf{\Sigma}}\widehat{\mathbf{v}} = \mathbf{v}^{*T}\mathbf{\Sigma}^*\mathbf{v}^* + o_{\mathbb{P}}(1)$ holds. Then, the weak convergence follows by Lemma F.1 and the Slutsky's Theorem. By Lemma E.1, it is easily seen that

$$\begin{aligned}|\widehat{\mathbf{v}}^T\widehat{\mathbf{\Sigma}}\widehat{\mathbf{v}} - \mathbf{v}^{*T}\mathbf{\Sigma}^*\mathbf{v}^*| &\leq |\widehat{\mathbf{v}}^T\widehat{\mathbf{\Sigma}}\widehat{\mathbf{v}} - \mathbf{v}^{*T}\widehat{\mathbf{\Sigma}}\mathbf{v}^*| + |\mathbf{v}^{*T}\mathbf{\Sigma}^*\mathbf{v}^* - \mathbf{v}^{*T}\widehat{\mathbf{\Sigma}}\mathbf{v}^*| \\ &\leq \|\widehat{\mathbf{\Sigma}}\|_{\max}\|\widehat{\mathbf{v}} - \mathbf{v}^*\|_1^2 + 2\|\widehat{\mathbf{\Sigma}}\mathbf{v}^*\|_\infty\|\widehat{\mathbf{v}} - \mathbf{v}^*\|_1 + \|\mathbf{v}^*\|_1^2\|\widehat{\mathbf{\Sigma}} - \mathbf{\Sigma}^*\|_{\max}\end{aligned}$$

Note that $||\widehat{\mathbf{\Sigma}}||_{\max}||\widehat{\mathbf{v}} - \mathbf{v}^*||_1^2 = \mathcal{O}_{\mathbb{P}}(\eta_2^2(n))$, and $||\widehat{\mathbf{\Sigma}}\mathbf{v}^*||_\infty ||\widehat{\mathbf{v}} - \mathbf{v}^*||_1 = o_{\mathbb{P}}(1)$. In addition,

$$||\mathbf{v}^*||_1^2 ||\widehat{\mathbf{\Sigma}} - \mathbf{\Sigma}^*||_{\max} = ||\mathbf{v}^*||_1^2 \cdot \mathcal{O}_{\mathbb{P}}(\eta_7(n)) = o_{\mathbb{P}}(1).$$

This complete the proof. □

## F.2 Illustration for Generalized Score Test

One example for the inference based on the general loss function is considered by Ning and Liu (2014). This paper studied the following semiparametric generalized linear model,

$$\mathbb{P}(Y_i \mid \boldsymbol{Q}_i; \boldsymbol{\beta}, f) = \exp\left\{\boldsymbol{\beta}^T \boldsymbol{Q}_i Y_i - b(\boldsymbol{\beta}^T \boldsymbol{Q}_i, f) + \log f(Y_i)\right\},$$

where $\boldsymbol{\beta} = (\theta, \boldsymbol{\gamma})$ is a $d$ dimensional parameter, $f(\cdot)$ is an unknown function and $b(a, f) = \log \int \exp(ay) f(y) dy$ is a normalizing function. Due to the intractability of the likelihood for $\boldsymbol{\beta}$, the generalized score test is constructed based on the following logistic loss,

$$\ell(\boldsymbol{\beta}) = \binom{n}{2}^{-1} \sum_{1 \leq i < j \leq n} \log\left(1 + R_{ij}(\boldsymbol{\beta})\right), \quad \text{where} \quad R_{ij}(\boldsymbol{\beta}) = \exp\{-(Y_i - Y_j)\boldsymbol{\beta}^T(\boldsymbol{X}_i - \boldsymbol{X}_j)\},$$

which is free of $f(\cdot)$ and satisfies $\boldsymbol{\beta}^* = \operatorname{argmin}_{\boldsymbol{\beta}} \mathbb{E}_{\boldsymbol{\beta}^*}(\ell(\boldsymbol{\beta}))$. The asymptotic properties of the generalized score test based on the loss function $\ell(\boldsymbol{\beta})$ has been established in Theorem 4.9 of Ning and Liu (2014). The technical details as well as the application of the semiparametric generalized linear model can be found in Ning and Liu (2014). To save space, we do not replicate the details.

# G  Proofs of Auxiliary Lemmas

In this appendix, we present the proofs for the auxiliary lemmas in Appendices A, B, C, D and E.

## G.1  Proofs of Auxiliary Lemmas in Appendix A

*Proof of Lemma A.1.* Similar to the proof of Theorem 3.5, we can obtain

$$\widehat{S}(0, \widehat{\boldsymbol{\gamma}}) = S(\boldsymbol{\beta}^*) + \underbrace{(\mathbf{w}^* - \widehat{\mathbf{w}})^T \nabla_{\boldsymbol{\gamma}} \ell(\boldsymbol{\beta}^*)}_{I_1} + \underbrace{\{\nabla^2_{\theta\boldsymbol{\gamma}} \ell(0, \widetilde{\boldsymbol{\gamma}}) - \widehat{\mathbf{w}}^T \nabla^2_{\boldsymbol{\gamma}\boldsymbol{\gamma}} \ell(0, \widetilde{\boldsymbol{\gamma}})\}(\widehat{\boldsymbol{\gamma}} - \boldsymbol{\gamma}^*)}_{I_2},$$

where $\widetilde{\boldsymbol{\gamma}} = v\boldsymbol{\gamma}^* + (1-v)\widehat{\boldsymbol{\gamma}}$ for some $v \in [0, 1]$. On the event $\mathcal{F}^{\boldsymbol{\beta}^*} = \mathcal{F}_1^{\boldsymbol{\beta}^*} \cap \mathcal{F}_2^{\boldsymbol{\beta}^*} \cap \mathcal{F}_3^{\boldsymbol{\beta}^*} \cap \mathcal{F}_4^{\boldsymbol{\beta}^*}$, we have that

$$|I_1| \leq ||\mathbf{w}^* - \widehat{\mathbf{w}}||_1 ||\nabla_{\boldsymbol{\gamma}} \ell(\boldsymbol{\beta}^*)||_\infty \lesssim \eta_2(n)\eta_3(n),$$

and

$$|I_2| \leq ||\nabla^2_{\theta\boldsymbol{\gamma}} \ell(0, \widetilde{\boldsymbol{\gamma}}) - \widehat{\mathbf{w}}^T \nabla^2_{\boldsymbol{\gamma}\boldsymbol{\gamma}} \ell(0, \widetilde{\boldsymbol{\gamma}})||_\infty ||\widehat{\boldsymbol{\gamma}} - \boldsymbol{\gamma}^*||_1 \lesssim \eta_1(n)\eta_4(n).$$

This implies that as $n \to \infty$,

$$\inf_{\boldsymbol{\beta}^* \in \Omega_0} \mathbb{P}_{\boldsymbol{\beta}^*}\left(\left|\widehat{S}(0, \widehat{\boldsymbol{\gamma}}) - S(\boldsymbol{\beta}^*)\right| \lesssim \left(\eta_2(n)\eta_3(n) + \eta_1(n)\eta_4(n)\right)\right)$$
$$\geq \inf_{\boldsymbol{\beta}^* \in \Omega_0} \mathbb{P}_{\boldsymbol{\beta}^*}(\mathcal{F}^{\boldsymbol{\beta}^*}) \geq 1 - \sum_{i=1}^{4} \sup_{\boldsymbol{\beta}^* \in \Omega_0} \mathbb{P}_{\boldsymbol{\beta}^*}(\bar{\mathcal{F}}_i^{\boldsymbol{\beta}^*}) \to 1,$$

where the last step follows by Assumption 3.9–3.12. This completes the proof of (A.1). For some constant $C > 0$, denote $q(n) = Cn^{1/2}\left(\eta_2(n)\eta_3(n) + \eta_1(n)\eta_4(n)\right)$. Note that

$$\mathbb{P}_{\boldsymbol{\beta}^*}\left(n^{1/2}\widehat{S}(0,\widehat{\boldsymbol{\gamma}})I_{\theta|\gamma}^{*-1/2} \leq t\right) \leq \mathbb{P}_{\boldsymbol{\beta}^*}\left(n^{1/2}\widehat{S}(0,\widehat{\boldsymbol{\gamma}})I_{\theta|\gamma}^{*-1/2} \leq t, \mathcal{F}^{\boldsymbol{\beta}^*}\right) + \mathbb{P}_{\boldsymbol{\beta}^*}(\bar{\mathcal{F}}^{\boldsymbol{\beta}^*})$$
$$\leq \mathbb{P}_{\boldsymbol{\beta}^*}\left(n^{1/2}S(\boldsymbol{\beta}^*)I_{\theta|\gamma}^{*-1/2} \leq t + q(n)\right) + \mathbb{P}_{\boldsymbol{\beta}^*}(\bar{\mathcal{F}}^{\boldsymbol{\beta}^*}).$$

To show (A.2), we have that

$$\mathbb{P}_{\boldsymbol{\beta}^*}\left(n^{1/2}\widehat{S}(0,\widehat{\boldsymbol{\gamma}})I_{\theta|\gamma}^{*-1/2} \leq t\right) - \Phi(t)$$
$$\leq \mathbb{P}_{\boldsymbol{\beta}^*}\left(n^{1/2}S(\boldsymbol{\beta}^*)I_{\theta|\gamma}^{*-1/2} \leq t + q(n)\right) - \Phi(t) + \mathbb{P}_{\boldsymbol{\beta}^*}(\bar{\mathcal{F}}^{\boldsymbol{\beta}^*})$$
$$= \left\{\mathbb{P}_{\boldsymbol{\beta}^*}\left(n^{1/2}S(\boldsymbol{\beta}^*)I_{\theta|\gamma}^{*-1/2} \leq t + q(n)\right) - \Phi(t + q(n))\right\} + \left\{\Phi(t + q(n)) - \Phi(t)\right\} + \mathbb{P}_{\boldsymbol{\beta}^*}(\bar{\mathcal{F}}^{\boldsymbol{\beta}^*}).$$

By Assumption 3.12, and the fact that $\Phi(t + q(n)) - \Phi(t) \leq \frac{q(n)}{\sqrt{2\pi}}$, we obtain

$$\limsup_{n \to \infty} \sup_{\boldsymbol{\beta}^* \in \Omega_0} \sup_{t \in \mathbb{R}} \left(\mathbb{P}_{\boldsymbol{\beta}^*}\left(n^{1/2}\widehat{S}(0,\widehat{\boldsymbol{\gamma}})I_{\theta|\gamma}^{*-1/2} \leq t\right) - \Phi(t)\right) \leq 0.$$

Following the similar arguments, it is seen that

$$\liminf_{n \to \infty} \inf_{\boldsymbol{\beta}^* \in \Omega_0} \inf_{t \in \mathbb{R}} \left(\mathbb{P}_{\boldsymbol{\beta}^*}\left(n^{1/2}\widehat{S}(0,\widehat{\boldsymbol{\gamma}})I_{\theta|\gamma}^{*-1/2} \leq t\right) - \Phi(t)\right) \geq 0.$$

Since $\lim_{n \to \infty} \sup_{\boldsymbol{\beta}^* \in \Omega_0} \mathbb{P}_{\boldsymbol{\beta}^*}(\bar{\mathcal{F}}^{\boldsymbol{\beta}^*}) = 0$, this completes the proof of (A.2). $\square$

*Proof of Lemma A.2.* By the mean value theorem,

$$\widehat{S}(0,\widehat{\boldsymbol{\gamma}}) = \widehat{S}(0,\boldsymbol{\gamma}^*) + \underbrace{\{\nabla_{\theta\gamma}^2 \ell(0, \widetilde{\boldsymbol{\gamma}}) - \widehat{\mathbf{w}}^T \nabla_{\gamma\gamma}^2 \ell(0, \widetilde{\boldsymbol{\gamma}})\}(\widehat{\boldsymbol{\gamma}} - \boldsymbol{\gamma}^*)}_{I_1}, \tag{G.1}$$

where $\widetilde{\boldsymbol{\gamma}} = v\boldsymbol{\gamma}^* + (1-v)\widehat{\boldsymbol{\gamma}}$ for some $v \in [0,1]$. For $\widehat{S}(0,\boldsymbol{\gamma})$, we have

$$\widehat{S}(0,\boldsymbol{\gamma}^*) = \nabla_\theta \ell(0,\boldsymbol{\gamma}^*) - \widehat{\mathbf{w}}^T \nabla_\gamma \ell(0,\boldsymbol{\gamma}^*)$$
$$= \nabla_\theta \ell(0,\boldsymbol{\gamma}^*) - \mathbf{w}^{*T} \nabla_\gamma \ell(0,\boldsymbol{\gamma}^*) + (\mathbf{w}^* - \widehat{\mathbf{w}})^T \nabla_\gamma \ell(0,\boldsymbol{\gamma}^*)$$
$$= S(\boldsymbol{\beta}^*) - I_{\theta|\gamma}^* \widetilde{C} n^{-\phi} + \underbrace{S(0,\boldsymbol{\gamma}^*) - S(\boldsymbol{\beta}^*) + I_{\theta|\gamma}^* \widetilde{C} n^{-\phi}}_{I_2} + \underbrace{(\mathbf{w}^* - \widehat{\mathbf{w}})^T \nabla_\gamma \ell(0,\boldsymbol{\gamma}^*)}_{I_3}. \tag{G.2}$$

On the event $\mathcal{F}^{\boldsymbol{\beta}^*} = \mathcal{F}_1^{\boldsymbol{\beta}^*} \cap ... \cap \mathcal{F}_6^{\boldsymbol{\beta}^*}$, by Assumption 3.19,

$$|I_1| \leq ||\nabla^2_{\theta\gamma}\ell(0,\widetilde{\boldsymbol{\gamma}}) - \widehat{\mathbf{w}}^T \nabla^2_{\gamma\gamma}\ell(0,\widetilde{\boldsymbol{\gamma}})||_\infty ||\widehat{\boldsymbol{\gamma}} - \boldsymbol{\gamma}^*||_1 \lesssim \eta_1(n)\eta_4(n).$$

By Assumption 3.16 and the triangle inequality, $|I_2| \lesssim \eta_6(n)n^{-1/2}$. For $I_3$, by Assumption 3.18,

$$|I_3| \leq ||\mathbf{w}^* - \widehat{\mathbf{w}}||_1 ||\nabla_\gamma \ell(0,\boldsymbol{\gamma}^*)||_\infty \lesssim \eta_2(n)\eta_3(n).$$

It is also seen that the event $\mathcal{F}^{\boldsymbol{\beta}^*}$ holds uniformly with probability tending to one. Combining the upper bounds for $I_1$, $I_2$ and $I_3$ with equations (G.1) and (G.2), we obtain (A.5).

Denote $q(n) = C(n^{1/2}\psi_n + \eta_6(n))$, for some constant $C > 0$. We shall show that (A.6) and (A.7) hold. Note that

$$\mathbb{P}_{\boldsymbol{\beta}^*}\left(n^{1/2}\widehat{S}(0,\widehat{\boldsymbol{\gamma}})I_{\theta|\gamma}^{*-1/2} \leq t\right) - \Phi(t)$$

$$\leq \mathbb{P}_{\boldsymbol{\beta}^*}\left(n^{1/2}S(\boldsymbol{\beta}^*)I_{\theta|\gamma}^{*-1/2} \leq t + \widetilde{C}I_{\theta|\gamma}^{*1/2}n^{1/2-\phi} + q(n)\right) - \Phi(t) + \mathbb{P}_{\boldsymbol{\beta}^*}(\bar{\mathcal{F}}^{\boldsymbol{\beta}^*})$$

$$= \left\{\mathbb{P}_{\boldsymbol{\beta}^*}\left(n^{1/2}S(\boldsymbol{\beta}^*)I_{\theta|\gamma}^{*-1/2} \leq t + \widetilde{C}I_{\theta|\gamma}^{*1/2}n^{1/2-\phi} + q(n)\right) - \Phi(t + \widetilde{C}I_{\theta|\gamma}^{*1/2}n^{1/2-\phi} + q(n))\right\}$$

$$+ \left\{\Phi(t + \widetilde{C}I_{\theta|\gamma}^{*1/2}n^{1/2-\phi} + q(n)) - \Phi(t)\right\} + \mathbb{P}_{\boldsymbol{\beta}^*}(\bar{\mathcal{F}}^{\boldsymbol{\beta}^*}). \tag{G.3}$$

By Assumption 3.20, and the fact that

$$\Phi\bigl(t + \widetilde{C}I_{\theta|\gamma}^{*1/2}n^{1/2-\phi} + q(n)\bigr) - \Phi(t) \leq \frac{1}{\sqrt{2\pi}}\bigl(q(n) + \widetilde{C}I_{\theta|\gamma}^{*1/2}n^{1/2-\phi}\bigr),$$

we obtain that if $\phi > 1/2$,

$$\limsup_{n\to\infty} \sup_{\boldsymbol{\beta}^* \in \Omega_1(\widetilde{C},\phi)} \sup_{t\in\mathbb{R}} \left(\mathbb{P}_{\boldsymbol{\beta}^*}\bigl(n^{1/2}\widehat{S}(0,\widehat{\boldsymbol{\gamma}})I_{\theta|\gamma}^{*-1/2} \leq t\bigr) - \Phi(t)\right) \leq 0.$$

The lower bound follows from the similar arguments, and this completes the proof of (A.6). Note that (A.7) can be established by essentially the same steps. To show (A.8), it is easily seen that on the event $\mathcal{F}^{\boldsymbol{\beta}^*}$, $|n^{1/2}\widehat{S}(0,\widehat{\boldsymbol{\gamma}})I_{\theta|\gamma}^{*-1/2}| \leq t$ implies that $A_{\min} \leq n^{1/2}S(\boldsymbol{\beta}^*)I_{\theta|\gamma}^{*-1/2} \leq A_{\max}$, where

$$A_{\min} = -t - q(n) + \widetilde{C}I_{\theta|\gamma}^{*1/2}n^{1/2-\phi}, \text{ and } A_{\max} = t + q(n) + \widetilde{C}I_{\theta|\gamma}^{*1/2}n^{1/2-\phi}.$$

Similar to (G.2), we can show that

$$\mathbb{P}_{\boldsymbol{\beta}^*}\bigl(|n^{1/2}\widehat{S}(0,\widehat{\boldsymbol{\gamma}})I_{\theta|\gamma}^{*-1/2}| \leq t\bigr) \leq \mathbb{P}_{\boldsymbol{\beta}^*}\bigl(A_{\min} \leq n^{1/2}S(\boldsymbol{\beta}^*)I_{\theta|\gamma}^{*-1/2} \leq A_{\max}\bigr) + \mathbb{P}_{\boldsymbol{\beta}^*}(\bar{\mathcal{F}}^{\boldsymbol{\beta}^*}).$$

Denote $Z \sim N(0,1)$. By Assumption 3.20,

$$\limsup_{n\to\infty} \sup_{\boldsymbol{\beta}^* \in \Omega_1(\widetilde{C},\phi)} \bigl|\mathbb{P}_{\boldsymbol{\beta}^*}\bigl(A_{\min} \leq n^{1/2}S(\boldsymbol{\beta}^*)I_{\theta|\gamma}^{*-1/2} \leq A_{\max}\bigr) - \mathbb{P}(A_{\min} \leq Z \leq A_{\max})\bigr| = 0.$$

Next, we will bound $\mathbb{P}(A_{\min} \leq Z \leq A_{\max})$. When $\phi < 1/2$, if $\widetilde{C} < 0$, we have $\mathbb{P}(A_{\min} \leq Z \leq A_{\max}) \leq \Phi(A_{\max}) \to 0$ uniformly over $\boldsymbol{\beta}^* \in \Omega_1(\widetilde{C},\phi)$, as $n \to 0$. The same result holds for $\widetilde{C} > 0$, since $\mathbb{P}(A_{\min} \leq Z \leq A_{\max}) \leq 1 - \Phi(A_{\min}) \to 0$ uniformly over $\boldsymbol{\beta}^* \in \Omega_1(\widetilde{C},\phi)$, as $n \to 0$. In addition, $\limsup_{n\to\infty} \sup_{\boldsymbol{\beta}^* \in \Omega_1(\widetilde{C},\phi)} \mathbb{P}_{\boldsymbol{\beta}^*}(\bar{\mathcal{F}}^{\boldsymbol{\beta}^*}) = 0$. This completes the proof.

$\square$

## G.2 Proofs of Auxiliary Lemmas in Appendix B

*Proof of Lemma B.1.* Since $\|\epsilon_i\|_{\psi_2} \leq C$ and $\|Q_{ij}\|_{\psi_2} \leq C$, by Lemma H.1, we have $\|\boldsymbol{X}_i\epsilon_i\|_{\psi_1} \leq 2C^2$. Since $X_{ij}\epsilon_i$ has mean 0, by Lemma H.2, we have for any $t > 0$ and $j = 1, ..., d$,

$$\mathbb{P}_{\boldsymbol{\beta}^*}\left(\frac{1}{n}\left|\sum_{i=1}^{n} X_{ij}\epsilon_i\right| \geq t\right) \leq 2\exp\left[-C''\min\left(\frac{t^2}{4C^4}, \frac{t}{2C^2}\right)n\right],$$

where $C''$ is given in Lemma H.2. By the union bound inequality,

$$\mathbb{P}_{\boldsymbol{\beta}^*}\left(\left\|\frac{1}{n}\sum_{i=1}^{n}\boldsymbol{X}_i\epsilon_i\right\|_{\infty} \geq t\right) \leq 2d\exp\left[-C''\min\left(\frac{t^2}{4C^4}, \frac{t}{2C^2}\right)n\right].$$

With $t = C\sqrt{\frac{\log d}{n}}$, for some constant $C > 0$, Lemma B.1 holds. □

*Proof of Lemma B.2.* By the definition of the Dantzig selector, it suffices to show that $\mathbf{w}^*$ is in the feasible set, i.e.,

$$\left\|\frac{1}{n}\sum_{i=1}^{n}(Z_i\boldsymbol{X}_i - \mathbf{w}^{*T}\boldsymbol{X}_i^{\otimes 2})\right\|_{\infty} \leq C\sqrt{\frac{\log d}{n}}, \quad (G.4)$$

with high probability. Note that for any $j = 1, ..., d$, by the triangle inequality

$$\|Z_iX_{ij} - \mathbf{w}^{*T}\boldsymbol{X}_iX_{ij}\|_{\psi_2} \leq \|Z_iX_{ij}\|_{\psi_2} + \|\mathbf{w}^{*T}\boldsymbol{X}_iX_{ij}\|_{\psi_2} \leq 4C^2.$$

By Lemma H.2 and the union bound inequality, we have for any $t > 0$,

$$\mathbb{P}_{\boldsymbol{\beta}^*}\left(\left\|\frac{1}{n}\sum_{i=1}^{n}(Z_i\boldsymbol{X}_i - \mathbf{w}^{*T}\boldsymbol{X}_i^{\otimes 2})\right\|_{\infty} \geq t\right) \leq 2d\exp\left[-C''\min\left(\frac{t^2}{16C^4}, \frac{t}{4C^2}\right)n\right],$$

where $C''$ is given in Lemma H.2. With $t = C\sqrt{\frac{\log d}{n}}$, for some constant $C$, Lemma B.2 holds. □

**Lemma G.1.** *Under the conditions of Theorem 4.1, we have* $\mathrm{RE} \geq \kappa$ *and* $\mathrm{RE}' \geq \kappa$ *with probability at least* $1 - C'd^{-1}$ *for some constant* $C' > 0$, *where*

$$\mathrm{RE} = \min\left\{\frac{\mathbf{v}^T\mathbf{H}_Q\mathbf{v}}{\|\mathbf{v}\|_2^2} : \mathbf{v} \in \mathbb{R}^d\setminus\{0\}, \|\mathbf{v}_{\bar{S}}\|_1 \leq 3\|\mathbf{v}_S\|_1\right\},$$

$$\mathrm{RE}' = \min\left\{\frac{\mathbf{v}^T\mathbf{H}_X\mathbf{v}}{\|\mathbf{v}\|_2^2} : \mathbf{v} \in \mathbb{R}^{d-1}\setminus\{0\}, \|\mathbf{v}_{\bar{S}'}\|_1 \leq 3\|\mathbf{v}_{S'}\|_1\right\},$$

*and* $\mathbf{H}_Q = n^{-1}\sum_{i=1}^{n}\boldsymbol{Q}_i^{\otimes 2}$ *and* $\mathbf{H}_X = n^{-1}\sum_{i=1}^{n}\boldsymbol{X}_i^{\otimes 2}$.

*Proof of Lemma G.1.* The proof is similar to that of Lemma G.4. See also Bickel et al. (2009). □

*Proof of Lemma B.3.* Let $S$ denote the support set of $\boldsymbol{\beta}^*$. Note that $\beta_j^* = 0$ if $j \in \bar{S}$. Denote $D(\boldsymbol{\beta}_1, \boldsymbol{\beta}) = (\boldsymbol{\beta}_1 - \boldsymbol{\beta})^T \mathbf{H}_Q (\boldsymbol{\beta}_1 - \boldsymbol{\beta})$ and $\widehat{\boldsymbol{\Delta}} = \widehat{\boldsymbol{\beta}} - \boldsymbol{\beta}^*$. Thus

$$
\begin{aligned}
D(\widehat{\boldsymbol{\beta}}, \boldsymbol{\beta}^*) &= \widehat{\boldsymbol{\Delta}}^T \left\{ -\frac{1}{n} \sum_{i=1}^n \boldsymbol{Q}_i (Y_i - \widehat{\boldsymbol{\beta}}^T \boldsymbol{Q}_i) + \frac{1}{n} \sum_{i=1}^n \boldsymbol{Q}_i (Y_i - \boldsymbol{\beta}^{*T} \boldsymbol{Q}_i) \right\} \\
&= -\widehat{\boldsymbol{\beta}}_{\bar{S}}^T \left\{ \frac{1}{n} \sum_{i=1}^n \boldsymbol{Q}_{i\bar{S}} (Y_i - \widehat{\boldsymbol{\beta}}^T \boldsymbol{Q}_i) \right\} - \widehat{\boldsymbol{\Delta}}_S^T \left\{ \frac{1}{n} \sum_{i=1}^n \boldsymbol{Q}_{iS} (Y_i - \widehat{\boldsymbol{\beta}}^T \boldsymbol{Q}_i) \right\} + \widehat{\boldsymbol{\Delta}}^T \left\{ \frac{1}{n} \sum_{i=1}^n \boldsymbol{Q}_i \epsilon_i \right\}.
\end{aligned}
$$

Denote $\mathbf{G}^* = \frac{1}{n} \sum_{i=1}^n \boldsymbol{Q}_i \epsilon_i$. By Lemma B.1, $\|\mathbf{G}^*\|_\infty \leq C \sqrt{\frac{\log d}{n}}$ with probability at least $1 - d^{-1}$. With the same $C$ and $\lambda = 2C \sqrt{\frac{\log d}{n}}$, by the Karush-Kuhn-Tucker (KKT) condition and Hölder inequality, we have

$$
D(\widehat{\boldsymbol{\beta}}, \boldsymbol{\beta}^*) \leq -\lambda \|\widehat{\boldsymbol{\Delta}}_{\bar{S}}\|_1 + \lambda \|\widehat{\boldsymbol{\Delta}}_S\|_1 + \|\widehat{\boldsymbol{\Delta}}\|_1 \|\mathbf{G}^*\|_\infty \leq C \sqrt{\frac{\log d}{n}} (3\|\widehat{\boldsymbol{\Delta}}_S\|_1 - \|\widehat{\boldsymbol{\Delta}}_{\bar{S}}\|_1). \quad \text{(G.5)}
$$

Since $D(\widehat{\boldsymbol{\beta}}, \boldsymbol{\beta}^*) \geq 0$, we have $\|\widehat{\boldsymbol{\Delta}}_{\bar{S}}\|_1 \leq 3\|\widehat{\boldsymbol{\Delta}}_S\|_1$. In addition, by the RE condition in Lemma G.1, $D(\widehat{\boldsymbol{\beta}}, \boldsymbol{\beta}^*) \geq \kappa \|\widehat{\boldsymbol{\Delta}}\|_2^2$. Since by (G.5) $D(\widehat{\boldsymbol{\beta}}, \boldsymbol{\beta}^*) \leq 3C \sqrt{\frac{s^* \log d}{n}} \|\widehat{\boldsymbol{\Delta}}\|_2$, this implies that $\|\widehat{\boldsymbol{\Delta}}\|_2 \leq \frac{3C}{\kappa} \sqrt{\frac{s^* \log d}{n}}$. Finally,

$$
\|\widehat{\boldsymbol{\Delta}}\|_1 \leq 4 \|\widehat{\boldsymbol{\Delta}}_S\|_1 \leq 4 s^{*1/2} \|\widehat{\boldsymbol{\Delta}}_S\|_2 \leq 4 s^{*1/2} \|\widehat{\boldsymbol{\Delta}}\|_2 \leq \frac{12 C s^*}{\kappa} \sqrt{\frac{\log d}{n}}.
$$

In addition, it is easily seen from (G.5) that $D(\widehat{\boldsymbol{\beta}}, \boldsymbol{\beta}^*) \leq \frac{36 C^2}{\kappa} \frac{s^* \log d}{n}$. This completes the proof. □

*Proof of Lemma B.4.* Let $S'$ denote the support set of $\mathbf{w}^*$. By the definition, we have $\|\mathbf{w}_{S'}^*\|_1 \geq \|\widehat{\mathbf{w}}_{S'}\|_1 + \|\widehat{\mathbf{w}}_{\bar{S}'}\|_1$. The triangle inequality yields $\|\widehat{\mathbf{w}}_{S'}\|_1 \geq \|\mathbf{w}_{S'}^*\|_1 - \|(\widehat{\mathbf{w}} - \mathbf{w}^*)_{S'}\|_1$. Together with $\mathbf{w}_{\bar{S}'}^* = 0$, these imply that $\|\widehat{\boldsymbol{\Delta}}_{S'}\| \geq \|\widehat{\boldsymbol{\Delta}}_{\bar{S}'}\|$, where $\widehat{\boldsymbol{\Delta}} = \widehat{\mathbf{w}} - \mathbf{w}^*$. In addition, with $\lambda' = C \sqrt{n^{-1} \log d}$ with $C$ given in Lemma B.2,

$$
\|\mathbf{H}_X \widehat{\boldsymbol{\Delta}}\|_\infty \leq \|\mathbf{H}_{XZ} - \mathbf{H}_X \widehat{\mathbf{w}}\|_\infty + \|\mathbf{H}_{XZ} - \mathbf{H}_X \mathbf{w}^*\|_\infty \leq 2C \sqrt{\frac{\log d}{n}}.
$$

Together with $\|\widehat{\boldsymbol{\Delta}}\|_1 \leq 2 \|\widehat{\boldsymbol{\Delta}}_{S'}\|_1 \leq 2 s'^{1/2} \|\widehat{\boldsymbol{\Delta}}_{S'}\|_2 \leq 2 s'^{1/2} \|\widehat{\boldsymbol{\Delta}}\|_2$, this implies that,

$$
\widehat{\boldsymbol{\Delta}}^T \mathbf{H}_X \widehat{\boldsymbol{\Delta}} \leq \|\widehat{\boldsymbol{\Delta}}\|_1 \|\mathbf{H}_X \widehat{\boldsymbol{\Delta}}\|_\infty \leq 2C \sqrt{\frac{\log d}{n}} \|\widehat{\boldsymbol{\Delta}}\|_1 \leq 4C \sqrt{\frac{s' \log d}{n}} \|\widehat{\boldsymbol{\Delta}}\|_2.
$$

In addition, by the RE condition $\widehat{\boldsymbol{\Delta}}^T \mathbf{H}_X \widehat{\boldsymbol{\Delta}} \geq \kappa \|\widehat{\boldsymbol{\Delta}}\|_2^2$. This implies that $\|\widehat{\boldsymbol{\Delta}}\|_2 \leq 4\kappa^{-1} C \sqrt{\frac{s' \log d}{n}}$. Finally,

$$
\|\widehat{\boldsymbol{\Delta}}\|_1 \leq 2 \|\widehat{\boldsymbol{\Delta}}_{S'}\|_1 \leq 2 s'^{1/2} \|\widehat{\boldsymbol{\Delta}}_{S'}\|_2 \leq 2 s'^{1/2} \|\widehat{\boldsymbol{\Delta}}\|_2 \leq 8 s' \kappa^{-1} C \sqrt{\frac{\log d}{n}}.
$$

This completes the proof.

□

*Proof of Lemma B.5.* By the Lyapunov Central Limit Theorem, it suffices to show that $\mathbb{E}_{\boldsymbol{\beta}^*}(|A_i|^3)$ is a constant, where $A_i = \sigma^{-2} I_{\theta|\boldsymbol{\gamma}}^{*-1/2} \epsilon_i(Z_i - \mathbf{w}^{*T}\boldsymbol{X}_i)$. Since $I_{\theta|\boldsymbol{\gamma}}^{*-1} = (\mathbf{I}^{*-1})_{\theta\theta}$, we have $I_{\theta|\boldsymbol{\gamma}}^{*-1/2} \leq \kappa^{-1/2}\sigma$. Moreover, by Lemma H.1 and the triangle inequality,

$$\|\epsilon_i(Z_i - \mathbf{w}^{*T}\boldsymbol{X}_i)\|_{\psi_1} \leq 2\|\epsilon_i\|_{\psi_2}\|Z_i - \mathbf{w}^{*T}\boldsymbol{X}_i\|_{\psi_2} \leq 2\|\epsilon_i\|_{\psi_2}(\|Z_i\|_{\psi_2} + \|\mathbf{w}^{*T}\boldsymbol{X}_i\|_{\psi_2}) \leq 4C^2.$$

By the definition of $\psi_1$ norm, $\|\epsilon_i(Z_i - \mathbf{w}^{*T}\boldsymbol{X}_i)\|_{\psi_1} \geq 3^{-1}(\mathbb{E}_{\boldsymbol{\beta}^*}|\epsilon_i(Z_i - \mathbf{w}^{*T})|^3)^{1/3}$. This implies that $\mathbb{E}_{\boldsymbol{\beta}^*}|\epsilon_i(Z_i - \mathbf{w}^{*T})|^3 \leq 12^3 C^6$, and thus $\mathbb{E}_{\boldsymbol{\beta}^*}(|A_i|^3) \leq 12^3 C^6 \sigma^{-3} \kappa^{-3/2}$. Hence, the Lyapunov condition holds, i.e.,

$$n^{-3/2} \sum_{i=1}^n \mathbb{E}_{\boldsymbol{\beta}^*}(|A_i|^3) \leq 12^3 C^6 \sigma^{-3} \kappa^{-3/2} n^{-1/2} = o(1).$$

In addition, together with the Berry-Esseen Theorem in Lemma H.4, we obtain (B.1). $\square$

*Proof of Lemma B.6.* Note that for any $i = 1, ..., n$, $j, k = 1, .., d$,

$$\|Q_{ij}Q_{ik} - \mathbb{E}(Q_{ij}Q_{ik})\|_{\psi_1} \leq 2\|Q_{ij}Q_{ik}\|_{\psi_1} \leq 4\|Q_{ij}\|_{\psi_2}\|Q_{ik}\|_{\psi_2} = 4C^2.$$

By Lemma H.2 and the union bound inequality, we have for any $t > 0$,

$$\mathbb{P}_{\boldsymbol{\beta}^*}\left(\left\|\mathbf{H}_Q - \mathbb{E}(\boldsymbol{Q}_i^{\otimes 2})\right\|_{\max} \geq t\right) \leq 2d^2 \exp\left[-C'' \min\left(\frac{t^2}{16C^4}, \frac{t}{4C^2}\right)n\right],$$

where $C''$ is given in Lemma H.2. With $t = C\sqrt{\frac{\log d}{n}}$, for some constant $C$, Lemma B.6 holds. $\square$

*Proof of Lemma B.7.* We start from the definition of $S(\theta^*, \boldsymbol{\gamma}^*)$ and $\mathbf{I}_{\theta|\boldsymbol{\gamma}}^*$,

$$\sqrt{n}|S(\theta^*, \boldsymbol{\gamma}^*) - S(0, \boldsymbol{\gamma}^*) - \theta^*\mathbf{I}_{\theta|\boldsymbol{\gamma}}^*| = \left|\frac{1}{\sqrt{n}\sigma^2}\sum_{i=1}^n \left\{Z_i(Z_i - \mathbf{w}^{*T}\boldsymbol{X}_i) - \mathbb{E}_{\boldsymbol{\beta}^*}(Z_i(Z_i - \mathbf{w}^{*T}\boldsymbol{X}_i))\right\}\right| \cdot |\theta^*|.$$

By the properties of the $\psi_1$ norm, we obtain,

$$\|Z_i(Z_i - \mathbf{w}^{*T}\boldsymbol{X}_i) - \mathbb{E}_{\boldsymbol{\beta}^*}(Z_i(Z_i - \mathbf{w}^{*T}\boldsymbol{X}_i)\|_{\psi_1} \leq 2\|Z_i(Z_i - \mathbf{w}^{*T}\boldsymbol{X}_i)\|_{\psi_1} \leq 8C^2.$$

By Lemma H.2 and the above inequality, we have,

$$\mathbb{P}_{\boldsymbol{\beta}^*}\left(\left|\frac{1}{n}\sum_{i=1}^n \left\{Z_i(Z_i - \mathbf{w}^{*T}\boldsymbol{X}_i) - \mathbb{E}_{\boldsymbol{\beta}^*} Z_i(Z_i - \mathbf{w}^{*T}\boldsymbol{X}_i)\right\}\right| \geq t\right) \leq 2\exp\left[-C''\min\left(\frac{t^2}{64C^4}, \frac{t}{8C^2}\right)n\right],$$

where $C''$ is given in Lemma H.2. Recall that $\theta^* = \widetilde{C}n^{-\phi}$. With $t = C\sqrt{\frac{\log n}{n}}$ for some constant $C > 0$, the with probability at least $1 - n^{-1}$,

$$\sqrt{n}|S(\theta^*, \boldsymbol{\gamma}^*) - S(0, \boldsymbol{\gamma}^*) - \theta^*\mathbf{I}_{\theta|\boldsymbol{\gamma}}^*| \leq \sigma^{-2} C\widetilde{C}n^{-\phi}\sqrt{\log n}.$$

$\square$

*Proof of Lemma B.8.* By definition, we have

$$\nabla_{\boldsymbol{\gamma}}\ell(0,\boldsymbol{\gamma}^*) = \underbrace{-\frac{1}{\sigma^2 n}\sum_{i=1}^n \boldsymbol{X}_i(Y_i - \theta^* Z_i - \boldsymbol{\gamma}^{*T}\boldsymbol{X}_i)}_{I_1} \underbrace{-\frac{\theta^*}{\sigma^2 n}\sum_{i=1}^n \boldsymbol{X}_i Z_i}_{I_2},$$

where $\theta^* = \widetilde{C}n^{-\phi}$. Under the probability measure $\mathbb{P}_{\boldsymbol{\beta}^*}$, $I_1 = -\frac{1}{\sigma^2 n}\sum_{i=1}^n \boldsymbol{X}_i \epsilon_i$. By Lemma B.1, we have $\|I_1\|_\infty \leq C\sigma^{-2}\sqrt{\frac{\log d}{n}}$ with probability at least $1 - d^{-1}$. Denote $\mathbf{W}_i = \boldsymbol{X}_i Z_i - \mathbb{E}(\boldsymbol{X}_i Z_i)$. Since $\|W_{ij}\|_{\psi_1} \leq 2\|X_{ij}Z_i\|_{\psi_1} \leq 4C^2$, Lemma H.2 together with the union bound implies that $\|\frac{1}{n}\sum_{i=1}^n \mathbf{W}_i\|_\infty \leq C\sqrt{\frac{\log d}{n}}$ for some constant $C > 0$, with probability at least $1 - d^{-1}$. By the definition of the sub-Gaussian norm, $\mathbb{E}(X_{ij}Z_i) \leq (\mathbb{E}X_{ij}^2)^{1/2}(\mathbb{E}Z_i^2)^{1/2} \leq 2C^2$. This implies that $\|I_2\|_\infty \leq (2C^2 + C\sqrt{\frac{\log d}{n}})\widetilde{C}\sigma^{-2}n^{-\phi}$, with probability at least $1 - d^{-1}$. Combining the bounds for $I_1$ and $I_2$, with probability tending to one, we have uniformly over $\boldsymbol{\beta}^* \in \Omega_1(\widetilde{C}, \phi)$,

$$\|\nabla_{\boldsymbol{\gamma}}\ell(0,\boldsymbol{\gamma}^*)\|_\infty \leq C\sigma^{-2}\sqrt{\frac{\log d}{n}} + \left(2C^2 + C\sqrt{\frac{\log d}{n}}\right)\frac{\widetilde{C}}{\sigma^2 n^\phi}.$$

□

*Proof of Lemma B.9.* Consider the singular value decomposition for $\mathbf{X}_S$, $\mathbf{X}_S = \mathbf{U}\boldsymbol{\Lambda}\mathbf{V}^T$, where $\mathbf{U} \in \mathbb{R}^{n \times n}$ is an orthogonal matrix, $\boldsymbol{\Lambda} \in \mathbb{R}^{n \times s^*}$ is a diagonal matrix and $\mathbf{V} \in \mathbb{R}^{s^* \times s^*}$ is an orthogonal matrix. Let $\boldsymbol{\Delta} = \boldsymbol{\Lambda}\boldsymbol{\Lambda}^T \in \mathbb{R}^{n \times n}$. Thus,

$$(k\mathbf{X}_S\mathbf{X}_S^T + \sigma^2\mathbf{I}_n)^{-1} = \mathbf{U}(k\boldsymbol{\Delta} + \sigma^2\mathbf{I}_n)^{-1}\mathbf{U}^T.$$

Note that $\boldsymbol{\Delta}$ is a diagonal matrix with first $s^*$ elements being positive. For any $n$, as $k \to \infty$, we have

$$\frac{1}{n}\mathbf{Z}^T(k\mathbf{X}_S\mathbf{X}_S^T + \sigma^2\mathbf{I}_n)^{-1}\mathbf{Z} = \frac{1}{n}\mathbf{Z}^T\mathbf{U}(\sigma^{-2}\mathbf{D})\mathbf{U}^T\mathbf{Z} + o_\mathbb{P}(1), \tag{G.6}$$

where $\mathbf{D}$ is a diagonal matrix with first $s^*$ elements being 0 and the last $(n - s^*)$ elements being 1. Consider the following identity, $\mathbf{Z} = \mathbf{Z}^{(1)} + \mathbf{Z}^{(2)}$, where $\mathbf{Z}^{(1)} = \mathbf{Z} - \mathbf{X}_S\mathbb{E}(\boldsymbol{X}_{i,S}^{\otimes 2})^{-1}\mathbb{E}(Z_i\boldsymbol{X}_{i,S})$ and $\mathbf{Z}^{(2)} = \mathbf{X}_S\mathbb{E}(\boldsymbol{X}_{i,S}^{\otimes 2})^{-1}\mathbb{E}(Z_i\boldsymbol{X}_{i,S})$. By definition of $\mathbf{Z}^{(2)}$ and the singular decomposition for $\mathbf{X}_S$, we obtain

$$\mathbf{Z}^{T(2)}\mathbf{U}\mathbf{D}\mathbf{U}^T\mathbf{Z}^{(2)} = \mathbb{E}(Z_i\boldsymbol{X}_{i,S}^T)\mathbb{E}(\boldsymbol{X}_{i,S}^{\otimes 2})^{-1}\mathbf{V}\boldsymbol{\Lambda}^T\mathbf{D}\boldsymbol{\Lambda}\mathbf{V}^T\mathbb{E}(\boldsymbol{X}_{i,S}^{\otimes 2})^{-1}\mathbb{E}(Z_i\boldsymbol{X}_{i,S}).$$

By the form of $\mathbf{D}$ and $\boldsymbol{\Lambda}$, it is easily seen that $\boldsymbol{\Lambda}^T\mathbf{D}\boldsymbol{\Lambda} = \mathbf{0}$. Following the similar arguments, we can show that $\mathbf{Z}^{T(1)}\mathbf{U}\mathbf{D}\mathbf{U}^T\mathbf{Z}^{(2)} = \mathbf{0}$ and $\mathbf{Z}^{T(2)}\mathbf{U}\mathbf{D}\mathbf{U}^T\mathbf{Z}^{(1)} = \mathbf{0}$. It remains to calculate $\mathbf{Z}^{T(1)}\mathbf{U}\mathbf{D}\mathbf{U}^T\mathbf{Z}^{(1)}$. Simple algebra shows that $\mathbb{E}(Z_i^{(1)}\boldsymbol{X}_{i,S}) = \mathbf{0}$. Since $Z_i$ and $\boldsymbol{X}_{i,S}$ are all Gaussian, it implies that $\mathbf{Z}^{(1)}$ are independent of $\mathbf{X}_S$ and therefore $\mathbf{Z}^{(1)}$ are independent of $\mathbf{U}$. Note that $Z_i^{(1)} \sim N(0, K)$, where $K = \mathbb{E}(Z_i^2) - \mathbb{E}(Z_i\boldsymbol{X}_{i,S}^T)\mathbb{E}(\boldsymbol{X}_{i,S}^{\otimes 2})^{-1}\mathbb{E}(\boldsymbol{X}_{i,S}Z_i)$. This implies that $\mathbf{Z}^{T(1)}\mathbf{U} \sim N(0, K\mathbf{I}_n)$. Thus,

$$\frac{1}{n}\mathbf{Z}^T\mathbf{U}(\sigma^{-2}\mathbf{D})\mathbf{U}^T\mathbf{Z} \sim \frac{1}{n}\sigma^{-2}K\chi^2_{(n-s^*)},$$

where $\chi_k^2$ is the $\chi^2$ distribution with $k$ degree of freedom. Hence, together with (G.6), we finally obtain
$$\frac{1}{n}\mathbf{Z}^T(k\mathbf{X}_S\mathbf{X}_S^T + \sigma^2 \mathbf{I}_n)^{-1}\mathbf{Z} = \sigma^{-2}K + o_{\mathbb{P}}(1) = I_{\theta|S}^* + o_{\mathbb{P}}(1).$$
$\square$

**Lemma G.2.** *Under the conditions of Proposition 4.12,*
$$\lim_{n\to\infty} \mathbb{P}_{\boldsymbol{\beta}^*}\left(\left\|\frac{1}{n}\sum_{i=1}^n (Z_i\boldsymbol{X}_{i,\widehat{S}} - \widehat{\mathbf{v}}^T \boldsymbol{X}_{i,\widehat{S}}^{\otimes 2})\right\|_\infty \geq C\sqrt{\frac{\log s^*}{n}}\right) = 0,$$
*for some constant $C > 0$.*

*Proof of Lemma G.2.* By Theorem 1 of Wainwright (2009), the event $\{\widehat{S} = S\}$ holds with high probability, i.e. $\lim_{n\to\infty} \mathbb{P}_{\boldsymbol{\beta}^*}(\widehat{S} \neq S) = 0$. Hence,

$$\begin{aligned}
&\mathbb{P}_{\boldsymbol{\beta}^*}\left(\left\|\frac{1}{n}\sum_{i=1}^n (Z_i\boldsymbol{X}_{i,\widehat{S}} - \widehat{\mathbf{v}}^T \boldsymbol{X}_{i,\widehat{S}}^{\otimes 2})\right\|_\infty \geq C\sqrt{\frac{\log s^*}{n}}\right) \\
&\leq \mathbb{P}_{\boldsymbol{\beta}^*}\left(\left\|\frac{1}{n}\sum_{i=1}^n (Z_i\boldsymbol{X}_{i,\widehat{S}} - \widehat{\mathbf{v}}^T \boldsymbol{X}_{i,\widehat{S}}^{\otimes 2})\right\|_\infty \geq C\sqrt{\frac{\log s^*}{n}}, \widehat{S} = S\right) + \mathbb{P}_{\boldsymbol{\beta}^*}(\widehat{S} \neq S) \\
&\leq \mathbb{P}_{\boldsymbol{\beta}^*}\left(\left\|\frac{1}{n}\sum_{i=1}^n (Z_i\boldsymbol{X}_{i,S} - \mathbf{v}^{*T} \boldsymbol{X}_{i,S}^{\otimes 2})\right\|_\infty \geq C\sqrt{\frac{\log s^*}{n}}\right) + \mathbb{P}_{\boldsymbol{\beta}^*}(\widehat{S} \neq S).
\end{aligned}$$

Similar to the proof of Lemma B.2, we can show that the first term goes to 0, as $n \to \infty$. To save space, we do not replicate the details. Thus, Lemma G.2 holds. $\square$

### G.3 Proofs of Auxiliary Lemmas in Appendix C

*Proof of Lemma C.1.* As shown in Assumption 5.1, $\mathbb{E}_{\boldsymbol{\beta}^*}(Y_i \mid \boldsymbol{Q}_i) = b'(\boldsymbol{Q}_i^T\boldsymbol{\beta}^*)$, $\max_{1\leq i\leq n} |Y_i - b'(\boldsymbol{Q}_i^T\boldsymbol{\beta}^*)| \leq K'$, and $\max_{i,j} |X_{ij}| \leq K$, by the Hoeffding inequality, we have for any $t > 0$,
$$\mathbb{P}_{\boldsymbol{\beta}^*}\left(\left|\frac{1}{n}\sum_{i=1}^n (-Y_i + b'(\boldsymbol{Q}_i^T\boldsymbol{\beta}^*))X_{ij}\right| \geq t\right) \leq 2\exp\left(-\frac{nt^2}{2(KK')^2}\right).$$

By the union bound inequality, with $t = 2KK'\sqrt{\frac{\log d}{n}}$, we have
$$\mathbb{P}_{\boldsymbol{\beta}^*}\left(\left\|\frac{1}{n}\sum_{i=1}^n (-Y_i + b'(\boldsymbol{Q}_i^T\boldsymbol{\beta}^*))\boldsymbol{X}_i\right\|_\infty \geq 2KK'\sqrt{\frac{\log d}{n}}\right) \leq 2d^{-1}.$$
$\square$

*Proof of Lemma C.2.* By the Lyapunov Central Limit Theorem, it suffices to show that $\mathbb{E}_{\boldsymbol{\beta}^*}(|A_i|^3)$ is a constant, where $A_i = I_{\theta|\gamma}^{*-1/2}(Y_i - b'(\boldsymbol{Q}_i^T\boldsymbol{\beta}^*))(Z_i - \mathbf{w}^{*T}\boldsymbol{X}_i)$. Since $I_{\theta|\gamma}^{*-1} = (\mathbf{I}^{*-1})_{\theta\theta}$, we have $I_{\theta|\gamma}^{*-1/2} \leq \kappa^{-1}$. By Assumption 5.1, $|A_i| \leq 2\kappa^{-1}K'K$. Hence, the Lyapunov condition holds, i.e.,
$$n^{-3/2}\sum_{i=1}^n \mathbb{E}_{\boldsymbol{\beta}^*}(|A_i|^3) \leq 2^3\kappa^{-3}K'^3K^3 n^{-1/2} = o(1).$$

In addition, together with the Berry-Esseen Theorem in Lemma H.4, we obtain the result. □

**Lemma G.3.** Denote $\mathbf{F}(\boldsymbol{\beta}) = \frac{1}{n}\sum_{i=1}^{n} b''(\boldsymbol{Q}_i^T\boldsymbol{\beta})\boldsymbol{X}_i^{\otimes 2}$, and

$$\kappa_D(s') = \min\left\{\frac{s'^{1/2}(\mathbf{v}^T\mathbf{F}(\widehat{\boldsymbol{\beta}})\mathbf{v})^{1/2}}{||\mathbf{v}_{\mathcal{S}'}||_1} : \mathbf{v} \in \mathbb{R}^{d-1}\backslash\{0\}, ||\mathbf{v}_{\bar{\mathcal{S}}'}||_1 \leq \xi||\mathbf{v}_{\mathcal{S}'}||_1\right\},$$

where $\xi$ is a positive constant. Assume that Assumptions 5.1 and 5.2 hold. Then, $\kappa_D(s') \geq \kappa/\sqrt{2}$, with probability tending to one.

*Proof of Lemma G.3.* By the definition of $\kappa_D(s')$ and the fact that $||\mathbf{v}_{\mathcal{S}'}||_1 \leq s'^{1/2}||\mathbf{v}_{\mathcal{S}'}||_2 \leq s'^{1/2}||\mathbf{v}||_2$, we have

$$\kappa_D^2(s') \geq \min\left\{\frac{\mathbf{v}^T\mathbf{F}(\widehat{\boldsymbol{\beta}})\mathbf{v}}{||\mathbf{v}||_2^2} : \mathbf{v} \in \mathbb{R}^{d-1}\backslash\{0\}, ||\mathbf{v}_{\bar{\mathcal{S}}'}||_1 \leq \xi||\mathbf{v}_{\mathcal{S}'}||_1\right\}.$$

For notational simplicity, we denote $\widehat{b}_i'' = b''(\widehat{\boldsymbol{\beta}}^T\boldsymbol{Q}_i)$ and $b_i'' = b''(\boldsymbol{\beta}^{*T}\boldsymbol{Q}_i)$. By the definition of $\mathbf{F}(\widehat{\boldsymbol{\beta}})$,

$$\frac{\mathbf{v}^T\mathbf{F}(\widehat{\boldsymbol{\beta}})\mathbf{v}}{||\mathbf{v}||_2^2} = \frac{\mathbf{v}^T\mathbf{F}(\boldsymbol{\beta}^*)\mathbf{v}}{||\mathbf{v}||_2^2} + \frac{1}{n}\sum_{i=1}^{n}\frac{(\boldsymbol{X}_i^T\mathbf{v})^2}{||\mathbf{v}||_2^2}(\widehat{b}_i'' - b_i'').$$

By Assumption 5.1, we have

$$\max_{1\leq i\leq n}|\widehat{b}_i'' - b_i''| = \max_{1\leq i\leq n}\left|\frac{\widehat{b}_i'' - b_i''}{\boldsymbol{Q}_i^T(\widehat{\boldsymbol{\beta}} - \boldsymbol{\beta}^*)}\right|\cdot|\boldsymbol{Q}_i^T(\widehat{\boldsymbol{\beta}} - \boldsymbol{\beta}^*)| \lesssim ||(\boldsymbol{\beta}^* - \widehat{\boldsymbol{\beta}})||_1 = o_{\mathbb{P}}(1).$$

Hence, $\frac{\mathbf{v}^T\mathbf{F}(\widehat{\boldsymbol{\beta}})\mathbf{v}}{||\mathbf{v}||_2^2} \geq \frac{3}{4}\frac{\mathbf{v}^T\mathbf{F}(\boldsymbol{\beta}^*)\mathbf{v}}{||\mathbf{v}||_2^2}$, with probability tending to one. Note that

$$\frac{\mathbf{v}^T\mathbf{F}(\widehat{\boldsymbol{\beta}})\mathbf{v}}{||\mathbf{v}||_2^2} \geq \frac{3}{4}\frac{\mathbf{v}^T\mathbf{F}(\boldsymbol{\beta}^*)\mathbf{v}}{||\mathbf{v}||_2^2} = \frac{3}{4}\left(\frac{\mathbf{v}^T\mathbf{I}_{\gamma\gamma}^*\mathbf{v}}{||\mathbf{v}||_2^2} + \frac{\mathbf{v}^T(\mathbf{F}(\boldsymbol{\beta}^*) - \mathbf{I}_{\gamma\gamma}^*)\mathbf{v}}{||\mathbf{v}||_2^2}\right)$$

$$\geq \frac{3}{4}\left(\lambda_{\min}(\mathbf{I}_{\gamma\gamma}^*) - \left|\frac{\mathbf{v}^T(\mathbf{F}(\boldsymbol{\beta}^*) - \mathbf{I}_{\gamma\gamma}^*)\mathbf{v}}{||\mathbf{v}||_2^2}\right|\right) \geq \frac{3}{4}\left(\kappa^2 - \frac{||\mathbf{v}||_1^2||\mathbf{F}(\boldsymbol{\beta}^*) - \mathbf{I}_{\gamma\gamma}^*||_{\max}}{||\mathbf{v}||_2^2}\right).$$

In addition, by $||\mathbf{v}||_1^2 \leq (\xi+1)^2||\mathbf{v}_{\mathcal{S}'}||_1^2 \leq s'(\xi+1)^2||\mathbf{v}||_2^2$, we can show that

$$\frac{\mathbf{v}^T\mathbf{F}(\widehat{\boldsymbol{\beta}})\mathbf{v}}{||\mathbf{v}||_2^2} \geq \frac{3}{4}\big(\kappa^2 - (\xi+1)^2 s'||\mathbf{F}(\boldsymbol{\beta}^*) - \mathbf{I}_{\gamma\gamma}^*||_{\max}\big).$$

Similar to the proof of Lemma G.4, we obtain $||\mathbf{F}(\boldsymbol{\beta}^*) - \mathbf{I}_{\gamma\gamma}^*||_{\max} = \mathcal{O}_{\mathbb{P}}(\sqrt{\frac{\log d}{n}})$. Since by assumption, $s'\sqrt{\frac{\log d}{n}} = o_{\mathbb{P}}(1)$, we obtain $s'||\mathbf{F}(\boldsymbol{\beta}^*) - \mathbf{I}_{\gamma\gamma}^*||_{\max} = o_{\mathbb{P}}(1)$. Therefore, for $n$ large enough, $s'||\mathbf{F}(\boldsymbol{\beta}^*) - \mathbf{I}_{\gamma\gamma}^*||_{\max} \leq \kappa^2/(3(\xi+1)^2)$. Hence, $\kappa_D(s') \geq \kappa/\sqrt{2}$, with probability tending to one, which completes the proof. □

*Proof of Lemma C.3.* We first show that $||(\widehat{\boldsymbol{\Delta}})_{\bar{\mathcal{S}}'}||_1 \leq ||(\widehat{\boldsymbol{\Delta}})_{\mathcal{S}'}||_1$, where $\widehat{\boldsymbol{\Delta}} = \widehat{\mathbf{w}} - \mathbf{w}^*$. Since $||\mathbf{w}^*||_1 \geq ||\widehat{\mathbf{w}}||_1$, we have

$$\sum_{j\in\mathcal{S}'}|w_j^*| \geq \sum_{j\in\mathcal{S}'}|\widehat{w}_j| + \sum_{j\in\bar{\mathcal{S}}'}|\widehat{w}_j| \geq \sum_{j\in\mathcal{S}'}|w_j^*| - \sum_{j\in\mathcal{S}'}|\widehat{\Delta}_j| + \sum_{j\in\bar{\mathcal{S}}'}|\widehat{w}_j|.$$

This implies that $\sum_{j\in\mathcal{S}'}|\widehat{\Delta}_j| \geq \sum_{j\in\bar{\mathcal{S}}'}|\widehat{w}_j| = \sum_{j\in\bar{\mathcal{S}}'}|\widehat{\Delta}_j|$. Denote $\mathbf{F}(\boldsymbol{\beta}) = \frac{1}{n}\sum_{i=1}^n b''(\boldsymbol{Q}_i^T\boldsymbol{\beta})\boldsymbol{X}_i^{\otimes 2}$. We consider the following quadratic function of $\widehat{\boldsymbol{\Delta}}$,

$$\begin{aligned}\widehat{\boldsymbol{\Delta}}^T\mathbf{F}(\widehat{\boldsymbol{\beta}})\widehat{\boldsymbol{\Delta}} &= \frac{1}{n}\sum_{i=1}^n b''(\boldsymbol{Q}_i^T\widehat{\boldsymbol{\beta}})\widehat{\boldsymbol{\Delta}}^T\boldsymbol{X}_i^{\otimes 2}\widehat{\boldsymbol{\Delta}} \\ &= \underbrace{\frac{1}{n}\sum_{i=1}^n b''(\boldsymbol{Q}_i^T\widehat{\boldsymbol{\beta}})\widehat{\boldsymbol{\Delta}}^T\boldsymbol{X}_i\{Z_i - \mathbf{w}^{*T}\boldsymbol{X}_i\}}_{I_1} - \underbrace{\frac{1}{n}\sum_{i=1}^n b''(\boldsymbol{Q}_i^T\widehat{\boldsymbol{\beta}})\widehat{\boldsymbol{\Delta}}^T\boldsymbol{X}_i\{Z_i - \widehat{\mathbf{w}}^T\boldsymbol{X}_i\}}_{I_2}.\end{aligned}$$

By assumption, for $I_1$,

$$I_1 \leq \|\widehat{\boldsymbol{\Delta}}\|_1\left\|\frac{1}{n}\sum_{i=1}^n b''(\boldsymbol{Q}_i^T\widehat{\boldsymbol{\beta}})\boldsymbol{X}_i\{Z_i - \mathbf{w}^{*T}\boldsymbol{X}_i\}\right\|_\infty \leq \lambda'\|\widehat{\boldsymbol{\Delta}}\|_1.$$

Similarly, for $I_2$, by the definition of the Dantzig selector,

$$I_2 \leq \|\widehat{\boldsymbol{\Delta}}\|_1\left\|\frac{1}{n}\sum_{i=1}^n b''(\boldsymbol{Q}_i^T\widehat{\boldsymbol{\beta}})\boldsymbol{X}_i\{Z_i - \widehat{\mathbf{w}}^T\boldsymbol{X}_i\}\right\|_\infty \leq \lambda'\|\widehat{\boldsymbol{\Delta}}\|_1.$$

Hence, $\widehat{\boldsymbol{\Delta}}^T\mathbf{F}(\widehat{\boldsymbol{\beta}})\widehat{\boldsymbol{\Delta}} = I_1 + I_2 \leq 2\lambda'\|\widehat{\boldsymbol{\Delta}}\|_1 \leq 4\lambda'\|(\widehat{\boldsymbol{\Delta}})_{\mathcal{S}'}\|_1$. On the other hand, by the definition of $\kappa_D(s')$ and Lemma G.3, with probability tending to one,

$$\widehat{\boldsymbol{\Delta}}^T\mathbf{F}(\widehat{\boldsymbol{\beta}})\widehat{\boldsymbol{\Delta}} \geq \kappa_D^2(s')\|(\widehat{\boldsymbol{\Delta}})_{\mathcal{S}'}\|_1^2 s'^{-1} \geq \frac{\kappa^2}{2}\|(\widehat{\boldsymbol{\Delta}})_{\mathcal{S}'}\|_1^2 s'^{-1}.$$

This implies that $\|(\widehat{\boldsymbol{\Delta}})_{\mathcal{S}'}\|_1 \leq 8\lambda' s'/\kappa^2$. Hence, $\|\widehat{\boldsymbol{\Delta}}\|_1 \leq 16\lambda' s'/\kappa^2$. By condition (C4) and the proof of Theorem 3.3 in van de Geer et al. (2014), we can take $\lambda' \asymp \sqrt{\frac{\log d}{n}}$. Thus, we obtain $\|\widehat{\boldsymbol{\Delta}}\|_1 \leq \frac{16s'}{\kappa^2}\sqrt{\frac{\log d}{n}}$ with high probability, and therefore, $\widehat{\boldsymbol{\Delta}}^T\mathbf{F}(\widehat{\boldsymbol{\beta}})\widehat{\boldsymbol{\Delta}} \leq \frac{32}{\kappa^2}\frac{s'\log d}{n}$. We complete the proof. □

*Proof of Lemma C.4.* Denote $\widetilde{\boldsymbol{\Delta}} = \widetilde{\mathbf{w}} - \mathbf{w}^*$, $\widehat{b}_i'' = b''(\boldsymbol{Q}_i^T\widehat{\boldsymbol{\beta}})$ and $b_i'' = b''(\boldsymbol{Q}_i^T\boldsymbol{\beta}^*)$. By Definition,

$$\frac{1}{n}\sum_{i=1}^n \widehat{b}_i''(Z_i - \widetilde{\mathbf{w}}^T\boldsymbol{X}_i)^2 + \lambda'\|\widetilde{\mathbf{w}}\|_1 \leq \frac{1}{n}\sum_{i=1}^n \widehat{b}_i''(Z_i - \mathbf{w}^{*T}\boldsymbol{X}_i)^2 + \lambda'\|\mathbf{w}^*\|_1.$$

By rearranging terms, we can show that it is equivalent to

$$\frac{1}{n}\sum_{i=1}^n \widehat{b}_i''(\widetilde{\boldsymbol{\Delta}}^T\boldsymbol{X}_i)^2 \leq \underbrace{\frac{2}{n}\sum_{i=1}^n \widehat{b}_i''(Z_i - \mathbf{w}^{*T}\boldsymbol{X}_i)\widetilde{\boldsymbol{\Delta}}^T\boldsymbol{X}_i}_{I_1} + \lambda'\|\mathbf{w}^*\|_1 - \lambda'\|\widetilde{\mathbf{w}}\|_1. \quad (G.7)$$

We now further bound the right hand side of (G.7). Since $\mathbf{w}_{\bar{\mathcal{S}}'}^* = 0$, we have

$$\lambda'\|\mathbf{w}^*\|_1 - \lambda'\|\widetilde{\mathbf{w}}\|_1 \leq \lambda'\|\mathbf{w}_{\mathcal{S}'}^*\|_1 - \lambda'\|\widetilde{\mathbf{w}}_{\mathcal{S}'}\|_1 - \lambda'\|\widetilde{\mathbf{w}}_{\bar{\mathcal{S}}'}\|_1 \leq \lambda'\|\widetilde{\boldsymbol{\Delta}}_{\mathcal{S}'}\|_1 - \lambda'\|\widetilde{\boldsymbol{\Delta}}_{\bar{\mathcal{S}}'}\|_1.$$

Note that for $I_1$, we have

$$I_1 = \frac{2}{n}\sum_{i=1}^{n} b_i''(Z_i - \mathbf{w}^{*T}\mathbf{X}_i)\widetilde{\boldsymbol{\Delta}}^T\mathbf{X}_i + \frac{2}{n}\sum_{i=1}^{n}(\widehat{b}_i'' - b_i'')(Z_i - \mathbf{w}^{*T}\mathbf{X}_i)\widetilde{\boldsymbol{\Delta}}^T\mathbf{X}_i.$$

By the Hölder inequality, and Lemma C.1, with probability tending to one,

$$\left|\frac{2}{n}\sum_{i=1}^{n} b_i''(Z_i - \mathbf{w}^{*T}\mathbf{X}_i)\widetilde{\boldsymbol{\Delta}}^T\mathbf{X}_i\right| \leq \left|\frac{2}{n}\sum_{i=1}^{n} b_i''(Z_i - \mathbf{w}^{*T}\mathbf{X}_i)\mathbf{X}_i\right|_\infty \cdot \|\widetilde{\boldsymbol{\Delta}}\|_1 \leq C\sqrt{\frac{\log d}{n}}\|\widetilde{\boldsymbol{\Delta}}\|_1,$$

for some constant $C$. For the second term, the Cauchy-Schwartz inequality yields,

$$\frac{2}{n}\sum_{i=1}^{n}(\widehat{b}_i'' - b_i'')(Z_i - \mathbf{w}^{*T}\mathbf{X}_i)\widetilde{\boldsymbol{\Delta}}^T\mathbf{X}_i \leq 2\left|\frac{1}{n}\sum_{i=1}^{n}\widehat{b}_i''(\widetilde{\boldsymbol{\Delta}}^T\mathbf{X}_i)^2\right|^{1/2} \cdot \left|\frac{1}{n}\sum_{i=1}^{n}\frac{(\widehat{b}_i'' - b_i'')^2}{\widehat{b}_i''}(Z_i - \mathbf{w}^{*T}\mathbf{X}_i)^2\right|^{1/2}.$$

Similar to the proof of Theorem 5.3, it is easily seen that

$$\left|\frac{1}{n}\sum_{i=1}^{n}\frac{(\widehat{b}_i'' - b_i'')^2}{\widehat{b}_i''}(Z_i - \mathbf{w}^{*T}\mathbf{X}_i)^2\right|^{1/2} \leq C\left|\frac{1}{n}\sum_{i=1}^{n}\left(\mathbf{Q}_i^T(\widehat{\boldsymbol{\beta}} - \boldsymbol{\beta}^*)\right)^2\right|^{1/2} \leq C'\sqrt{s^*}\lambda,$$

for some constants $C$ and $C'$. Denote $V = \frac{1}{n}\sum_{i=1}^{n}\widehat{b}_i''(\widetilde{\boldsymbol{\Delta}}^T\mathbf{X}_i)^2$. Plugging these results into (G.7) and taking $\lambda' = 2C\sqrt{\frac{\log d}{n}}$, we obtain

$$V \leq C'\sqrt{\frac{s^*\log d}{n}}V^{1/2} + 3C\sqrt{\frac{\log d}{n}}\|\widetilde{\boldsymbol{\Delta}}_{S'}\|_1 - C\sqrt{\frac{\log d}{n}}\|\widetilde{\boldsymbol{\Delta}}_{\bar{S}'}\|_1, \tag{G.8}$$

for some constants $C$ and $C'$. Now, we consider two situations. If $V^{1/2} \leq C'\sqrt{\frac{s^*\log d}{n}}$, the result (C.2) holds trivially. Otherwise, we have $V^{1/2} > C'\sqrt{\frac{s^*\log d}{n}}$. In this case, $V - C'\sqrt{\frac{s^*\log d}{n}}V^{1/2} > 0$, together with (G.8), this implies $3\|\widetilde{\boldsymbol{\Delta}}_{S'}\|_1 \geq \|\widetilde{\boldsymbol{\Delta}}_{\bar{S}'}\|_1$. Due to this cone condition for $\widetilde{\boldsymbol{\Delta}}$, we can apply Lemma G.3 with $\xi = 3$ to conclude that $\|\widetilde{\boldsymbol{\Delta}}_{S'}\|_1 \leq C''\sqrt{s'}V^{1/2}$ for some constant $C''$. By (G.8), $V \leq C'\sqrt{\frac{s^*\log d}{n}}V^{1/2} + 3CC''\sqrt{\frac{s'\log d}{n}}V^{1/2}$, which implies that $V^{1/2} = \mathcal{O}_{\mathbb{P}}(\sqrt{\frac{(s^*\vee s')\log d}{n}})$. We complete the proof of (C.2). To prove (C.1), we still consider two situations. First, if $6\|\widetilde{\boldsymbol{\Delta}}_{S'}\|_1 \geq \|\widetilde{\boldsymbol{\Delta}}_{\bar{S}'}\|_1$, then we have $\|\widetilde{\boldsymbol{\Delta}}\|_1 \leq 7\|\widetilde{\boldsymbol{\Delta}}_{S'}\|_1 \leq C''\sqrt{s'}V^{1/2}$, by Lemma G.3 with $\xi = 6$. Therefore, by (C.2), we obtain $\|\widetilde{\boldsymbol{\Delta}}\|_1 = \mathcal{O}_{\mathbb{P}}((s^* \vee s')\sqrt{\frac{\log d}{n}})$. Otherwise, we have $6\|\widetilde{\boldsymbol{\Delta}}_{S'}\|_1 < \|\widetilde{\boldsymbol{\Delta}}_{\bar{S}'}\|_1$. Together with (G.8), this implies that $V \leq C'\sqrt{\frac{s^*\log d}{n}}V^{1/2} - 3C\sqrt{\frac{\log d}{n}}\|\widetilde{\boldsymbol{\Delta}}_{S'}\|_1$. Hence,

$$\|\widetilde{\boldsymbol{\Delta}}\|_1 \leq \frac{7}{6}\|\widetilde{\boldsymbol{\Delta}}_{\bar{S}'}\|_1 \leq \frac{7}{6}\left(3C\sqrt{\frac{\log d}{n}}\right)^{-1}\left(C'\sqrt{\frac{s^*\log d}{n}}V^{1/2} - V\right) \leq \frac{7C'}{18C}\sqrt{s^*}V^{1/2}.$$

Therefore, by (C.2), we obtain $\|\widetilde{\boldsymbol{\Delta}}\|_1 = \mathcal{O}_{\mathbb{P}}((s^* \vee s')\sqrt{\frac{\log d}{n}})$. The proof of (C.1) is finished.

□

**Lemma G.4.** Denote $\mathbf{D}(\boldsymbol{\beta}) = \frac{1}{n}\sum_{i=1}^n (Y_i - b'(\boldsymbol{\beta}^T \mathbf{Q}_i))^2 \mathbf{X}_i^{\otimes 2}$, and

$$\kappa_R(s') = \min\left\{\frac{s'^{1/2}(\mathbf{v}^T \mathbf{D}(\widehat{\boldsymbol{\beta}})\mathbf{v})^{1/2}}{\|\mathbf{v}_{\mathcal{S}'}\|_1} : \mathbf{v} \in \mathbb{R}^{d-1}\setminus\{0\}, \|\mathbf{v}_{\bar{\mathcal{S}}'}\|_1 \leq \xi\|\mathbf{v}_{\mathcal{S}'}\|_1\right\},$$

for some constant $\xi > 0$. Under the conditions in Theorem 5.5, we obtain $\kappa_R(s') \geq C\kappa$, for some constant $C > 0$, with probability tending to one.

*Proof of Lemma G.4.* By the definition of $\kappa_R(s')$ and the fact that $\|\mathbf{v}_{\mathcal{S}'}\|_1 \leq s'^{1/2}\|\mathbf{v}_{\mathcal{S}'}\|_2 \leq s'^{1/2}\|\mathbf{v}\|_2$, we have

$$\kappa_R^2(s') \geq \min\left\{\frac{\mathbf{v}^T \mathbf{D}(\widehat{\boldsymbol{\beta}})\mathbf{v}}{\|\mathbf{v}\|_2^2} : \mathbf{v} \in \mathbb{R}^{d-1}\setminus\{0\}, \|\mathbf{v}_{\bar{\mathcal{S}}'}\|_1 \leq \xi\|\mathbf{v}_{\mathcal{S}'}\|_1\right\}.$$

For notational simplicity, we denote $\widehat{b}'_i = b'(\widehat{\boldsymbol{\beta}}^T \mathbf{Q}_i)$ and $b'_i = b'(\boldsymbol{\beta}^{*T} \mathbf{Q}_i)$. By the definition of $\mathbf{D}(\widehat{\boldsymbol{\beta}})$,

$$\frac{\mathbf{v}^T \mathbf{D}(\widehat{\boldsymbol{\beta}})\mathbf{v}}{\|\mathbf{v}\|_2^2} = \frac{1}{n}\sum_{i=1}^n \frac{(\mathbf{X}_i^T \mathbf{v})^2}{\|\mathbf{v}\|_2^2}\left\{(Y_i - b'_i)^2 + (b'_i - \widehat{b}'_i)^2 + (Y_i - b'_i)(b'_i - \widehat{b}'_i)\right\}.$$

By assumptions in Theorem 5.5, we have $\max_{1\leq i\leq n}(Y_i - b'_i)^2 \geq C > 0$ for some constant $C$. In addition,

$$\max_{1\leq i\leq n}(b'_i - \widehat{b}'_i)^2 = \max_{1\leq i\leq n} b''^2(\mathbf{Q}_i^T \widetilde{\boldsymbol{\beta}})\{\mathbf{Q}_i^T(\boldsymbol{\beta}^* - \widehat{\boldsymbol{\beta}})\}^2 \lesssim \|(\boldsymbol{\beta}^* - \widehat{\boldsymbol{\beta}})\|_1^2 = o_{\mathbb{P}}(1),$$

where $\widetilde{\boldsymbol{\beta}}$ is an intermediate value between $\boldsymbol{\beta}^*$ and $\widehat{\boldsymbol{\beta}}$. Similarly,

$$\max_{1\leq i\leq n}|(Y_i - b'_i)(b'_i - \widehat{b}'_i)| \lesssim \max_{1\leq i\leq n} b''(\mathbf{Q}_i^T \widetilde{\boldsymbol{\beta}})|\mathbf{Q}_i^T(\boldsymbol{\beta}^* - \widehat{\boldsymbol{\beta}})| \lesssim \|(\boldsymbol{\beta}^* - \widehat{\boldsymbol{\beta}})\|_1 = o_{\mathbb{P}}(1).$$

Therefore, we get, with probability tending to one,

$$\frac{\mathbf{v}^T \mathbf{D}(\widehat{\boldsymbol{\beta}})\mathbf{v}}{\|\mathbf{v}\|_2^2} \geq C\frac{1}{n}\sum_{i=1}^n \frac{(\mathbf{X}_i^T \mathbf{v})^2}{\|\mathbf{v}\|_2^2} \geq C'\frac{1}{n}\sum_{i=1}^n \frac{b''(\mathbf{Q}_i^T \boldsymbol{\beta}^*)(\mathbf{X}_i^T \mathbf{v})^2}{\|\mathbf{v}\|_2^2}.$$

for some constant $C > 0$. Here, we use the assumption that $\max_{1\leq i\leq n} b''(\mathbf{Q}_i^T \boldsymbol{\beta}^*) \leq C$ for some constant $C$. Recall that $\mathbf{F}(\boldsymbol{\beta}) = \frac{1}{n}\sum_{i=1}^n b''(\mathbf{Q}_i^T \boldsymbol{\beta})\mathbf{X}_i^{\otimes 2}$. Then,

$$\frac{1}{n}\sum_{i=1}^n \frac{b''(\mathbf{Q}_i^T \boldsymbol{\beta}^*)(\mathbf{X}_i^T \mathbf{v})^2}{\|\mathbf{v}\|_2^2} = \frac{\mathbf{v}^T \mathbf{I}^*_{\gamma\gamma}\mathbf{v}}{\|\mathbf{v}\|_2^2} + \frac{\mathbf{v}^T (\mathbf{F}(\boldsymbol{\beta}^*) - \mathbf{I}^*_{\gamma\gamma})\mathbf{v}}{\|\mathbf{v}\|_2^2}$$

$$\geq \lambda_{\min}(\mathbf{I}^*_{\gamma\gamma}) - \left|\frac{\mathbf{v}^T (\mathbf{F}(\boldsymbol{\beta}^*) - \mathbf{I}^*_{\gamma\gamma})\mathbf{v}}{\|\mathbf{v}\|_2^2}\right| \geq \kappa^2 - \frac{\|\mathbf{v}\|_1^2 \|\mathbf{F}(\boldsymbol{\beta}^*) - \mathbf{I}^*_{\gamma\gamma}\|_{\max}}{\|\mathbf{v}\|_2^2},$$

where in the last step, we use the fact that $\lambda_{\min}(\mathbf{I}^*_{\gamma\gamma}) \geq \lambda_{\min}(\mathbf{I}^*) = \kappa^2$. For any $\mathbf{v} \in \mathbb{R}^{d-1}\setminus\{0\}$ satisfying $\|\mathbf{v}_{\bar{\mathcal{S}}'}\|_1 \leq \xi\|\mathbf{v}_{\mathcal{S}'}\|_1$, we have $\|\mathbf{v}\|_1^2 \leq (1+\xi)^2\|\mathbf{v}_{\mathcal{S}'}\|_1^2 \leq (1+\xi)^2 s'\|\mathbf{v}\|_2^2$. Hence, with probability tending to one,

$$\frac{\mathbf{v}^T \mathbf{D}(\widehat{\boldsymbol{\beta}})\mathbf{v}}{\|\mathbf{v}\|_2^2} \geq C\big(\kappa^2 - (1+\xi)^2 s'\|\mathbf{F}(\boldsymbol{\beta}^*) - \mathbf{I}^*_{\gamma\gamma}\|_{\max}\big). \tag{G.9}$$

It is easily seen that $|\mathbf{F}(\boldsymbol{\beta}^*)_{jk}| = |b''(\boldsymbol{Q}_i^T\boldsymbol{\beta}^*)| \cdot |X_{ij}X_{ik}| \leq CK^2$, where $j, k = 1, ..., (d-1)$. By the Hoeffding inequality and the union bound inequality, for any $t > 0$, we get

$$\mathbb{P}_{\boldsymbol{\beta}^*}\left(||\mathbf{F}(\boldsymbol{\beta}^*) - \mathbf{I}^*_{\gamma\gamma}||_{\max} \geq t\right) \leq \sum_{1 \leq j,k \leq (d-1)} \mathbb{P}_{\boldsymbol{\beta}^*}\left(|(\mathbf{F}(\boldsymbol{\beta}^*) - \mathbf{I}^*_{\gamma\gamma})_{jk}| \geq t\right) \leq 2d^2 \exp\left(-\frac{nt^2}{8C^2K^4}\right).$$

Thus, we obtain $||\mathbf{F}(\boldsymbol{\beta}^*) - \mathbf{I}^*_{\gamma\gamma}||_{\max} = \mathcal{O}_{\mathbb{P}}(\sqrt{\frac{\log d}{n}})$. We can plug it into (G.9) and by $s'\sqrt{\frac{\log d}{n}} = o(1)$ we get $\frac{\mathbf{v}^T\mathbf{D}(\widehat{\boldsymbol{\beta}})\mathbf{v}}{||\mathbf{v}||_2^2} \geq C\kappa^2$, with probability tending to one. This completes the proof. □

*Proof of Lemma C.5.* Denote $\bar{\boldsymbol{\Delta}} = \bar{\mathbf{w}} - \mathbf{w}^*$, $\widehat{\epsilon}_i = Y_i - b'(\boldsymbol{Q}_i^T\widehat{\boldsymbol{\beta}})$ and $\epsilon_i = Y_i - b'(\boldsymbol{Q}_i^T\boldsymbol{\beta}^*)$. By Definition,

$$\frac{1}{n}\sum_{i=1}^n \widehat{\epsilon}_i^2(Z_i - \bar{\mathbf{w}}^T\boldsymbol{X}_i)^2 + \lambda'||\bar{\mathbf{w}}||_1 \leq \frac{1}{n}\sum_{i=1}^n \widehat{\epsilon}_i^2(Z_i - \mathbf{w}^{*T}\boldsymbol{X}_i)^2 + \lambda'||\mathbf{w}^*||_1.$$

By rearranging terms, we can show that it is equivalent to

$$\frac{1}{n}\sum_{i=1}^n \widehat{\epsilon}_i^2(\bar{\boldsymbol{\Delta}}^T\boldsymbol{X}_i)^2 \leq \underbrace{\frac{2}{n}\sum_{i=1}^n \widehat{\epsilon}_i^2(Z_i - \mathbf{w}^{*T}\boldsymbol{X}_i)\bar{\boldsymbol{\Delta}}^T\boldsymbol{X}_i + \lambda'||\mathbf{w}^*||_1 - \lambda'||\bar{\mathbf{w}}||_1}_{I_1}. \qquad (\text{G}.10)$$

We now further bound the right hand side of (G.7). Since $\mathbf{w}^*_{\bar{S}'} = 0$, similar to the proof of Lemma C.4, we can show that $\lambda'||\mathbf{w}^*||_1 - \lambda'||\bar{\mathbf{w}}||_1 \leq \lambda'||\bar{\boldsymbol{\Delta}}_{S'}||_1 - \lambda'||\bar{\boldsymbol{\Delta}}_{\bar{S}'}||_1$. Note that for $I_1$, we have

$$I_1 = \frac{2}{n}\sum_{i=1}^n \epsilon_i^2(Z_i - \mathbf{w}^{*T}\boldsymbol{X}_i)\bar{\boldsymbol{\Delta}}^T\boldsymbol{X}_i + \frac{2}{n}\sum_{i=1}^n (\widehat{\epsilon}_i^2 - \epsilon_i^2)(Z_i - \mathbf{w}^{*T}\boldsymbol{X}_i)\bar{\boldsymbol{\Delta}}^T\boldsymbol{X}_i.$$

By the Hölder inequality, and Lemma C.1,

$$\left|\frac{2}{n}\sum_{i=1}^n \epsilon_i^2(Z_i - \mathbf{w}^{*T}\boldsymbol{X}_i)\bar{\boldsymbol{\Delta}}^T\boldsymbol{X}_i\right| \leq \left|\frac{2}{n}\sum_{i=1}^n \epsilon_i^2(Z_i - \mathbf{w}^{*T}\boldsymbol{X}_i)\boldsymbol{X}_i\right|_\infty \cdot ||\bar{\boldsymbol{\Delta}}||_1 \leq C\sqrt{\frac{\log d}{n}}||\bar{\boldsymbol{\Delta}}||_1,$$

for some constant $C$. Here, we use the fact that $\mathbb{E}_{\boldsymbol{\beta}^*}(\epsilon_i^2 \mid \boldsymbol{Q}_i) = b''(\boldsymbol{Q}_i^T\boldsymbol{\beta}^*)$ and the definition of $\mathbf{w}^*$. For the second term, the Cauchy-Schwartz inequality yields,

$$\frac{2}{n}\sum_{i=1}^n (\widehat{\epsilon}_i^2 - \epsilon_i^2)(Z_i - \mathbf{w}^{*T}\boldsymbol{X}_i)\bar{\boldsymbol{\Delta}}^T\boldsymbol{X}_i \leq 2\left|\frac{1}{n}\sum_{i=1}^n \widehat{\epsilon}_i^2(\bar{\boldsymbol{\Delta}}^T\boldsymbol{X}_i)^2\right|^{1/2}\left|\frac{1}{n}\sum_{i=1}^n \frac{(\widehat{\epsilon}_i^2 - \epsilon_i^2)^2}{\widehat{\epsilon}_i^2}(Z_i - \mathbf{w}^{*T}\boldsymbol{X}_i)^2\right|^{1/2}.$$

Since $\widehat{\epsilon}_i^2 - \epsilon_i^2 = (b'(\boldsymbol{Q}_i^T\widehat{\boldsymbol{\beta}}) - b'(\boldsymbol{Q}_i^T\boldsymbol{\beta}^*))(2Y_i - b'(\boldsymbol{Q}_i^T\widehat{\boldsymbol{\beta}}) - b'(\boldsymbol{Q}_i^T\boldsymbol{\beta}^*))$ and $b'(\boldsymbol{Q}_i^T\widehat{\boldsymbol{\beta}}) - b'(\boldsymbol{Q}_i^T\boldsymbol{\beta}^*) = b''(\boldsymbol{Q}_i^T\bar{\boldsymbol{\beta}})\boldsymbol{Q}_i^T(\widehat{\boldsymbol{\beta}} - \boldsymbol{\beta}^*)$, for some $\bar{\boldsymbol{\beta}}$, we can show that

$$\left|\frac{1}{n}\sum_{i=1}^n \frac{(\widehat{\epsilon}_i^2 - \epsilon_i^2)^2}{\widehat{\epsilon}_i^2}(Z_i - \mathbf{w}^{*T}\boldsymbol{X}_i)^2\right| \lesssim \frac{1}{n}\sum_{i=1}^n \{\boldsymbol{Q}_i^T(\widehat{\boldsymbol{\beta}} - \boldsymbol{\beta}^*)\}^2 \lesssim s^*\frac{\log d}{n}.$$

Denote $W = \frac{1}{n}\sum_{i=1}^n \widehat{\epsilon}_i^2 (\bar{\boldsymbol{\Delta}}^T \boldsymbol{X}_i)^2$. Plugging these results into (G.10) and taking $\lambda' = 2C\sqrt{\frac{\log d}{n}}$, we obtain

$$W \leq C'\sqrt{\frac{s^* \log d}{n}} W^{1/2} + 3C\sqrt{\frac{\log d}{n}} \|\bar{\boldsymbol{\Delta}}_{S'}\|_1 - C\sqrt{\frac{\log d}{n}} \|\bar{\boldsymbol{\Delta}}_{\bar{S}'}\|_1, \qquad (\text{G.11})$$

for some constants $C$ and $C'$. Given (G.11), we can use the same arguments as those in the proof of Lemma C.4 to conclude that $W^{1/2} = \mathcal{O}_{\mathbb{P}}(\sqrt{\frac{(s^* \vee s')\log d}{n}})$. Here, the compatibility factor condition is true by invoking Lemma G.4. Since $\epsilon_i^2 \geq C > 0$ for some constant $C$ with probability tending to one, we have $\frac{1}{n}\sum_{i=1}^n (\bar{\boldsymbol{\Delta}}^T \boldsymbol{X}_i)^2 \lesssim W$. Thus, we complete the proof of the second result in (C.3). Similarly, the first result in (C.3) can be also proved.

□

### G.4 Proofs of Auxiliary Lemmas in Appendix D

*Proof of Lemma D.1.* By the proof of Theorem 3.5, we obtain

$$\left| \|\widehat{\boldsymbol{T}}\|_\infty - \|\boldsymbol{T}^*\|_\infty \right| \leq \|\widehat{\boldsymbol{T}} - \boldsymbol{T}^*\|_\infty \leq Cq(n),$$

for some constant $C$, with probability tending to one. This implies that

$$\mathbb{P}_{\boldsymbol{\beta}^*}\left( \|\widehat{\boldsymbol{T}}\|_\infty \leq t \right) - \mathbb{P}\left( \|\boldsymbol{N}\|_\infty \leq t \right) \leq \underbrace{\mathbb{P}_{\boldsymbol{\beta}^*}\left( \|\boldsymbol{T}^*\|_\infty \leq t + Cq(n) \right) - \mathbb{P}\left( \|\boldsymbol{N}\|_\infty \leq t + Cq(n) \right)}_{I_1}$$
$$+ \underbrace{\mathbb{P}\left( \|\boldsymbol{N}\|_\infty \leq t + Cq(n) \right) - \mathbb{P}\left( \|\boldsymbol{N}\|_\infty \leq t \right)}_{I_2}.$$

By Assumption 6.4 and Lemma H.6, $\limsup_{n\to\infty} \sup_{t\in\mathbb{R}} I_1 \leq 0$. By the Gaussian anti-concentration inequality in Lemma H.5, we have $\sup_{t\in\mathbb{R}} I_2 \leq Cq(n)\sqrt{1 \vee \log(d_0/q(n))}$, for some constant $C$. Therefore,

$$\limsup_{n\to\infty} \sup_{t\in\mathbb{R}} \left( \mathbb{P}_{\boldsymbol{\beta}^*}(\|\widehat{\boldsymbol{T}}\|_\infty \leq t) - \mathbb{P}(\|\boldsymbol{N}\|_\infty \leq t) \right) \leq 0.$$

Similarly,

$$\liminf_{n\to\infty} \inf_{t\in\mathbb{R}} \left( \mathbb{P}_{\boldsymbol{\beta}^*}(\|\widehat{\boldsymbol{T}}\|_\infty \leq t) - \mathbb{P}(\|\boldsymbol{N}\|_\infty \leq t) \right) \geq 0.$$

This completes the proof. □

*Proof of Lemma D.2.* The proof of the first result follows from that of Lemma D.1. For the second result, we obtain

$$\left| \|\widehat{\boldsymbol{N}}_e\|_\infty - \|\boldsymbol{N}_e\|_\infty \right| \leq \|\widehat{\boldsymbol{N}}_e - \boldsymbol{N}_e\|_\infty = \max_{1 \leq j \leq d_0} \left| \frac{1}{\sqrt{n}} \sum_{i=1}^n e_i (\widehat{S}_{ij} - S_{ij}) \right|.$$

Since $e_i$ is Gaussian, the Hoeffding inequality implies that

$$\mathbb{P}_e\left(\left|\frac{1}{\sqrt{n}}\sum_{i=1}^{n}e_i(\widehat{S}_{ij}-S_{ij})\right|\geq Cq'(n)\right)\leq 2\exp\left(-\frac{nC^2q'^2(n)}{2\sum_{i=1}^{n}\left(\widehat{S}_{ij}-S_{ij}\right)^2}\right)\leq 2\exp\left(-C_3\log d_0\right),$$

where $C_3$ is some sufficiently large constant. Then, by the union bound inequality,

$$\mathbb{P}_e\left(\max_{1\leq j\leq d_0}\left|\frac{1}{\sqrt{n}}\sum_{i=1}^{n}e_i(\widehat{S}_{ij}-S_{ij})\right|\geq Cq'(n)\right)\leq 2\exp\left(-(C_3-1)\log d_0\right).$$

This implies that $\mathbb{P}_e(|\|\widehat{\mathbf{N}}_e\|_\infty - \|\mathbf{N}_e\|_\infty| \geq Cq'(n)) = o_\mathbb{P}(1)$. $\square$

*Proof of Lemma D.3.* By the proof of Lemma B.2, there exists a constant $C$ such that for any $j=1,...,d_0$,

$$\mathbb{P}_{\boldsymbol{\beta}^*}\left(\left\|\frac{1}{n}\sum_{i=1}^{n}(Z_{ij}\boldsymbol{X}_i - \mathbf{W}_{*j}^{*T}\boldsymbol{X}_i^{\otimes 2})\right\|_\infty \geq C\sqrt{\frac{\log d}{n}}\right) \leq d^{-2},$$

The union bound inequality yields,

$$\mathbb{P}_{\boldsymbol{\beta}^*}\left(\max_{1\leq j\leq d_0}\left\|\frac{1}{n}\sum_{i=1}^{n}(Z_{ij}\boldsymbol{X}_i - \mathbf{W}_{*j}^{*T}\boldsymbol{X}_i^{\otimes 2})\right\|_\infty \geq C\sqrt{\frac{\log d}{n}}\right) \leq d^{-1}, \tag{G.12}$$

Given (G.12), the remaining proof is identical to that of Lemma B.4.

$\square$

*Proof of Lemma D.4.* This is implied by the definition of the Dantzig selector and inequality (G.12).

$\square$

*Proof of Lemma D.5.* By the Hölder inequality,

$$\max_{1\leq j\leq d_0}\sqrt{\frac{1}{n}\sum_{i=1}^{n}\left(\widehat{S}_{ij}-S_{ij}\right)^2} \leq \max_{1\leq j\leq d_0}\max_{1\leq i\leq n}|\widehat{S}_{ij}-S_{ij}|.$$

By applying the triangular inequality repeatedly, we obtain,

$$\begin{aligned}|\widehat{S}_{ij}-S_{ij}| &= \left|\widehat{\epsilon}_i(\boldsymbol{Z}_i-\widehat{\mathbf{W}}_{*j}^T\boldsymbol{X}_i)-\epsilon_i(\boldsymbol{Z}_i-\mathbf{W}_{*j}^{*T}\boldsymbol{X}_i)\right|\\ &\leq \left|(\widehat{\epsilon}_i-\epsilon_i)Z_{ij}\right|+\left|(\widehat{\epsilon}_i-\epsilon_i)\widehat{\mathbf{W}}_{*j}^T\boldsymbol{X}_i\right|+\left|\epsilon_i(\widehat{\mathbf{W}}_{*j}-\mathbf{W}_{*j}^*)^T\boldsymbol{X}_i\right|,\end{aligned} \tag{G.13}$$

where $\widehat{\epsilon}_i = Y_i - \widehat{\boldsymbol{\gamma}}^T\boldsymbol{X}_i$. Note that $\widehat{\epsilon}_i - \epsilon_i = (\widehat{\boldsymbol{\gamma}}-\boldsymbol{\gamma}^*)^T\boldsymbol{X}_i$. Hence,

$$\max_{1\leq j\leq d_0}\max_{1\leq i\leq n}\left|(\widehat{\epsilon}_i-\epsilon_i)Z_{ij}\right| \leq \|\widehat{\boldsymbol{\gamma}}-\boldsymbol{\gamma}^*\|_1 \max_{1\leq j\leq d}\max_{1\leq i\leq n}Q_{ij}^2 = \mathcal{O}_\mathbb{P}\left(s^*\sqrt{\frac{(\log(nd))^3}{n}}\right), \tag{G.14}$$

where the last step follows from the $s^*\sqrt{\log d/n}$ convergence of $\|\widehat{\boldsymbol{\gamma}}-\boldsymbol{\gamma}^*\|_1$ and $\max_{i,j}|Q_{ij}| = \mathcal{O}_\mathbb{P}(\sqrt{\log(nd)})$, due to the sub-Gaussian property of $Q_{ij}$. Since by assumption $\mathbf{W}_{*j}^{*T}\boldsymbol{X}_i$ is also

sub-Gaussian, following the similar arguments, we obtain

$$\max_{1\leq j\leq d_0} \max_{1\leq i\leq n} \left|(\widehat{\epsilon}_i - \epsilon_i)\widehat{\mathbf{W}}_{*j}^T \mathbf{X}_i\right|$$
$$\leq \max_{1\leq j\leq d_0} \max_{1\leq i\leq n} \left|(\widehat{\epsilon}_i - \epsilon_i)\mathbf{W}_{*j}^{*T} \mathbf{X}_i\right| + \max_{1\leq j\leq d_0} \max_{1\leq i\leq n} \left|(\widehat{\epsilon}_i - \epsilon_i)(\widehat{\mathbf{W}}_{*j} - \mathbf{W}_{*j}^*)^T \mathbf{X}_i\right|$$
$$= \mathcal{O}_{\mathbb{P}}\left(s^* \sqrt{\frac{(\log(nd))^3}{n}}\right). \tag{G.15}$$

Similarly, we can show that

$$\max_{i,j} \left|\epsilon_i(\widehat{\mathbf{W}}_j - \mathbf{W}_j^*)^T \mathbf{X}_i\right| \leq \max_j \|\widehat{\mathbf{W}}_{*j} - \mathbf{W}_{*j}^*\|_1 \cdot \max_{i,j} |\epsilon_i| \cdot |X_{ij}| = \mathcal{O}_{\mathbb{P}}\left(s'\sqrt{\frac{(\log(nd))^3}{n}}\right).$$

Together with G.13 G.14 and G.15, we have completed the proof. $\square$

*Proof of Lemma D.6.* Note that $\mathbb{E}_{\boldsymbol{\beta}^*}(S_{ij}^2) = D_{jj}^* = (\mathbf{I}_{\boldsymbol{\theta}|\boldsymbol{\gamma}}^*)_{jj}$. Hence, by Assumption 6.4, we obtain $\max_{1\leq j\leq d_0} |D_{jj}^*| \geq C_{\min} > 0$. Similar to the proof of Corollary 3.7, we can show that

$$\max_{1\leq j\leq d_0} |(\widehat{\mathbf{I}}_{\boldsymbol{\theta}|\boldsymbol{\gamma}})_{jj} - (\mathbf{I}_{\boldsymbol{\theta}|\boldsymbol{\gamma}}^*)_{jj}| \leq \max_{1\leq j\leq d_0} |\nabla^2_{\theta_j\theta_j}\ell(\widehat{\boldsymbol{\beta}}) - I^*_{\theta_j\theta_j}| + \max_{1\leq j\leq d_0} \|\mathbf{W}_{*j}^*\|_1 \max_{1\leq j\leq d_0} \|\mathbf{I}^*_{\theta_j\boldsymbol{\gamma}} - \nabla^2_{\theta_j\boldsymbol{\gamma}}\ell(\widehat{\boldsymbol{\beta}})\|_\infty$$
$$+ \max_{1\leq j\leq d_0} \|\widehat{\mathbf{W}}_{*j} - \mathbf{W}_{*j}^*\|_1 \max_{1\leq j\leq d_0} \|\nabla^2_{\theta_j\boldsymbol{\gamma}}\ell(\widehat{\boldsymbol{\beta}})\|_\infty$$
$$= \mathcal{O}_{\mathbb{P}}(\eta_5(n)) + \mathcal{O}_{\mathbb{P}}(C_W\eta_5(n)) + \mathcal{O}_{\mathbb{P}}(\eta_2(n)C_I) = o_{\mathbb{P}}(1),$$

where $C_W = \max_{1\leq j\leq d_0} \|\mathbf{W}_{*j}^*\|_1$ and $C_I = \max_{1\leq j\leq d_0} \|\mathbf{I}^*_{\theta_j\boldsymbol{\gamma}}\|_\infty$. Since $\max_{1\leq j\leq d_0} |D_{jj}^*| \geq C_{\min} > 0$, by the assumptions in Lemma D.6, we have

$$\|\widehat{\mathbf{D}}^{-1/2} - \mathbf{D}^{*-1/2}\|_{\max} = \max_{1\leq j\leq d_0} \frac{|(\widehat{\mathbf{I}}_{\boldsymbol{\theta}|\boldsymbol{\gamma}})_{jj} - (\mathbf{I}_{\boldsymbol{\theta}|\boldsymbol{\gamma}}^*)_{jj}|}{(\widehat{\mathbf{I}}_{\boldsymbol{\theta}|\boldsymbol{\gamma}})_{jj}^{1/2} + (\mathbf{I}_{\boldsymbol{\theta}|\boldsymbol{\gamma}}^*)_{jj}^{1/2}}$$
$$= \mathcal{O}_{\mathbb{P}}(\eta_5(n)) + \mathcal{O}_{\mathbb{P}}(C_W\eta_5(n)) + \mathcal{O}_{\mathbb{P}}(\eta_2(n)C_I). \tag{G.16}$$

Similar to the proof of Theorem 6.6, we obtain $\left|\|\widehat{\mathbf{T}}_R\|_\infty - \|\mathbf{T}_R^*\|_\infty\right| \leq \|\widehat{\mathbf{T}}_R - \mathbf{T}_R^*\|_\infty$. Together with

$$\|\widehat{\mathbf{T}}_R - \mathbf{T}_R^*\|_\infty \leq \left\|\sqrt{n}\mathbf{D}^{*-1/2}(\widehat{\mathbf{S}}(\mathbf{0},\widehat{\boldsymbol{\gamma}}) - \mathbf{S}(\mathbf{0},\boldsymbol{\gamma}^*))\right\|_\infty + \left\|\sqrt{n}(\widehat{\mathbf{D}}^{-1/2} - \mathbf{D}^{*-1/2})\widehat{\mathbf{S}}(\mathbf{0},\widehat{\boldsymbol{\gamma}})\right\|_\infty,$$

we conclude that $\left|\|\widehat{\mathbf{T}}_R\|_\infty - \|\mathbf{T}_R^*\|_\infty\right| = \mathcal{O}_{\mathbb{P}}(q(n)) + \mathcal{O}_{\mathbb{P}}(r(n))$, where $r(n) = (C_W\eta_5(n) + C_I\eta_2(n))\log d_0$. Here, we use the fact that $\max_{1\leq j\leq d_0} |D_{jj}^*| \geq C_{\min} > 0$ and equation (G.16). This implies that

$$\mathbb{P}_{\boldsymbol{\beta}^*}\left(\|\widehat{\mathbf{T}}_R\|_\infty \leq t\right) - \mathbb{P}\left(\|\mathbf{N}_R\|_\infty \leq t\right)$$
$$\leq \underbrace{\mathbb{P}_{\boldsymbol{\beta}^*}\left(\|\mathbf{T}_R^*\|_\infty \leq t + C(q(n) + r(n))\right) - \mathbb{P}\left(\|\mathbf{N}_R\|_\infty \leq t + C(q(n) + r(n))\right)}_{I_1}$$
$$+ \underbrace{\mathbb{P}\left(\|\mathbf{N}_R\|_\infty \leq t + C(q(n) + r(n))\right) - \mathbb{P}\left(\|\mathbf{N}_R\|_\infty \leq t\right)}_{I_2}.$$

By Assumption 6.4 and Lemma H.6, $\limsup_{n\to\infty} \sup_{t\in\mathbb{R}} I_1 \leq 0$. By the Gaussian anti-concentration inequality in Lemma H.5, we have $\sup_{t\in\mathbb{R}} I_2 \leq C(q(n)+r(n))\sqrt{1 \vee \log(d_0/(q(n)+r(n)))}$, for some constant $C$. The remaining proof is identical to that of Theorem 6.6. □

*Proof of Lemma D.7.* The proof of the first result follows from that of Lemma D.6. For the second result, we obtain $|\|\widehat{\mathbf{N}}_e^R\|_\infty - \|\mathbf{N}_e^R\|_\infty| \leq \|\widehat{\mathbf{N}}_e^R - \mathbf{N}_e^R\|_\infty$. Moreover,

$$\|\widehat{\mathbf{N}}_e^R - \mathbf{N}_e^R\|_\infty \leq \underbrace{\max_{1\leq j \leq d_0}\left|\frac{1}{\sqrt{n}}\sum_{i=1}^n e_i(\widehat{S}_{ij} - S_{ij})(\mathbf{I}^*_{\boldsymbol{\theta}|\boldsymbol{\gamma}})_{jj}^{-1/2}\right|}_{I_1} + \underbrace{\max_{1\leq j \leq d_0}\left|\frac{1}{\sqrt{n}}\sum_{i=1}^n e_i\widehat{S}_{ij}\left((\widehat{\mathbf{I}}_{\boldsymbol{\theta}|\boldsymbol{\gamma}})_{jj}^{-1/2} - (\mathbf{I}^*_{\boldsymbol{\theta}|\boldsymbol{\gamma}})_{jj}^{-1/2}\right)\right|}_{I_2}.$$

Since $e_i$ is Gaussian, the Hoeffding inequality implies that

$$\mathbb{P}_e\left(\left|\frac{1}{\sqrt{n}}\sum_{i=1}^n e_i(\widehat{S}_{ij} - S_{ij})(\mathbf{I}^*_{\boldsymbol{\theta}|\boldsymbol{\gamma}})_{jj}^{-1/2}\right| \geq Cq'(n)\right) \leq 2\exp\left(-\frac{nC^2 q'^2(n)}{2\sum_{i=1}^n (\widehat{S}_{ij} - S_{ij})^2 (\mathbf{I}^*_{\boldsymbol{\theta}|\boldsymbol{\gamma}})_{jj}^{-1}}\right)$$
$$\leq 2\exp\left(-C_3 \log d_0\right),$$

where $C_3$ is some sufficiently large constant. Here, we use the fact that $\max_{1\leq j \leq d_0}|(\mathbf{I}^*_{\boldsymbol{\theta}|\boldsymbol{\gamma}})_{jj}| \geq C_{\min}$. By the union bound inequality, we obtain $\mathbb{P}_e(I_1 \geq Cq'(n)) = o_\mathbb{P}(1)$, for some constant $C$ sufficiently large. We use the similar methods to handle the $I_2$ term. In particular,

$$\mathbb{P}_e\left(\left|\frac{1}{\sqrt{n}}\sum_{i=1}^n e_i\widehat{S}_{ij}\left((\widehat{\mathbf{I}}_{\boldsymbol{\theta}|\boldsymbol{\gamma}})_{jj}^{-1/2} - (\mathbf{I}^*_{\boldsymbol{\theta}|\boldsymbol{\gamma}})_{jj}^{-1/2}\right)\right| \geq Cr'(n)\right) \leq 2\exp\left(-\frac{C^2 r'^2(n)}{2\delta_j}\right),$$

where $\delta_j = n^{-1}\sum_{i=1}^n \widehat{S}_{ij}^2((\widehat{\mathbf{I}}_{\boldsymbol{\theta}|\boldsymbol{\gamma}})_{jj}^{-1/2} - (\mathbf{I}^*_{\boldsymbol{\theta}|\boldsymbol{\gamma}})_{jj}^{-1/2})^2$. It is easily seen that $\max_{1\leq j\leq d_0} \delta_j \leq Cr^2(n)$, where $r(n)$ is given in Lemma D.6. The proof is complete by applying the union bound inequality. □

### G.5 Proofs of Auxiliary Lemmas in Appendix E

*Proof of Lemma E.2.* Since $\|Y_i - \boldsymbol{\gamma}^{oT}\boldsymbol{X}_i\|_{\psi_2} \leq C$ and $\|Q_{ij}\|_{\psi_2} \leq C$, by Lemma H.1, we have $\|X_{ij}(Y_i - \boldsymbol{\gamma}^{oT}\boldsymbol{X}_i)\|_{\psi_1} \leq C^2$, for any $j = 1, ..., (d-1)$. By the definition of the oracle parameter, we have $\mathbb{E}^*(X_{ij}(Y_i - \boldsymbol{\gamma}^{oT}\boldsymbol{X}_i)) = 0$. By Lemma H.2, we have for any $t > 0$ and $j = 1, ..., d$,

$$\mathbb{P}^*\left(\frac{1}{n}\left|\sum_{i=1}^n X_{ij}(Y_i - \boldsymbol{\gamma}^{oT}\boldsymbol{X}_i)\right| \geq t\right) \leq 2\exp\left[-C'' \min\left(\frac{t^2}{C^4}, \frac{t}{C^2}\right)n\right].$$

By the union bound inequality,

$$\mathbb{P}^*\left(\left\|\frac{1}{n}\sum_{i=1}^n \boldsymbol{X}_i(Y_i - \boldsymbol{\gamma}^{oT}\boldsymbol{X}_i)\right\|_\infty \geq t\right) \leq 2d\exp\left[-C'' \min\left(\frac{t^2}{C^4}, \frac{t}{C^2}\right)n\right].$$

With $t = C\sqrt{\frac{\log d}{n}}$, for some constant $C$, Lemma E.2 holds.

□

*Proof of Lemma E.3.* The proof is similar to that of Lemma B.3. For reader's convenience, we give a rigorous proof. Recall that $D(\boldsymbol{\beta}_1, \boldsymbol{\beta}) = (\boldsymbol{\beta}_1 - \boldsymbol{\beta})^T \mathbf{H}_Q (\boldsymbol{\beta}_1 - \boldsymbol{\beta})$. Denote the support set of $\boldsymbol{\beta}^o$ by $S^o$ and $\widehat{\boldsymbol{\Delta}} = \widehat{\boldsymbol{\beta}} - \boldsymbol{\beta}^o$. Note that $\boldsymbol{\beta}_j^o = 0$ if $j \in \bar{S}^o$. Thus

$$D(\widehat{\boldsymbol{\beta}}, \boldsymbol{\beta}^o) = \widehat{\boldsymbol{\Delta}}^T \left\{ -\frac{1}{n} \sum_{i=1}^n \boldsymbol{Q}_i (Y_i - \widehat{\boldsymbol{\beta}}^T \boldsymbol{Q}_i) + \frac{1}{n} \sum_{i=1}^n \boldsymbol{Q}_i (Y_i - \boldsymbol{\beta}^{oT} \boldsymbol{Q}_i) \right\}$$

$$= -\widehat{\boldsymbol{\beta}}_{\bar{S}^o}^T \left\{ \frac{1}{n} \sum_{i=1}^n \boldsymbol{Q}_{i\bar{S}^o} (Y_i - \widehat{\boldsymbol{\beta}}^T \boldsymbol{Q}_i) \right\} - \widehat{\boldsymbol{\Delta}}_{S^o}^T \left\{ \frac{1}{n} \sum_{i=1}^n \boldsymbol{Q}_{iS^o} (Y_i - \widehat{\boldsymbol{\beta}}^T \boldsymbol{Q}_i) \right\}$$

$$+ \widehat{\boldsymbol{\Delta}}^T \left\{ \frac{1}{n} \sum_{i=1}^n \boldsymbol{Q}_i (Y_i - \boldsymbol{\beta}^{oT} \boldsymbol{Q}_i) \right\}.$$

Denote $\mathbf{G}^o = \frac{1}{n} \sum_{i=1}^n \boldsymbol{Q}_i (Y_i - \boldsymbol{\beta}^{oT} \boldsymbol{Q}_i)$. Under the null hypothesis $\theta^o = 0$, Lemma E.2 implies that, $\|\mathbf{G}^o\|_\infty \leq C \sqrt{\frac{\log d}{n}}$ with probability at least $1 - d^{-1}$. With the same $C$ and $\lambda = 2C \sqrt{\frac{\log d}{n}}$, by the Karush-Kuhn-Tucker (KKT) condition and Hölder inequality, we have

$$D(\widehat{\boldsymbol{\beta}}, \boldsymbol{\beta}^o) \leq -\lambda \|\widehat{\boldsymbol{\Delta}}_{\bar{S}^o}\|_1 + \lambda \|\widehat{\boldsymbol{\Delta}}_{S^o}\|_1 + \|\widehat{\boldsymbol{\Delta}}\|_1 \|\mathbf{G}^o\|_\infty \leq C \sqrt{\frac{\log d}{n}} (3\|\widehat{\boldsymbol{\Delta}}_{S^o}\|_1 - \|\widehat{\boldsymbol{\Delta}}_{\bar{S}^o}\|_1). \quad \text{(G.17)}$$

Since $D(\widehat{\boldsymbol{\beta}}, \boldsymbol{\beta}^o) \geq 0$, we have $\|\widehat{\boldsymbol{\Delta}}_{\bar{S}^o}\|_1 \leq 3\|\widehat{\boldsymbol{\Delta}}_{S^o}\|_1$. In addition, by the RE condition in Lemma G.1, $D(\widehat{\boldsymbol{\beta}}, \boldsymbol{\beta}^o) \geq \kappa \|\widehat{\boldsymbol{\Delta}}\|_2^2$. Since by (G.17), we have that $D(\widehat{\boldsymbol{\beta}}, \boldsymbol{\beta}^o) \leq 3C\sqrt{\frac{s^* \log d}{n}} \|\widehat{\boldsymbol{\Delta}}\|_2$, this implies that $\|\widehat{\boldsymbol{\Delta}}\|_2 \leq \frac{3C}{\kappa} \sqrt{\frac{s^* \log d}{n}}$. Finally,

$$\|\widehat{\boldsymbol{\Delta}}\|_1 \leq 4\|\widehat{\boldsymbol{\Delta}}_{S^o}\|_1 \leq 4s^{*1/2} \|\widehat{\boldsymbol{\Delta}}_{S^o}\|_2 \leq 4s^{*1/2} \|\widehat{\boldsymbol{\Delta}}\|_2 \leq \frac{12Cs^*}{\kappa} \sqrt{\frac{\log d}{n}}.$$

It is easily seen from (G.17) that $D(\widehat{\boldsymbol{\beta}}, \boldsymbol{\beta}^o) \leq \frac{36C^2}{\kappa} \frac{s^* \log d}{n}$. This completes the proof. □

*Proof of Lemma E.4.* The proof is identical to that of Lemma B.4. □

*Proof of Lemma E.5.* The proof is identical to that of Lemma B.2. □

*Proof of Lemma E.6.* We now verify the Lyapunov condition. Note that

$$\|(Y_i - \boldsymbol{\gamma}^{oT} \boldsymbol{X}_i)(Z_i - \mathbf{w}^{oT} \boldsymbol{X}_i)\|_{\psi_1} \leq 2\|Y_i - \boldsymbol{\gamma}^{oT} \boldsymbol{X}_i\|_{\psi_2} \|Z_i - \mathbf{w}^{oT} \boldsymbol{X}_i\|_{\psi_2} \leq 4C^2.$$

By the definition of $\psi_1$ norm,

$$\|(Y_i - \boldsymbol{\gamma}^{oT} \boldsymbol{X}_i)(Z_i - \mathbf{w}^{oT} \boldsymbol{X}_i)\|_{\psi_1} \geq 3^{-1} (\mathbb{E}_{\boldsymbol{\beta}^*} |(Y_i - \boldsymbol{\gamma}^{oT} \boldsymbol{X}_i)(Z_i - \mathbf{w}^{oT} \boldsymbol{X}_i)|^3)^{1/3}.$$

This implies that $\mathbb{E}^* |(Y_i - \boldsymbol{\gamma}^{oT} \boldsymbol{X}_i)(Z_i - \mathbf{w}^{oT} \boldsymbol{X}_i)|^3 \leq 12^3 C^6$. On the other hand, it is easily seen that $\mathbf{v}^{oT} \boldsymbol{\Sigma}^o \mathbf{v}^o \geq \lambda_{\min}(\boldsymbol{\Sigma}^o) \|\mathbf{v}^o\|_2^2 \geq \lambda_{\min}(\boldsymbol{\Sigma}^o)$. This implies that

$$n^{-3/2} (\mathbf{v}^{oT} \boldsymbol{\Sigma}^o \mathbf{v}^o)^{-3/2} \sum_{i=1}^n \mathbb{E}^* |(Y_i - \boldsymbol{\gamma}^{oT} \boldsymbol{X}_i)(Z_i - \mathbf{w}^{oT} \boldsymbol{X}_i)|^3 \leq \frac{C}{n^{1/2} \lambda_{\min}^{3/2}(\boldsymbol{\Sigma}^o)} \to 0,$$

as $n \to \infty$. Hence, the Lyapunov condition holds. This completes the proof. □

*Proof of Lemma E.7.* Note that

$$\|\widehat{\boldsymbol{\Sigma}} - \boldsymbol{\Sigma}^o\|_{\max} \leq \|\widetilde{\boldsymbol{\Sigma}} - \boldsymbol{\Sigma}^o\|_{\max} + \|\widehat{\boldsymbol{\Sigma}} - \widetilde{\boldsymbol{\Sigma}}\|_{\max},$$

where $\widetilde{\boldsymbol{\Sigma}} = \frac{1}{n}\sum_{i=1}^n \left(\boldsymbol{Q}_i^{\otimes 2}(Y_i - \boldsymbol{\beta}^{oT}\boldsymbol{Q}_i)^2\right)$. For the first term, it is easily seen that Lemma H.3 holds with $r = 1/2$. By the union bound inequality, we can show that with $x = C\sqrt{\frac{(\log d)^5}{n}}$ for some sufficiently large constant $C'$,

$$\mathbb{P}^*\big(\|\widetilde{\boldsymbol{\Sigma}} - \boldsymbol{\Sigma}^o\|_{\max} \geq x\big) \leq 4d^2 \exp\big(-\frac{1}{8}n^{1/5}x^{2/5}\big) + 4nd^2 C_1 \exp\big(-C_2 n^{1/5}x^{2/5}\big),$$

for some constants $C_1$ and $C_2$. Hence, with high probability, $\|\widetilde{\boldsymbol{\Sigma}} - \boldsymbol{\Sigma}^o\|_{\max} \leq C\sqrt{\frac{(\log d)^5}{n}}$, for some constant $C$. For the second term, for any $(j,k)$,

$$\begin{aligned}
|\widehat{\boldsymbol{\Sigma}}_{jk} - \widetilde{\boldsymbol{\Sigma}}_{jk}| &= \left|\frac{1}{n}\sum_{i=1}^n Q_{ij}Q_{ik}\boldsymbol{Q}_i^T(\widehat{\boldsymbol{\beta}} - \boldsymbol{\beta}^o)(2Y_i - \boldsymbol{Q}_i^T(\widehat{\boldsymbol{\beta}} + \boldsymbol{\beta}^o))\right| \\
&\leq \underbrace{\left|\frac{2}{n}\sum_{i=1}^n Q_{ij}Q_{ik}\boldsymbol{Q}_i^T(\widehat{\boldsymbol{\beta}} - \boldsymbol{\beta}^o)(Y_i - \boldsymbol{Q}_i^T\boldsymbol{\beta}^o)\right|}_{I_{1jk}} + \underbrace{\left|\frac{1}{n}\sum_{i=1}^n Q_{ij}Q_{ik}\{\boldsymbol{Q}_i^T(\widehat{\boldsymbol{\beta}} - \boldsymbol{\beta}^o)\}^2\right|}_{I_{2jk}}.
\end{aligned}$$

We now separately bound $I_{1jk}$ and $I_{2jk}$. The Cauchy-Schwartz inequality and Lemma E.3 yields,

$$\max_{1\leq j,k\leq d} |I_{1jk}| \lesssim \max_{1\leq j,k\leq d}\sqrt{\frac{1}{n}\sum_{i=1}^n Q_{ij}^2 Q_{ik}^2} \cdot \sqrt{\frac{1}{n}\sum_{i=1}^n \left(\boldsymbol{Q}_i^T(\widehat{\boldsymbol{\beta}} - \boldsymbol{\beta}^o)\right)^2} = \mathcal{O}_\mathbb{P}\left(\sqrt{\frac{s^* \log d}{n}}\right).$$

In the last step, we claim that $\max_{1\leq j,k\leq d}\sqrt{\frac{1}{n}\sum_{i=1}^n Q_{ij}^2 Q_{ik}^2} = \mathcal{O}_\mathbb{P}(1)$. This is true, since Lemma H.3 with $r = 1/2$ implies

$$\max_{1\leq j,k\leq d}\left|\frac{1}{n}\sum_{i=1}^n Q_{ij}^2 Q_{ik}^2 - \mathbb{E}^*(Q_{ij}^2 Q_{ik}^2)\right| \leq C\sqrt{\frac{(\log d)^5}{n}},$$

with high probability and $\mathbb{E}^*(Q_{ij}^2 Q_{ik}^2)$ are bounded by a universal constant. By the sub-Gaussian property, we can show that $\max_{1\leq i\leq n}\max_{1\leq j,k\leq d}\left|Q_{ij}^2 Q_{ik}^2\right| = \mathcal{O}_\mathbb{P}((\log(nd))^2)$. The Hölder inequality and Lemma E.3 yields,

$$\max_{1\leq j,k\leq d}|I_{2jk}| \leq \max_{1\leq i\leq n}\max_{1\leq j,k\leq d}\left|Q_{ij}^2 Q_{ik}^2\right| \cdot \frac{1}{n}\sum_{i=1}^n\{\boldsymbol{Q}_i^T(\widehat{\boldsymbol{\beta}} - \boldsymbol{\beta}^o)\}^2 = \mathcal{O}_\mathbb{P}\left(\frac{s^*(\log(nd))^3}{n}\right).$$

Hence, $\|\widehat{\boldsymbol{\Sigma}} - \widetilde{\boldsymbol{\Sigma}}\|_{\max} = \mathcal{O}_\mathbb{P}(\sqrt{\frac{s^* \log d}{n}})$, and

$$\|\widehat{\boldsymbol{\Sigma}} - \boldsymbol{\Sigma}^o\|_{\max} = \mathcal{O}_\mathbb{P}\left(\sqrt{\frac{s^* \log d}{n}} \vee \sqrt{\frac{(\log d)^5}{n}}\right).$$

Similarly, we have
$$\|(\widehat{\boldsymbol{\Sigma}} - \boldsymbol{\Sigma}^o)\mathbf{v}^o\|_\infty \leq \|(\widetilde{\boldsymbol{\Sigma}} - \boldsymbol{\Sigma}^o)\mathbf{v}^o\|_\infty + \|(\widehat{\boldsymbol{\Sigma}} - \widetilde{\boldsymbol{\Sigma}})\mathbf{v}^o\|_\infty.$$

To bound the first term, we apply Lemma H.3 with $r = 1/2$. This yields, $\|(\widetilde{\boldsymbol{\Sigma}} - \boldsymbol{\Sigma}^o)\mathbf{v}^o\|_\infty = \mathcal{O}_\mathbb{P}(\sqrt{\frac{(\log d)^5}{n}})$. For the second term, we have, for any $j = 1, ..., d$,

$$\begin{aligned}
|\{(\widehat{\boldsymbol{\Sigma}} - \widetilde{\boldsymbol{\Sigma}})\mathbf{v}^o\}_j| &= \left|\frac{1}{n}\sum_{i=1}^n Q_{ij}\boldsymbol{Q}_i^T\mathbf{v}^o\boldsymbol{Q}_i^T(\widehat{\boldsymbol{\beta}} - \boldsymbol{\beta}^o)(2Y_i - \boldsymbol{Q}_i^T(\widehat{\boldsymbol{\beta}} + \boldsymbol{\beta}^o))\right| \\
&\leq \left|\frac{2}{n}\sum_{i=1}^n Q_{ij}\boldsymbol{Q}_i^T\mathbf{v}^o\boldsymbol{Q}_i^T(\widehat{\boldsymbol{\beta}} - \boldsymbol{\beta}^o)(Y_i - \boldsymbol{Q}_i^T\boldsymbol{\beta}^o)\right| + \left|\frac{1}{n}\sum_{i=1}^n Q_{ij}\boldsymbol{Q}_i^T\mathbf{v}^o\{\boldsymbol{Q}_i^T(\widehat{\boldsymbol{\beta}} - \boldsymbol{\beta}^o)\}^2\right|.
\end{aligned}$$

We can use the similar arguments to bound the last two terms separately. This gives $\|(\widehat{\boldsymbol{\Sigma}} - \widetilde{\boldsymbol{\Sigma}})\mathbf{v}^o\|_\infty = \mathcal{O}_\mathbb{P}(\sqrt{\frac{s^* \log d}{n}})$, and therefore

$$\|(\widehat{\boldsymbol{\Sigma}} - \boldsymbol{\Sigma}^o)\mathbf{v}^o\|_{\max} = \mathcal{O}_\mathbb{P}\left(\sqrt{\frac{s^* \log d}{n}} \vee \sqrt{\frac{(\log d)^5}{n}}\right).$$

This completes the proof of Lemma E.7.

□

## H  Supplementary Lemmas

For the reader's convenience, in this appendix, we present some existing results frequently used in our proofs. The first two Lemmas are the basic properties for the sub-Gaussian or sub-exponential variables. Lemma H.3 is a generalization of the Bernstein inequality in Lemma H.2. A similar version for U-statistic is proved by Ning and Liu (2014). For completeness, we give the proof of this lemma. Lemma H.4 also known as Berry-Esseen Theorem can be found in Theorem 5.4 of Petrov (1995). The last three Lemmas H.5, H.6 and H.7 are adapted from Chernozhukov et al. (2013).

**Lemma H.1.** Assume that $X$ and $Y$ are sub-Gaussian. Then $\|XY\|_{\psi_1} \leq 2\|X\|_{\psi_2}\|Y\|_{\psi_2}$

**Lemma H.2** (Bernstein Inequality). Let $X_1, ..., X_n$ be independent mean 0 sub-exponential random variables and let $K = \max_i \|X_i\|_{\psi_1}$. The for any $t > 0$, we have

$$\mathbb{P}_{\boldsymbol{\beta}^*}\left(\frac{1}{n}\left|\sum_{i=1}^n X_i\right| \geq t\right) \leq 2\exp\left[-C''\min\left(\frac{t^2}{K^2}, \frac{t}{K}\right)n\right],$$

where $C'' > 0$ is a universal constant.

**Lemma H.3.** (Ning and Liu, 2014, General Concentration Inequality) Let $X_1, ..., X_n$ be i.i.d mean 0 random variables. If there exist constants $L_1$ and $L_2$, such that $\mathbb{P}(|X_i| \geq x) \leq L_1 \exp(-L_2 x^r)$, for some $r > 0$, then for $x \geq \sqrt{8\mathbb{E}(X_i^2)/n}$,

$$\mathbb{P}\left(\left|\frac{1}{n}\sum_{i=1}^n X_i\right| \geq x\right) \leq 4\exp\left(-\frac{1}{8}n^{r/(2+r)}x^{2r/(2+r)}\right) + 4nL_1\exp\left(-\frac{L_2 n^{r/(2+r)}x^{2r/(2+r)}}{2^r}\right).$$

*Proof of Lemma H.3.* Denote $U_n = \frac{1}{n} \sum_{i=1}^n X_i$. By the Markov inequality, for $x \geq \sqrt{8\mathbb{E}(X_i^2)/n}$,

$$\mathbb{P}(|U_n| \geq x/2) \leq \frac{4\mathbb{E}(U_n^2)}{x^2} = \frac{4\mathbb{E}(X_i^2)}{nx^2} \leq \frac{1}{2}. \tag{H.1}$$

Let $Y_1, ..., Y_n$ be independent copies of $X_1, ..., X_n$. Denote $V_n = \frac{1}{n} \sum_{i=1}^n Y_i$. Then for $x \geq \sqrt{8\mathbb{E}(X_i^2)/n}$, by (H.1), we get

$$\mathbb{P}(|U_n| \geq x) \leq \frac{\mathbb{P}(|U_n - V_n| \geq x/2)}{\mathbb{P}(|V_n| \leq x/2)} \leq 2\mathbb{P}(|U_n - V_n| \geq x/2). \tag{H.2}$$

For some $c > 0$, define $(X_i - Y_i)^{(c)} = (X_i - Y_i) I(|X_i - Y_i| \leq c)$. By (H.2), $\mathbb{P}(|U_n| \geq x) \leq 2(I_1 + I_2)$, where

$$I_1 = \mathbb{P}\left(\left|\frac{1}{n}\sum_{i=1}^n (X_i - Y_i)^{(c)}\right| \geq \frac{x}{2}\right), \text{ and } I_2 = \mathbb{P}\left(\max_i \left|X_i - Y_i - (X_i - Y_i)^{(c)}\right| > 0\right).$$

Note that $\mathbb{E}(X_i - Y_i)^{(c)} = 0$. By the Hoeffding inequality, $I_1 \leq 2\exp\{-nx^2/(8c^2)\}$, and

$$I_2 \leq 2\sum_{i=1}^n \mathbb{P}(|X_i| > c/2) \leq 2nL_1 \exp(-L_2 c^r/2^r).$$

Lemma H.3 follows by taking $c = n^{1/(2+r)} x^{2/(2+r)}$. □

**Lemma H.4** (Berry-Esseen Theorem). Assume that $X_1, ..., X_n$ are independent random variable with $\mathbb{E}(X_i) = 0$ and $\mathbb{E}(X_i^2) = \sigma_i^2$ and $\mathbb{E}(|X_i|^3) = \rho_i$. Then

$$\sup_{x \in \mathbb{R}} \left|\mathbb{P}\left(\frac{1}{s_n}\sum_{i=1}^n X_i \leq x\right) - \Phi(x)\right| \leq C\left(\sum_{i=1}^n \sigma_i^2\right)^{-3/2} \left(\sum_{i=1}^n \rho_i\right),$$

where $s_n^2 = \sum_{i=1}^n \sigma_i^2$ and $C$ is an absolute positive constant.

**Lemma H.5.** (Chernozhukov et al., 2013, Gaussian Anti-Concentration) Assume that $\boldsymbol{X} \sim N_d(0, \boldsymbol{\Sigma})$. Let $\sigma_j^2 = \Sigma_{jj}$ and define $\sigma_{\min} = \min_j \sigma_j$ and $\sigma_{\max} = \max_j \sigma_j$. Then

$$\sup_{t \in \mathbb{R}} \left|\mathbb{P}(\max_j |X_j| \leq t + \epsilon) - \mathbb{P}(\max_j |X_j| \leq t)\right| \leq C\epsilon \sqrt{1 \vee \log(d/\epsilon)},$$

where $C$ is a constant depending on $\sigma_{\min}$ and $\sigma_{\max}$.

**Lemma H.6.** (Chernozhukov et al., 2013) Let $\boldsymbol{X}_1, ..., \boldsymbol{X}_n$ be $n$ i.i.d random vectors, where $\boldsymbol{X}_i = (X_{i1}, ..., X_{id})$. Assume that there are some constants $0 < c_1 < C_1$ such that $\mathbb{E}(X_{ij}^2) \geq c_1$ and $X_{ij}$ is sub-exponential with $\|X_{ij}\|_{\psi_1} \leq C_1$ for all $1 \leq j \leq d$. If $(\log(dn))^7/n = o(1)$, then

$$\lim_{n \to \infty} \sup_{t \in \mathbb{R}} \left|\mathbb{P}\left(\left\|\frac{1}{\sqrt{n}}\sum_{i=1}^n \boldsymbol{X}_i\right\|_\infty \leq t\right) - \mathbb{P}\left(\|\boldsymbol{N}\|_\infty \leq t\right)\right| = 0,$$

where $\boldsymbol{N} \sim N_d(\boldsymbol{0}, \mathbb{E}(\boldsymbol{X}_i^{\otimes 2}))$.

**Lemma H.7.** (Chernozhukov et al., 2013) Let $\boldsymbol{X}_1, ..., \boldsymbol{X}_n$ be $n$ i.i.d random vectors, where $\boldsymbol{X}_i = (X_{i1}, ..., X_{id})$. Assume that there are some constants $0 < c_1 < C_1$ such that $\mathbb{E}(X_{ij}^2) \geq c_1$ and $X_{ij}$ is sub-exponential with $\|X_{ij}\|_{\psi_1} \leq C_1$ for all $1 \leq j \leq d$. If $(\log(dn))^9/n = o(1)$, then

$$\lim_{n\to\infty} \sup_{\alpha \in (0,1)} \left| \mathbb{P}\bigg( \|\boldsymbol{T}\|_\infty \leq c_N(\alpha) \bigg) - \alpha \right| = 0,$$

where $\boldsymbol{T} = \frac{1}{\sqrt{n}} \sum_{i=1}^n \boldsymbol{X}_i$, $c_N(\alpha) = \inf\{t \in \mathbb{R} : \mathbb{P}_e(\|\boldsymbol{N}_e\|_\infty \leq t) \geq \alpha\}$, and $\boldsymbol{N}_e = \frac{1}{\sqrt{n}} \sum_{i=1}^n e_i \boldsymbol{X}_i$ and $e_1, ..., e_n$ are independent copies of $N(0,1)$. In addition, assume that there exist statistics $\widehat{\boldsymbol{T}}$ and $\widehat{\boldsymbol{N}}_e$ such that

$$\mathbb{P}(|\|\widehat{\boldsymbol{T}}\|_\infty - \|\boldsymbol{T}\|_\infty| \geq \xi_1) \to 0, \quad \text{and} \quad \mathbb{P}_e(|\|\widehat{\boldsymbol{N}}_e\|_\infty - \|\boldsymbol{N}_e\|_\infty| \geq \xi_2) \to_p 0,$$

for some $\xi_1$ and $\xi_2$ depending on $n$ and $\xi = \xi_1 \vee \xi_2$. If $\xi \sqrt{1 \vee \log(d/\xi)} = o(1)$, then

$$\lim_{n\to\infty} \sup_{\alpha \in (0,1)} \left| \mathbb{P}\bigg( \|\widehat{\boldsymbol{T}}\|_\infty \leq c'_N(\alpha) \bigg) - \alpha \right| = 0,$$

where

$$c'_N(\alpha) = \inf\{t \in \mathbb{R} : \mathbb{P}_e(\|\widehat{\boldsymbol{N}}_e\|_\infty \leq t) \geq \alpha\}.$$